\documentclass{article}
\usepackage{frExamplee}
\usepackage{graphics} 
\usepackage{graphicx}
\usepackage[caption=false, font=footnotesize]{subfig}
\usepackage{apalike}
\usepackage{setspace}
\usepackage[nolist,nohyperlinks]{acronym}
\usepackage{url}
\usepackage{todonotes}
\usepackage{siunitx}
\usepackage{algorithm}
\usepackage[noend]{algpseudocode}
\usepackage{comment}
\usepackage{float}
\usepackage{placeins}
\usepackage{booktabs}
\usepackage{amsmath} 
\usepackage{amssymb}
\usepackage{xcolor}
\usepackage{bm}
\usepackage{hyperref}
\usepackage{svg}
\usepackage{cite}
\usepackage{caption}
\usepackage{makecell}
\usepackage{multicol}
\usepackage{verbatim}
\usepackage[toc,page]{appendix}

\title{Towards Autonomous Robotic Precision Harvesting: Mapping, Localization, Planning and Control for a Legged Tree Harvester}
\author{
Edo Jelavic\thanks{Corresponding author: Edo Jelavic. Alternative email address: edo.jelavic@gmail.com} \\
Robotic Systems Lab\\
ETH Zurich\\
8092 Zurich, Switzerland \\
\texttt{edo.jelavic@mavt.ethz.ch} \\
\And
Dominic Jud \\
Robotic Systems Lab\\
ETH Zurich\\
8092 Zurich, Switzerland \\
\texttt{dominic.jud@mavt.ethz.ch} \\
\AND
Pascal Egli \\
Robotic Systems Lab\\
ETH Zurich\\
8092 Zurich, Switzerland \\
\texttt{pascalarturo.egli@mavt.ethz.ch} \\
\And
Marco Hutter \\
Robotic Systems Lab\\
ETH Zurich\\
8092 Zurich, Switzerland \\
\texttt{marco.hutter@mavt.ethz.ch} \\
}

\begin{document}
\maketitle
\begin{abstract}
This paper presents an integrated system for performing precision harvesting missions using a legged harvester. Our harvester performs a challenging task of autonomous navigation and tree grabbing in a confined, \ac{GPS} denied forest environment. Strategies for mapping, localization, planning, and control are proposed and integrated into a fully autonomous system. The mission starts with a human mapping the area of interest using a custom-made sensor module. Subsequently, a human expert selects the trees for harvesting. The sensor module is then mounted on the machine and used for localization within the given map. A planning algorithm searches for both an approach pose and a path in a single path planning problem. We design a path following controller leveraging the legged harvester's capabilities for negotiating rough terrain. Upon reaching the approach pose, the machine grabs a tree with a general-purpose gripper. This process repeats for all the trees selected by the operator. Our system has been tested on a testing field with tree trunks and in a natural forest. To the best of our knowledge, this is the first time this level of autonomy has been shown on a full-size hydraulic machine operating in a realistic environment.
\end{abstract}
\begin{acronym}
\acro{DoF}{Degrees of Freedom}
\acro{GDP}{Gross Domestic Product}
\acro{UAV}{Unmanned Aerial Vehicle}
\acro{LIDAR}{Light Detection And Ranging}
\acro{UGV}{Unmanned Ground Vehicle}
\acro{SLAM}{Simultaneous Localization and Mapping}
\acro{EKF}{Extended Kalman Filter}
\acro{GPS}{Global Positioning System}
\acro{ICP}{Iterative closest point}
\acro{MPC}{Model Predictive Control}
\acro{ATV}{All Terrain Vehicle}
\acro{RTK}{Real-time Kinematic}
\acro{RRT}{Rapidly-exploring Random Tree}
\acro{HEAP}{Hydraulic Excavator for Autonomous Purpose}
\acro{IMU}{Inertial Measurement Unit}
\acro{GNSS}{Global Navigation Satellite System}
\acro{ROS}{Robot Operating System}
\acro{VIO}{Visual Inertial Odometry}
\acro{OS}{Operating System}
\acro{PTP}{Precision Time Protocol}
\acro{CPU}{Central Processing Unit}
\acro{DoF}{Degrees of Freedom}
\acro{NTP}{Network Time Protocol}
\acro{CAD}{Computer Aided Design}
\acro{GUI}{Graphical User Interface}
\acro{HBC}{Hip Balancing Controller}
\acro{IK}{Inverse Kinematics}
\acro{PCA}{Principal Component Analysis}
\acro{USB}{Universal Serial Bus}
\acro{HO}{Hierarchical Optimization}
\acro{FOV}{Field of View}
\acro{PRM}{Probabilistic Roadmap}
\acro{CAN}{Controller Area Network}
\acro{ID}{Inverse Dynamics}
\acro{OMPL}{Open Motion Planning Library}
\acro{RS}{Reeds-Shepp}
\acro{ODE}{Open Dynamics Engine}
\acro{ZMP}{Zero Moment Point}
\acro{DBH}{Diameter at Breast Height}
\end{acronym}
\section{Introduction}
\label{sec::introduction}
Forests cover roughly 30\% of the world's land surface \cite{FOA}. Trees provide the human population with raw materials, and they are a significant source of energy. Besides, woodlands provide a habitat for animal wildlife and contribute to the fight against climate change. Therefore, efficient forest management is of interest to all humankind.

The forestry industry has a share of 1.2\% in the world's \ac{GDP} and employs about 30-45 million people \cite{renner2008green}. In some countries, forest industry products have a significant share in the total value of exported goods, e.g., Finland(18\%), Latvia(16\%) \cite{swedishForestAgency}. Given the growing labor, shortage \cite{hawkinson2017forestry}, and classification of tree harvesting as a \emph{3D} job (dirty, difficult, and dangerous), automation of forestry operations will be of high importance in the future.
\subsection{Automation in Forestry}
\begin{figure*}[tb]
\centering
\subfloat[Komatsu-895 forwarder \label{fig::big_forwarder}]{
       \includegraphics[width=0.515\textwidth]{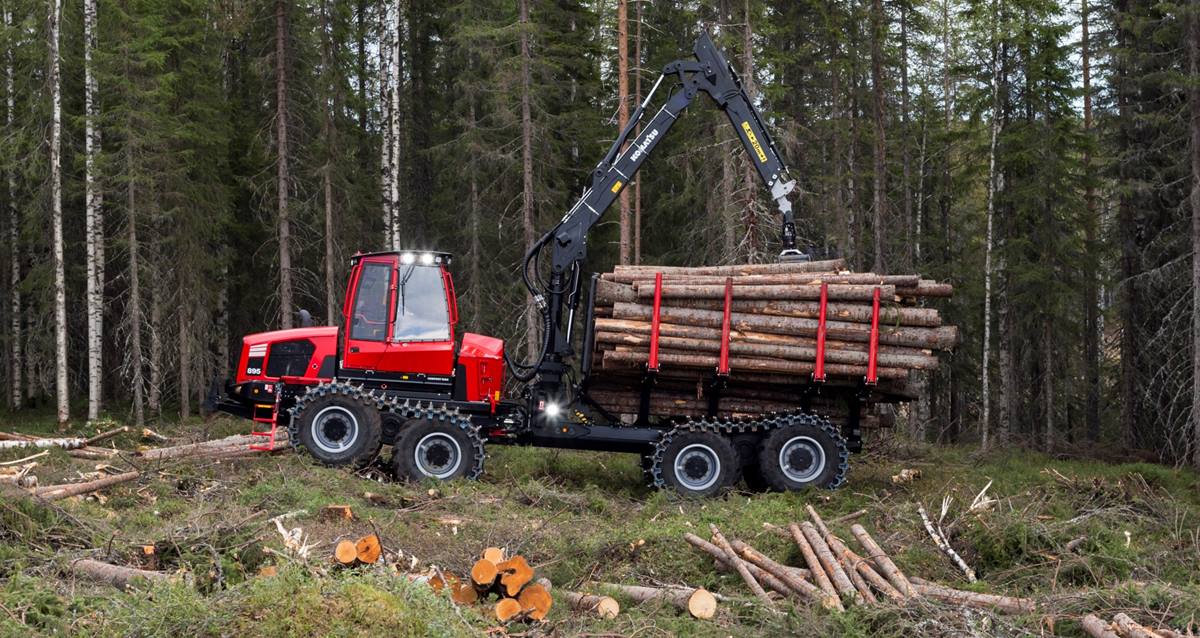}}
     \subfloat[Komatsu-951 harvester  \label{fig::big_harvester}]{
       \includegraphics[width=0.485\textwidth]{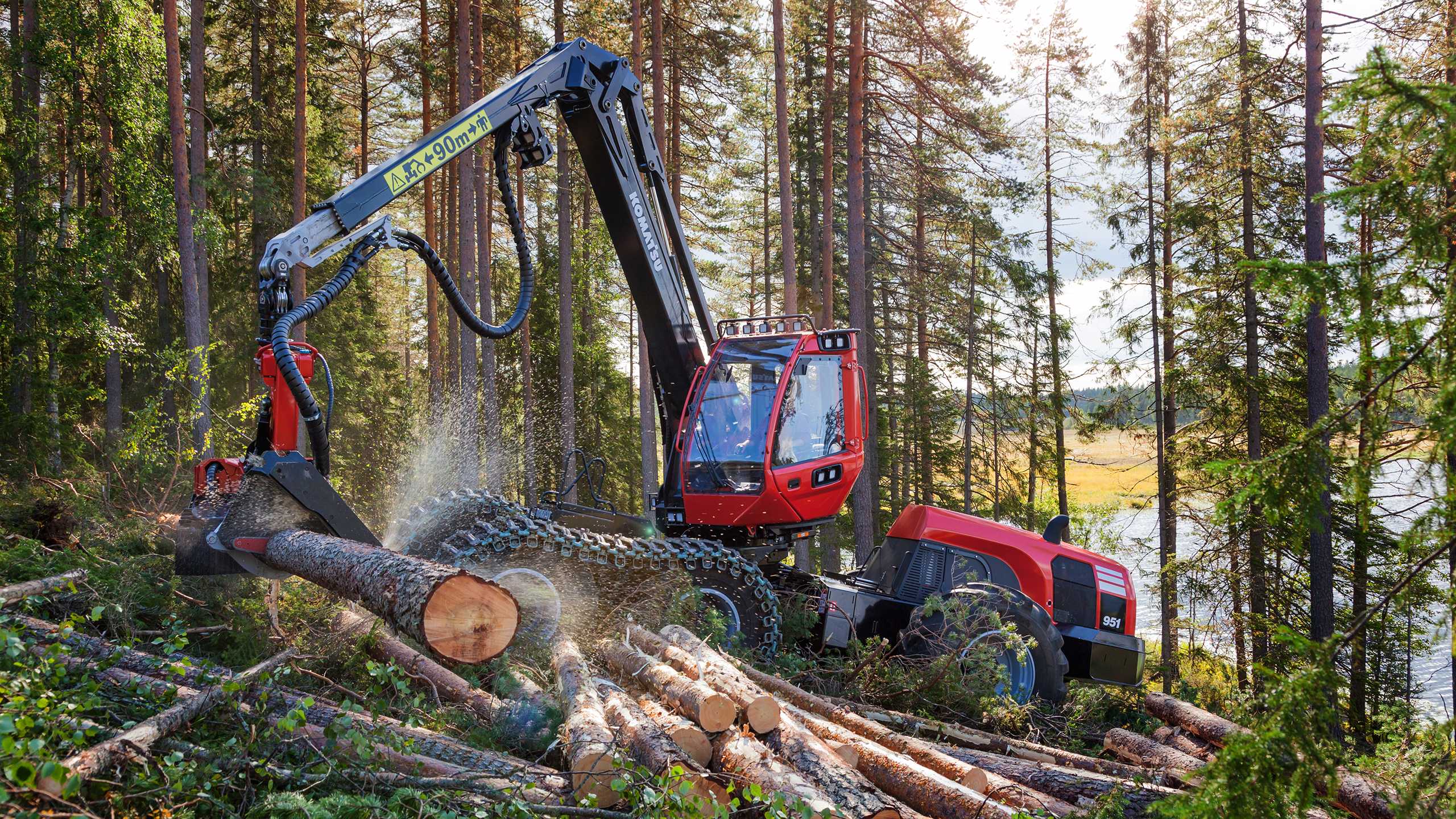}
     }
    \caption{Different machines are typically used in modern forestry operations. They all have operator cabins to increase the comfort level in a possibly wet and muddy forest environment. Images were taken from \cite{komatsuProducts}}
    \label{fig::big_machines}
\end{figure*}

Mechanization in forestry has brought greater productivity; up to the point where the human operator has become the bottleneck in the whole process \cite{parker2016robotics}. Since modern machines have many \acp{DoF}, training good operators takes a long time, which presents a serious impediment to productivity increase. To mitigate the problem, researchers have tried to develop \ac{IK} controlled, semi-autonomous modules for the crane operation in modern machines such as ones shown in Fig.~\ref{fig::big_forwarder} and Fig.~\ref{fig::big_harvester}. Examples of such work can be found in \cite{westerberg2014semi, ortiz2014increasing, hyyti2018forestry, hellstrom2008autonomous}.

\begin{figure*}[tb]
\centering
\subfloat[RCM Harveri harvester \label{fig::harveri}]{
       \includegraphics[width=0.47\textwidth]{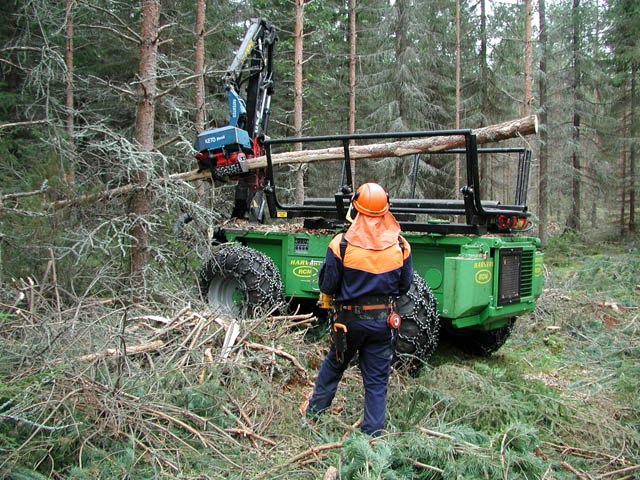}}
     \subfloat[eBeaver harvester \label{fig::eBeaver}]{
       \includegraphics[width=0.475\textwidth]{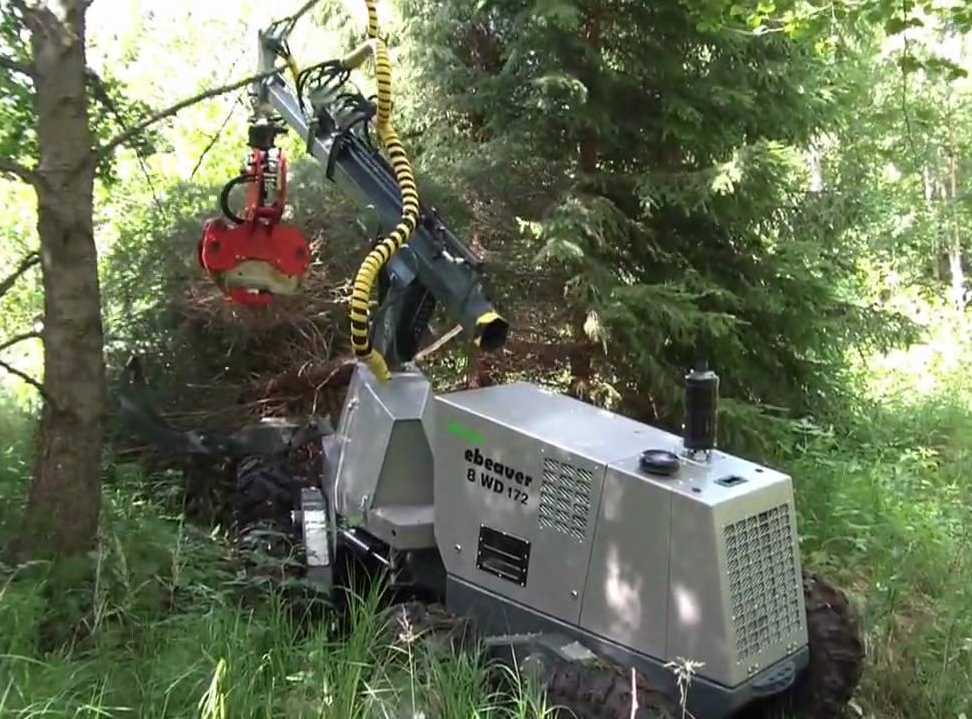}}
    \caption{Smaller harvesters that are typically used for thinning operations. They do not have a cabin for the operator and are remotely controlled within the operator's line of sight. Images taken from \cite{harveriImage} and \cite{eBeaverImage}.}
    \label{fig::small_machines}
\end{figure*}

Besides harvesting, a common task in forestry is thinning. Thinning is a process where we selectively remove some trees to make more space for others to grow \cite{forestryFocus}. Thinning requires negotiating tight spaces (as opposed to traditional clear-cutting) and can be done with smaller machines that are remotely controlled within the operator's line of sight. Two examples: Harveri and eBeaver, both shown in Fig.~\ref{fig::small_machines}. Despite the lower cost and less damage to the forest ground than large harvesters, machines such as Harveri are not very popular since operators are reluctant to give up a cabin's comfort and safety (personal communication with \cite{silvere}).
\subsection{Precision Forestry}
\label{sec::precision_forestry}
Traditional forestry operations are still primarily based on fundamentals developed about 300 years ago \cite{von1732sylvicultura}. Nowadays, the trend is to move towards precision forestry. Precision forestry uses automated data collection (not manual measuring) and analysis (software or machine intelligence aided) to allow site-specific and tree specific management of forests.  Instead of compartment-based management based on human experience and qualitative judgment, the forest's massive digital data enables much more granular and quantitative management. With the newest sensing technologies (e.g. \ac{LIDAR}), precision forestry can even be done on the single tree level \cite{mcKinsey2018revolution}, \cite{holopainen2014outlook}. Compared to the traditional techniques, precision forestry diminishes the error in inventory estimates by 400 percent \cite{mcKinsey2018revolution}. Furthermore, there are other benefits such as granular fertilizing, fire monitoring, pest and disease monitoring that can increase yields and labor productivity up to 10 times \cite{silvere}.
\subsection{Scope}
\label{sec::scope}
\begin{figure}[tbh]
\centering
    \includegraphics[width=0.65\textwidth]{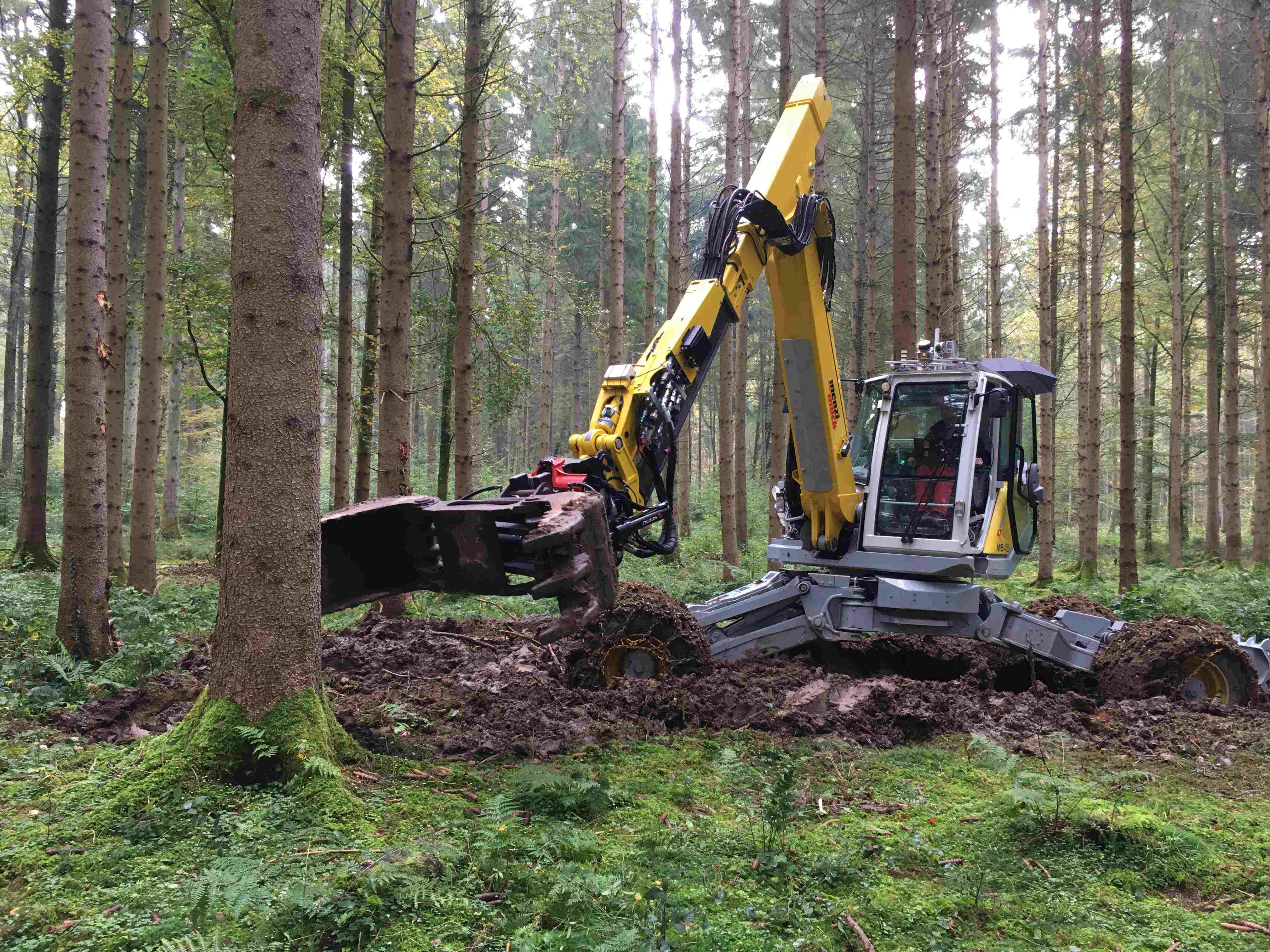}
    \caption{Legged excavator navigating to a tree and performing a grab. Human is in the cabin for safety reasons.}
    \label{fig::menzi_forest}
\end{figure}

This article presents an approach for carrying out autonomous forestry missions, particularly harvesting and thinning (referred to as harvesting in other text). We primarily focus on executing an autonomous mission and all the aspects that executing such a task entails, namely mapping, localization, planning, control, and tool positioning. To this end, we develop a versatile sensor module for data collection and localization of the harvesting machine. We present our hardware platform, an automated Menzi Muck M545 harvester \cite{jud2021heap} equipped with custom sensors and actuators (see Fig.~\ref{fig::menzi_forest}), and outline the technical approach used to carry out a harvesting mission planned by an expert (a human or a data-supported algorithm). Furthermore, we present experiments of a potential harvesting mission in the forest, and we state our findings and recommendations for tuning and system integration. 

High-level decisions on which trees to cut and how to manage the forest inventory are beyond this article's scope. Nowadays, companies can create forest inventories and recommend future actions with aid from machine intelligence (e.g., \cite{silvere}), and our pipeline assumes using such a service. We focus on automating the workflow until the moment of cutting the tree. Cutting and debranching are not investigated since there are existing harvesting tools for this process.

This is the first field report of deploying large-scale machines for fully autonomous forestry missions to the best of our knowledge. Our approach enables operations under the forest canopy, which are significantly harder than clear-cutting since the machine has to drive between the trees where \ac{GPS} signal may be unreliable. In summary, our work extends state of the art presented in Section \ref{sec::related_work}  with the following contributions:
\begin{itemize}
    \item Design of a portable sensor module with a sensor setup that enables mapping and localization. We test our sensor module in handheld operation, and we deploy it on a robotic platform.
    \item Development of a robust algorithm for converting raw point clouds into 2.5D elevation maps. The applicability of the algorithm is shown in a real forest and in several other environments. Our implementation is available as open-source\footnote{\url{https://github.com/ANYbotics/grid_map/tree/master/grid_map_pcl}}.
    \item We develop an approach pose planning algorithm suited for tight spaces and both structured and unstructured environments. The algorithm is evaluated on synthetic and maps produced from real data in a planning scenario including large-scale harvesters. 
    \item Development of a control approach for driving in rough terrain with a legged harvester. The algorithm performs path tracking and chassis stabilization at the same time.
    \item Both planning and path tracking algorithm implementations have been made publicly available for the community\footnote{\url{https://github.com/leggedrobotics/se2_navigation}}.
    \item We develop a lightweight method for tree detection based on \ac{LIDAR} scans of local forest patches. Our method is purely geometric based and operates directly on point clouds. It is suitable for online operation on the robot and we make it available for the community \footnote{\url{https://github.com/leggedrobotics/tree_detection}}
    \item We integrate all system components into a functioning autonomous system. Experimental verification of the proposed pipeline is done on a full-size legged harvester in a realistic environment
    \item An evaluation of system components in the real-world scenario is presented. The evaluation includes chassis control, tree detection, approach pose planning, mapping and localization, and point cloud to elevation map conversion.
\end{itemize}

The rest of the paper is organized as follows: Chapter \ref{sec::related_work} introduces the related work. Chapter \ref{sec::hardware} describes the robotic platform and hardware used in the experiments. In Chapter \ref{sec::system_overview}, we give a high-level overview of the proposed harvesting system with brief explanations for each of the submodules. Chapter \ref{sec::mapping_and_localization} describes mapping and localization, while Chapter \ref{sec::tree_detection} describes tree detection procedure. Control and planning are described in Chapters \ref{sec::control} and \ref{sec::planning}, respectively. Finally, Chapter \ref{sec::results} presents results and evaluation and Chapter \ref{sec::conclusion_outlook} discusses conclusions and future work.
\section{Related Work}
\label{sec::related_work}
In this section, we give a brief overview of the work relevant for precision harvesting missions; a more in-depth survey on forestry robotics can be found in  \cite{oliveira2021advances}. To the best of our four knowledge, no autonomous system for thinning or harvesting missions has been presented in the literature so far. The harvester presented in \cite{rossmann2009realization} is the closest to our work in the sense that developed methods are showcased on a full-size machine. The authors develop a particle filter-based localization for forest environments; however, no attempt to automate other machine parts was made. Similarly, in \cite{li2020localization}, a full-size harvester is fitted with a 3D \ac{LIDAR} the authors develop a place recognition algorithm based on tree stems.

To acquire an overview about the forest and plan for harvesting, maps are typically first acquired by airborne surveying \cite{naesset1997determination}, \cite{rossmann2009mapping}. For mapping and localization, \ac{LIDAR} sensors are a popular choice. \ac{GPS} based localization can be inaccurate under a tree canopy, and vision-based sensors (cameras) can be sensitive to illumination changes in outdoor environments. For \acp{LIDAR}, \ac{ICP} is a popular method for scan registration and has been used for mapping in forest surveying \cite{morita2018development, yue2018online, tsubouchi2014forest}. In \cite{babin2019large}, \ac{ICP} is fused with \ac{GPS} for mapping subarctic forests with sparse canopy. The \ac{ICP} is also used for mapping in \cite{tremblay2020automatic} and the authors propose a method for determining \ac{DBH}. Another possibility is to use filtering with trees as landmarks. \cite{rossmann2009mapping, rossmann2010navigation} build a map using an airborne laser sensor and localize a harvester in \emph{SE(2)} space using a particle filter. In \cite{miettinen2007simultaneous}, an \ac{EKF} based \ac{SLAM} is used on a skid steer robot to build a map and localize within it. The downside of relying on tree landmarks is that it can be susceptible to false positives. 

However, \ac{LIDAR} mapping does not have to rely on tree landmarks. Thus, mapping algorithms without tree stem detection can potentially handle the most general forest types (sparse, dense, thick vegetation). Many researchers have used LOAM (\cite{ zhang2014loam,zhang2015visual}) for \ac{LIDAR} mapping, and numerous variants of LOAM have been tested in urban environments. LOAM is a \ac{LIDAR} odometry and accumulates drift which may be problematic for large areas. Recently, loop closing has been added to LOAM (\cite{shan2018lego, shan2020lio}), which enabled the algorithm to correct for the accumulated drift. The loop closure mechanism is based on \ac{ICP} to map matching and cannot handle large drift. In \cite{chen2020sloam, nevalainen2020navigation} LOAM has also been used in a forest. An alternative to LOAM is Cartographer \cite{hess2016real} which has been shown to work well for large-scale mapping. We use Cartographer as a mapping algorithm of choice.

Once in possession of a map, one typically wants to estimate the biomass or classify the tree species, which requires segmentation of both the canopy and the stems. The forestry community has extensively studied techniques for tree segmentation and classification. Example of model-based approaches include \cite{zhang2019novel}, \cite{burt2019extracting}. Model-based approaches typically rely on pointcloud processing such as euclidean clustering, surface normal computation, and RANSAC model fitting to segment out the trees. On the other hand, learning-based approaches train networks end to end to segment trees from point clouds \cite{ayrey2017layer}, \cite{chen2021individual},\cite{bryson20177}. The need to accurately segment out whole trees in cluttered scenes renders most of the approaches complicated and with many steps. Hence, all algorithms are designed for offline operation, and some require powerful computational resources. In this work, a lightweight model-based approach operating online on the robot is used for tree detection.

A harvester can be treated as a big mobile manipulator. Planning and control for mobile manipulators is a well-studied problem in robotics. One can either treat the robot in a whole-body fashion \cite{kim2019whole, giftthaler2017efficient, gawel2019fully} or decouple the planning and control for the arm and the base \cite{carius2018deployment, schwarz2017nimbro}. The latter approach has often been used for forestry automation. Controlling the harvester's (or forwarder's) crane is an integral part of forestry operations. Examples can be found in \cite{lindroos2015estimating,westerberg2014semi,la2009modeling, kalmari2014nonlinear}. A major effort was put towards semi-automating the crane operation since coordinating many \acp{DoF} is one of the hardest tasks for a human operator. Most crane control algorithms are based on \ac{IK} or \ac{ID} \cite{siciliano2010robotics}. Compared to classical robotic manipulators, the biggest difference is that harvester cranes were seldom designed with autonomous operations in mind. They often come without sensing capabilities which means they have to be retrofitted with sensors to estimate the end-effector position. Moreover, the sensors have to be robust to withstand operation outdoors—such retrofitting results in increased automation effort. An additional difficulty is that hydraulic actuators are harder to control than their electric motors counterparts (nonlinearities, valve overlap, less bandwidth). To minimize the automation effort, \cite{morales2011open, ortiz2014increasing} investigate possibilities for purely open-loop control. More recently, an attempt has been made to develop arm motion planning for a feller-buncher machine \cite{song2020time}, \cite{song2021stability}. The authors rely on \ac{ZMP} to compute a stable trajectory guaranteeing the machine's (tipping over) stability. The stability of the harvester becomes essential as soon as one uses an actuated end-effector for tree manipulation (as opposed to passive tools that let trees fall freely).

Little has been done to develop a fully autonomous system for forest environment missions; the robotic community has mostly focused on urban environments. \cite{mikhaylov2018control} shows a very simplistic architecture without any validation. In \cite{georgsson2005development}, a \ac{GPS} based tracking approach is presented and validated on a forwarder machine outside of a forest environment. \cite {tominaga2018development} show experiments on an \ac{ATV} without an arm in a small scale environment using \ac{GPS} with \ac{RTK} correction. The authors use a global graph-based mission planner and a local state space sampling planner. A pure-pursuit algorithm is used for tracking. The proposed planning and control strategy cannot handle backward driving, limiting its ability to negotiate confined spaces. \cite{zhang2019rubber} presents an integrated system for navigation in a forest. The authors present a planner and a path follower that run on a small skid-steer robot without an arm. They show results in a structured forest (a rubber tree farm) where trees form a grid. This is reflected in the path generation algorithm, which involves heuristics to exploit the environment's structure and generates only straight-line paths. 

\cite{wooden2010autonomous} develop a navigation system for the Big Dog robot; it was tested in a forest, and it features a local planner based on a graph search A* algorithm. \cite{hellstrom2008autonomous} discusses different planning algorithms (A*, elastic bands, potential fields) for forwarder machines; however, no results are shown since it is a pre-study only. More examples of forest navigation using graph search algorithms to compute paths can be found in \cite{tanaka2017study,mowshowitz2018robot}. Compared to the \acp{UGV}, much more work has been done for \ac{UAV} path planning in forests, and examples can be found in, e.g., \cite{cui2014autonomous,liao20163d,pizetta2018control}. There is a research gap for \ac{UGV} path planning in natural environments, which this article tries to bridge. We compute plans for a non-holonomic constrained vehicle using RRT* \cite{karaman2011sampling}.
\section{Hardware}
\label{sec::hardware}
The Hardware section introduces the robot used and it describes the sensor module we developed for the autonomous missions.
\subsection{Platform}
\label{sec::platform}
Our robotic platform is \ac{HEAP}. It is based on a Menzi Muck M545 multi-purpose legged machine often used for harvesting in challenging terrain.  It is customized for teleoperation and fully autonomous operations. Besides forestry work \ac{HEAP}, can be used for digging, landscaping and manipulation tasks as well (see \cite{jud2019autonomous},\cite{johns2020autonomous}) . We use the terms \ac{HEAP} and harvester interchangeably in further text. Our machine is fitted with custom hydraulic actuators with pressure sensors and a high-performance servo valve. The actuators allow for precise force and position control, thus enabling active chassis balancing and adaptation to the uneven ground (see \cite{hutter2016force}).

\emph{Proprioceptive Sensing}: Two SBG Ellipse2-A \ac{IMU}'s (one in the cabin and one in the chassis) gather the inertial data that are primarily used to determine the chassis' roll and pitch angle. \ac{HEAP} is equipped with a series of \acp{IMU} rigidly attached to each arm link. Measurements from these \acp{IMU} are fused together to estimate end-effector pose and joint angles in the machine frame. Note that, unlike in \cite{jud2019autonomous}, we do not use externally mounted draw-wire encoders on the arm. This way, we avoid a potential entanglement of draw wires with tree branches.

\begin{figure*}[tb]
\centering
\subfloat[Front sensors on the cabin \label{fig::heap_front}]{
       \includegraphics[width=0.4\textwidth]{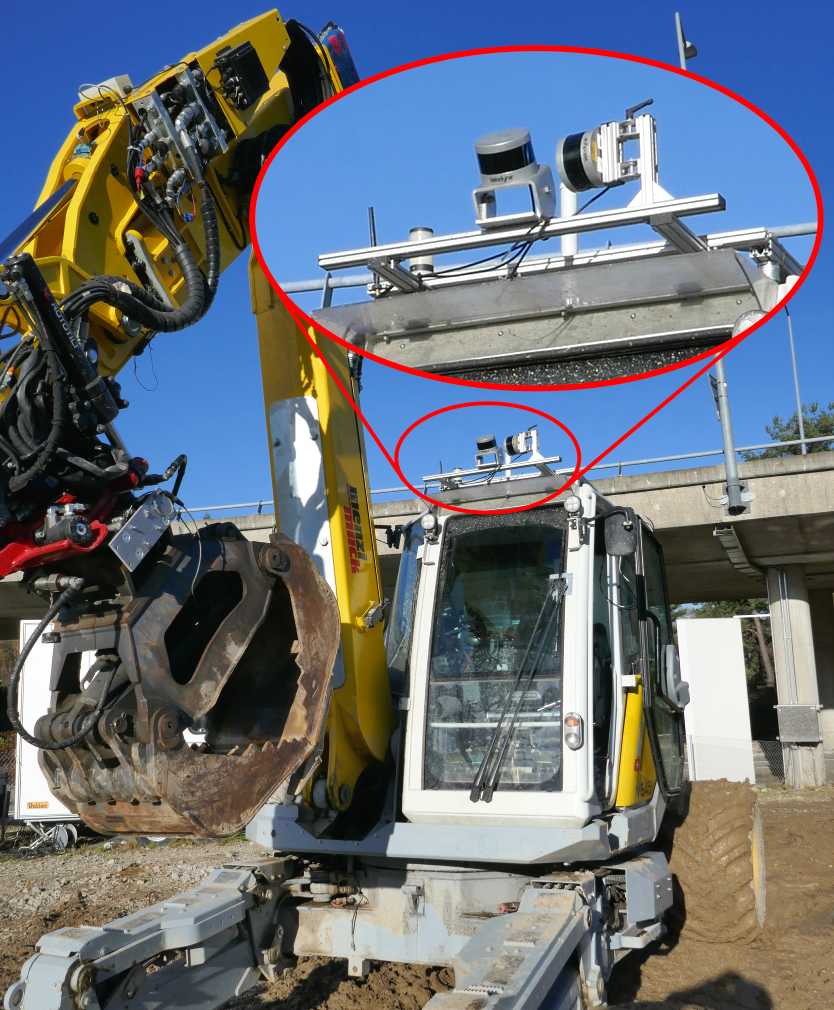}}
     \subfloat[Localization sensor module in the back  \label{fig::heap_back}]{
       \includegraphics[width=0.4\textwidth]{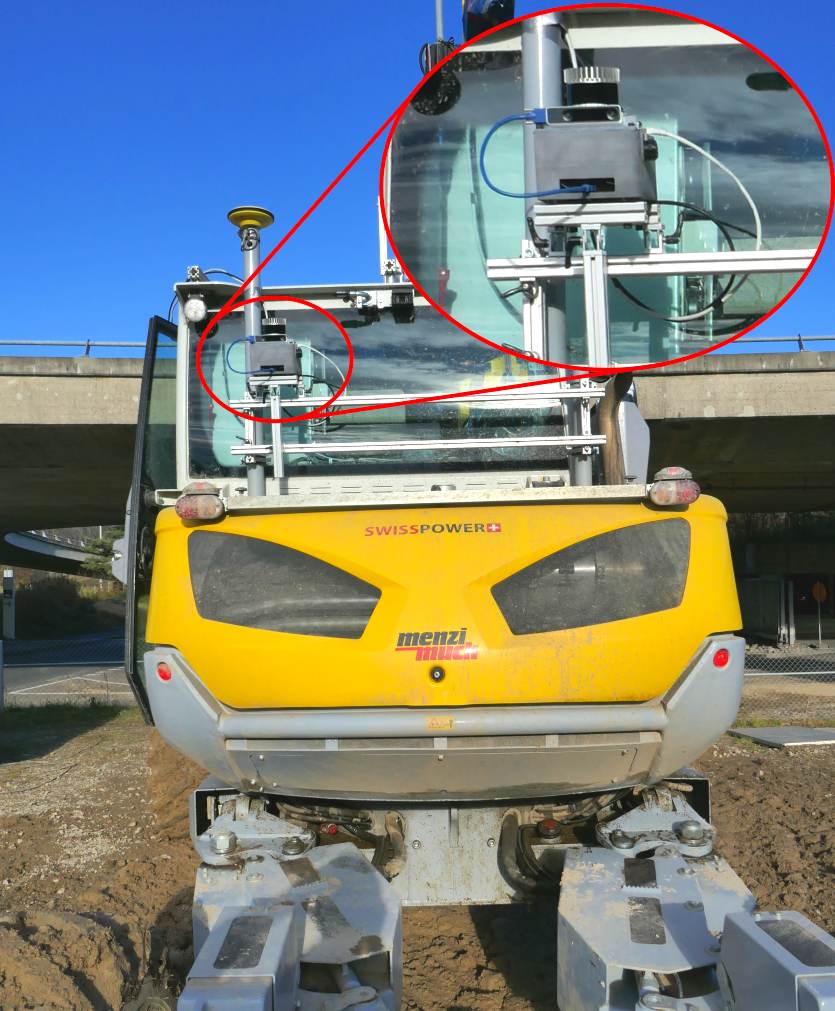}
     }
    \caption{\textbf{\textit{Left:}} two \ac{LIDAR} sensors mounted on top of the cabin and used for scanning the area in front of the machine. They are primarily used for tree detection. \textbf{\textit{Right:}} The localization module introduced in Sec.~\ref{sec::shpherds_crook} mounted in the back of the machine.}
    \label{fig::heap_front_back}
\end{figure*}

\begin{figure*}[tb]
\centering
\subfloat[Sensor module mounts \label{fig::crook_mounts}]{
       \includegraphics[width=0.45\textwidth]{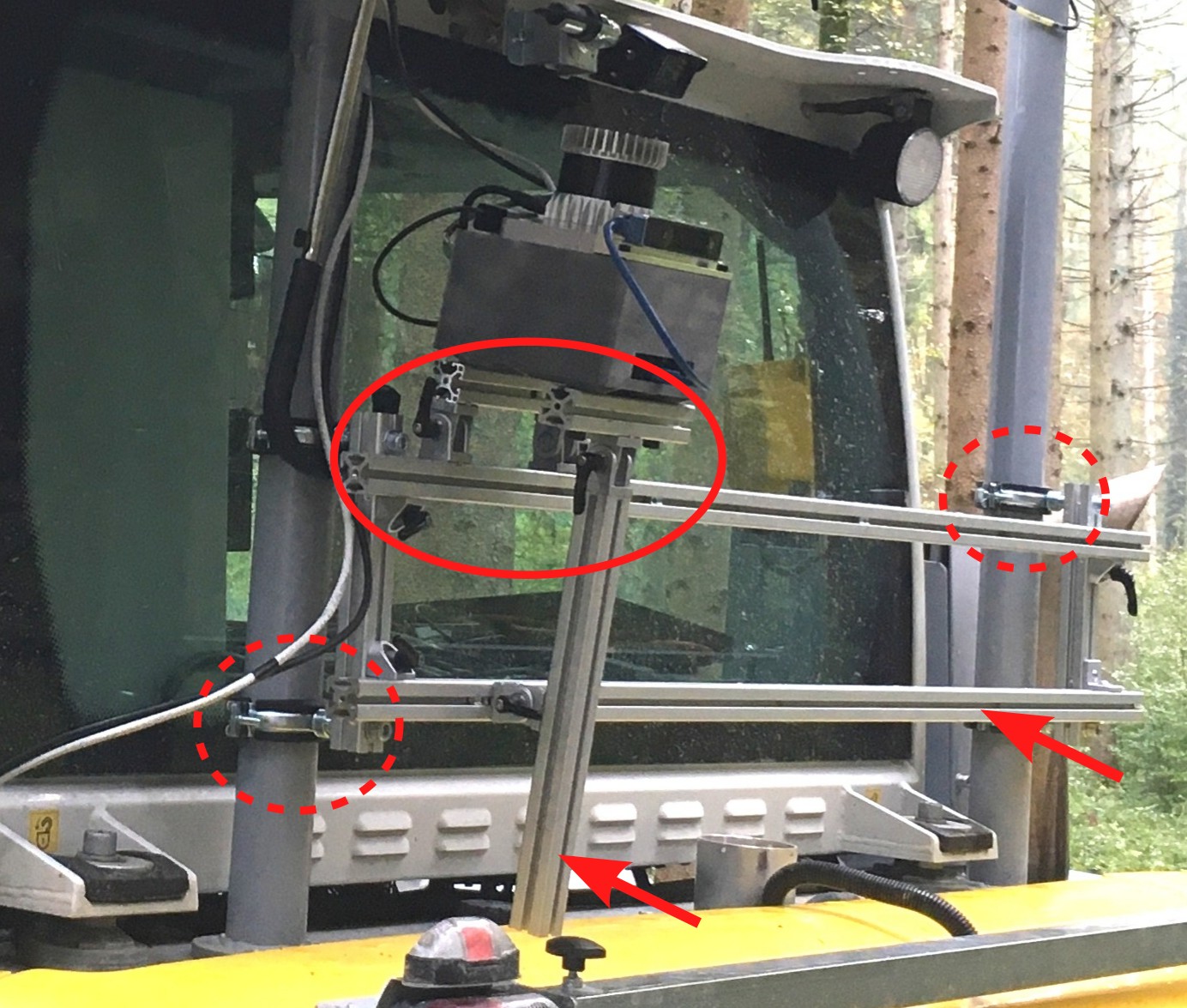}}
     \subfloat[Sensor module packed  \label{fig::crook_packed}]{
       \includegraphics[width=0.435\textwidth]{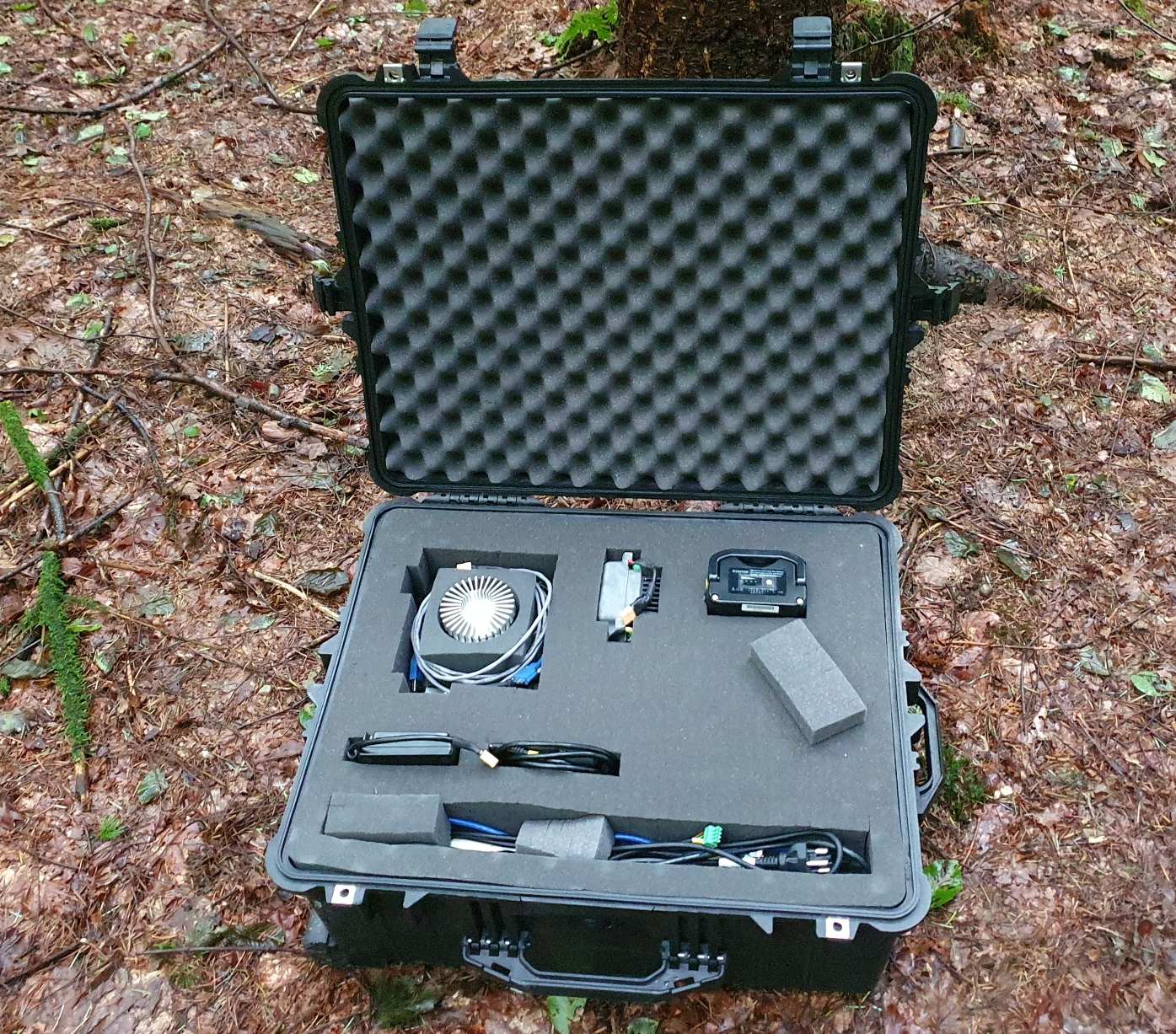}
     }
    \caption{\textbf{\textit{Left:}} Close-up photo of sensor module mounts. The module is mounted onto the aluminum profiles (shown with arrows), clamped to the machine using the o-ring clamps (encircled with dotted line). The lever fasteners are used to adjust the angle of the module (encircled with full line). \textbf{\textit{Right:}} The sensor module dismantled and packed after deployment.}
    \label{fig::shepherds_croook_mounts}
\end{figure*}

\emph{Exteroceptive Sensing:} Two Velodyne Puck VLP-16 \ac{LIDAR}'s are used for tree detection. It is important to note that one of the Pucks is rotated by \SI{90}{\degree} around the rolling axis (see Figure \ref{fig::heap_front}) in order to get a better resolution when mapping the area in front of the machine. A sensor module (see Sec.~\ref{sec::shpherds_crook}) is mounted on the back of the machine, as shown in Figure \ref{fig::heap_back}. A close-up image of the sensor mounts is shown in Fig.~\ref{fig::crook_mounts}. The sensor module is mounted using aluminum profiles from \emph{item}\cite{itemProfile}, and lever hinges from the same company. We attach it to the machine using o-ring clamps with rubber to mitigate the effect of vibrations. The mounts are rigid; hence the whole module does not move w.r.t to the cabin frame. Note that the module is not mounted in the middle to reduce further engine vibrations (the engine is just below the sensor module).

Trees can snap when manipulated, resulting in heavy debris falling on the machine. Hence, for a tree felling application (as opposed to tree grabbing), one needs to make the sensor module rugged and add mechanical protection (e.g., bulletproof glass). Another option would be to place the sensors directly inside the cabin. It is important to place the module such that it has as large \ac{FOV} as possible which is beneficial for the localization. Alternatively, one can use multiple synchronized sensors, each with a smaller \ac{FOV}. For \ac{HEAP}, the sensor module placement in the back as seen in Fig.~\ref{fig::heap_back} has an effective \ac{FOV} of about \SI{180}{\degree} (instead of \SI{360}{\degree}), which was enough for localizing.

The planning, control, and tree detection software stack runs on one computer (Intel i7-5820K, 6x3.60GHz, Ubuntu 18.04, 32 GB RAM) installed in the cabin. The control loops work at \SI{100}{\hertz} and are triggered by the \ac{CAN} driver. All the algorithms presented are implemented using C++ with \ac{ROS} as integration middleware. For more details on \ac{HEAP}, please refer to \cite{jud2021heap}.
\subsection{Sensor Module} 
\label{sec::shpherds_crook}
\begin{figure*}[tbh]
\centering
        \subfloat[Sensor module \label{fig::sensor_head}]{
       \includegraphics[width=0.48\textwidth]{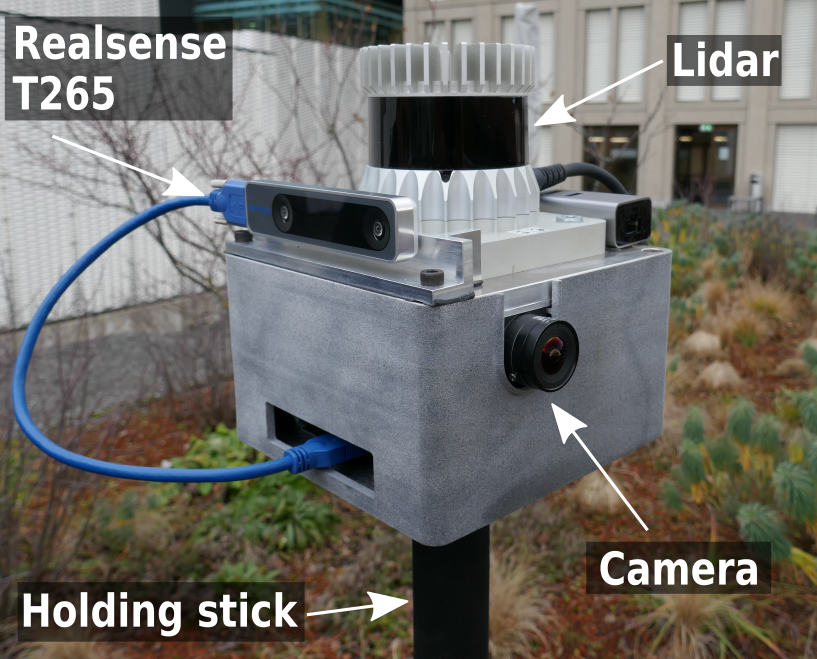}} \hfill
     \subfloat[Backpack with power  \label{fig::sensor_backpack}]{
       \includegraphics[width=0.48\textwidth]{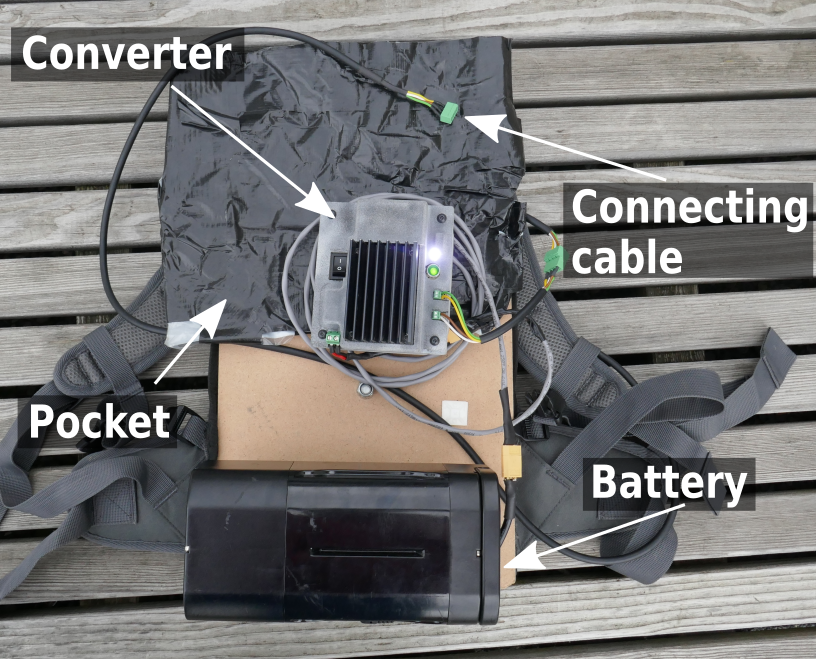}
     }
    \caption{\textbf{\textit{Left:}} Head of the sensor module. The \ac{CPU} and the \ac{IMU} are inside the grey box. \textbf{\textit{Right:}} Power supply for the sensor module mounted on the backpack.}
    \label{fig::shepherds_crook}
\end{figure*}

\begin{figure*}[tbh]
\centering
\subfloat[Sensor module in action \label{fig::crook_in_action}]{
       \includegraphics[width=0.35\textwidth]{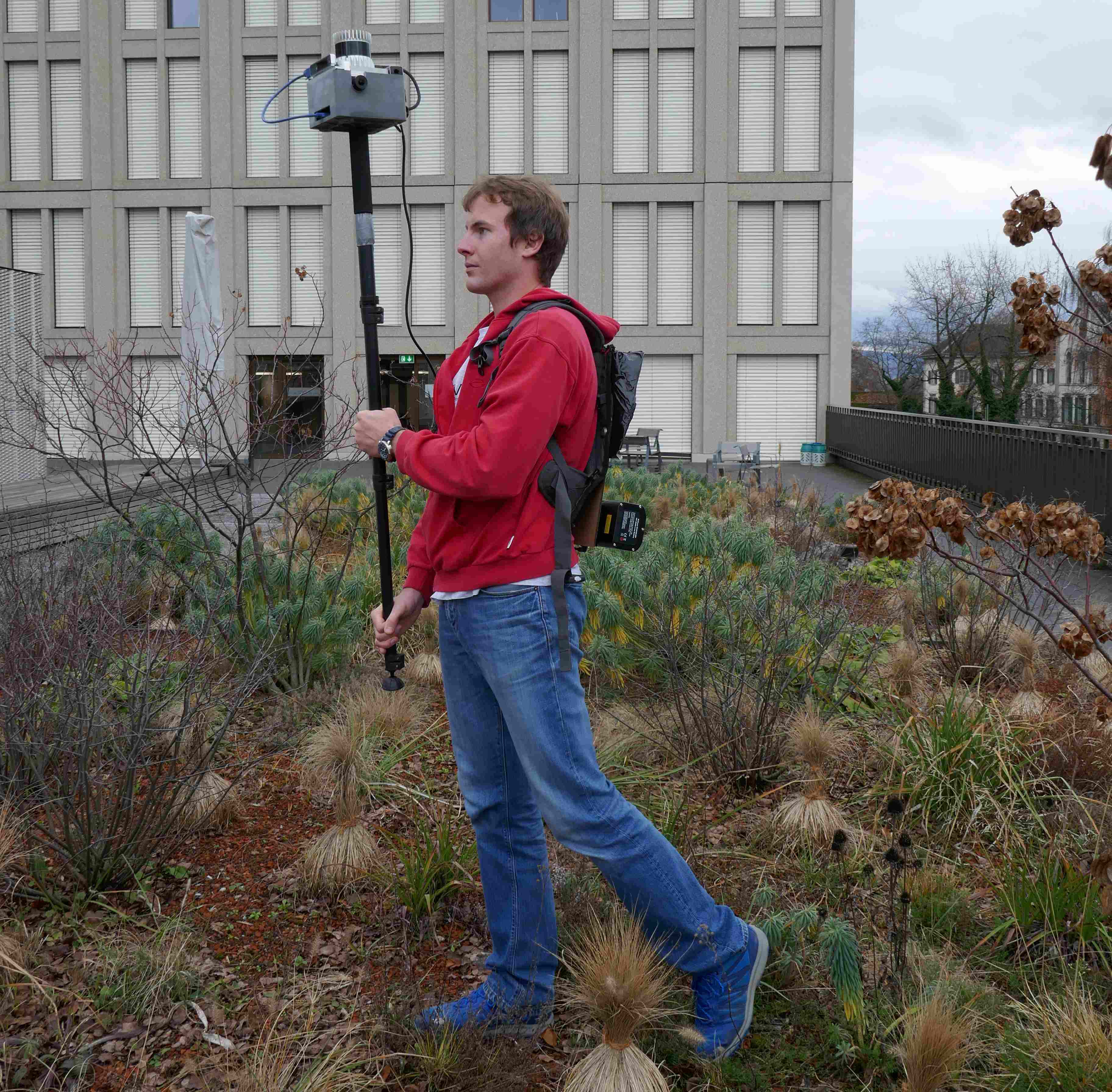}} 
     \subfloat[Sensor module schematic   \label{fig::crook_system}]{
       \includegraphics[width=0.63\textwidth]{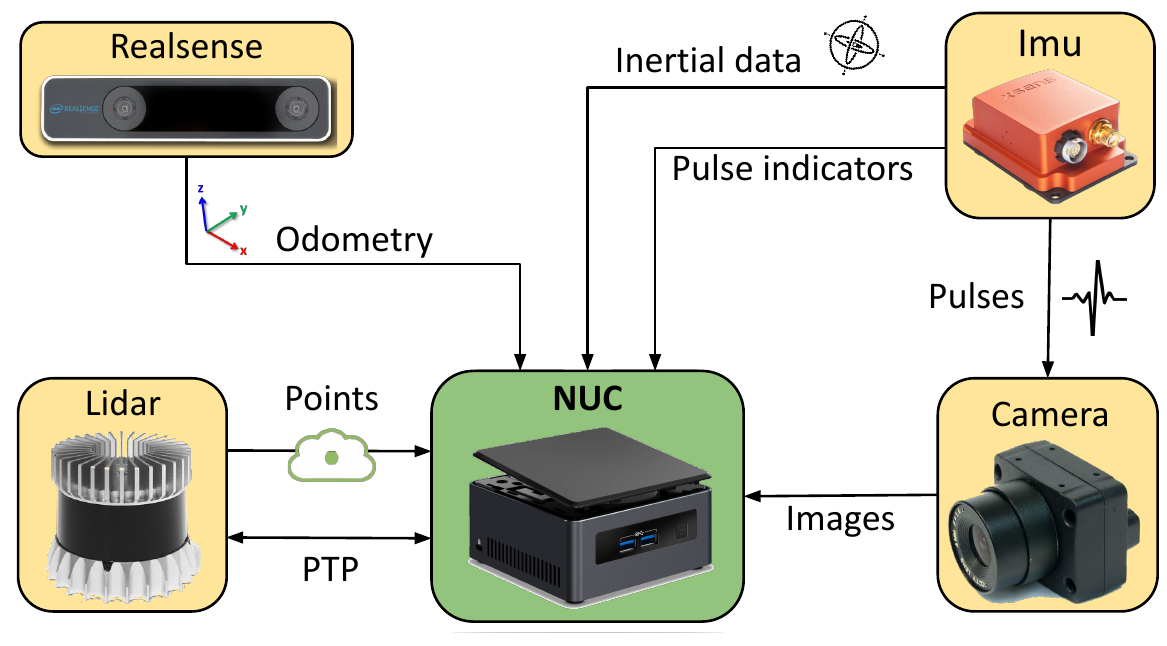}
     }
    \caption{\textbf{\textit{Left:}}Human operator mapping an area of interest. \textbf{\textit{Right:}} Schematic of the sensor module. It is composed of a camera, tracking camera (realsense), \ac{LIDAR} and an \ac{IMU}.}
    \label{fig::shepherds_crook_action_system}
\end{figure*}

We develop a sensor module for mapping and localization as displayed in Figure \ref{fig::shepherds_crook}. The module can be used in the first step by a human to map the area of interest (see Fig.~\ref{fig::crook_in_action}). In a second step, the system can be mounted on \ac{HEAP} (Fig.~\ref{fig::heap_back}) to localize the machine in the previously built map without GPS information. The whole module can be dismantled and packed, as shown in Fig.~\ref{fig::crook_packed}.

We design the sensor module to be lightweight such that handheld operation is possible. It has an integrated computer such that mission data can be collected and saved without an external laptop.  Besides, one can run the \ac{SLAM} directly on the module in real-time. The module has a \ac{LIDAR}, an \ac{IMU}, and two visual sensors. \ac{LIDAR} is used as a primary sensing modality with aid from visual and inertial sensors. This way, the motion distortion in the point cloud can be corrected. Extrinsic calibration between the sensors is obtained from \emph{Kalibr}\footnote{\url{https://github.com/ethz-asl/kalibr}} \cite{furgale2013unified} and \emph{lidar\_align}\footnote{\url{https://github.com/ethz-asl/lidar\_align}}, both of which are available open-source. Lastly, all sensors installed on the module are time-synchronized. Time synchronization is essential for state estimation and mapping algorithms to function correctly.

The sensor module in operation is shown in Fig.~\ref{fig::crook_in_action} and sensor components in Fig.~\ref{fig::sensor_head}. We use an Ouster OS-1 \ac{LIDAR} with 64 channels, an Intel Realsense T265 tracking camera, and a FLIR camera model BFS-U3-04S2M-CS. Inside the grey box in Fig.~\ref{fig::sensor_head}, one can find a computing unit (Intel NUC with i7-8650U processor) and an \ac{IMU} (Xsens MTI-100). Total cost of our sensor module is about 10500 USD. We implemented two complementary solutions for \ac{VIO}: FLIR camera synced with \ac{IMU} and the Intel Realsense T265 tracking camera. The FLIR camera has an excellent low light performance, while the T265 has almost \SI{180}{\degree} \ac{FOV} and higher frequency. Hence, one can choose the \ac{VIO} sensor used depending on the application. In this work we use T265 because of large \ac{FOV} and because higher frequency odometry estimates were beneficial for the \ac{LIDAR} based mapping.
%
%
The sensor module can be mounted on a stick shown in Figure \ref{fig::sensor_head}. In this way, a human can walk around with the module and point it to the places of interest to map it in sufficient detail. The sensor module is powered from the backpack (shown in Fig.~\ref{fig::sensor_backpack}), equipped with a \SI{36}{\volt} battery. We choose to use a \SI{300}{\watt\hour} e-bike lithium-ion battery (BiX Power BX3632H) with an integrated battery management system. All smaller sensors like cameras and the \ac{IMU} are powered through \ac{USB}.
%
%
%
%

The schematic of the sensor module is shown in Fig.~\ref{fig::crook_system}. The Intel Realsense camera provides software-synchronized \ac{VIO} at a frequency of \SI{200}{\hertz}. The FLIR camera (20Hz) and the \ac{IMU} (400 Hz) are synced using customized software. The \ac{LIDAR} is connected directly to the computer, and it is configured to use \ac{PTP} for time synchronization \cite{eidson2006measurement}. We modify the \ac{LIDAR}'s driver to send the packets as soon as they are received instead of batching them until the scan is completed. This way, the motion distortion is mitigated. We use the TICsync package \cite{harrison2011ticsync} to synchronize the \ac{IMU} with computer's clock. TICsync estimates the bias and the drift between the \ac{IMU}'s internal clock and the computer's clock to recover the exact time when the measurement happened. To increase the robustness, it is beneficial to assign the real-time priority to the \ac{IMU} driver (minimizes the timestamp jitter).
\section{Approach Overview}
\label{sec::system_overview}
\begin{figure}[tbh]
\centering
    \includegraphics[width=\textwidth]{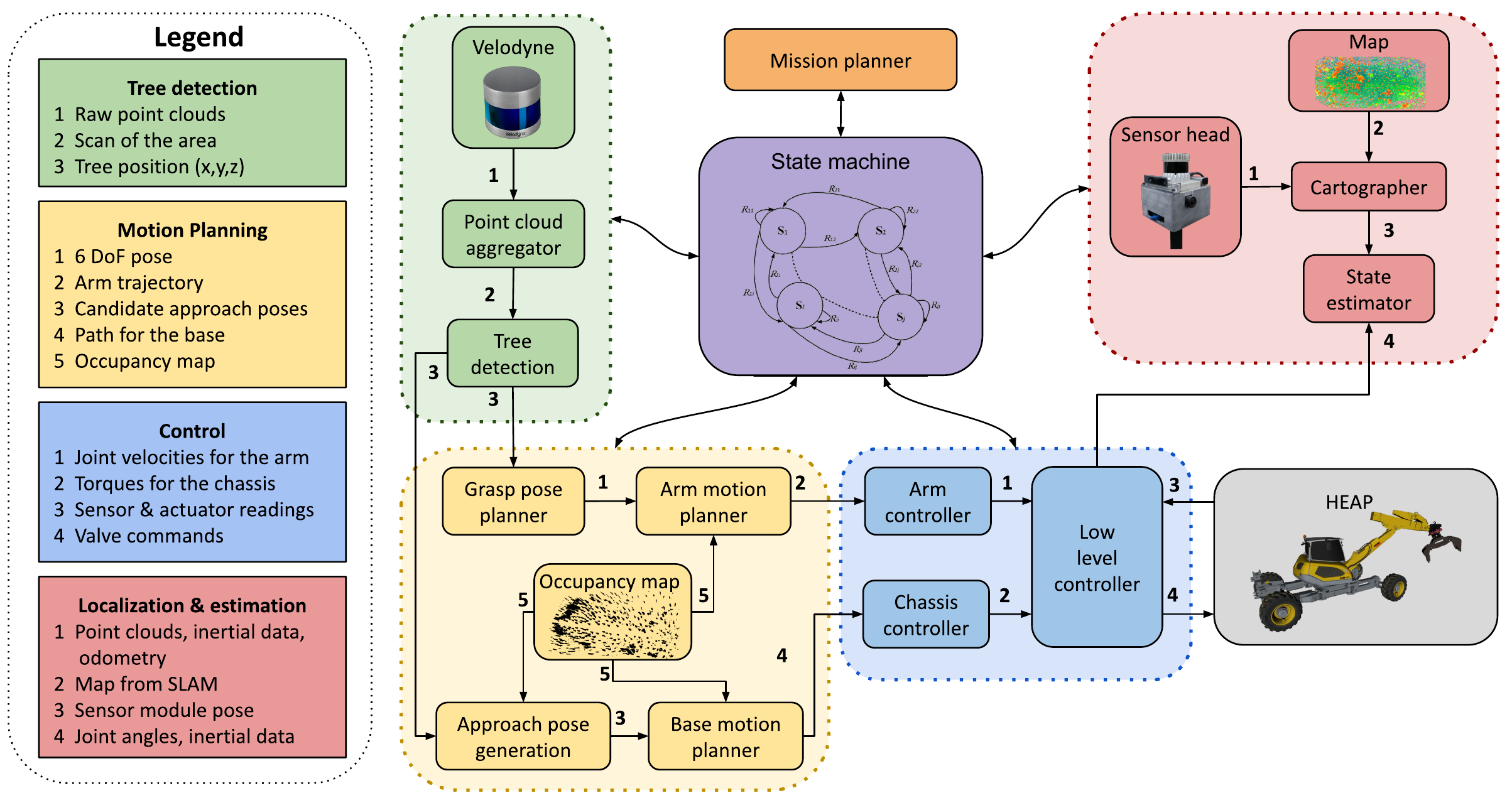}
    \caption{Overview of the system architecture deployed on \ac{HEAP}. Components belonging to the same subsystem are shown in the same color. Different subsystems are shown in different colors. Arrows depict communication channels in the system, with some communication lines omitted for the sake of brevity.}
    \label{fig::whole_system}
\end{figure}

Fig.~\ref{fig::whole_system} presents an overview of the autonomous harvesting system and serves as a visual outline for the chapters in this paper. We briefly describe each subsystem and give a more detailed description in their respective chapters.

\emph{Mapping and Localization:} We use a \ac{LIDAR} based \ac{SLAM} together with visual and inertial measurements to correct for scan distortion. The map is built by processing the data offline. At mission time, scans are registered against the existing map to compute the sensor module's pose.  
 
\emph{Tree Detection:} The \ac{LIDAR} sensors mounted on the top of the cabin (see Figure \ref{fig::heap_front}) deliver point clouds at \SI{20}{\hertz}. An intermediate node aggregates them and passes them to the tree detection module. Detection is done purely based on geometric features. The tree detection module forwards the detected tree's coordinates to the grasp pose planner and the approach pose planner.

\emph{Control:} The control subsystem is split into two parts. The arm controller ensures trajectory tracking by sending velocity commands to the arm joints (\ac{IK} controller). The base controller ensures path tracking, and it regulates contact forces such that the base stays upright when driving over uneven terrain. The low-level controller transcribes higher level references (velocity, torque, positions) to the valve commands.

\emph{Motion Planning (Arm and Base):} The planning stack comprises three planners. The base motion planner plans a path and an approach pose for the base of the harvester. The grasp pose planner computes a gripper pose in 3D space that encompasses the tree trunk. Finally, the arm motion planner produces a collision-free spline trajectory to move the gripper into the desired pose. 

\emph{Mission Planner:} This module determines which tree to grab next, and it sends the target position to the state machine, which then sends it to the approach pose planner.

\emph{State Estimation:} The state estimator uses the sensor module pose and proprioceptive measurements (\ac{IMU}, joint angles) to compute the complete state of the system (6 \ac{DoF} base pose + joint angles).

\emph{State Machine:} The state machine coordinates the execution of different tasks. It works on a handshaking principle where the state machine sends a request to a subsystem, and the subsystem responds with an acknowledgment once the task requested has been completed. Tasks that the state machine can request and the order of operations to grab a single tree are shown in Tab.~\ref{tab::tree_grab_seq}. 
\begin{table}[tb]
\centering
\renewcommand{\arraystretch}{1.1}
\caption{Order of operations and subsystems involved to grab a single tree}
\begin{tabular}{ |p{1.7cm}||p{9.5cm}|p{3.8cm}|  }
 \hline
 \centering Task order & \centering Task Description & Subsystems responsible\\
 \hline
 \centering 1 & Get the next tree position.	& Mission planner\\
 \centering 2 & Plan an approach pose and a path. If it fails, go to step 1. & Planning\\
 \centering 3 & Drive the machine to the approach pose found in step 2. In case tracking fails, go to step 2	& Control\\
 \centering 4 & Retract the arm (in case it is not already retracted)	& Planning, Control\\
 \centering 5 & Scan the area around the expected tree location.	& Control\\
 \centering 6 & Run tree detection algorithm and plan a grasp pose	& Tree detection, Planning\\
 \centering 7 & Plan and track an arm trajectory to reach the grasp pose	& Planning, Control\\
 \centering 8 & Plan and track arm trajectory back to default position (arm retracted all the way).	& Planning, Control\\
 \centering 9 & Go to step 1.	& State Machine\\
  \hline
 \end{tabular}
 \label{tab::tree_grab_seq}
 \end{table}
\FloatBarrier

\section{Mapping and Localization}
\label{sec::mapping_and_localization}
Precision forestry requires precise and consistent 3D geometric information about the forest. This data can be collected by a harvester itself, a smaller \ac{UGV} or by a human carrying the sensors. The collected information can then be used for mission planning and forest inventory management. Moreover, the map serves as a reference for the harvester to localize. To retain flexibility, we would like to make no assumptions on the surroundings and allow the harvester to work in different types of forests (e.g., sparse, with vegetation, non-flat). Therefore we avoid using tree stems as landmarks (e.g., \cite{rossmann2009mapping}) since they can be hard to detect (vegetation) or they may be scarce (e.g., a glade inside the forest).

We selected Google Cartographer \cite{hess2016real} as the backbone of our mapping and localization pipeline. Cartographer is a grid-based \ac{SLAM} approach that uses the scan to sub-map matching for loop closure detection and discards unlikely matches using the branch and bound method. At mission time, localization can be achieved simply by scan to sub-map matching. Compared to other \ac{SLAM} systems available at the time, Cartographer has the advantage that it is fully open source, can cover large spaces, features loop closures that work robustly, and has a large community of users. Below we provide a brief description of the working principle.

Cartographer divides the world into a raster of submaps where a number of accumulated scans determines the size of the submap. The submaps are grids (3D) where each cell is assigned a probability of being occupied. Before inserting a scan into a submap, the scan is voxelized, and each grid cell inside the scan is classified as \emph{occupied} of \emph{free}. Subsequently, the range scans is registered within the submap by solving a nonlinear least-squares problem which maximizes the probabilities at the scan points in the submap. This local, grid-based \ac{SLAM} relies on a good initial guess which is achieved by extrapolating the previous pose using the inertial/odometry data.

Cartographer follows the Sparse Pose Adjustment approach of optimizing all scans, and submaps \cite{konolige2010efficient}. Each range scan is associated with a trajectory node in the pose graph. In the background, all scans are matched to nearby submaps to create loop closure constraints. Suppose a sufficiently good match (user-defined minimum score) is found in a search window around the currently estimated pose. In that case, it is added as a loop closing constraint to the global optimization problem. The constraint graph of submap and scan poses is periodically optimized in the background (every few seconds). When localizing in a known map, Cartographer keeps only the latest \emph{N} submaps, which are then considered for loop closures against the submaps in the known map. The known map is not updated.

\begin{figure}[tbh]
\centering
    \subfloat[Testing forest patch \label{fig::map_pbstream}]{
       \includegraphics[width=0.98\textwidth]{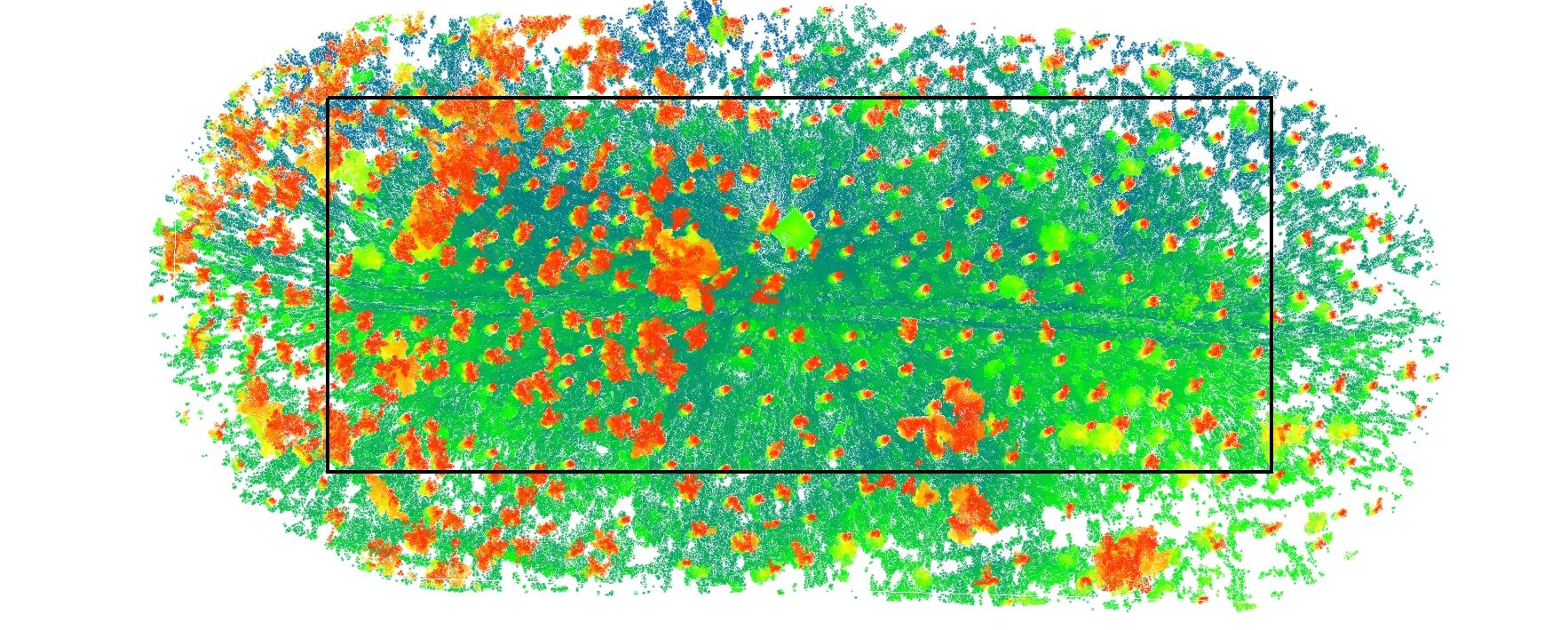}}\\
     \subfloat[Large forest area  \label{fig::large_map_pbstream}]{
       \includegraphics[width=0.99\textwidth]{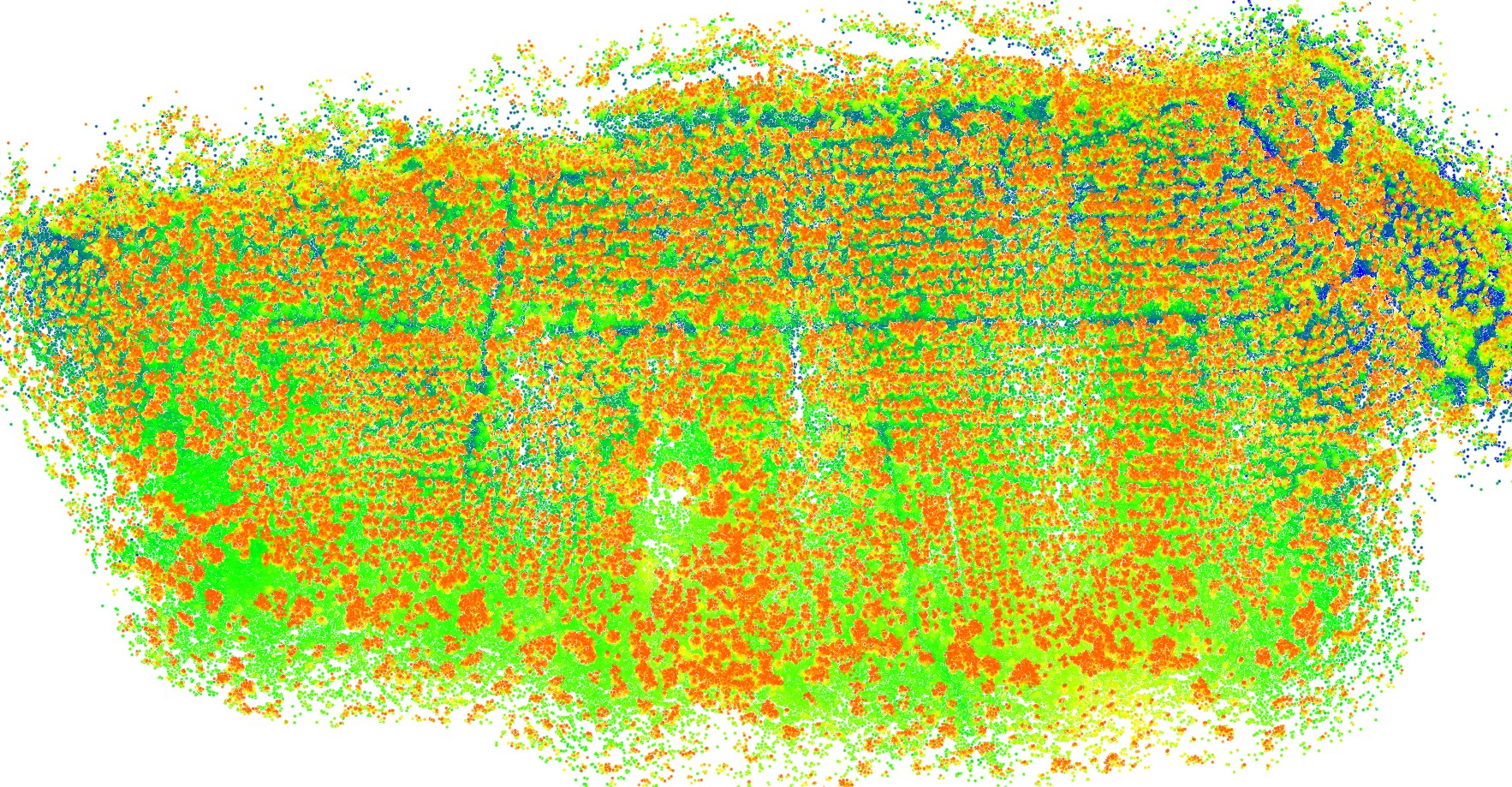}
     }
    \caption{Bird view of forest maps. Both maps are represented as point clouds. \textit{\textbf{Top:}} Small forest patch where we conducted the experiments. Approximate size \SI{140}{\meter} x \SI{60}{\meter} The area encircled with a black rectangle was converted to an elevation map (Fig.~\ref{fig::forest_elevation_map}). \textit{\textbf{Bottom:}} Map of a larger forest area (top view), where range, odometry and inertial data has been collected in multiple tours, concatenated and the processed with Cartographer, data courtesy of \cite{silvere}. Approximate map dimensions: \SI{340}{\meter} x \SI{170}{\meter}.}
    \label{fig::maps_pbstream}
\end{figure}

\subsection{Mapping}
\label{sec::mapping}
Examples of maps generated with Cartographer are shown in Fig.~\ref{fig::maps_pbstream}.

The map in Fig.~\ref{fig::map_pbstream} shows a point cloud of a forest viewed from above. Blue and green colors correspond to a lower elevation (ground), whereas yellow and red colors correspond to a higher elevation (vertical structures like tree trunks and canopy). The map is about \SI{140}{\meter} long and \SI{60}{\meter} wide, and it shows a part of the forest where we conducted the experiments with \ac{HEAP}. Another, larger map (\SI{340}{\meter} x \SI{170}{\meter}) is shown in Fig.~\ref{fig::large_map_pbstream}; the size of this map shows that Cartographer can map areas sufficiently large for an autonomous harvesting mission. Both maps have been processed offline and bundle adjusted. Compared to online processing, building the maps offline allows us to use more points from the \ac{LIDAR} which results in denser maps, use finer resolution voxels, and sample for constraints between submaps more often. Furthermore, we can increase the search radius for loop closures and run the scan matcher optimization more often. All the changes above result in less drift and more loop closures, which produce more consistent maps.

Once we have a consistent 3D map of the space, we use it for localization at mission time. Since the planning is done in \emph{SE(2)}, we convert the 3D map into a 2.5D map which the planner then uses. The 3D map is first converted into an elevation map, and from the elevation map, we compute a traversability map. Both maps are functions $f:(x,y) \rightarrow \mathbb{R}$ mapping the 2D coordinates to height or traversability.  Below, we detail the conversion of the point cloud into an elevation map used by the planner.

\begin{figure}[htb]
\centering
    \centering
    \includegraphics[width=0.98\textwidth]{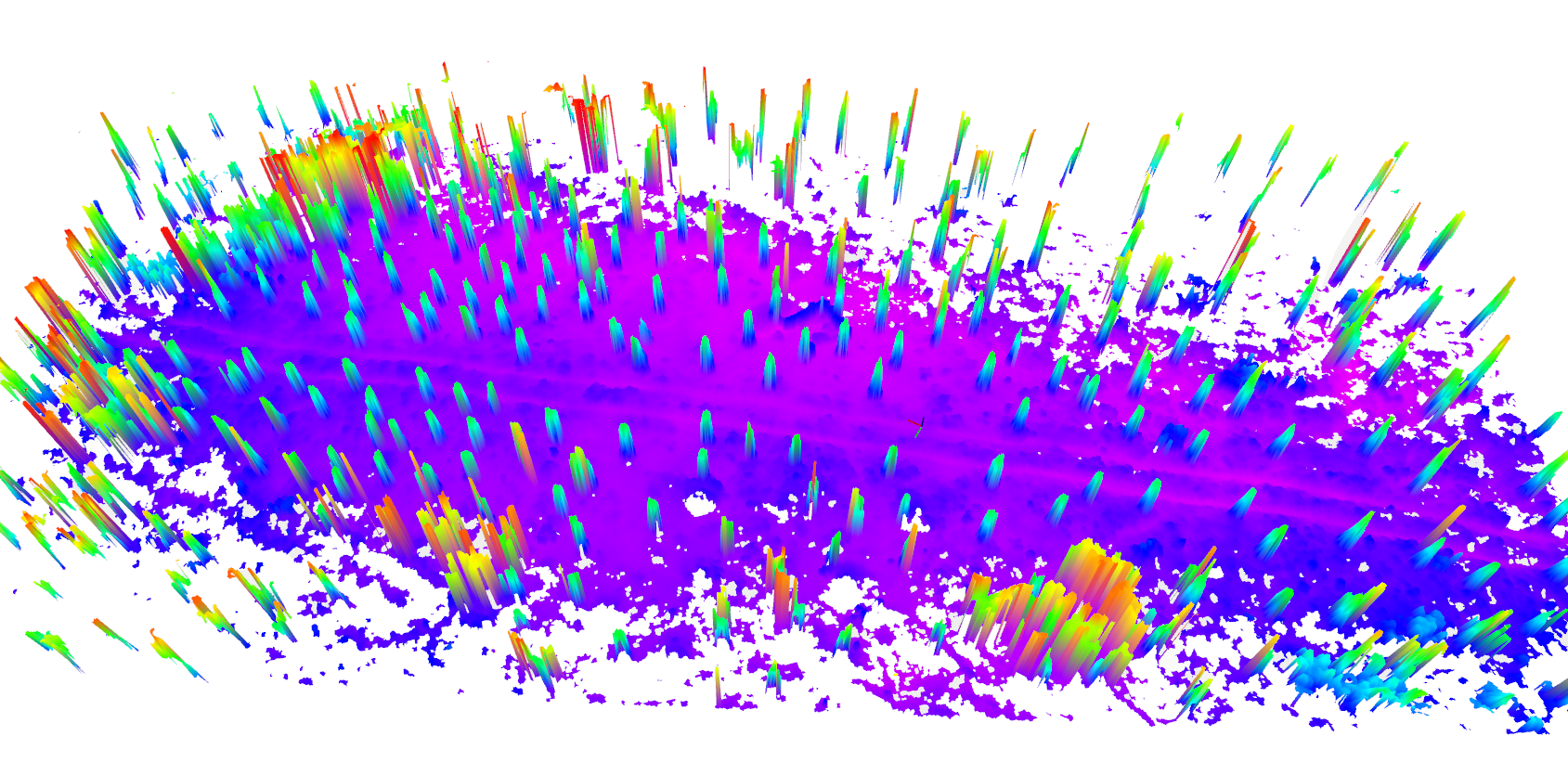}
    \caption{Elevation map of a forest patch encircled black in Fig.~\ref{fig::map_pbstream}. Purple color corresponds to the lowest elevation and red to the highest. The resulting grid map size is 1419 by 1318 cells with a resolution of \SI{10}{\centi\meter}. Note how tree trunks are clearly visible in the elevation map.}
    \label{fig::forest_elevation_map}
\end{figure}

A grid map data structure \cite{fankhauser2016universal} is used to store the elevation map. From the raw point cloud in Fig.~\ref{fig::map_pbstream}, our algorithm builds a 2.5D elevation map of the area shown in Fig.~\ref{fig::forest_elevation_map}. Conceptually, the algorithm is similar to the one used in \cite{Fankhauser2018ProbabilisticTerrainMapping}. In contrast to \cite{Fankhauser2018ProbabilisticTerrainMapping}, our algorithm can handle multiple elevations in the cell, i.e., multiple points with the same $(x,y)$ coordinates and different $z$ coordinates. Thus, it can filter out vegetation and clutter and recover true ground elevation more robustly. We successfully used the point cloud to elevation map conversion algorithm in both planar and non-planar environments. Implementation is available as an open-source package\footnote{\url{https://github.com/ANYbotics/grid_map/tree/master/grid_map_pcl}}.
\FloatBarrier

\begin{figure}[tbh]
\centering
    \includegraphics[width=0.5\textwidth]{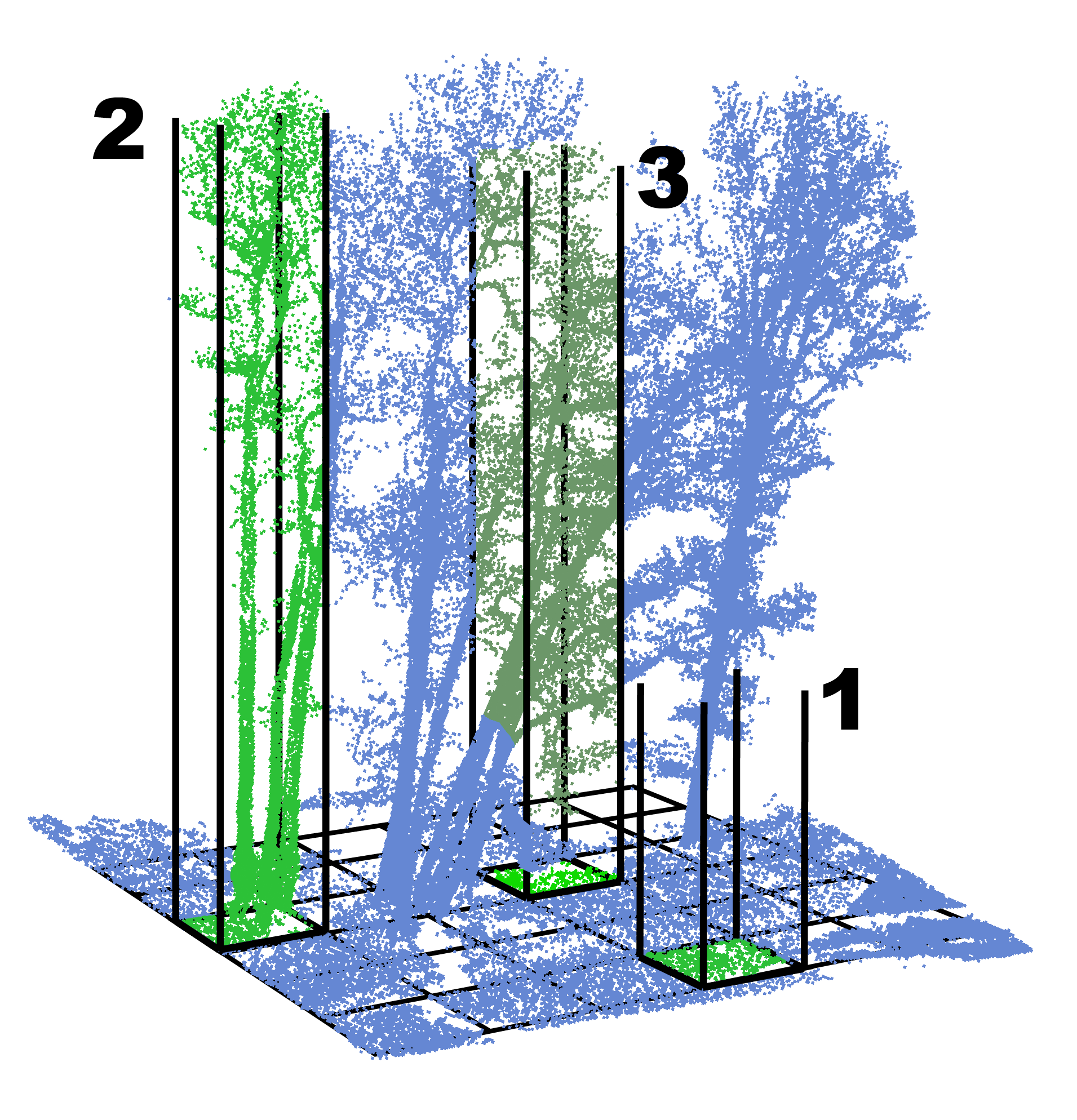}
    \caption{Generating a grid map from a raw point cloud. The grid is shown with black lines, and the input point cloud is shown in blue color. Clusters of points that the conversion algorithm might extract are shown in green. Note that image is not drawn to scale.}
    \label{fig::point_cloud_column}
\end{figure}

The point cloud to elevation map conversion algorithm starts by removing the outliers and downsampling the input point cloud (implementation based on \cite{rusu20113d}). The main loop is parallelized to speed up computation. For each cell in the grid map, we pre-compute all the points contained within that cell (this enables lookup in $O(1)$ time), thus vastly speeding up the algorithm. Each cell is a rectangle in $x,y$ coordinates. Once we have fetched the points inside the grid map cell, Euclidean clusters are computed. The cluster size is regulated with a cluster tolerance parameter (see \cite{rusu2010semantic}). Any points withn the tolerance distance are considered the same cluster. After clustering, we calculate the centroid of each cluster. Finally, the centroid with the smallest $z$ coordinate is deemed to be the ground elevation. Illustration of the process is shown in Fig.~\ref{fig::point_cloud_column}.
In Fig.~\ref{fig::point_cloud_column}, grid map cells are extended in the $z$ direction and form columns 1,2, and 3. Column 1 has one cluster which contains only the ground points (shown in light green), and the algorithm correctly extracts the ground height. In column 3, the algorithm extracts at least two clusters. The lower cluster is the ground (light green color), while branches and leaves form the upper cluster (dark green). Clusters are disjoint, and the algorithm correctly recovers the ground elevation as a mean value of the lower cluster. Column 2 contains one large cluster composed of both ground and the tree. In this case, we cannot recover ground information since the resolution of the map is too coarse; the height of the tree trunk determines the height. In this work, we use the resolution of \SI{10}{\centi \meter} for the grid. We analyze algorithm's behaviour in presence of vegetation and clutter in the Appendix~\ref{sec::apendix_pointcloud_to_map}.

Qualitative runtimes of our algorithm are shown in Table~\ref{table::conversion_times}. The foreseen use case is to run the algorithm once offline to construct a global elevation map that can be used for planning. The times shown in Table \ref{table::conversion_times} were obtained on an Intel Xeon E3-1535M (\SI{2.9}{\giga\hertz}, 4 cores). The algorithm is not limited to application in forests only but also generalizes to non-forest environments (see \url{https://github.com/ANYbotics/grid_map/tree/master/grid_map_pcl}). 
\begin{table}[tbh]
\caption{Qualitative runtimes of the point cloud to elevation map conversion. All the conversions were done on the same map with the size of 1419 by 1318 cells at \SI{10}{\centi \meter} resolution.}
\label{table::conversion_times}
\begin{tabular}{@{}l|llll@{}}
\toprule
Point cloud size       & less than 10 million points & 40-60 million points & 100 - 140 million points &  \\ \midrule
Algorithm runtime & 1-2 minutes                                                & 5-15 minutes                                  & 30-60 min                                               &  \\ \bottomrule
\end{tabular}
\end{table}
\subsection{Localization}
\label{sec::localization}
Apart from mapping, the sensor module from Sec.~\ref{sec::shpherds_crook} is used to localize \ac{HEAP} within the map. The localization pipeline consists of components encircled in red color in Figure \ref{fig::whole_system}. The sensor module is mounted on the excavator (as shown in Fig.~\ref{fig::heap_back}). The sensor module's clock is synchronized to the harvester's computer clock using \ac{NTP}. \ac{PTP} and \ac{NTP} run on different networks. We forward all the measurements to the main computer running Google Cartographer in the localization mode (see \cite{catographer2017online}). 

Cartographer gives a pose estimate of the sensor module in the map frame used to compute the excavator's entire state. Extrinsic calibration of the complete sensor module mounted on \ac{HEAP} is obtained from \ac{CAD} model and manual measurement. Accurately calibrating sensors mounted on heavy machinery remains a challenging problem, and to increase overall localization accuracy one should use more advanced methods (e.g. \cite{APIcalibration}). Note that sensors on the sensor module itself are calibrated w.r.t. each other using techniques mentioned in Section \ref{sec::shpherds_crook}. The setup with the sensor module mounted in the back resulted in end-effector position accuracy (coupled errors from sensor module localization and robot kinematics) of about \SI{30}{\centi \meter}. For evaluation, we have asked the harvester to grab the same tree blindly multiple times. Such a level of accuracy may not be enough for high precision harvesting, and we mitigate the problem by detecting the grabbing target in a locally built map (see Sec.~\ref{sec::tree_detection}).
\FloatBarrier
\clearpage
\section{Tree Detection}
\label{sec::tree_detection}
\begin{figure}[tbh]
\centering
    \includegraphics[width=\textwidth]{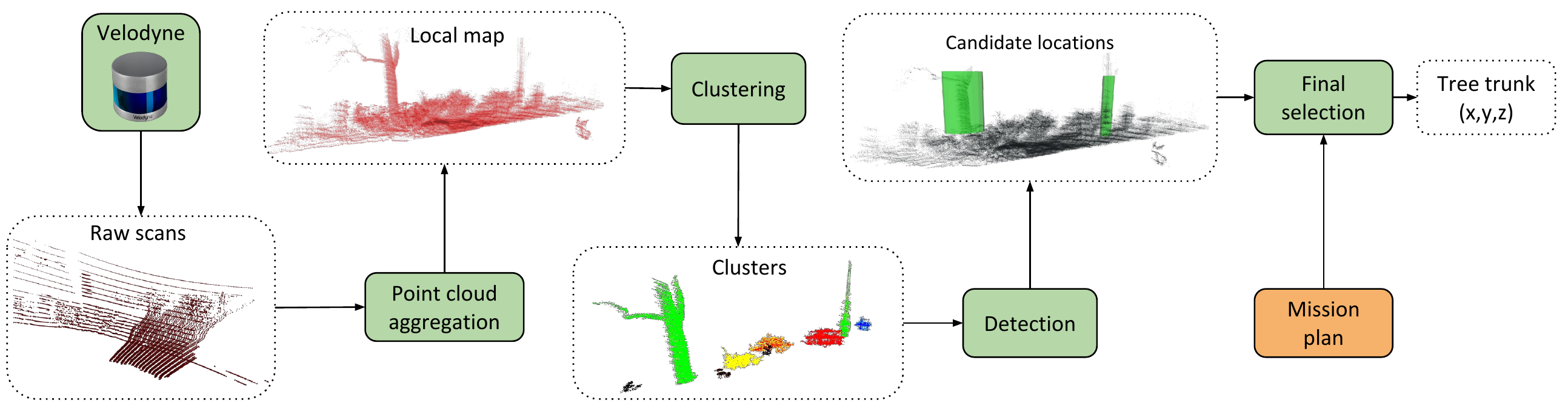}
    \caption{Flowchart of the tree detection subsystem procedure.}
    \label{fig::tree_detection_system}
\end{figure}

The tree detection subsystem's responsibility is to detect a tree trunk and compute its position. n contrast to the methods mentioned in Chapter~\ref{sec::related_work}, our method is lightweight, suitable for online operation, and can be implemented in a dozen lines of code (available as open source\footnote{\url{https://github.com/leggedrobotics/tree_detection}}). While learning-based methods can still be fast to evaluate, we opted for a model-based method to avoid data labeling and for ease of implementation. Note that we are not interested in complete tree segmentation like most of the related work, but we are only interested in detecting a good grabbing spot. This circumstance allows simplifying the detection algorithm. A schematic of the subsystem with some intermediate processing steps is shown in Fig.~\ref{fig::tree_detection_system}. Once \ac{HEAP} is positioned close to a tree, the state machine initiates a scanning maneuver to create a map of the scene in the local frame. Tree detection in the local frame is less affected by inaccuracies in the global localization system. It is worth noting that the tree detector also works on global maps and can be used to aid mission planning; this is discussed in Sec.~\ref{sec::res_tree_detection_mission_planner}.  

Tree detection starts with a scanning maneuver by rotating the cabin while the chassis is kept steady. The cabin is initially pointing in the direction where the harvester expects the target tree to be (known from the mission planner). The two \ac{LIDAR}'s mounted on the front of the machine (see Fig.~\ref{fig::heap_front}) stream the point clouds at \SI{20}{Hz}. Note that the harvester's arm (or legs) might appear in the point cloud which is undesired. We use the \emph{robot\_self\_filter} package from the \ac{ROS} software stack to filter them out. The \emph{robot\_self\_filter} uses robot's geometry (meshes or geometric primitives), state (joint angles, pose) and range sensor's extrinsic calibration to identify points belonging to robot's links in the point cloud. Subsequently, these points are filtered out.

\begin{figure*}[tbh]
\centering
     \subfloat[Uncropped local map \label{fig::non_cropped}]{
       \includegraphics[width=0.45\textwidth]{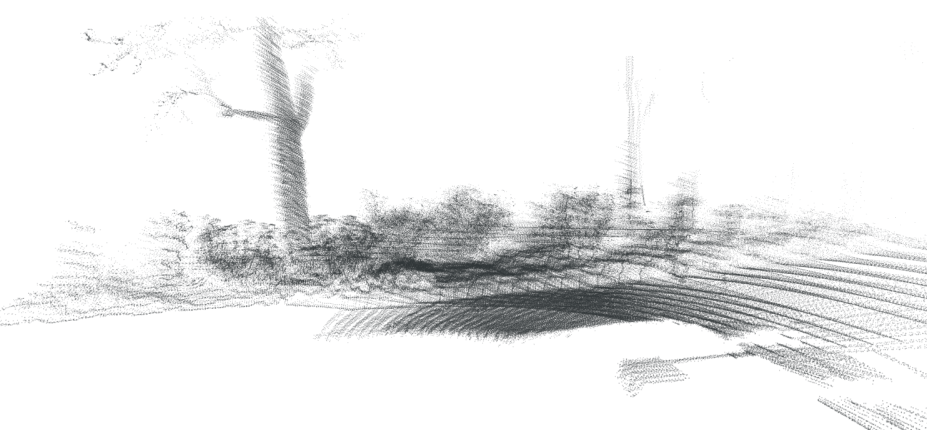}
     }
     \hfill
     \subfloat[Cropped and gravity aligned \label{fig::cropped_and_non_cropped}]{
       \includegraphics[width=0.45\textwidth]{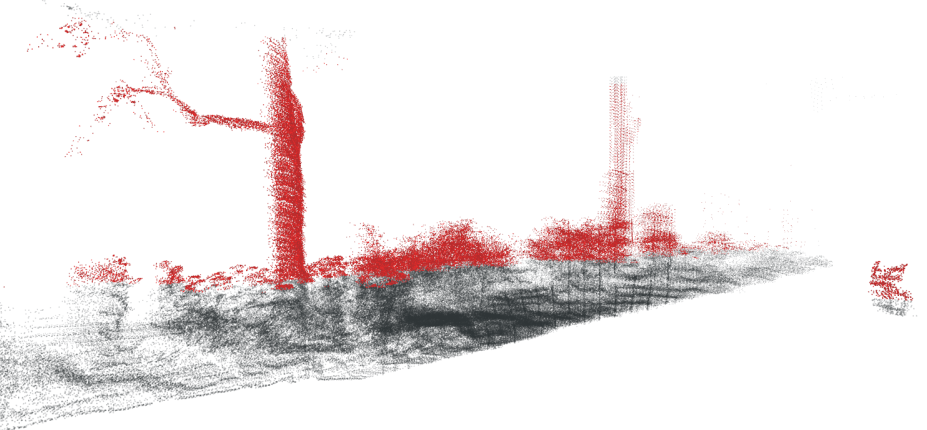}
     }
     \vskip\baselineskip
     \subfloat[Clusters in the scene \label{fig::trees_segmented}]{
       \includegraphics[width=0.45\textwidth]{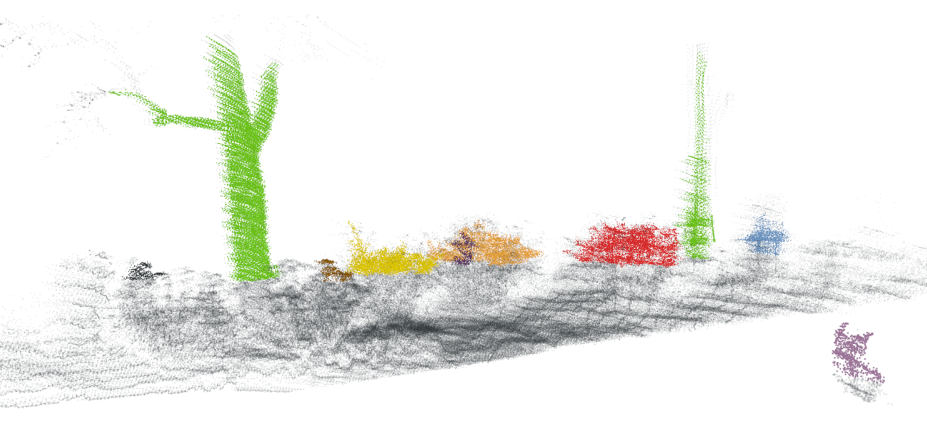}
     }
     \hfill
     \subfloat[Trees and bounding ellipsoids \label{fig::trees_detected}]{
       \includegraphics[width=0.45\textwidth]{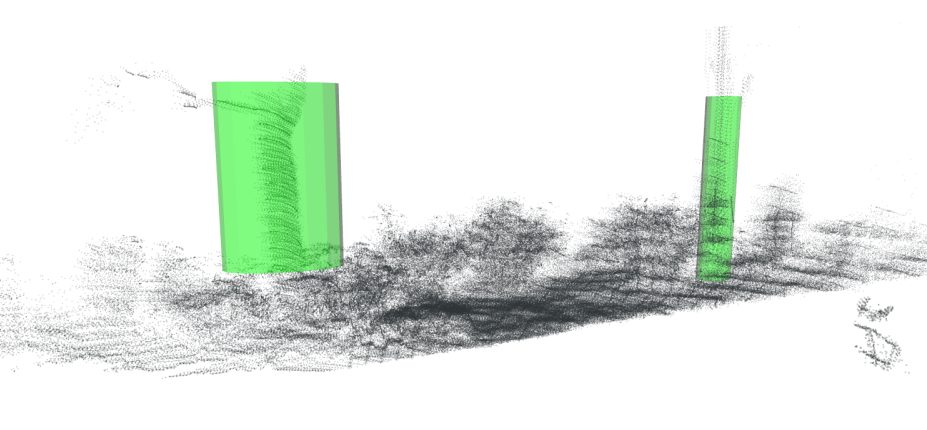}
     }
    \caption{Intermediate results in the tree detection pipeline. \textbf{\textit{Top Left:}} Scan of the area, without any cropping. One can observe the large ground plane in the point cloud. \textbf{\textit{Top Right:}} Cropped point cloud after first and second cropping (red color). \textbf{\textit{Bottom Left:}} Clusters in the scene. Each of these clusters is considered by the tree detection module. \textbf{\textit{Bottom Right:}} Final tree detection with the resulting bounding ellipsoids. }
    \label{fig::stitched_clouds}
\end{figure*}

A receiver node stitches filtered point clouds together into a local map. The relative transformation between subsequent scans is recovered from the chassis's roll and pitch angle together with the cabin joint angle measurement. Note that in the absence of joint angle measurements or odometry, once could use point cloud registration methods (e.g. \ac{ICP}). An example of a local map is shown in Figure \ref{fig::non_cropped}. The area scanned in Fig.~\ref{fig::non_cropped} is somewhat larger than necessary to visualize steps in the tree detection algorithm better. We turn the cabin in its yaw angle $\pm 30^{\circ}$ to scan the area and build a local map during the deployment. This corresponds to double the horizontal \ac{FOV} of the tilted Velodyne \ac{LIDAR} (see Fig.~\ref{fig::heap_front}) such that a dense local map can be built. We found that vertical Velodyne was more important for building dense maps, hence if one sensor is used, recommendation is to use it tilted.

The filtered scans from \acp{LIDAR} are transformed into the cabin frame and cropped to speed up the subsequent steps. We use a box filter to crop the point clouds. The $x-y$ limits are set to drop all the points farther than the arm reach. Once cropped, the scans are transformed into a gravity-aligned frame where they are concatenated. The maximal density of the concatenated cloud is constrained to further speed up the computation. We use \emph{libpointmatcher} library for cropping, density filtering and concatenation \cite{Pomerleau12comp}. Point cloud assembled from cropped scans is shown in black in Fig.~\ref{fig::cropped_and_non_cropped} (black). The assembled cloud is then gravity aligned and cropped again to filter out the ground plane and the tree crown (shown in red color, Fig.~\ref{fig::cropped_and_non_cropped}). Again, we use a box filter to remove all the that are lower than the center of the highest wheel. The red point cloud is sent to the tree detection module, which selects all prominent trees in the scan, as shown in Fig.~\ref{fig::trees_detected}.

The tree detection algorithm first computes Euclidean clusters in the input point cloud (shown in Fig.~\ref{fig::trees_segmented}). Euclidean clusters are searched using the algorithm from \cite{rusu2010semantic}. We discard the clusters with too few points. Since most trees grow vertically, a point cloud of a tree trunk should have a majority of the points spread out along the $z$ axis. Hence, \ac{PCA} is performed on each cluster, and we only keep clusters with a significant principal component along the $z$ axis. Note that we can exploit the verticality assumption since our point cloud is gravity-aligned. The gravity alignment score ($\in [0,1]$) is defined as the dot product of the largest principal component with a $[0 \: 0 \: 1]^T$ vector. Lastly, we check for the minimum height of the tree. The final detection result is shown in Fig.~\ref{fig::trees_detected}. Sizes of bounding ellipsoids are computed based on principal components' eigenvalues in $x$ and $y$ direction. The final tree location is the ellipsoid's center. In the case of multiple tree detection (such as in Fig.~\ref{fig::trees_detected}), the algorithm extracts coordinates of the tree closest to the expected tree position (from the mission planner).
\FloatBarrier
\clearpage
\section{Control}
\label{sec::control}
In this work, chassis control is responsible for locomotion and arm control for tree grabbing (harvesting). The respective control subsystems (shown in blue, Fig.~\ref{fig::whole_system}) are described in more detail in this section.
\subsection{Chassis Control}
\label{sec::chassis_control}
\begin{figure*}[tbh]
\centering
    \subfloat[Chassis schematic\label{fig::chassis}]{
       \includegraphics[width=0.41\textwidth]{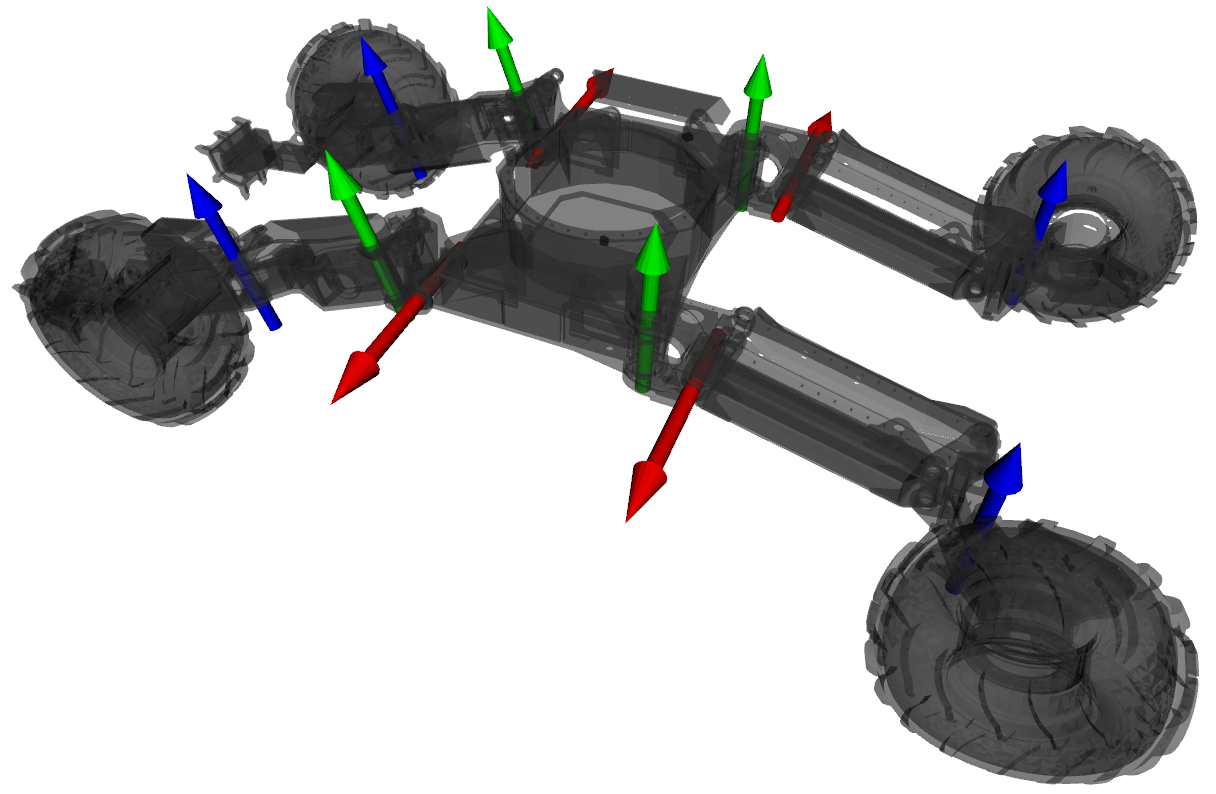}}
       \hfill
     \subfloat[Chassis in action   \label{fig::chassis_in_action}]{
       \includegraphics[width=0.5\textwidth]{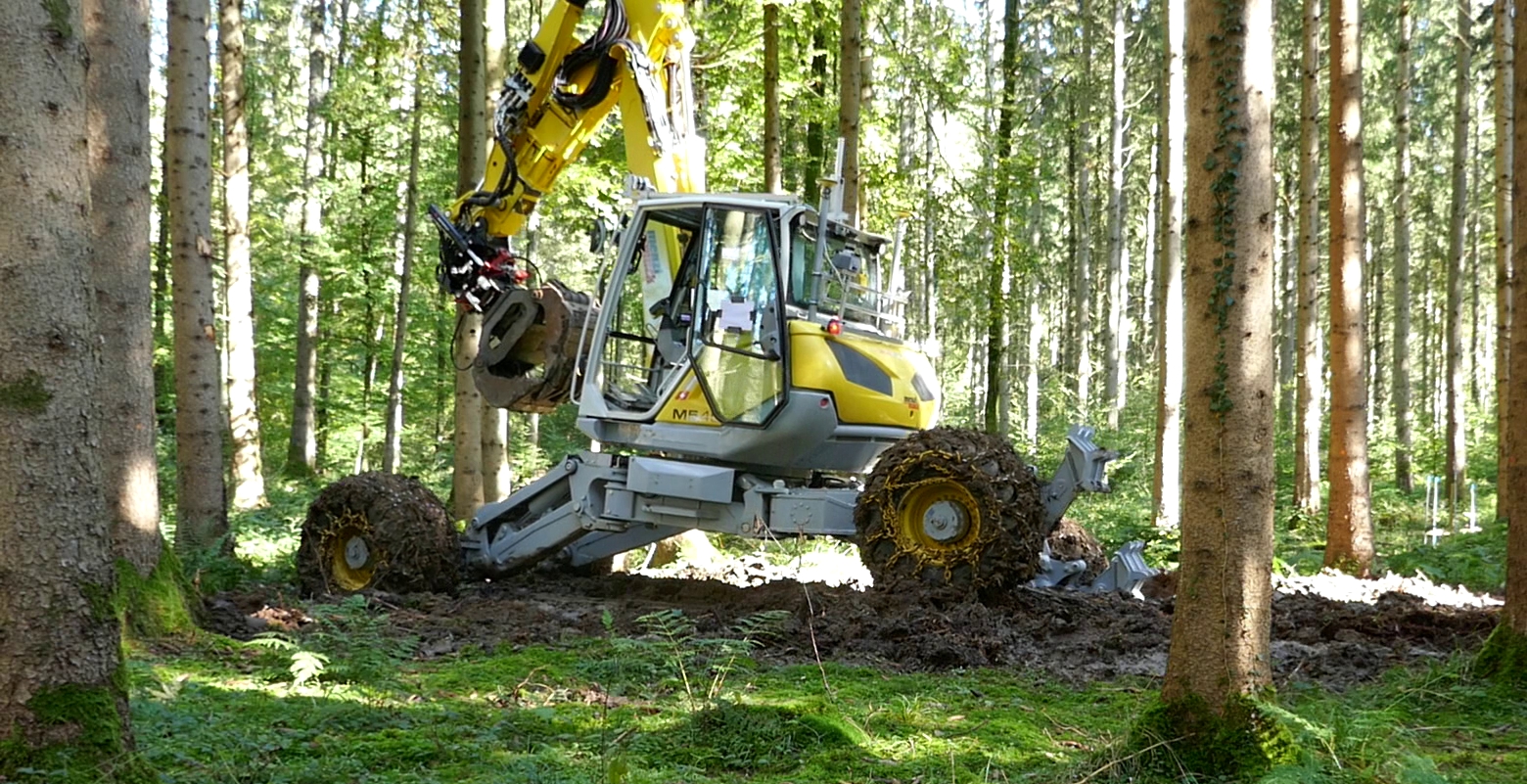}
     }
    \caption{\textit{\textbf{Left:}} Illustration of the \ac{HEAP}'s chassis. The cabin and the arm are not shown for the sake of clarity. Steering joints axes are shown with blue arrows, flexion joint axes with red, and abduction joint axes with green color arrows. The path following controller actuates the steering joints while the \ac{HBC} actuates the flexion joints. Abductions joints are not used. \textit{\textbf{Right:}} Chassis control system overcoming a stump during the deployment.}
    \label{fig::chassis_and_action}
\end{figure*}

\ac{HEAP} has four legs with wheels allowing it to drive and adapt to the terrain (see Fig.~\ref{fig::chassis_and_action}.) Thereby, the goal is to optimally distribute the four wheels' load to ensure traction and minimize terrain damage.

Terrain adaptation controller (also named \ac{HBC}) is based on a virtual model control principle where the controller computes a net force/torque on the chassis to achieve the desired orientation (roll, pitch) and height. An optimal contact force distribution is computed from the net force/torque for all the legs. Contact force tracking is achieved via force tracking on the hydraulic actuator level. For more details, refer to \cite{hutter2016force}. The described chassis controller can keep the base leveled while overcoming large irregularities in the terrain without getting stuck. Terrain adaptation is achieved using proprioceptive measurements only (joint sensing, \ac{IMU}). We have extensively tested \ac{HBC} performance in our previous work (\cite{hutter2015towards}, \cite{hutter2016force}). A video showing the machine overcoming challenging terrain can be found online\footnote{\url{https://youtu.be/5_Eq8CxKkvM}}.

\begin{figure}[tbh]
\centering
    \includegraphics[width=0.8\textwidth]{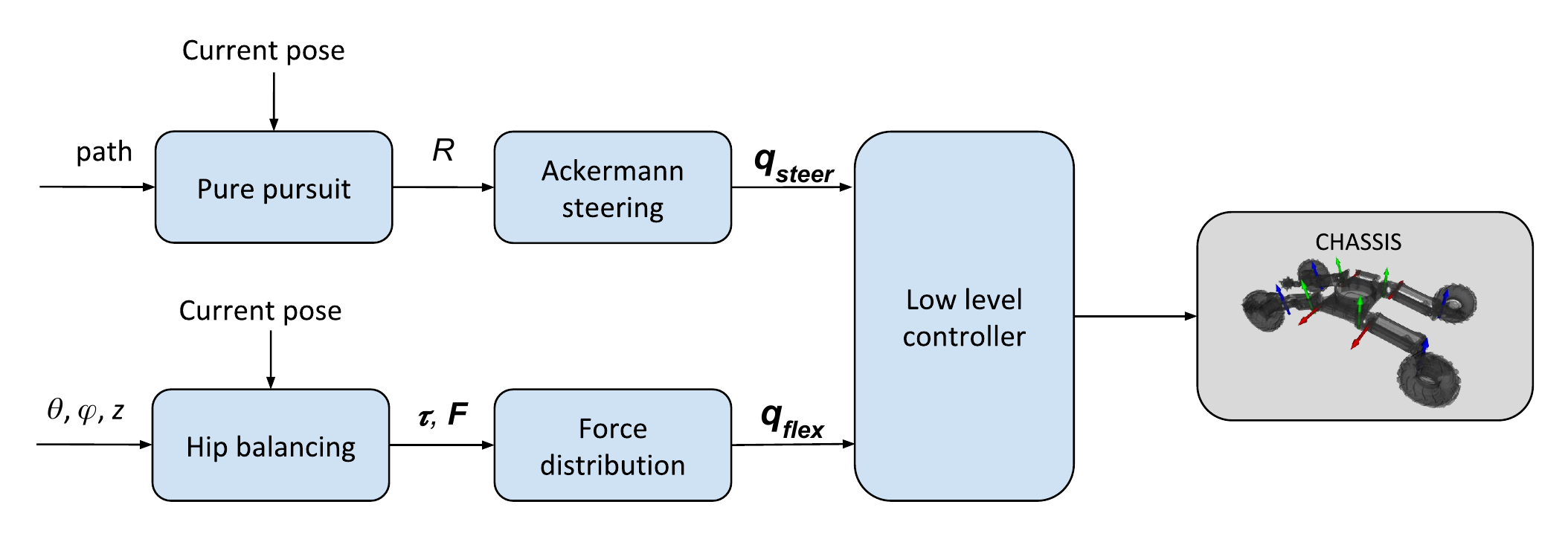}
    \caption{System diagram of the chassis control module deployed on \ac{HEAP}.}
    \label{fig::chassis_control_system}
\end{figure}
For path tracking, we use a pure pursuit controller similar to the one outlined in \cite{kuwata2008motion}. The pure pursuit controller calculates the turning radius (denoted with $R$) required to bring the machine on the path computed by the planner. The Ackermann steering module then calculates steering joint angles for each leg. Steering angles are computed such that all four wheels point to the same center of rotation with the radius $R$. The computation is implemented as a quadratic optimization that satisfies all the joint limits and finds a feasible center of rotation. For more details on the Ackermann steering module, the reader is referred to \cite{jud2021heap}.

The \ac{HBC} and the pure pursuit controller are combined to locomote the harvester, as shown in Fig.~\ref{fig::chassis_control_system}. Both controllers close the feedback loop over pose: \ac{HBC} regulates the roll, pitch, and height, whereas pure pursuit ensures that $x,y$, and yaw are tracked correctly. The Ackermann steering module computes position reference for the steering joints ($\boldsymbol{q}_{steer}$ in Fig.~\ref{fig::chassis_control_system}), while the force distribution module computes joint torques assuming quasi-static conditions ($\boldsymbol{q}_{flex}$). The low-level controller translates the joint quantities (torques, positions) into the valve commands. In Fig.~\ref{fig::chassis_in_action}, the proposed controller is driving over a stump. Note how the controller retracts the left hind leg to maximize the traction. In parallel, the pure pursuit controller controls the driving direction.
\subsection{Arm Control}
\label{sec::arm_control}
Reaching the end-effector target position determined from the tree detection module is achieved by following a trajectory from the planner described in Section~\ref{sec::arm_grasp_pose_planning}. The trajectory following is done using an \ac{IK} controller (see \cite{siciliano2010robotics}) which uses the \ac{HO}, based on the implementation from \cite{bellicoso2016perception}. The main difference is that we tailor the tasks for \ac{HEAP} instead of a quadruped robot. Besides tracking, \ac{HO} computes joint velocities, enforces kinematic limits, and ensures that all flow constraints for hydraulic actuators are satisfied. The set of tasks optimized by the \ac{HO} is given in Table~\ref{tab::arm_controller_tasks}, where \emph{1} is the highest priority task. Opening and closing the gripper is controlled directly by the state machine and it is done in a purely open-loop fashion.

\begin{table}[tbh]
\centering
\caption{Task priority inside the hierarchical optimization for the arm inverse kinematics controller.}
\label{tab::arm_controller_tasks}
\begin{tabular}{|l|l|}
\hline
Priority & Task                     \\ \hline \hline
1        & Equations of motion      \\
2        & Pump flow limit          \\
3        & Cylinder force limits    \\
3        & Cylinder velocity limits \\
3        & Cylinder position limits \\
4        & End-effector orientation \\
4        & End-effector position    \\ \hline
\end{tabular}
\end{table}
\clearpage
\section{Planning}
\label{sec::planning}
In this section we describe the mission planner and the motion planning stack in more detail. The motion planing stack is divided into three components shown in Fig.~\ref{fig::whole_system}: base motion planner, grasp pose planner and the arm motion planner.

\subsection{Mission planner}
\label{sec::mission_planner}
\begin{figure}[tbh]
\centering
    \includegraphics[width=0.9\textwidth]{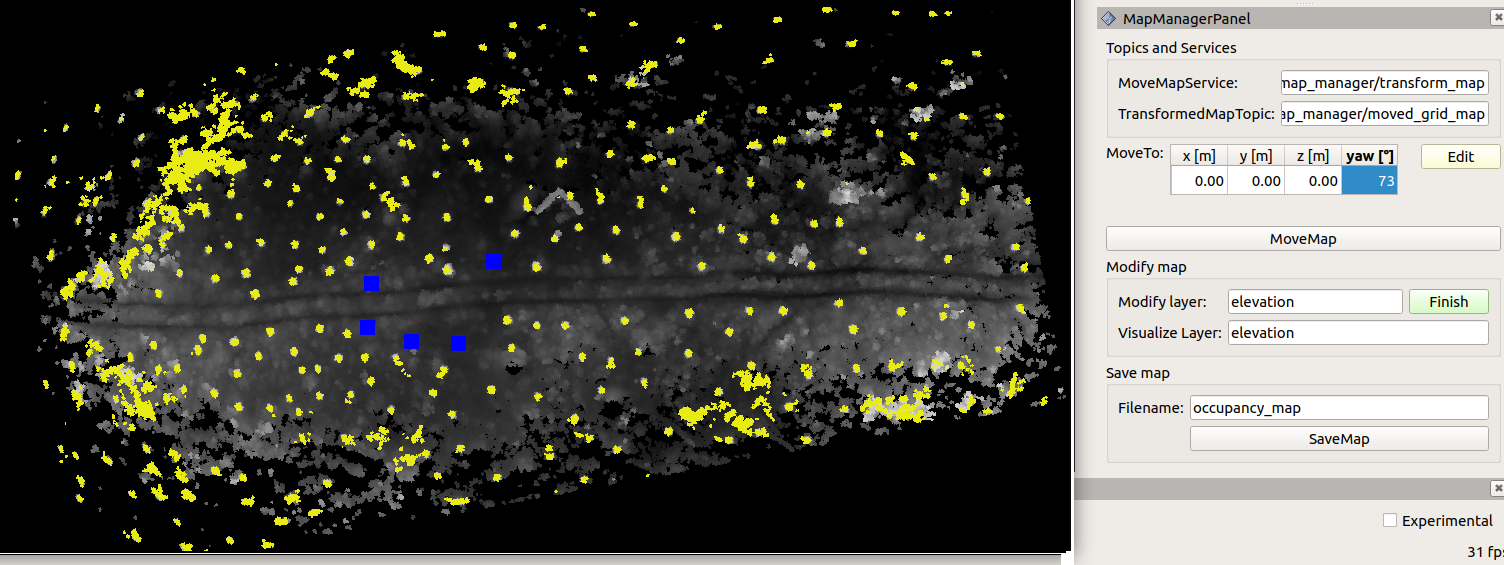}
    \caption{\ac{GUI} used for mission planning. It is a panel in \emph{Rviz} and it uses a \emph{Qt} front-end. An elevation map of previously mapped environment is shown in gray-scale. All points higher than \SI{2}{\meter} are colored in yellow; hence, the yellow color corresponds to the tree trunks and parts of the canopy. In this example, a total of five trees (marked with blue squares) are selected to be cut.}
    \label{fig::gui}
\end{figure}

To conduct the experiments, we design a mission planner which determines which tree to grab next. Before the mission, a human manually selects the trees to be cut. Selection is accomplished using the \ac{GUI} shown in Fig.~\ref{fig::gui}, thus mimicking an algorithm for tree inventory management. When clicked on, a tree gets added to the tree list for harvesting with position coordinates extracted in the map frame. The mission planner passes the tree coordinates to the state machine in the same order as they were selected. A mission planner for optimizing some user-given objective and finding an optimal cutting order remains to be investigated in the future.

\subsection{Base Pose Planning}
\label{sec::base_planning}
We do not make any assumptions on the terrain properties; thus, our planner accounts for irregular terrain. Path planning in rough terrain is a complex problem, and in general, it has not been solved yet. In this work, we split it into two more manageable subproblems: planning in \emph{SE(2)} (instead of \emph{SE(3)}) and traversability estimation.
\subsubsection{Traversability Estimation} 
\label{sec::traversability_estimation}
The traversability estimation module discerns which areas can be driven versus which ones should be avoided. From the elevation map we compute a function $f:(x,y) \rightarrow \; [0,1]$ that tells us for each coordinate $(x,y)$ where \ac{HEAP} can go (e.g. 1 - safe to drive, 0 - not safe). Traversability estimation has been previously studied, and different approaches for various sensing modalities exist. In this work, we use purely geometric traversability estimation, which is directly applied to elevation maps without any additional processing. We follow the approach presented in \cite{wermelinger2016navigation} that evaluates three criteria: terrain slope, roughness, and step height for local terrain patches (\SI{0.3}{\meter} radius). The final result is a weighted sum of all three components. We chose to mainly rely on step criterion (\SI{80}{\percent}) since the chassis control system introduced in Sec.~\ref{sec::chassis_control} can overcome slopes and drive over rough terrain. The remaining \SI{20}{\percent} was assigned to the slope criterion to prevent the harvester from driving on very steep slopes, which could result in slipping. 

\begin{figure*}[htb]
\centering
    \subfloat[Traversability Classification errors \label{fig::editing_traversability}]{
       \includegraphics[width=0.48\textwidth]{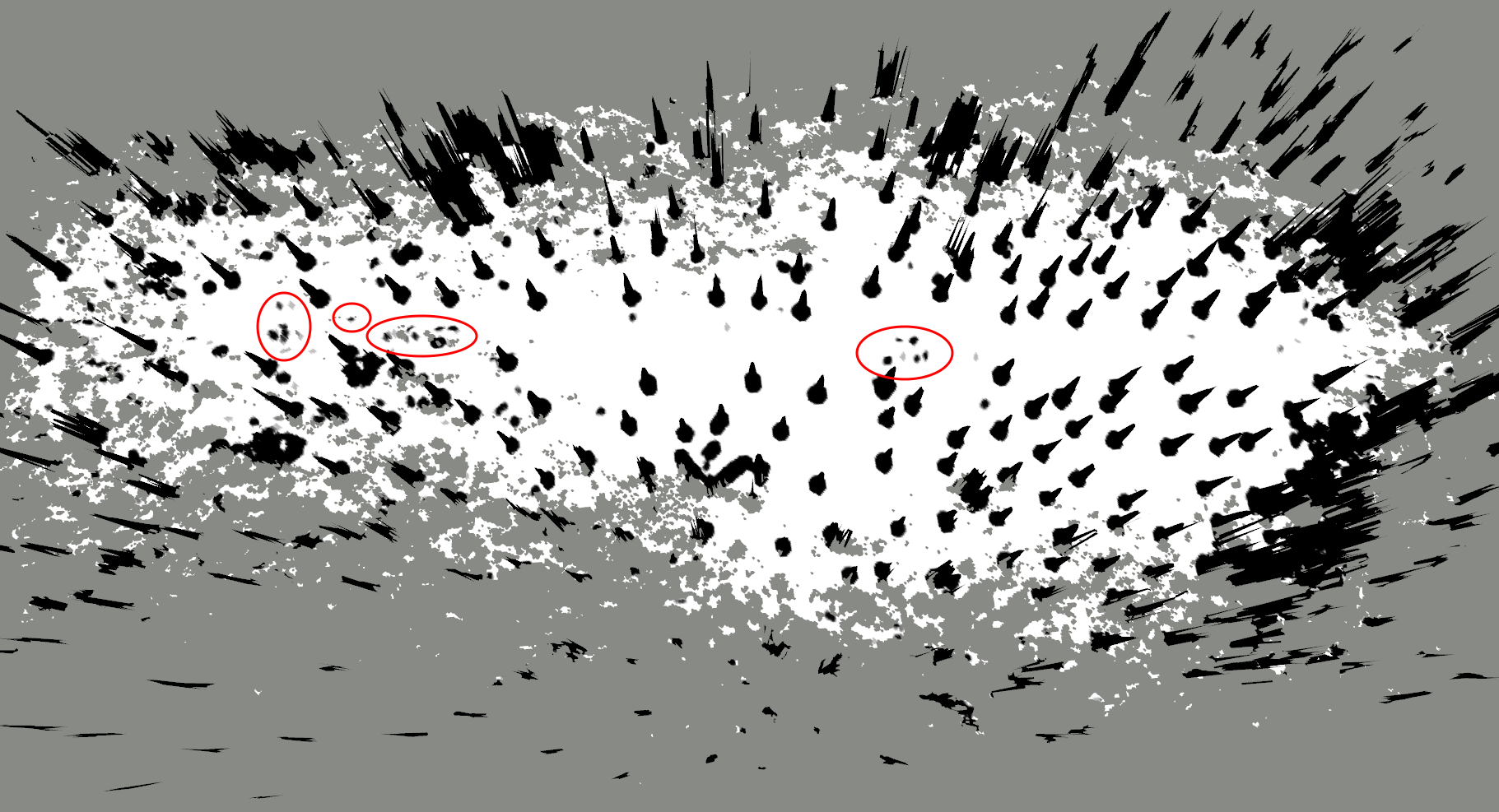}}
     \subfloat[Corrected traversability \label{fig::traversability_corrected}]{
       \includegraphics[width=0.48\textwidth]{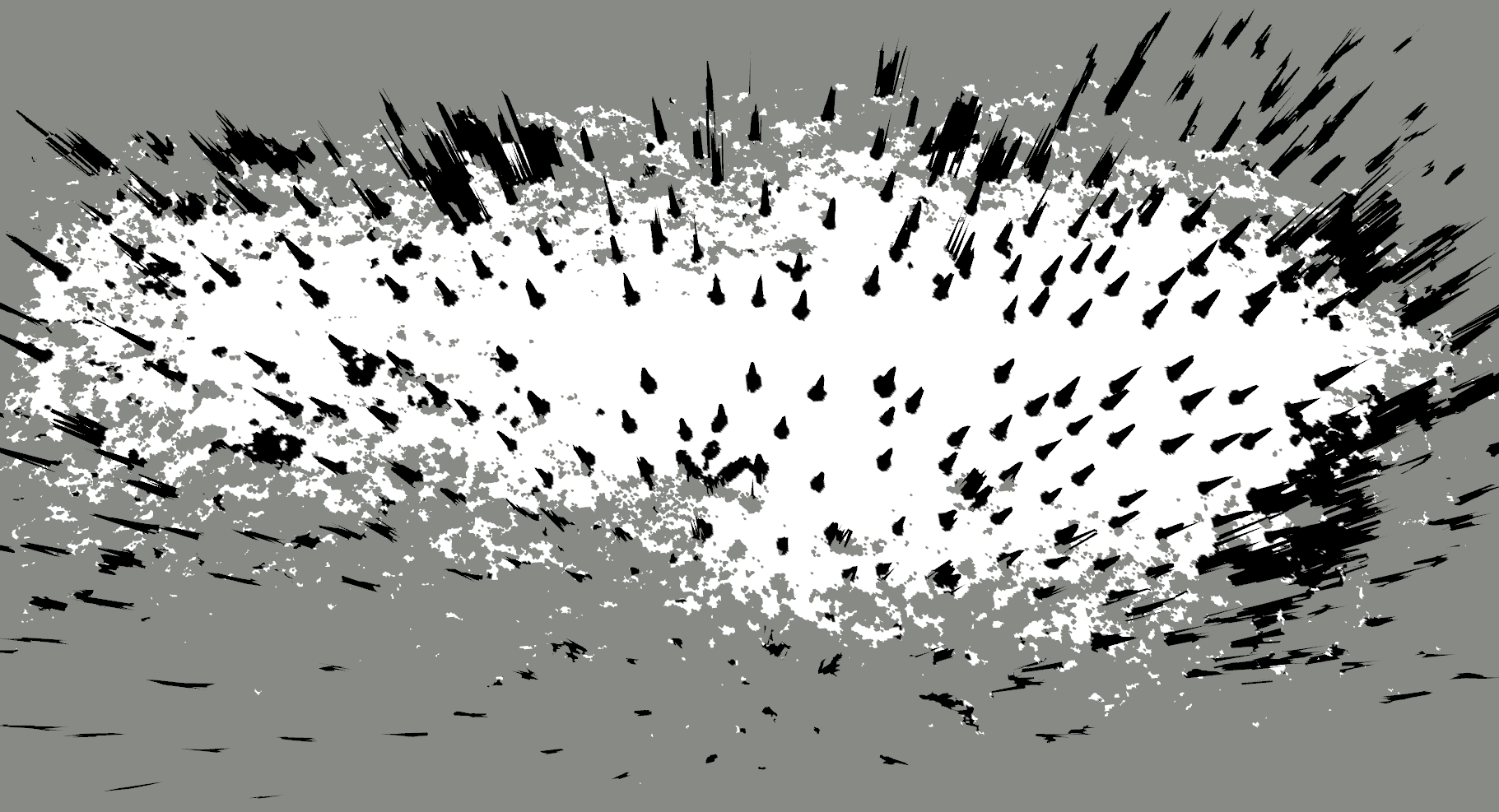}
     }
    \caption{Traversability maps computed by our approach shown against gray background. The maps shown are 2.5D maps with different layers (traversability layer displayed). White areas represent fully traversable terrain, whereas black areas represent terrain that is not traversable. \textbf{\textit{Left:}} Red areas have been incorrectly classified as untraversable by the algorithm (false negatives). \textbf{\textit{Right:}} Traversability map after applying the manual correction.}
    \label{fig::traversability_correction}
\end{figure*}

For storing the traversability map we use the grid map \cite{fankhauser2016universal} data structure. Fig.~\ref{fig::editing_traversability} shows the traversability map, a 2.5D map with a traversability layer; incorrectly classified areas (classified untraversable while it is traversable) are encircled red. Thick vegetation that occludes the ground from the \ac{LIDAR} sensor during the mapping phase is the main culprit for false negatives. Since the overall misclassified area is small, a human can manually correct the errors. We used the \ac{GUI} from Fig.~\ref{fig::gui} and the correction lasted about 2 minutes.  Fig.~\ref{fig::traversability_corrected} shows the corrected traversability map; note how the algorithm automatically classifies tree trunks as untraversable. Presented traversability estimation has an upside that is easy to implement; there is no need for elevation map processing or segmenting out the tree trunks explicitly. Fully automating the proposed pipeline would require additional work to eliminate the correction step and be implemented in future work.

The traversability map (see Fig.~\ref{fig::traversability_corrected}) is converted into an occupancy map. Any traversability value smaller than 0.5 is deemed an obstacle, and higher values are regarded as free space. Motion planners use the occupancy map, a 2.5D map with an occupancy layer.
\subsubsection{Base Approach Pose and Path Planning}
\label{sec::path_approach_pose_planning}
The planning subsystem (shown in yellow color in Fig.~\ref{fig::whole_system}) gets a tree position from the mission planner. The tree position is an approximate target location for the end-effector. The harvester should not reach the tree position itself because this would result in a collision with the tree. To this end, we develop an algorithm for joint path and approach pose planning. We evaluate the algorithm in both simulations and natural environments. Functionality described in this subsection corresponds to blocks \emph{Approach Pose Generation} and \emph{Base Motion Planner} in Fig.~\ref{fig::whole_system}.

\begin{figure*}[htb]
\centering
    \subfloat[Path planning footprint \label{fig::path_planning_footprint}]{
       \includegraphics[width=0.3\textwidth]{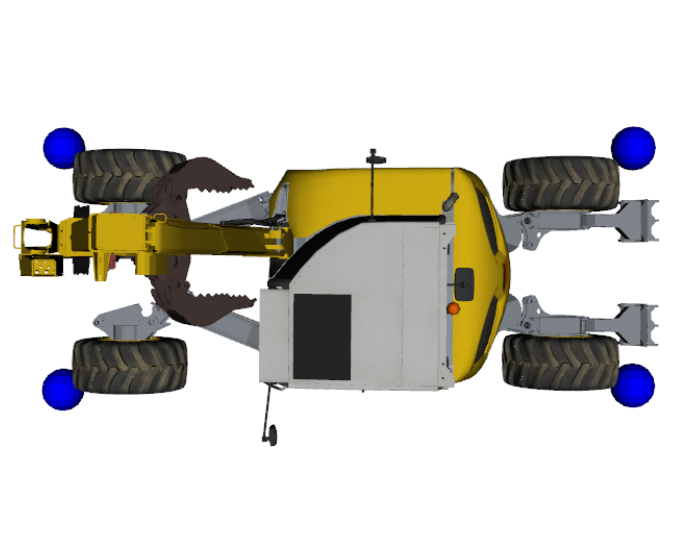}} \hspace{0.7cm}
     \subfloat[Approach pose planning footprint \label{fig::arm_planning_footprint}]{
       \includegraphics[width=0.3\textwidth]{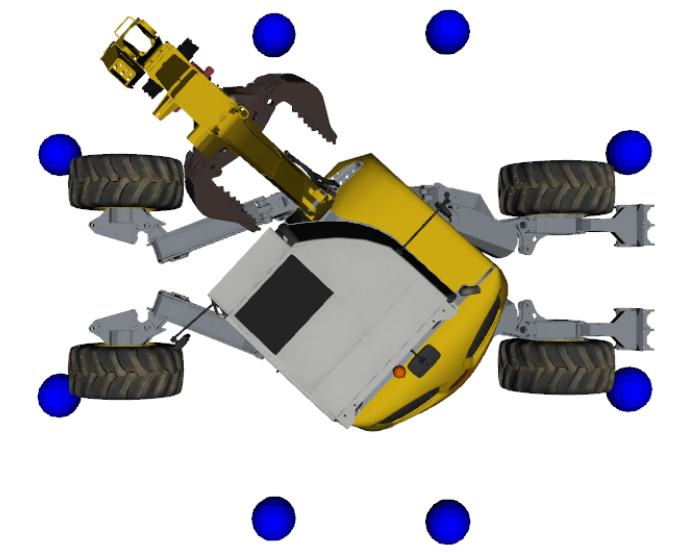}
     }
    \caption{Top view of \ac{HEAP} with the arm retracted with the footprint vertices shown in blue. Note: not drawn to scale. \textbf{\textit{Left:}} Footprint used for path planning (see Alg.~\ref{alg::approach_path_approach_pose_planning}). \textbf{\textit{Right:}} Footprint used for the approach pose generation (see Alg.~\ref{alg::approach_pose_candidate}) is wider in the middle, thus allowing for cabin turning.}
    \label{fig::footprints}
\end{figure*}

Existing planning algorithms typically plan from starting pose to the goal pose. However, in our case, we do not know the goal pose (only approximate end-effector position in $x$ and $y$). Hence, the proposed planning algorithm is split into two stages: first, the planner generates candidate poses and checks for their feasibility. Secondly, the planner attempts to compute a path to any of the feasible candidates. Thus the approach pose planning is reduced to a common path planning problem, and we leverage a \ac{RRT}* algorithm for a joint path and approach pose computation. Our implementation leverages \ac{OMPL}, \cite{sucan2012open}. We chose the \ac{OMPL} because it is available as open-source and comes with efficient implementations of various sampling-based planners. We do not rely on optimization-based planners since forests are cluttered environments with many local minima, which present a challenge for optimization.

The approach pose generating subroutine is shown in Alg.~\ref{alg::approach_pose_candidate}. It starts by getting a target/tree location $\boldsymbol{x}_T \in \mathbb{R}^2$ (line~\ref{alg::getNextTarget}), and computing a set of approach poses around it (lines \ref{alg::generateApproachPosesStart} - \ref{alg::generateApproachPosesEnd}). We compute $M \times N$ approach positions in polar coordinates around $\boldsymbol{x}_T$ by pairing $M$ different distances to the tree with $N$ uniformly distributed polar angles. Lastly, the approach positions are paired with $K$ yaw angles (lines 20-22) to generate a total of $N \times M \times K$ candidate approach poses in \emph{SE(2)}. To be a valid candidate, an approach pose must be collision-free (lines 6-7), the harvester's arm must be able to reach the target (lines 6-9), and all heuristics must be satisfied (lines 10-11).

\begin{algorithm}[htb]
\caption{Candidate approach pose generator}
\label{alg::approach_pose_candidate}
\begin{algorithmic}[1]
\State $x_T,y_T \gets \textsc{getNextTargetLocation}()$ \label{alg::getNextTarget}
\State $isUseHeuristic \gets \textsc{readFromConfigFile}()$
\Procedure{computeCandidateApproachPoses}{$x_T, y_T$}
    \State $candidateApproachPoses \gets \textsc{getApproachPosesAroundTarget}(x_T,y_T)$
    \For{\texttt{each} $pose$ \texttt{in} $candidateApproachPoses$}
        \If{$\textsc{isInCollision}(pose)$} 
            \State $candidateApproachPoses.\textsc{delete}(pose)$
        \EndIf
        \If{$\neg \textsc{isTargetReachableFrom}(pose)$} \label{alg::targetReachable}
            \State $candidateApproachPoses.\textsc{delete}(pose)$
        \EndIf
        \If{$ isUseHeuristic \; \textsc{AND} \; \neg \textsc{isHeuristicValid}(pose)$}  
            \State $candidateApproachPoses.\textsc{delete}(pose)$
        \EndIf
    \EndFor
    \State $\textbf{return} \; candidateApproachPoses$
\EndProcedure
\Procedure{getApproachPosesAroundTarget}{$x_T,y_T$} \label{alg::generateApproachPosesStart}
    \State $distances \gets \{ d_1,d_2,...,d_M\}$
    \State $polarAngles \gets \{ \phi_1, \phi_2,...,\phi_N \}$
    \State $headings \gets \{ \psi_1, \psi_2,...,\psi_K \}$
    \State $approachPoses \gets \emptyset$
    \For{\texttt{each} $d_i$ \texttt{in} $distances$}
        \For{\texttt{each} $\phi_i$ \texttt{in} $polarAngles$}
            \State $(x_i, y_i) \gets (x_T + d \cos(\phi_i),\: y_T + d \sin(\phi_i))$
                \For{\texttt{each} $\psi_i$ \texttt{in} $headings$}
                    \State $approachPoses.\textsc{ADD}([x_i,y_i,\psi_i])$
                \EndFor
        \EndFor
    \EndFor
    \State $\textbf{return} \; approachPoses$ \label{alg::generateApproachPosesEnd}
\EndProcedure
\end{algorithmic}
\end{algorithm}

The blue points shown in Fig.~\ref{fig::footprints} represent the harvester's collision footprint. The harvester is in a collision if there is an obstacle inside the blue points convex hull. Collision checks inside the Alg.~\ref{alg::approach_pose_candidate} use the footprint shown in Fig.~\ref{fig::arm_planning_footprint}. Note how the hull is wider in the middle to allow for cabin turns. The real-world map is abstracted away from the planner, and it only sees the obstacles computed by the traversability estimation step. Such a simplification is warranted by using a control system able to overcome slopes and rough terrain.

\begin{figure*}[htb]
\centering
    \subfloat[All approach poses \label{fig::all_approach_poses}]{
       \includegraphics[width=0.32\textwidth]{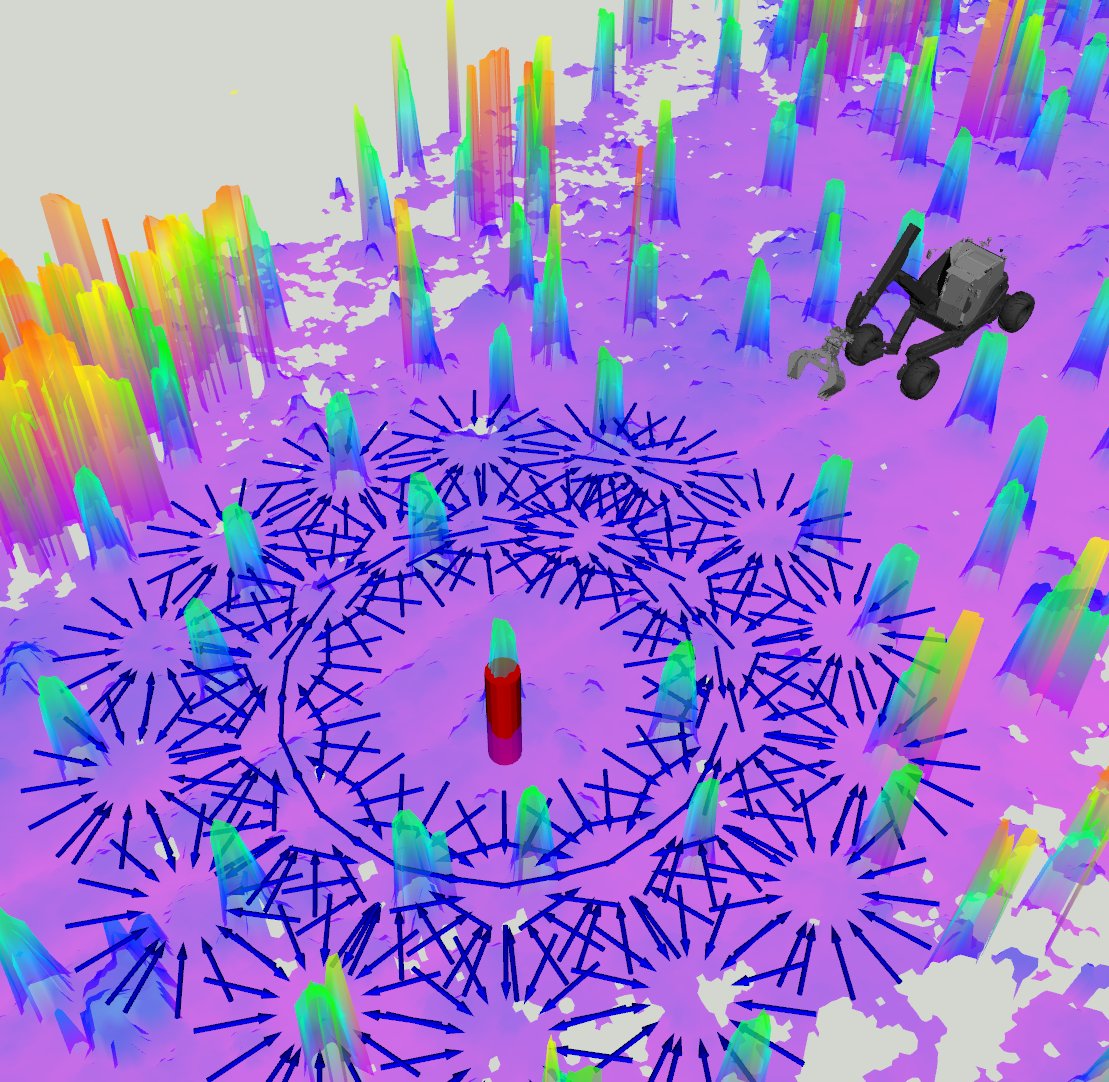}}
     \subfloat[Goal target reachable \label{fig::reachability_criterion}]{
       \includegraphics[width=0.32\textwidth]{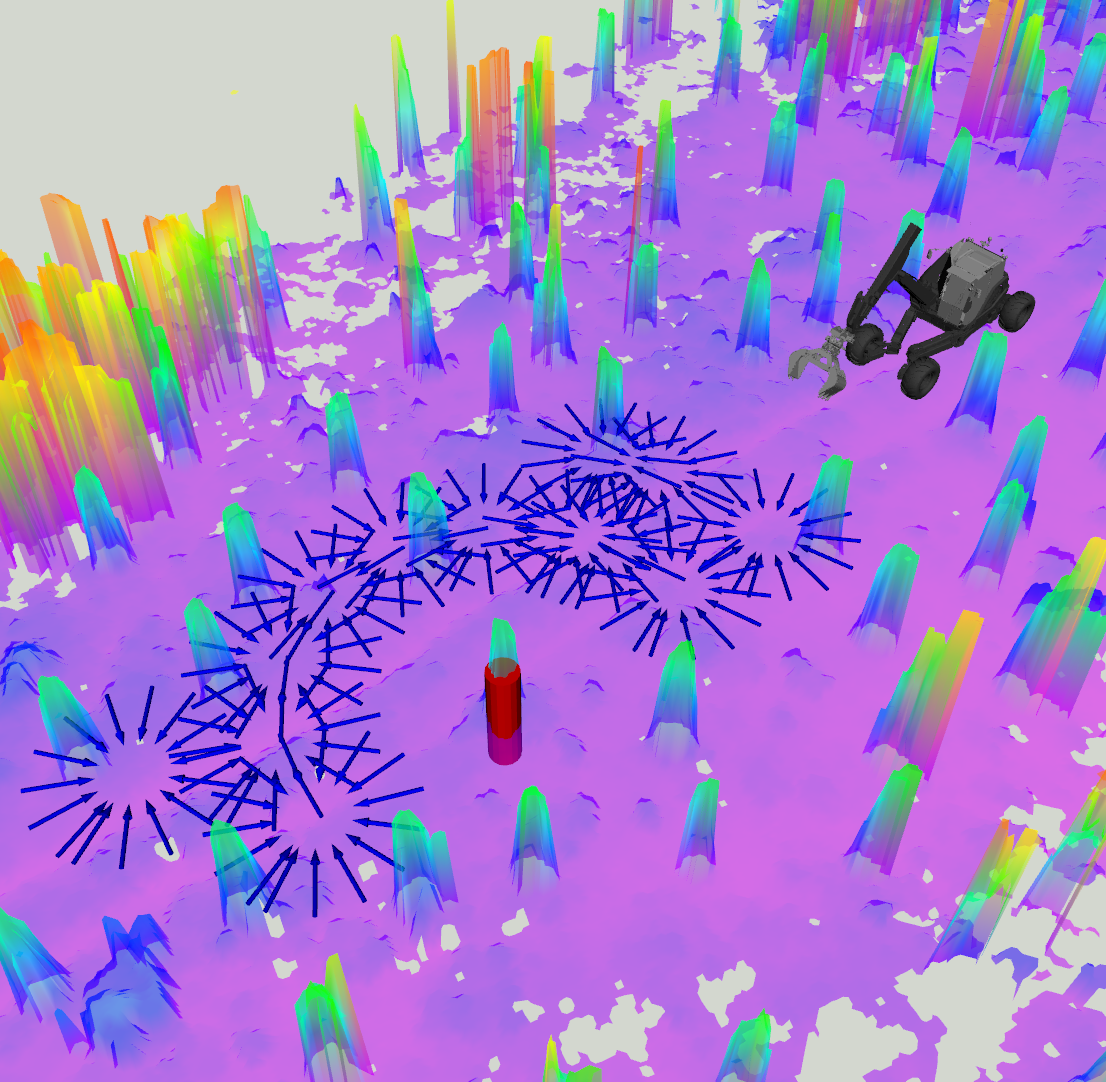}
     }
     \subfloat[Goal reachable \& no collision \label{fig::both_criterions}]{
       \includegraphics[width=0.32\textwidth]{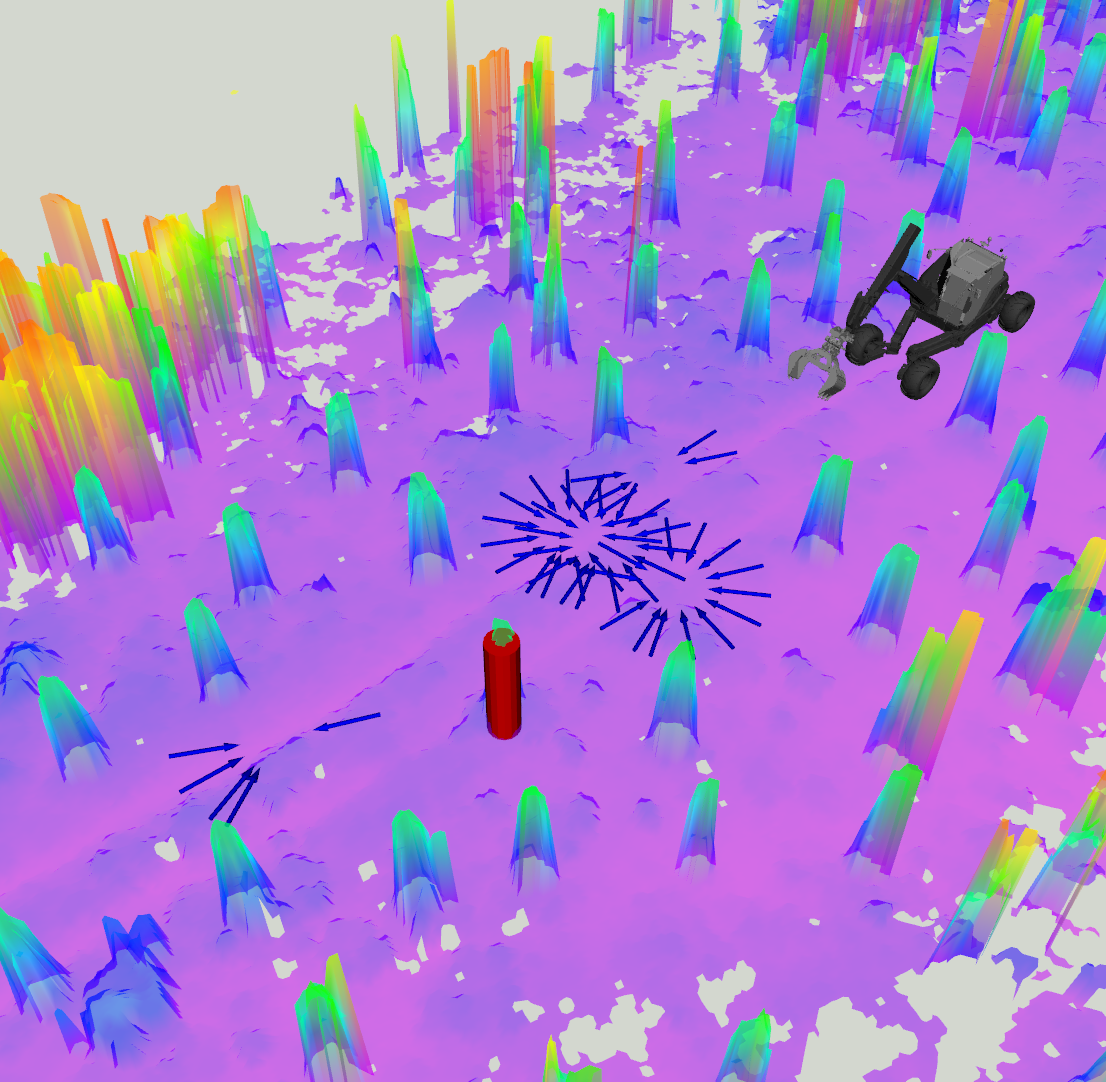}
     }
    \caption{Approach pose generation in the forest environment. The elevation map with the target tree (red cylinder) and candidate approach poses are shown with blue arrows. Some approach poses have been omitted for the sake of clarity. \textbf{\textit{Left:}} In total, 450 goal approach poses are generated. So far they have not been checked for feasibility and this example we do not apply any heuristics. \textbf{\textit{Middle:}} Out of 450 poses 155 do not satisfy the target reachability criterion (line 8 in Algorithm~\ref{alg::approach_pose_candidate}). The arm can extend and grab the target tree from 195 remaining poses. \textbf{\textit{Right:}}. Out of 195 poses, 168 of them are in collision with the environment (line 6 in Algorithm~\ref{alg::approach_pose_candidate}). The remaining 27 approach poses both satisfy the reachability criterion and are not in collision. These remaining 27 approach poses are then sampled inside the RRT* to determine which ones are attainable. }
    \label{fig::candidate_approach_poses}
\end{figure*}

The target being reachable from an approach pose means that the harvester can safely extend the arm to reach the goal position (see Alg.~\ref{alg::approach_pose_candidate}, line \ref{alg::targetReachable}). We check whether there is a collision-free line of sight from the base to the target goal. Because of \ac{HEAP}'s kinematic structure, the arm always stays within a slab in the $x-z$ plane (in the cabin frame). Hence, we require no obstacles inside the slab spanned by the target tree position and the base position. Therefore, there is no need for more complicated algorithms that are popular in mobile manipulation literature (e.g., \cite{zucker2013chomp}). An example of the approach pose generation in a forest environment is shown in Fig.~\ref{fig::candidate_approach_poses}.

\begin{figure}[htb]
\centering
    \includegraphics[width=0.5\columnwidth]{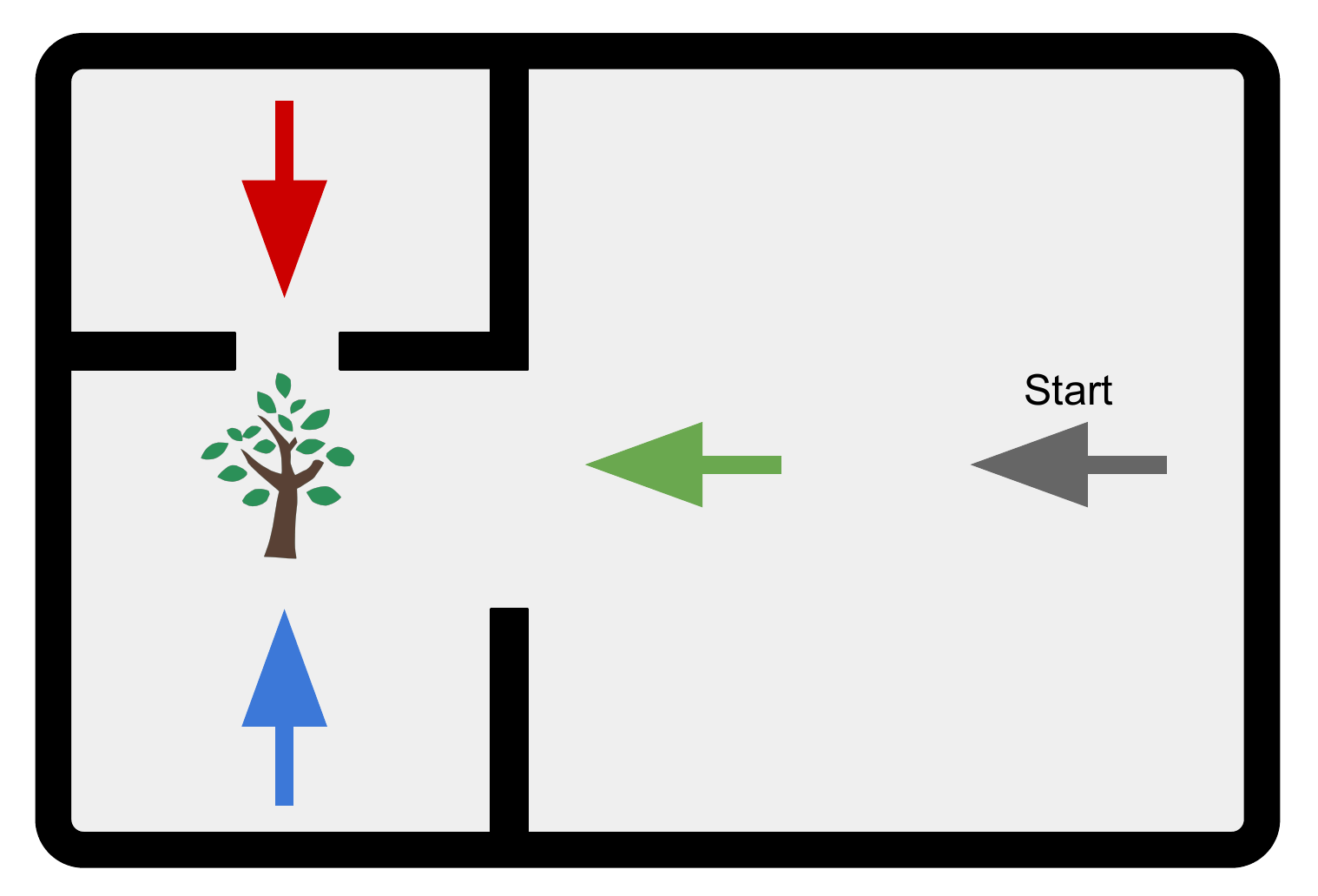}
    \caption{Example environment for approach pose planning. All approach poses are feasible; however, the red pose is not reachable from the start. The green pose is better than the blue pose since it is easier to reach.}
    \label{fig::good_and_bad_approach_poses}
\end{figure}

In the second stage, the planner checks whether the feasible approach poses are attainable, i.e., can the harvester drive to them. Note that a feasible approach poses such as the red one in Fig.~\ref{fig::good_and_bad_approach_poses} is not necessarily attainable. The subroutine for attainability checking is summarized in Alg.~\ref{alg::approach_path_approach_pose_planning}. We build upon a standard \ac{RRT}* planner that grows a random tree, as described in \cite{karaman2011sampling}. Instead of trying to connect the single goal pose (as standard \ac{RRT}*) to the rest of the tree, we attempt to connect every candidate approach pose. Planning terminates when the allotted planning time runs out. We found \SI{5}{\second} to be an adequate compromise between computation time vs mission progress time. The footprint used for collision checking inside Alg.~\ref{alg::approach_path_approach_pose_planning} is shown in Fig.~\ref{fig::path_planning_footprint}. As a steering function inside the \ac{RRT}*, we use the \ac{RS} curves \cite{reeds1990optimal} which allows us to respect the minimal turning radius constraint (just like any car, HEAP cannot turn in place). The turning radius parameter was set to \SI{8.3}{\meter}.

The rationale behind the planner's second stage is that an attainable approach pose has a high probability of having a collision-free connection to the rest of the random tree. For example, the green pose is a better approach pose compared to the blue one in Fig.~\ref{fig::good_and_bad_approach_poses}. The environment around the green pose is less cluttered, so the probability of a successful connection is higher. A beneficial feature of the described approach (Alg.~\ref{alg::approach_path_approach_pose_planning}) is outsourcing the final approach pose selection to the \ac{RRT}*. Note that using the \ac{RRT}* in its standard form would require the user to pick an approach pose, a challenging task since we do not know which ones are attainable \emph{a priori}. Since fewer approach poses ensure faster convergence, the proposed framework allows the use of heuristics for pruning the approach pose candidate set. Pruning heuristics can be a wide array of constraints and rules, thus allowing for great flexibility (e.g., discard approach poses where heading changes more than \SI{90}{\degree}).
\begin{algorithm}[htb]
\caption{Path and approach pose planner}
\label{alg::approach_path_approach_pose_planning}
\begin{algorithmic}[1]
\Procedure{computePathAndApproachPoses}{$x_T, y_T$}
    \State $x_T,y_T \gets \textsc{getNextTargetLocation}()$
    \State $\boldsymbol{p}_s \gets startingPose$
    \State $rrt.\textsc{initialize}(\boldsymbol{p}_s)$
    \State $candidateApproachPoses \gets \textsc{computeCandidateApproachPoses}(x_T,y_T)$
    \While{$t_{cpu} < T_{max}$}{
        \State $rrt.\textsc{growTree}()$
        \For{\texttt{each} $\boldsymbol{p}_i$ \texttt{in} $candidateApproachPoses$}
            \State $rrt.\textsc{connectToTree}(\boldsymbol{p_i})$
        \EndFor
    \EndWhile}
\EndProcedure
\State $\textbf{return} \; \big( \boldsymbol{p_i}, rrt.\textsc{getPath}() \big)$
\end{algorithmic}
\end{algorithm}

The proposed approach pose planning takes into account almost all \acp{DoF} that \ac{HEAP} has to offer. It allows the machine to turn the arm and approach trees from any angle. Furthermore, the harvester can utilize both driving directions when navigating to the goal target. One could still improve the approach pose generation to anticipate the arm turning direction. Algorithm \ref{alg::approach_pose_candidate} uses collision footprint shown in Fig.~\ref{fig::arm_planning_footprint} which is clearly conservative since the arm doesn't have to make a full \SI{360}{\degree} turn. Anticipating the turning direction would allow shrinking the collision footprint, which is very beneficial in cluttered environments such as one shown in Fig.~\ref{fig::candidate_approach_poses}. Another possible improvement is to adapt the number of approach pose candidates based on the environment. Generating too many approach pose candidates slows down the planning while marginally contributing to finding better solutions when the obstacle density is low.

\subsection{Arm Grasp Pose and Motion Planning}
\label{sec::arm_grasp_pose_planning}
\begin{figure}[tbh]
\centering
    \includegraphics[width=0.6\textwidth]{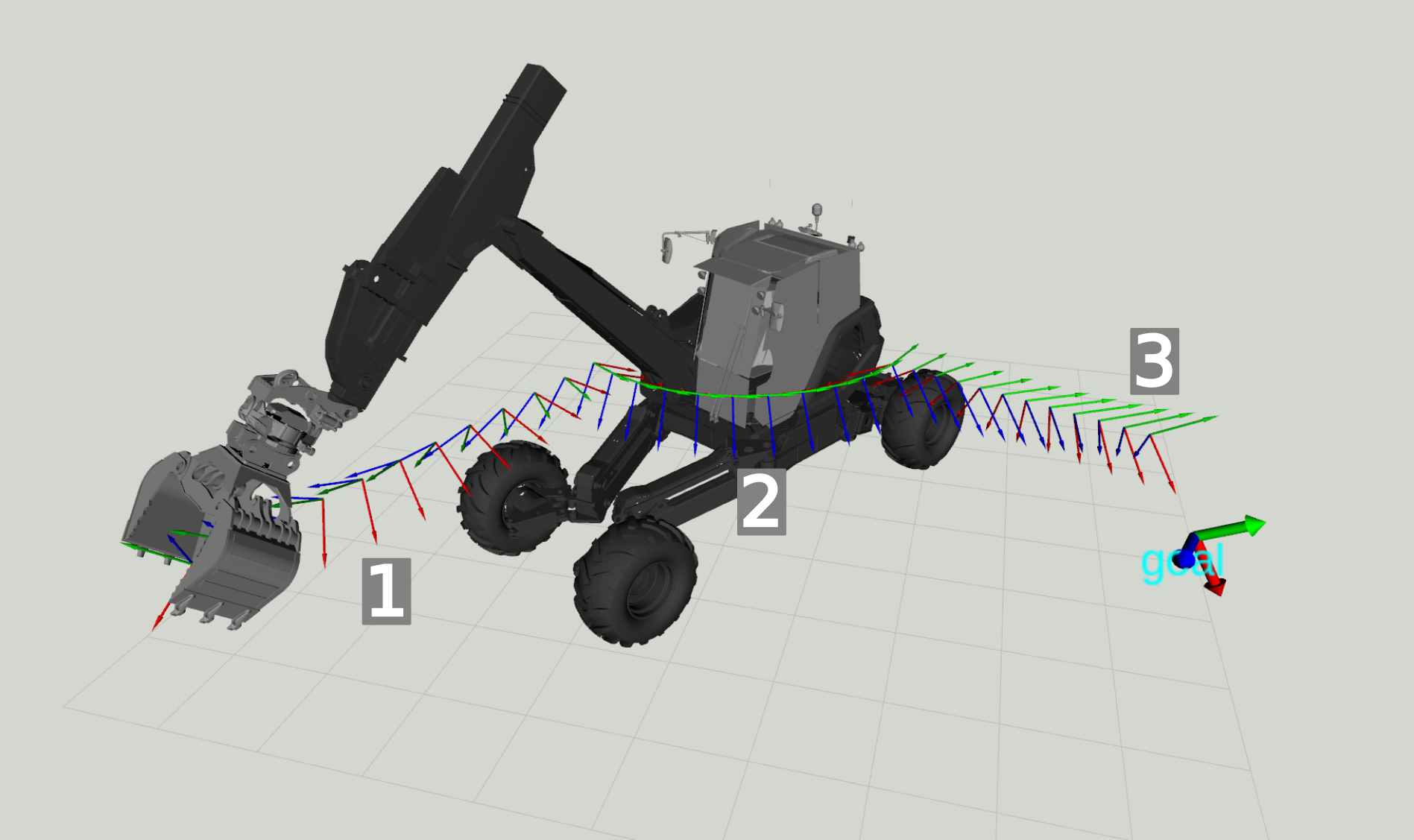}
    \caption{Plan for the arm end effector. Intermediate poses are visualized with coordinate systems. Each axis $(x,y,z)$ corresponds to a different color (red, green, blue). A thick coordinate system and "goal" sign denote the final desired pose. Numbers correspond to different maneuver stages: 1) retract, 2) turn, 3) extend.}
    \label{fig::arm_plan}
\end{figure}

The grasp pose planner receives a tree position from the tree detection subsystem and computes the desired gripper pose. Functionality in this section corresponds to \emph{Approach pose generation} and \emph{Base motion planner} blocks in Fig.~\ref{fig::whole_system}. Since for our demonstration, we use a gripper instead of a standard tree cutting tool (such as \cite{menziMuckHarvester}), we emulate the same behavior by fixing the roll and pitch of the gripper and by computing the yaw angle such that the cabin faces the tree. The kinematic properties of \ac{HEAP} and harvester machines in general with an arm that only moves in a plane allow us to come up with a simple arm planning algorithm. To reach the desired grasp pose, we design a three-stage maneuver that requires minimal space:
\begin{enumerate}
	\item Retract the arm
	\item Turn the cabin
	\item Extend the arm
\end{enumerate}
The approach pose planner (see Sec.~\ref{sec::path_approach_pose_planning}) ensures that there is enough space for the whole arm maneuver. An exemplary arm plan is shown in Fig.~\ref{fig::arm_plan} (stages of the maneuver are indicated with numbers). Intermediate poses (waypoints) are visualized with coordinate systems. We compute Hermite polynomials between the adjacent waypoints to form a trajectory, and we limit the average linear and angular velocity along the spline. The Hermite polynomial tracjecory is then tracked using the inverse kinematics controller, as described in Section~\ref{sec::arm_control}.

For tree felling, the proposed planning method would have to be augmented to consider tree geometry when the harvester is holding one. This could be achieved by adding visual sensors and tracking the falling tree such that the planner can optimize the pulling direction. Besides, one could also estimate the weight and adapt the arm controller tuning or add a feedforward command. The current arm controller is robust concerning weight changes in the gripper. We experimented with stone stacking in \cite{johns2020autonomous} and we were able to manipulate stones between \SI{400}{\kilo \gram} and \SI{2400}{\kilo \gram} (\SI{1040}{\kilo \gram} on average) which is enough to harvest a small tree. Lastly, one could extend approaches such as \cite{song2020time} to prevent the machine from tipping over when holding a tree.
\FloatBarrier
\clearpage
\section{Results}
\label{sec::results}
The presented system components are individually tested and evaluated for complete harvesting missions in the forest and on our testing field. It is worth noting that we implemented the whole system in simulation first to catch as many mistakes as possible before field testing. As the simulation environment, we use \emph{Gazebo} (with \ac{ROS} integration) which is based on the \ac{ODE} physics engine. For visualizing (e.g. trajectories, point clouds) we rely on \emph{Rviz}, a \ac{ROS} tool for visualization.
   
\subsection{Path Tracking}
\label{sec::res_tracking}
\begin{figure*}[tbh]
\centering
    \subfloat[Oberglatt site \label{fig::oberglatt_site}]{
       \includegraphics[width=0.35\textwidth]{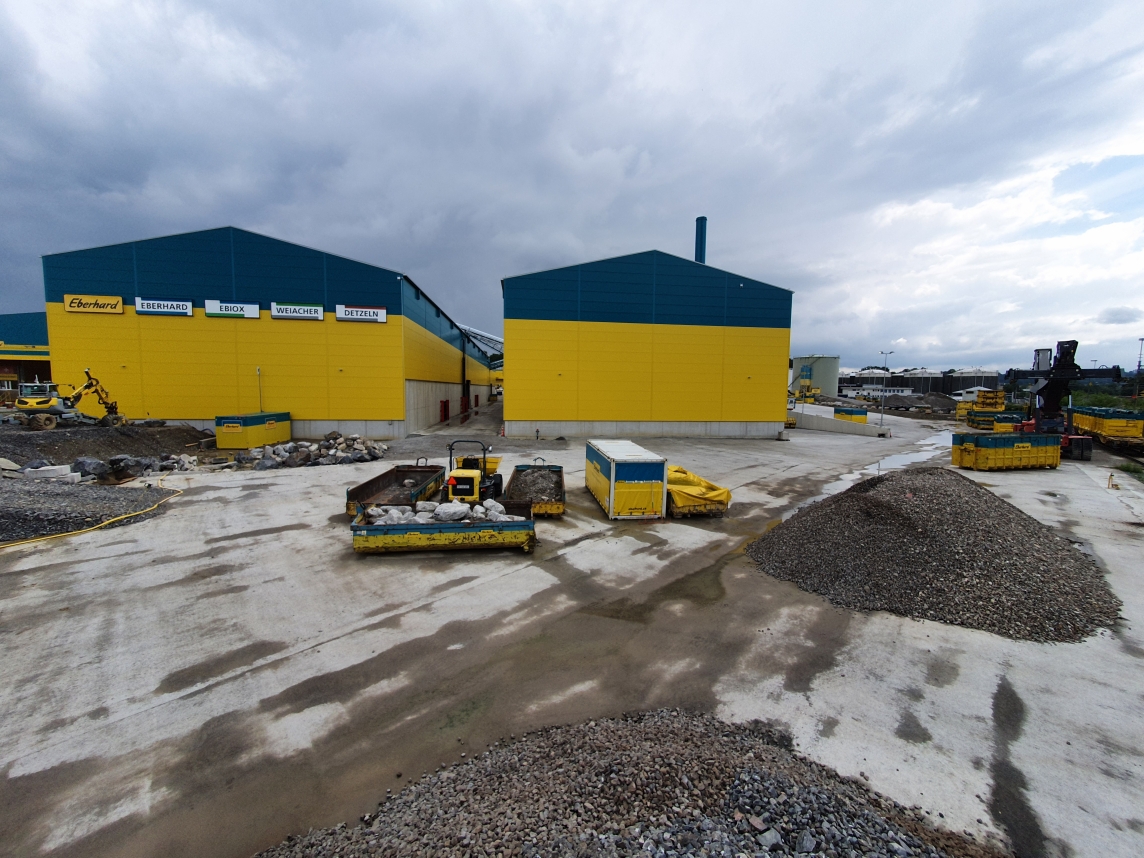}}
    \subfloat[Site map with overlaid paths \label{fig::oberglatt_map_plans}]{
       \includegraphics[width=0.5\textwidth]{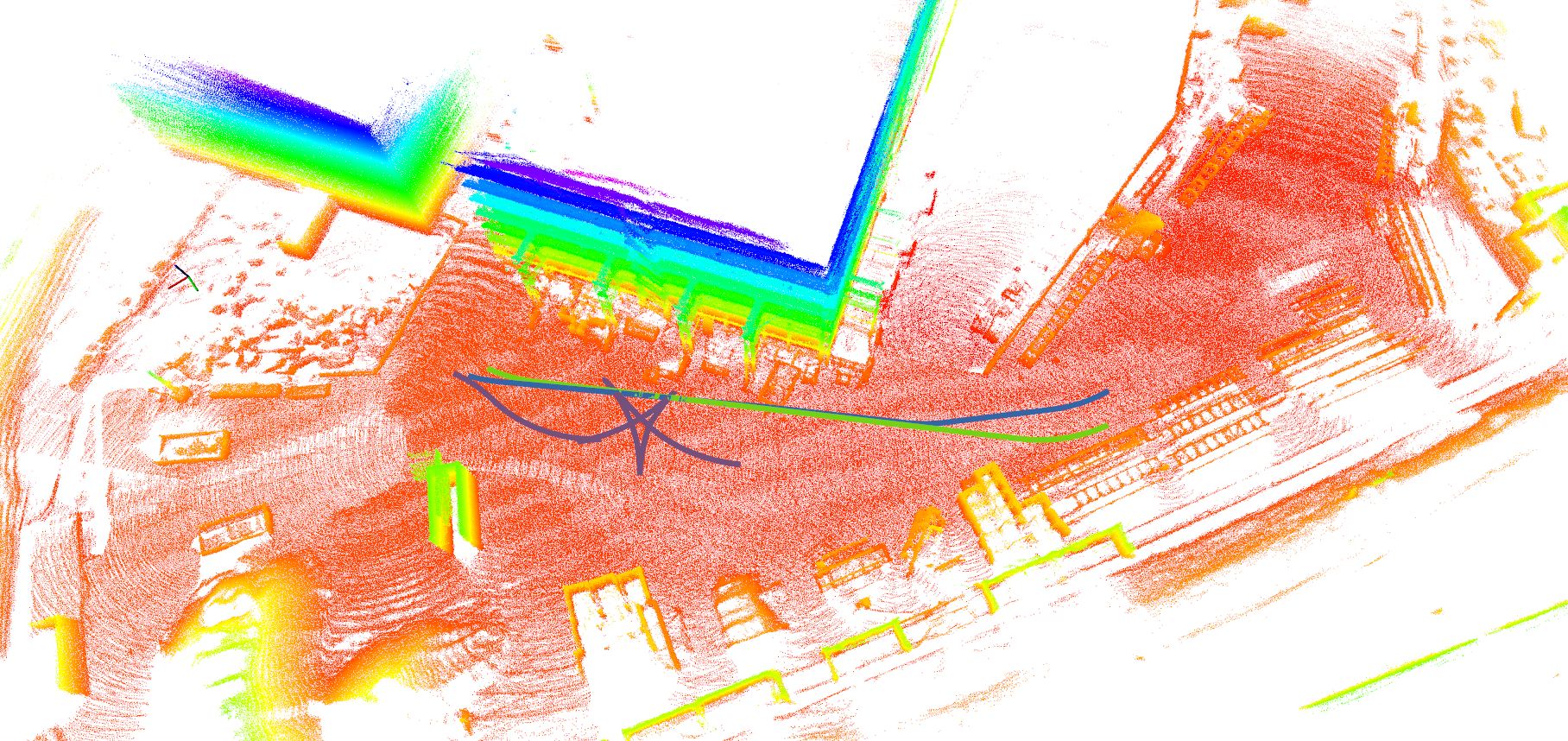}}
    \caption{Testing site where tracking experiments were performed. The image and the map were taken few weeks apart. \textit{\textbf{Left:}} Photo of the testing site where \ac{HEAP} was located at the time of writing this paper. We were allowed to drive on the site to test the tracking controller's performance. \textit{\textbf{Right:}} Map of the testing site with some example plans overlaid. The green path corresponds to the one in Fig.~\ref{fig::tracking_fwd}, the blue path can also be seen in Fig.~\ref{fig::tracking_bck} and the purple one can be seen in Fig.~\ref{fig::tracking_cusps}.}
    \label{fig::oberglatt}
\end{figure*}
\begin{figure*}[tbh]
\centering
    \subfloat[Forward driving \label{fig::tracking_fwd}]{
       \includegraphics[width=0.33\textwidth]{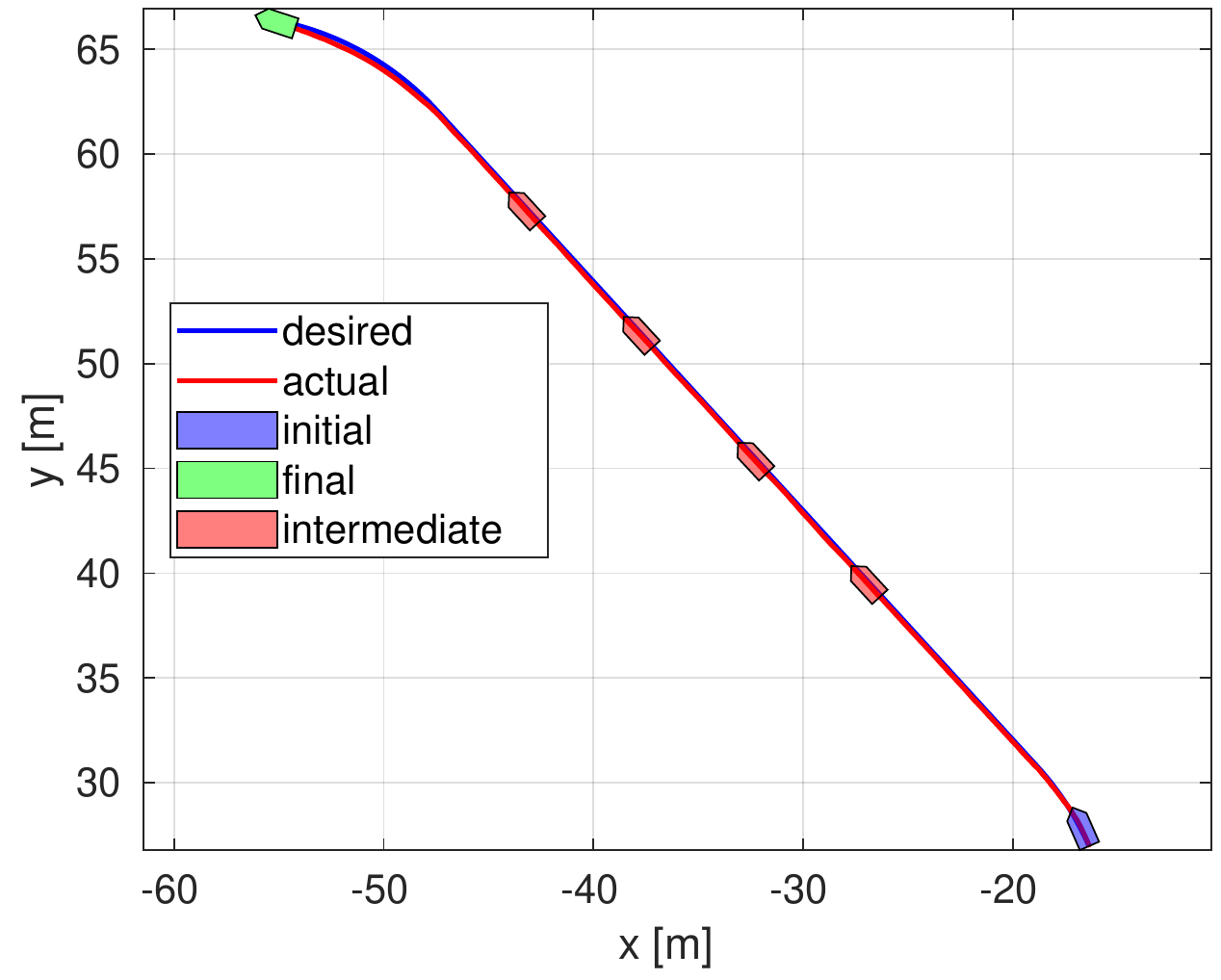}}
     \subfloat[Reverse driving \label{fig::tracking_bck}]{
       \includegraphics[width=0.33\textwidth]{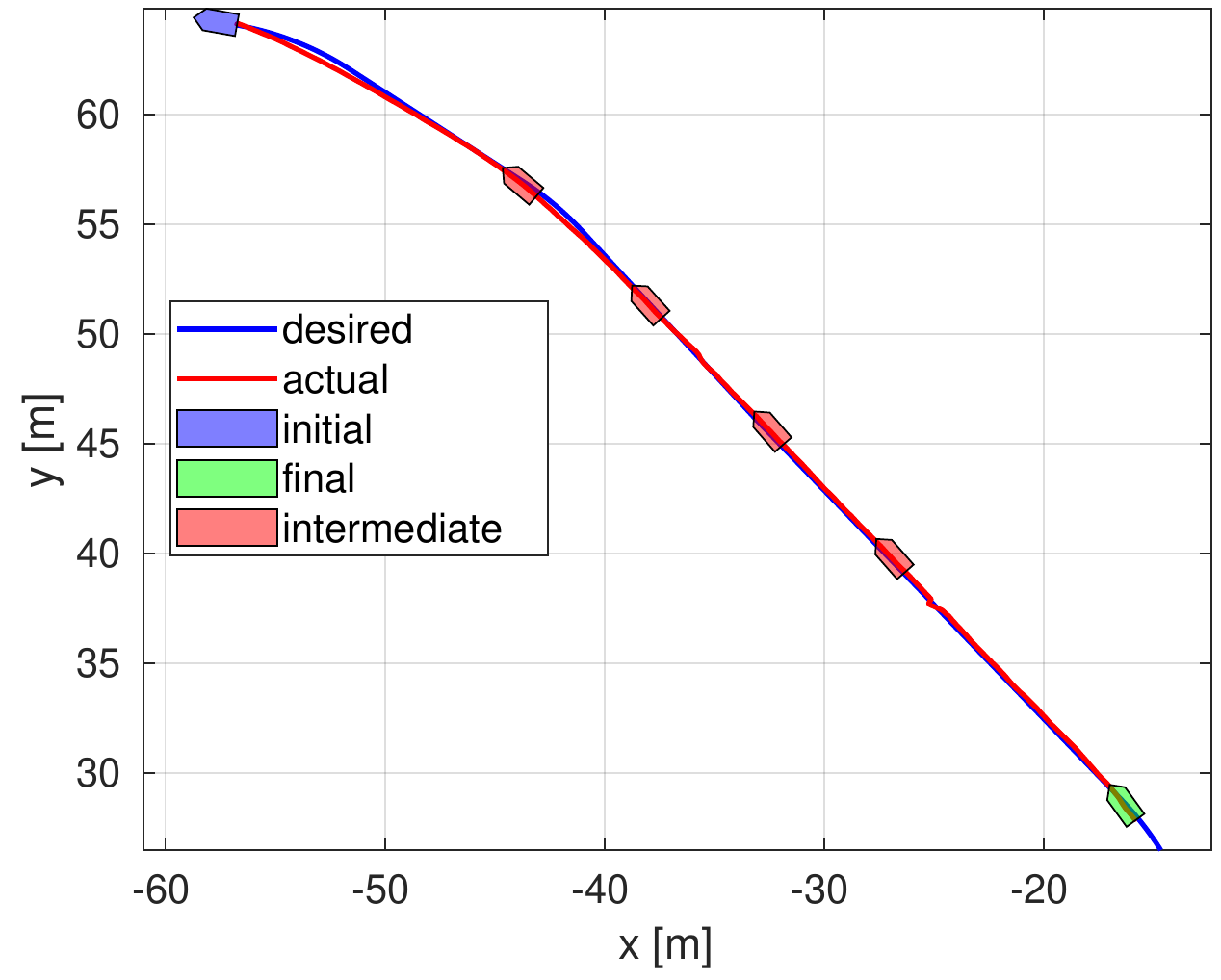}}
     \subfloat[Tight maneuvering \label{fig::tracking_cusps}]{
       \includegraphics[width=0.33\textwidth]{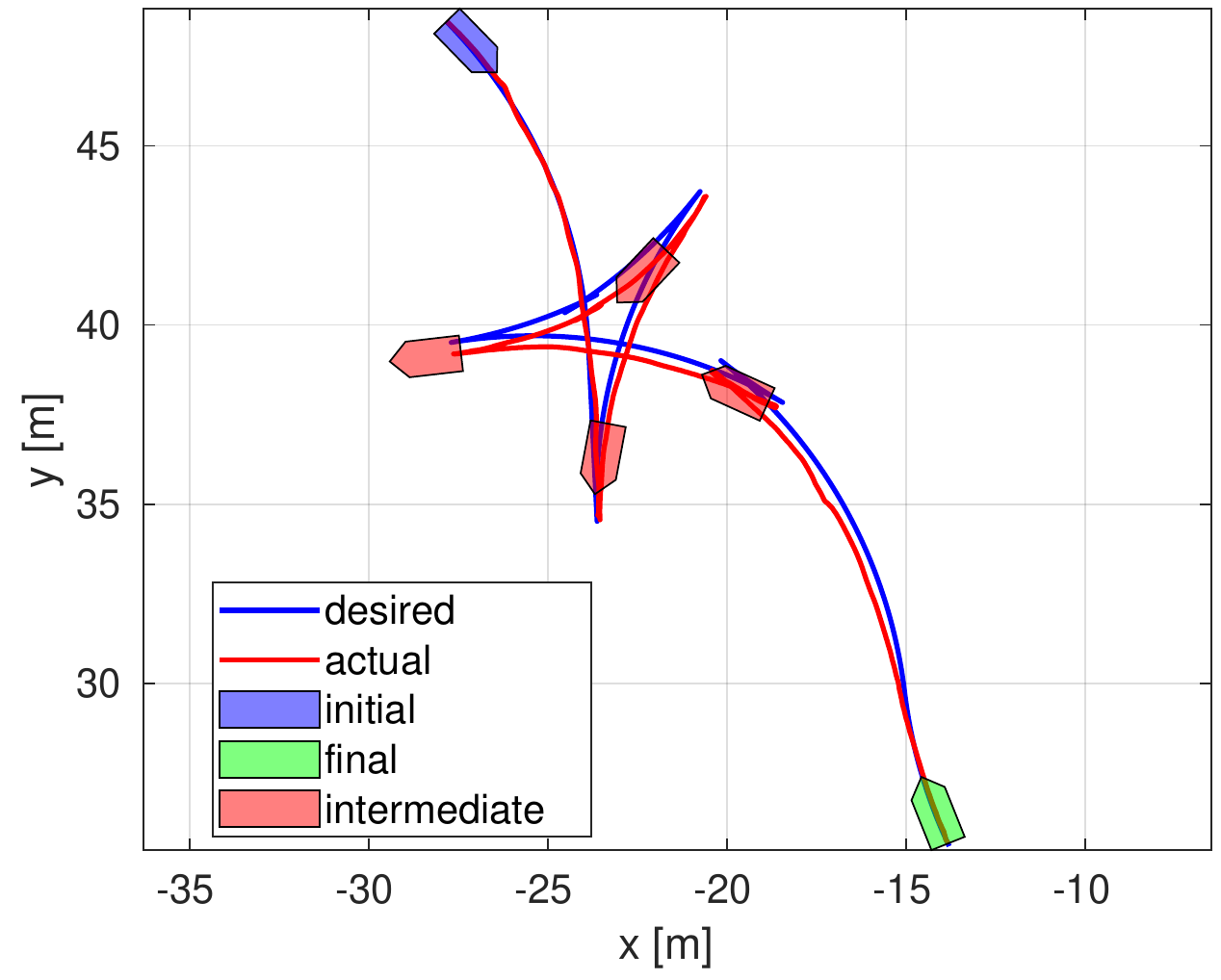}}
    \caption{Example trajectories from tracking accuracy evaluation. Blue color shows output from the planner; red color denotes the actual tracked path. \ac{HEAP}'s initial orientation is shown in blue color and the final one in green. The size of the vehicle is not drawn to scale. \textit{\textbf{Left:}} \ac{HEAP} achieved \SI{9}{\centi \meter} average tracking error over \SI{55}{\meter} of forward driving distance. \textit{\textbf{Middle:}} \ac{HEAP} achieved \SI{13}{\centi \meter} average error over \SI{57}{\meter} of backward driving distance. \textit{\textbf{Right:}} \ac{HEAP} achieved an average error of \SI{18}{\centi \meter} over \SI{62}{\meter} distance for tight maneuvering scenario. In this particular example, \ac{HEAP} does 7 cusps (changes of driving direction).}
    \label{fig::tracking_experiment_oberglatt}
\end{figure*}

We evaluated the tracking performance of the proposed control approach in Oberglatt (Switzerland), where \ac{HEAP} was located for another project at the time of writing this paper. The site is shown Fig.~\ref{fig::oberglatt}; it is a construction site with a mix of concrete and gravel surfaces. We evaluate the tracking performance using RTK \ac{GPS} to measure \ac{HEAP}'s position accurately. \ac{HEAP} was asked to track three different types of paths which consisted of: mostly forward driving, mostly backward driving, and tight maneuvering. An example of the forward driving path is shown in Fig.~\ref{fig::tracking_fwd}, the backward driving path is shown in Fig.~\ref{fig::tracking_bck} and tight maneuvering path is shown in Fig.~\ref{fig::tracking_cusps}. The maneuvering scenarios require the machine to change its orientation in a tight space. Hence, the planner comes up with paths containing cusps and tight turns; we asked \ac{HEAP} to change its orientation (heading) for either $90^{\circ}$ or $180^{\circ}$.

Our controller only tracks the path in $x-y$ coordinates, and hence, all the errors are computed in $x-y$ coordinates. The controller does not track the $z$ coordinate since the planner plans in \emph{SE(2)}. The tracking performance was evaluated over about 30 trials and about \SI{1.5}{\kilo \meter} of driving. We summarize the tracking performance in Table~\ref{table::tracking_eval}. Across all trials, the path tracking algorithm achieves the tracking error of about \SI{17}{\centi \meter}.  We note that the tracking errors achieved in the Oberglatt site are somewhat optimistic since tests took place on flat surfaces that have good traction. It is to be expected that the tracking performance deteriorates as the surfaces get more slippery. This effect we illustrate in a set of experiments conducted on our test field.
\begin{table}[tbh]
\centering
\caption{Path tracking performance evaluation in Oberglatt site. All the errors are transitional errors in $x-y$ coordinates. \emph{Avg length fwd} denotes the average distance traveled forward during the maneuver (analogously for the backward distance, \emph{avg length bck}). \emph{Traveled total} is the cumulative distance traveled across all trials.}
\label{table::tracking_eval}
\begin{tabular}{|l|l|l|l|l|l|l|l|}
\hline
scenario &
  \begin{tabular}[c]{@{}l@{}}num \\ trials\end{tabular} &
  \begin{tabular}[c]{@{}l@{}}avg track \\ error {[}m{]}\end{tabular} &
  \begin{tabular}[c]{@{}l@{}}standard\\ deviation {[}m{]}\end{tabular} &
  \begin{tabular}[c]{@{}l@{}}avg length\\ fwd {[}m{]}\end{tabular} &
  \begin{tabular}[c]{@{}l@{}}avg length \\ bck {[}m{]}\end{tabular} &
  \begin{tabular}[c]{@{}l@{}}avg num\\ cusps\end{tabular} &
  \begin{tabular}[c]{@{}l@{}}traveled \\ total {[}m{]}\end{tabular} \\ \hline
fwd driving   & 10 & 0.23 & 0.04 & 56.05 & 0.52  & 0.6  & 565.84 \\
bck driving & 8  & 0.16 & 0.04 & 0.76  & 55.93 & 0.75 & 453.56 \\
maneuvering     & 14 & 0.13 & 0.03 & 17.99 & 17.44 & 2.93 & 496.19 \\ \hline \hline
total     & 32 & 0.17 & 0.02 & 24.93 & 24.63 & 1.43 & 1515.59 \\ \hline
\end{tabular}
\end{table}
\begin{figure*}[tbh]
\centering
    \subfloat[test field SE view \label{fig::field_and_mahicne_close}]{
       \includegraphics[width=0.45\textwidth]{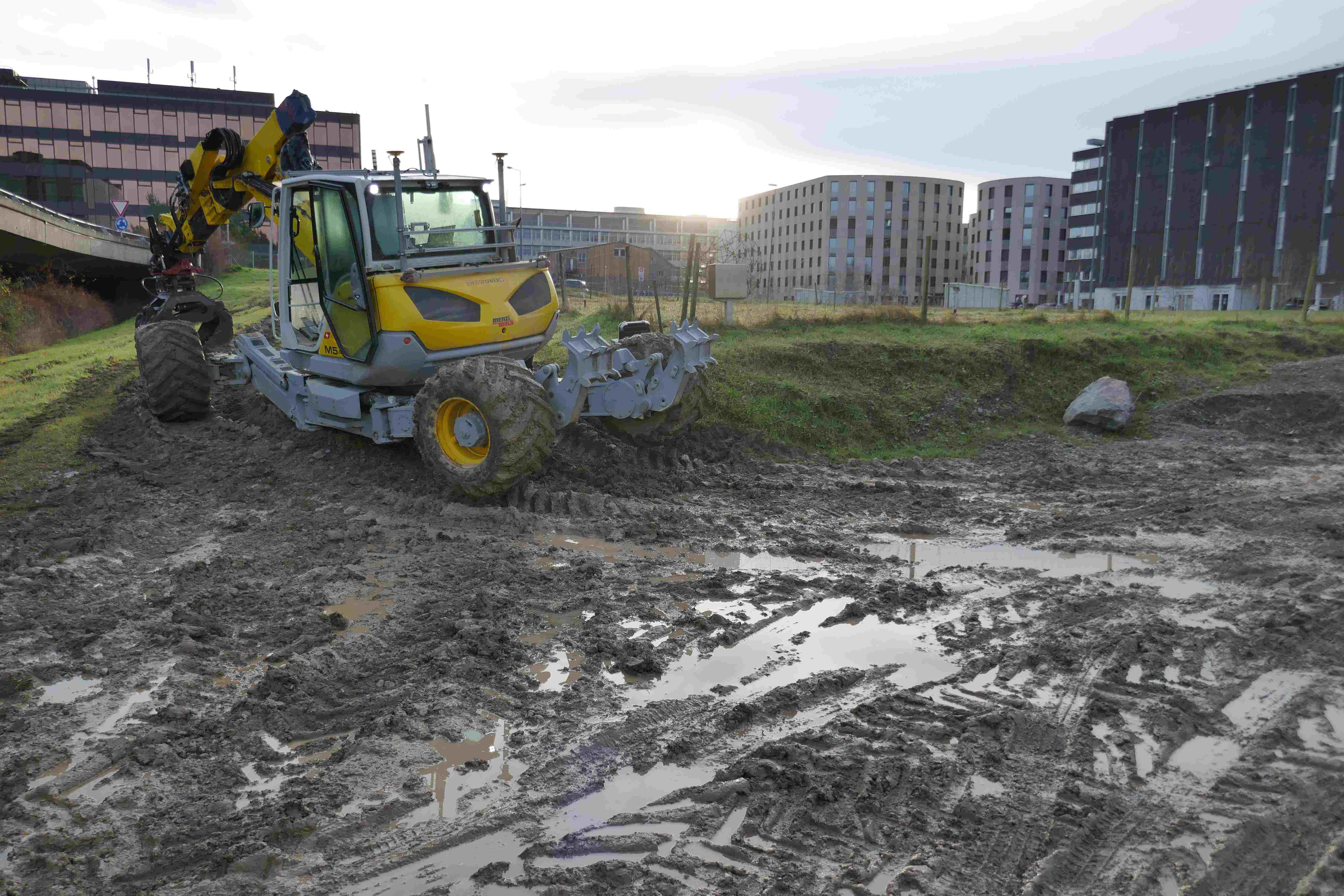}}
    \subfloat[aerial view \label{fig::field_aerial}]{
       \includegraphics[width=0.487\textwidth]{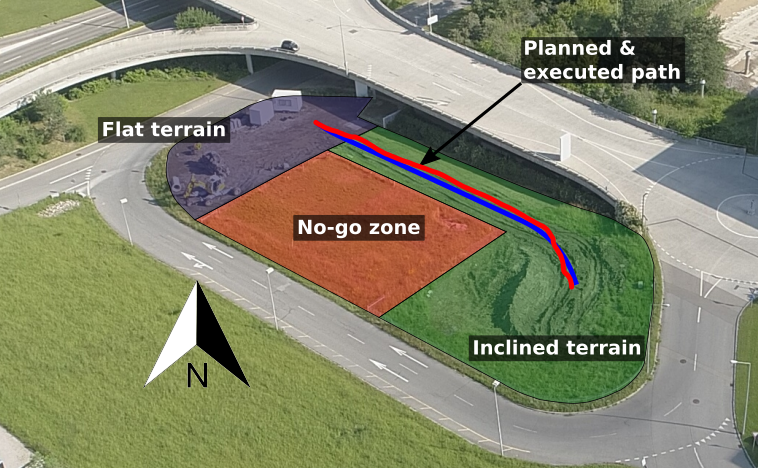}}
    \caption{Photos of our testing field, not all images were taken at the same time. \textit{\textbf{Left:}} View looking southeast with the machine parked. The area shown is flat.  \textit{\textbf{Right:}} Aerial view of the field with different classes of terrain labeled. We overlay the planned (blue) and executed (red) path from one tracking experiment (see Fig.~\ref{fig::tracking_run2}).}
    \label{fig::testing_filed}
\end{figure*}
\begin{figure*}[tbh]
\centering
    \subfloat[Forward on lateral slope \label{fig::tracking_run2}]{
       \includegraphics[width=0.33\textwidth]{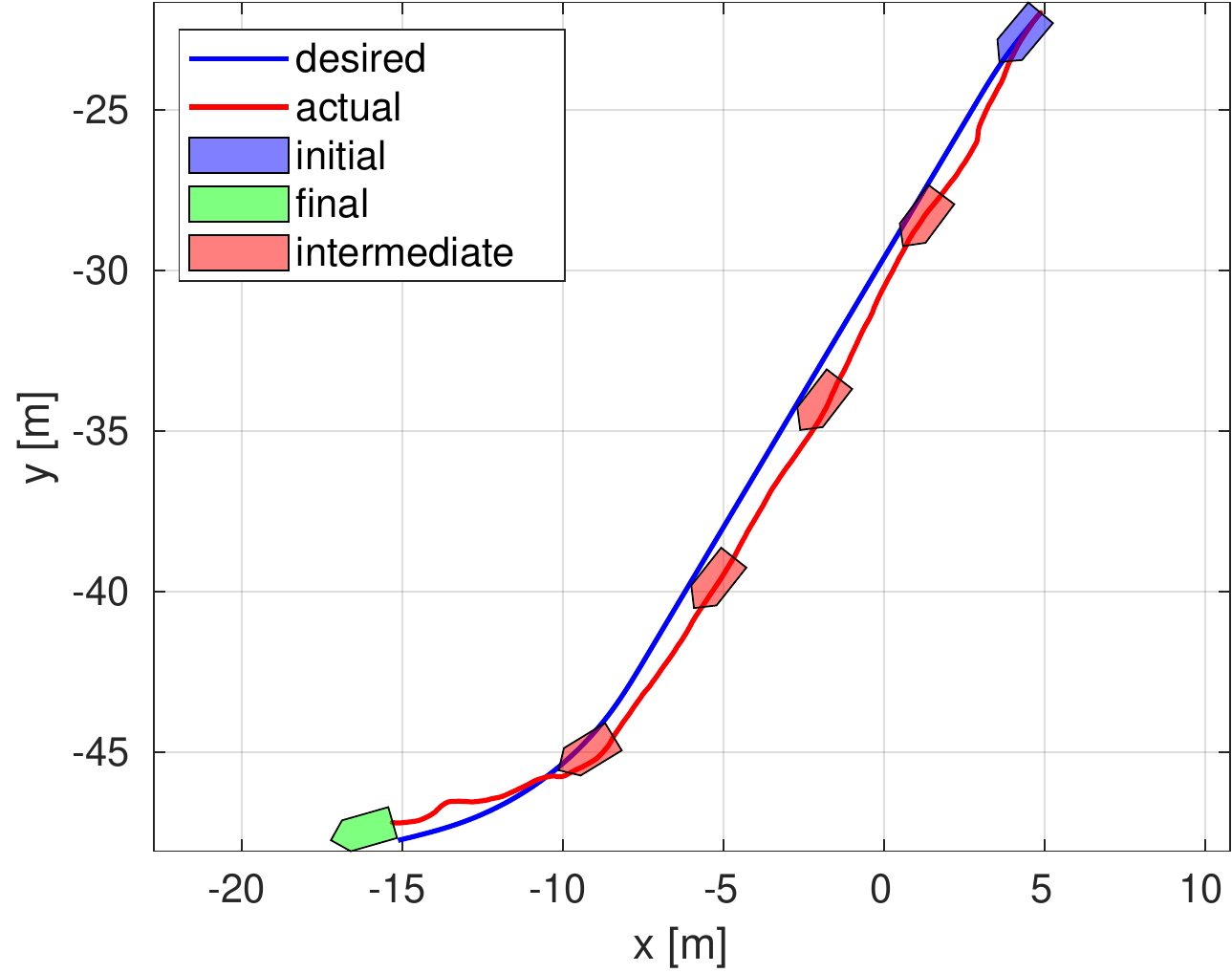}}
     \subfloat[Reverse on lateral slope \label{fig::tracking_run4}]{
       \includegraphics[width=0.33\textwidth]{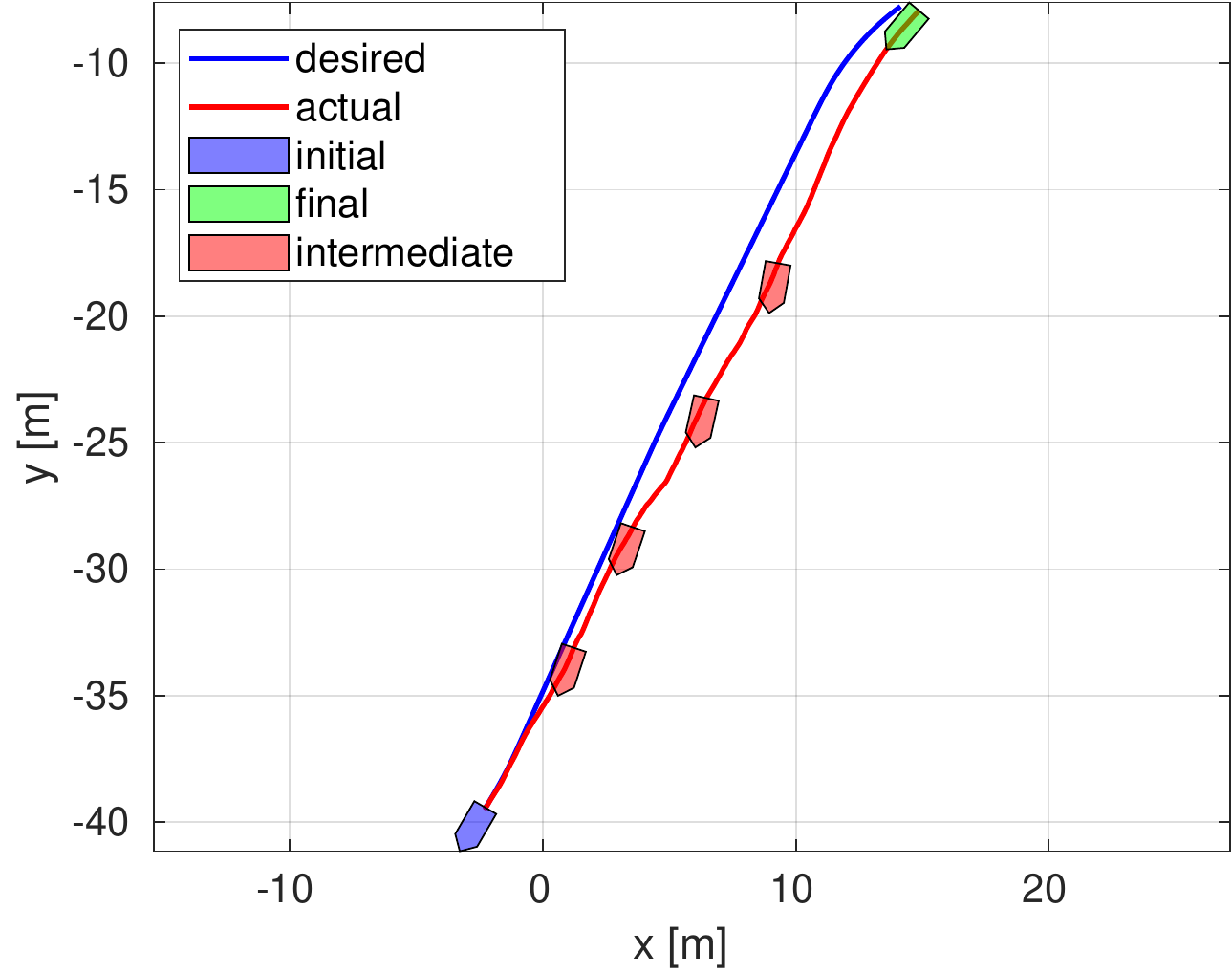}}
     \subfloat[Turns and cups on even ground \label{fig::tracking_run5}]{
       \includegraphics[width=0.33\textwidth]{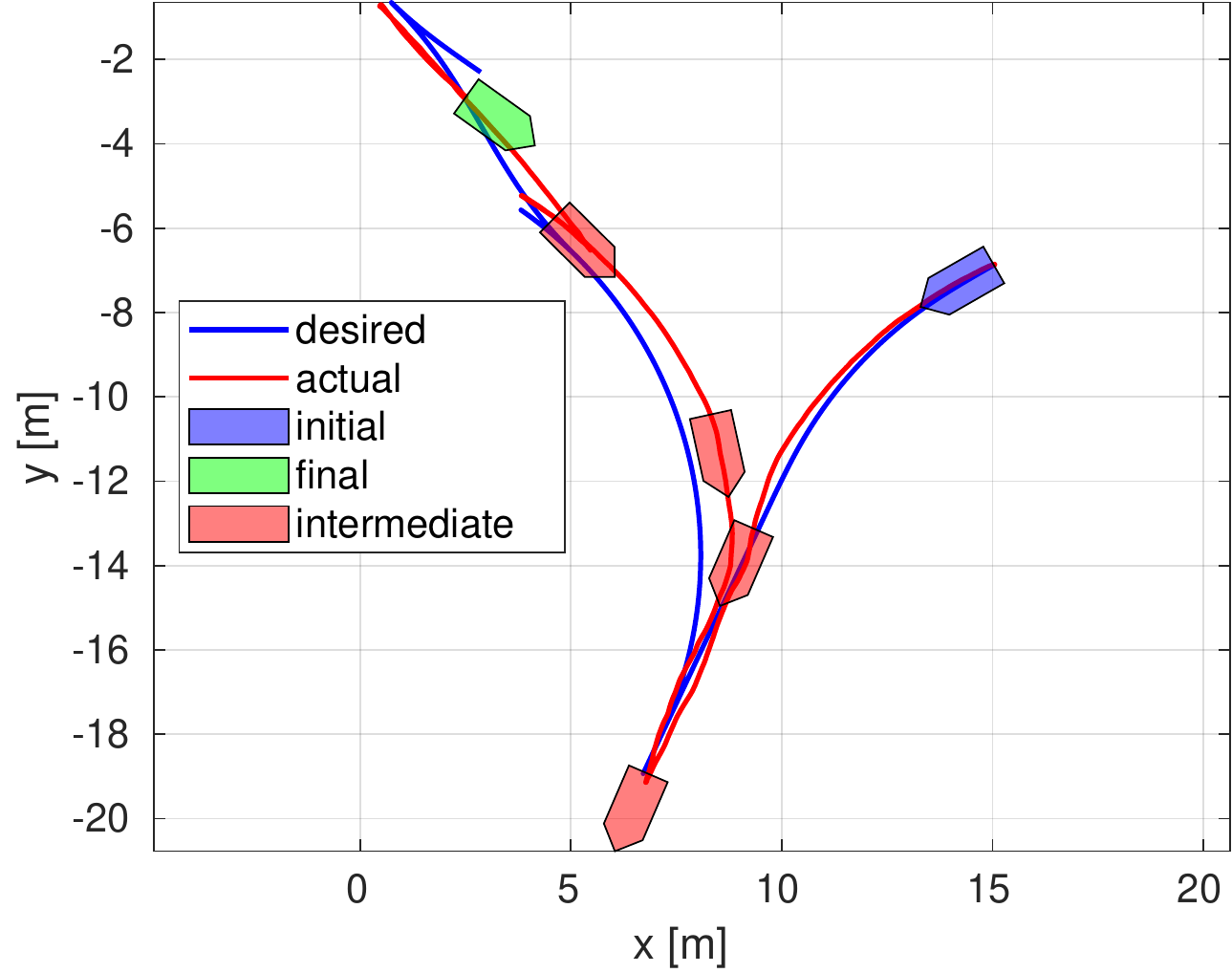}}
    \caption{Experiments were conducted to determine the tracking accuracy of our chassis control approach. In blue color, the planner's desired plan has been shown; red color denotes the actual tracked path. The initial orientation of the excavator is shown in blue color and the final one in red. The size of the vehicle is not drawn to scale.}
    \label{fig::tracking_experiment}
\end{figure*}

The testing field is shown in Fig.~\ref{fig::testing_filed}. \ac{HEAP} was commanded to follow three different paths shown in Fig.~\ref{fig::tracking_experiment}. Fig.~\ref{fig::tracking_run2} and Fig.~\ref{fig::tracking_run4} show tracking a long path on an inclined terrain (see Fig.~\ref{fig::field_aerial}). A combination of inclined and wet terrain caused the \ac{HEAP} to slide sideways. This can also be observed in the path visualization: the heading is not tangential to the path. The machine is pointing towards the slope to compensate for sliding to the side.  In Fig.~\ref{fig::tracking_run5} we asked \ac{HEAP} to reorient itself on the flat part of the testing field. 

The chassis controller was able to achieve mean absolute tracking error of \SI{0.461}{\meter} over \SI{34.1}{\meter} distance for the run shown in Fig.~\ref{fig::tracking_run2} and for the run shown in Fig.~\ref{fig::tracking_run4} the error was \SI{0.684}{\meter} over distance of \SI{44.28}{\meter}. While performing reorientation on the flat terrain, the mean absolute error was \SI{0.222}{\meter} over distance traveled of \SI{44.48}{\meter} (with all direction changes). One can observe that inclined terrain presents a bigger challenge for the tracking controller than direction changes. We note that the test field was fairly wet during the experiments, which negatively affected the amount of traction available. Forestry operations are typically performed when the ground is either dry or frozen, which reduces these slipping issues. Hence, the set of results from our test field presents a hard case scenario which is not often encountered in regular harvesting operations.

\subsection{Planning}
\label{sec::res_planning}
We evaluate the approach pose planning pipeline proposed in Sec.~\ref{sec::path_approach_pose_planning} in simulated scenarios and in two experiments on different terrain: test field and forest patch. We ask the planner to compute both path and an approach pose to grab the selected target trees. All the map visualizations in this section have been created using \emph{Rviz}.

\begin{figure*}[tbh]
\centering
    \subfloat[Simulated forest alley for tree density 0.1 \label{fig::forest_alley_sim}]{
       \includegraphics[width=0.62\textwidth]{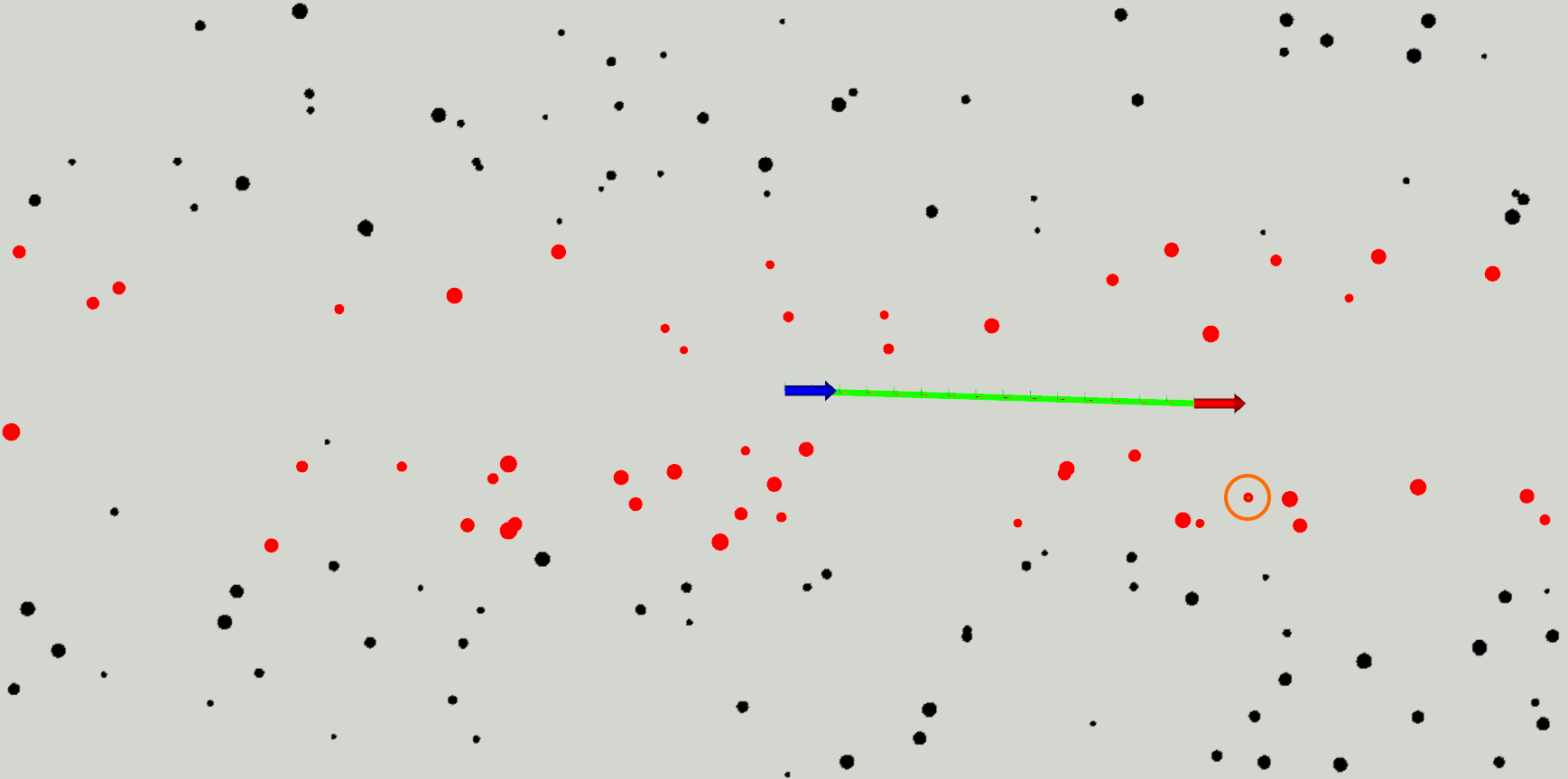}}
    \subfloat[Success rate \label{fig::forest_alley_success}]{
       \includegraphics[width=0.38\textwidth]{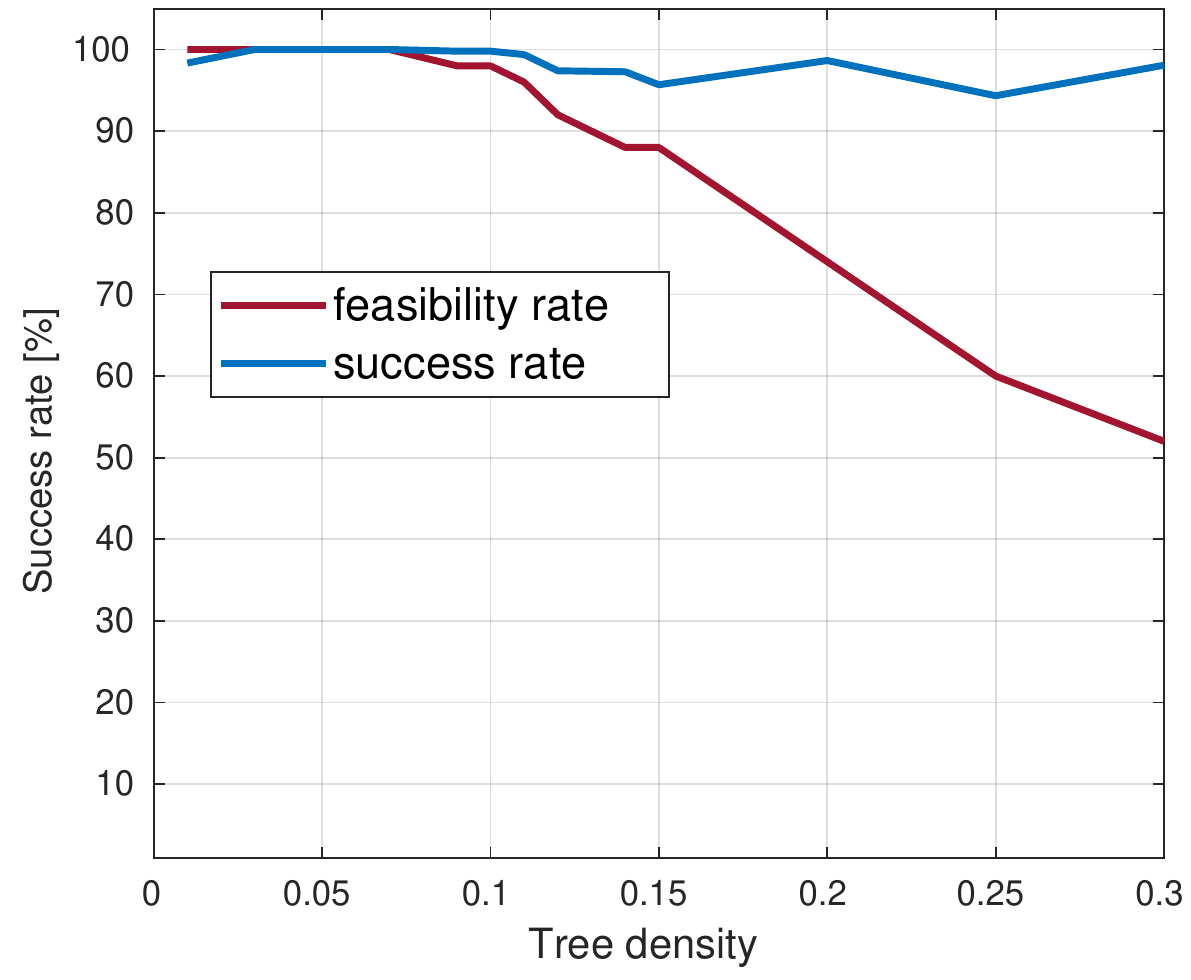}} \\
    \subfloat[Planning times \label{fig::forest_alley_planning_times}]{
       \includegraphics[width=0.5\textwidth]{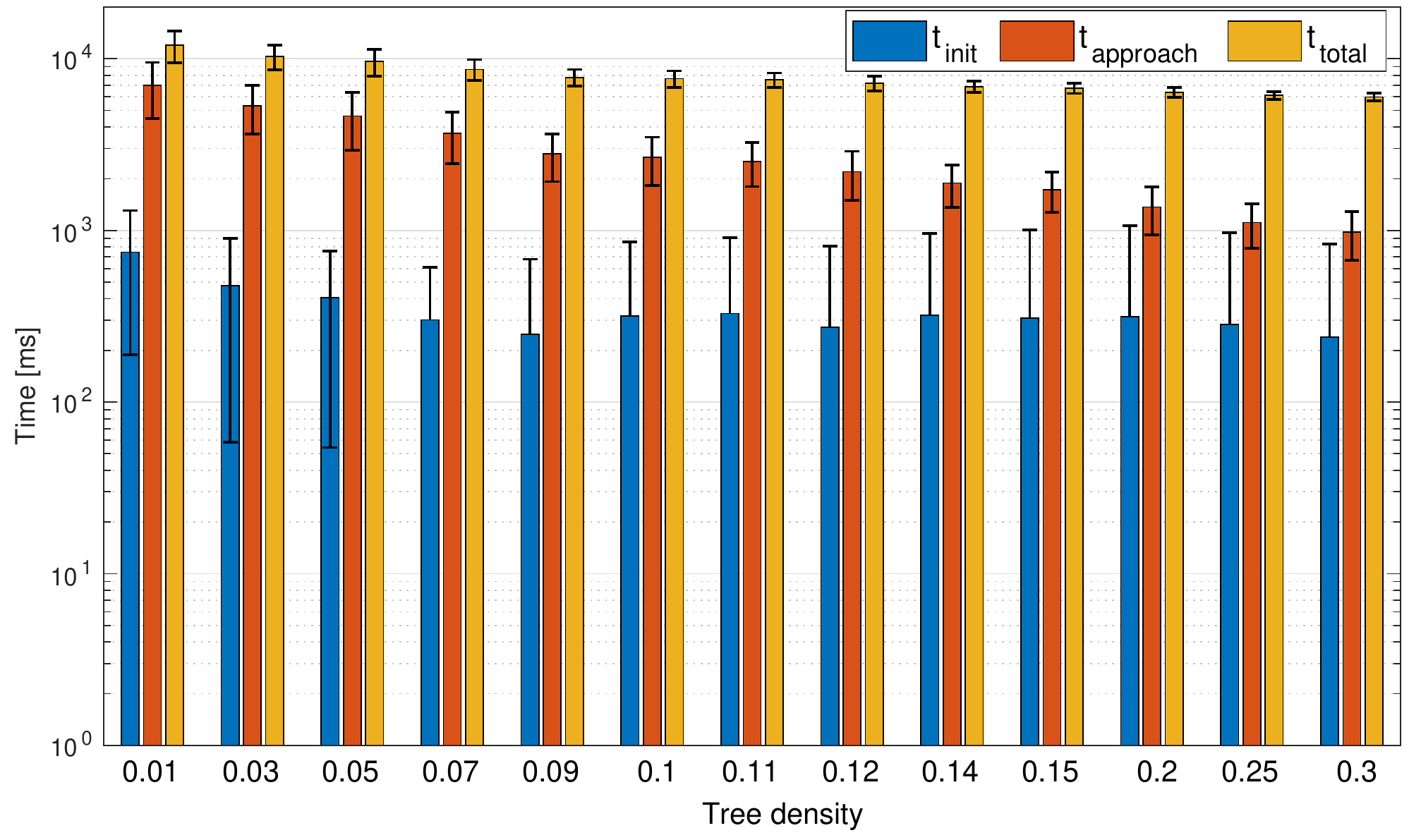}
     }
    \subfloat[Path lengths \label{fig::forest_alley_plan_lengths}]{
       \includegraphics[width=0.5\textwidth]{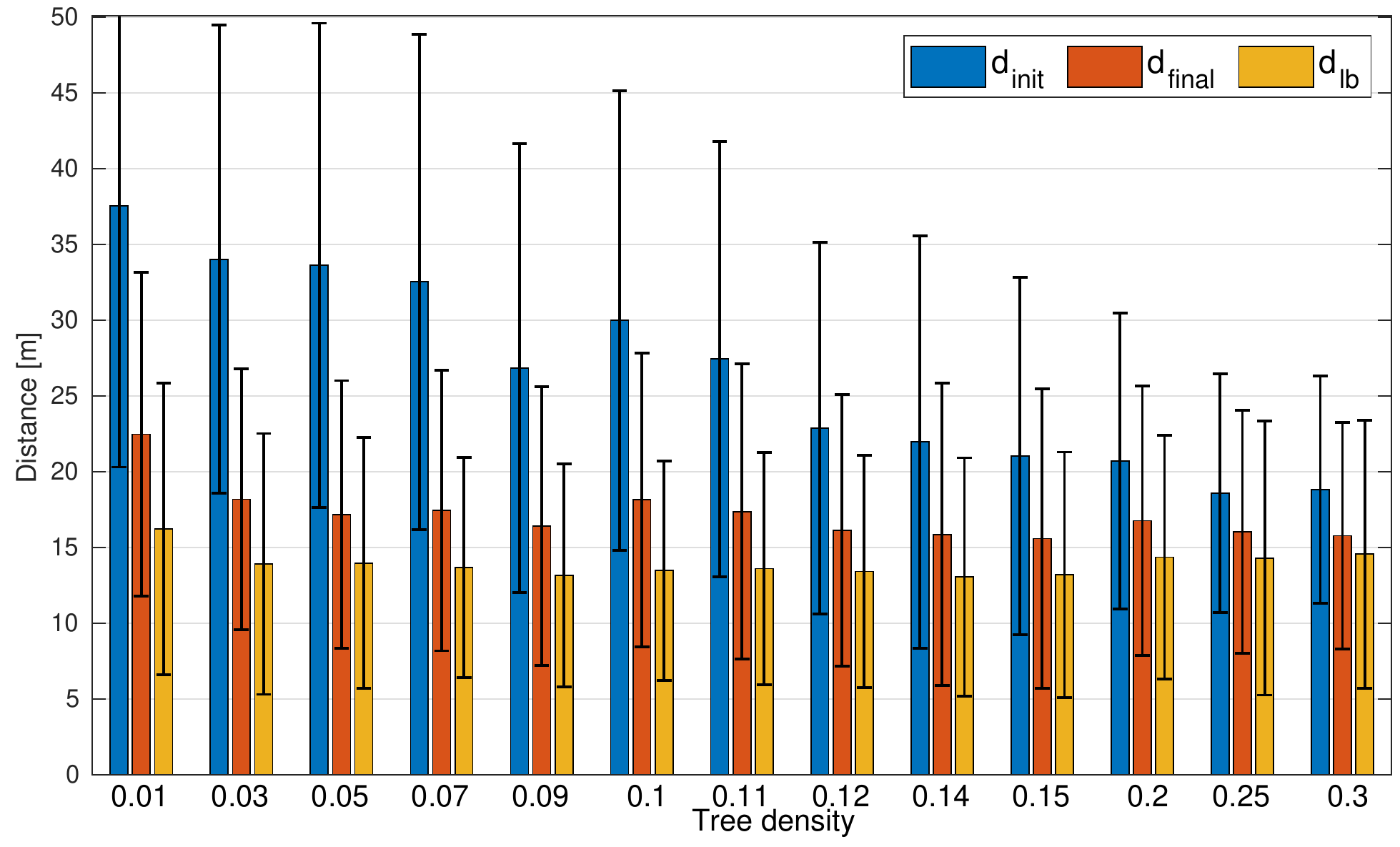}
     }
    \caption{Joint path and approach pose planning for the forest alley scenario. Tree density is defined as a number of trees per $m^2$. \textbf{\textit{Top Left:}} Top view of the random forest with the path and approach pose planned. The harvester starts from the blue arrow to grab the tree encircled orange. Path is shown in green and the final approach pose in red. Black dots represent the trees and red dots are the remaining target trees.  \textbf{\textit{Top Right:}} Percentage of feasible targets and the success rate as the forest density increases. Success rate is calculated as number of successful planning attempts divided by the number of feasible targets. \textbf{\textit{Bottom Left:}} Planning times against varying forest density, note the logarithmic scale on the $y$ axis \textbf{\textit{Bottom Right:}} Path lengths as a function of varying forest density.}
    \label{fig::forest_alley_scenario}
\end{figure*}

\begin{figure*}[tbh]
\centering
    \subfloat[Simulated unstructured forest for tree density 0.1 \label{fig::forest_unst_sim}]{
       \includegraphics[width=0.62\textwidth]{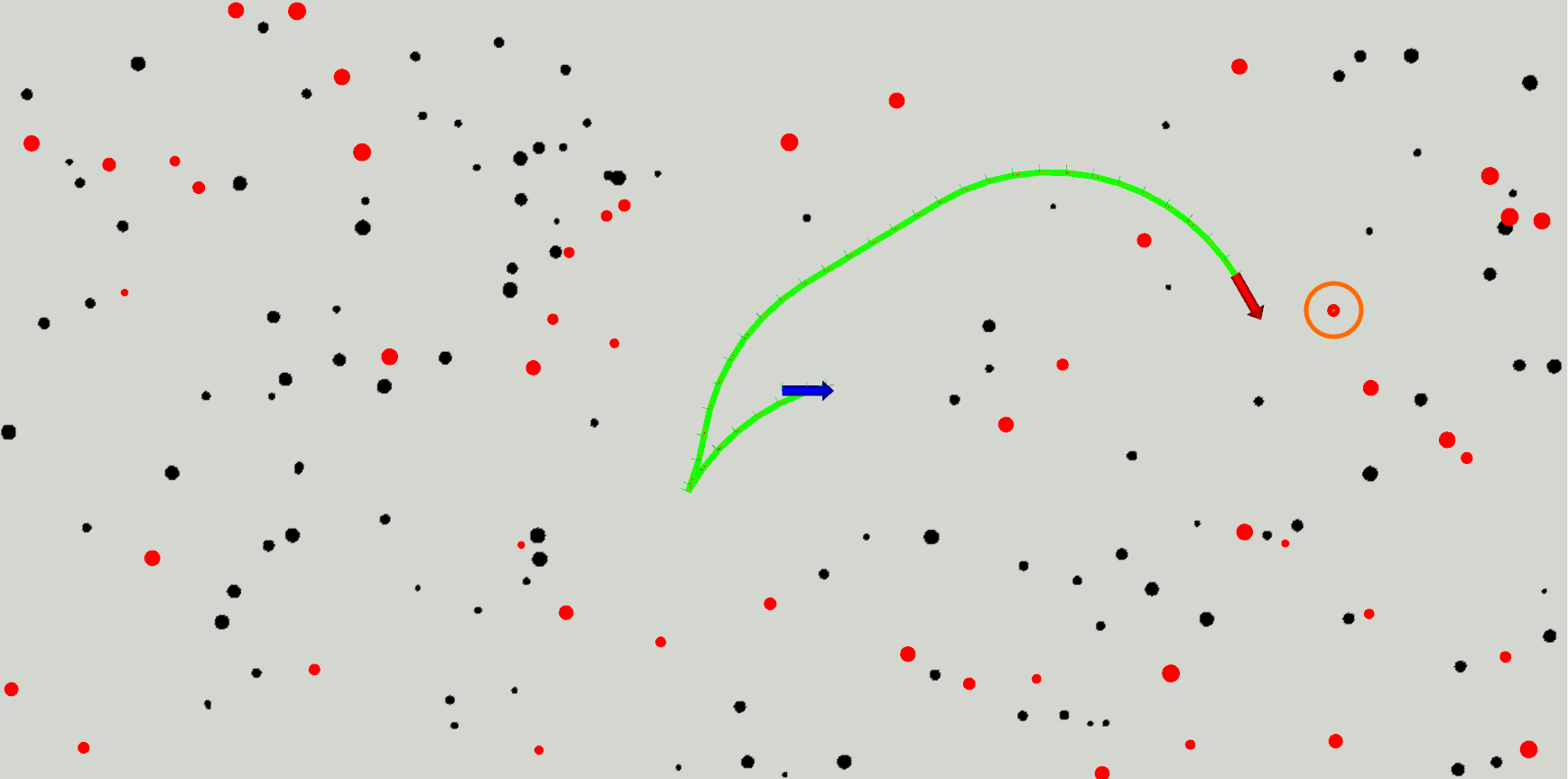}}
    \subfloat[Success rate \label{fig::forest_unst_success}]{
       \includegraphics[width=0.38\textwidth]{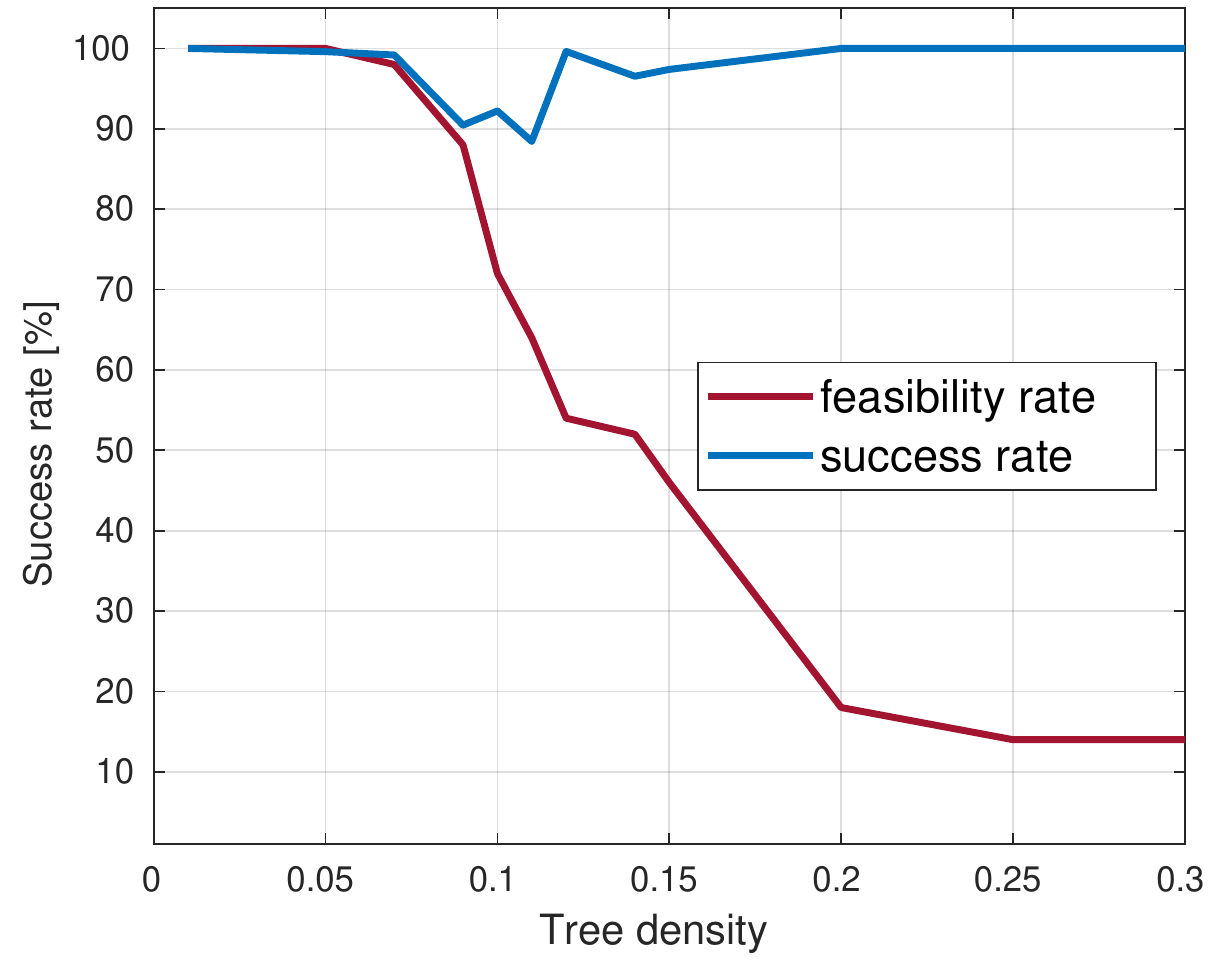}} \\
    \subfloat[Planning times \label{fig::forest_unst_planning_times}]{
       \includegraphics[width=0.5\textwidth]{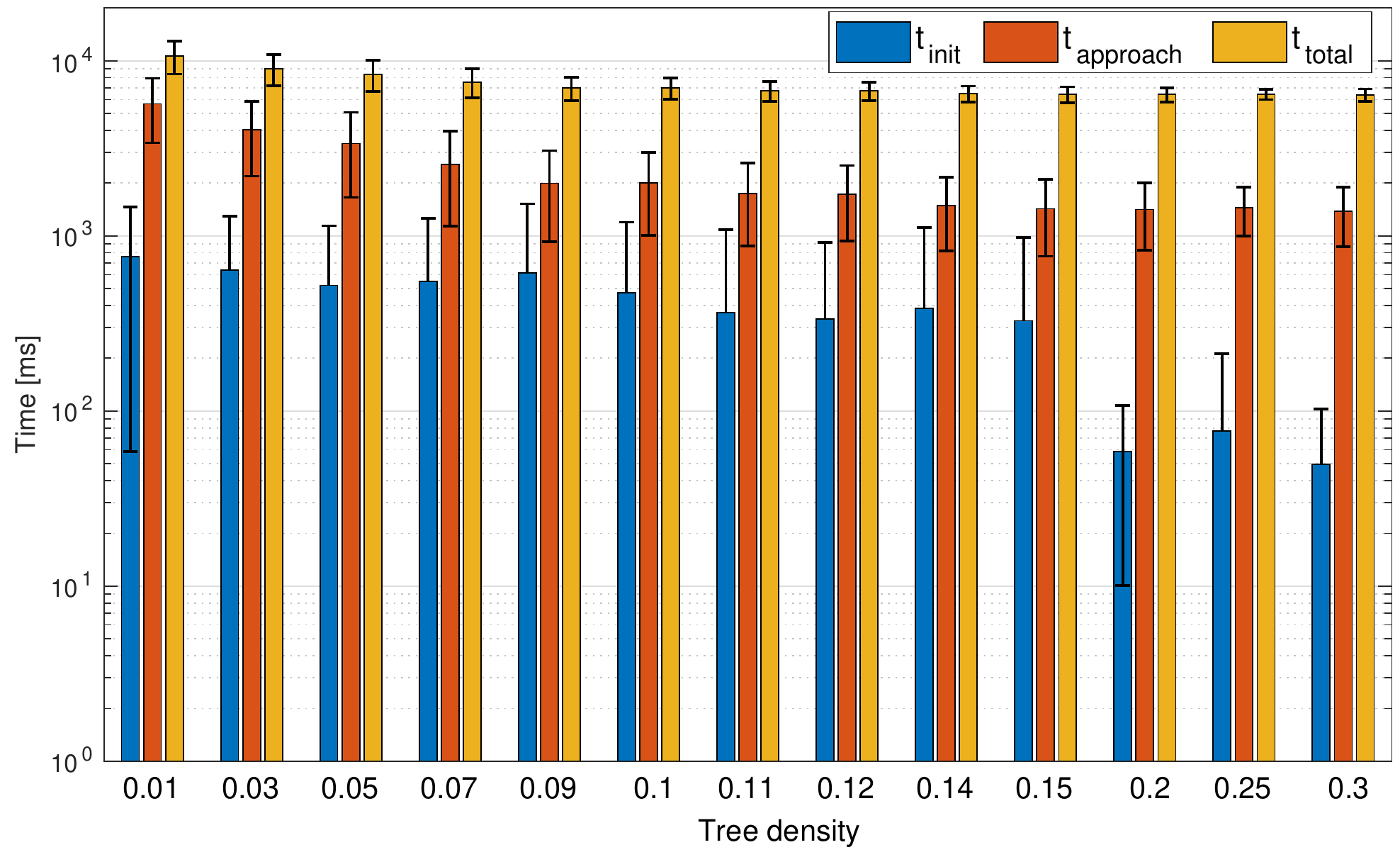}
     }
    \subfloat[Path lengths \label{fig::forest_unst_plan_lengths}]{
       \includegraphics[width=0.5\textwidth]{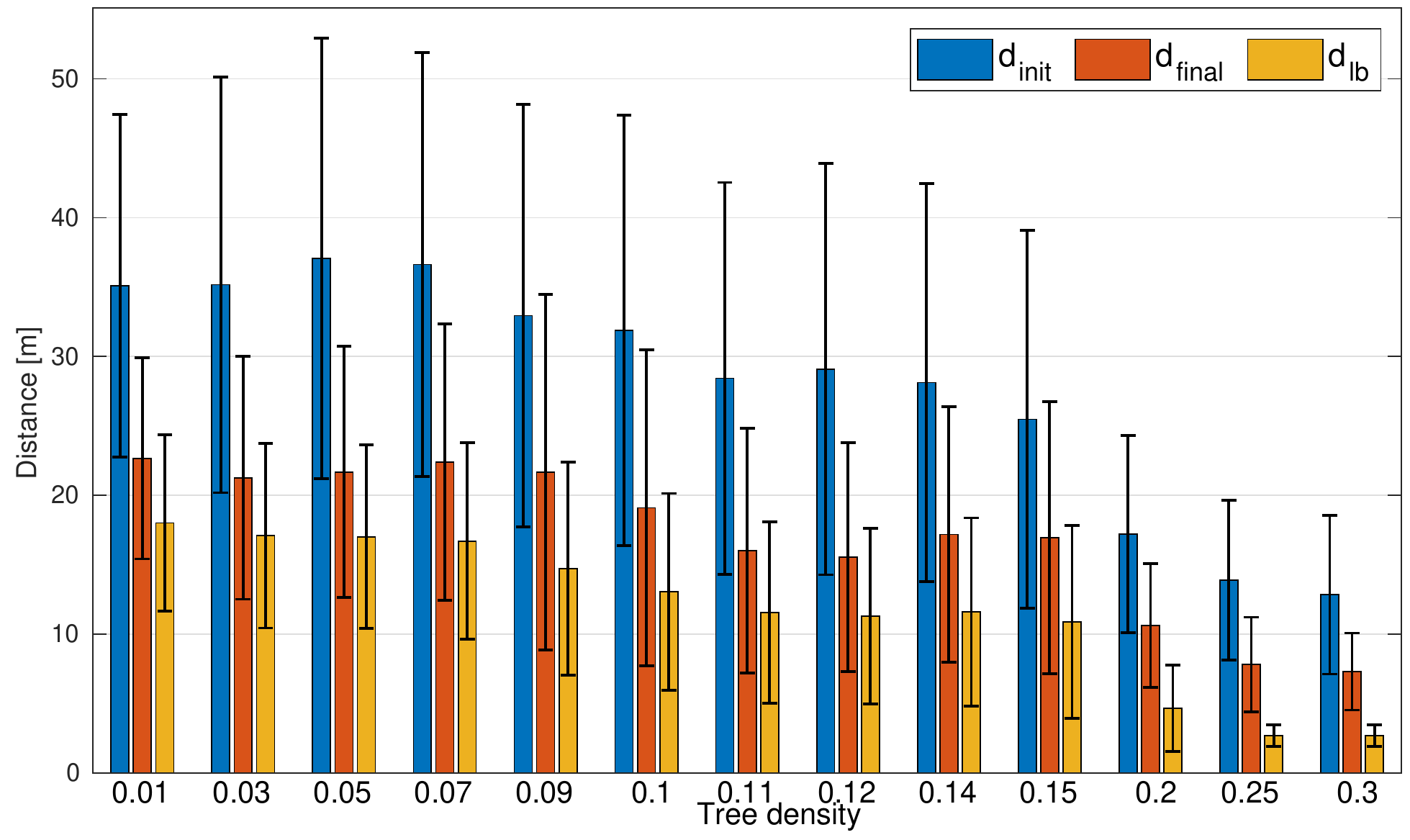}
     }
    \caption{Joint path and approach pose planning for the unstructured forest scenario. \textbf{\textit{Top Left:}} Top view of the random forest with the path and approach pose planned. See Fig.~\ref{fig::forest_alley_scenario} for explanations of the colors in the image. \textbf{\textit{Top Right:}} Success rate as the forest density increases. Feasibility rate and success rate are calculated the same way as for Fig.~\ref{fig::forest_alley_scenario}. \textbf{\textit{Bottom Left:}} Planning times against varying forest density, note the logarithmic scale on the $y$ axis \textbf{\textit{Bottom Right:}} Path lengths as a function of varying forest density.}
    \label{fig::forest_unst_scenario}
\end{figure*}

We created a simulated forest by sampling the number of trees from a Poisson distribution and their positions and radii from a uniform distribution. We consider two scenarios: in the first one, there is a forest alley that can be used for driving (see Fig.~\ref{fig::forest_alley_sim}). This resembles the situation encountered during the field experiments. There is no forest alley in the second scenario, and the harvester has to navigate between the trees (see Fig.~\ref{fig::forest_unst_sim}). Such a scenario is common when using a smaller machine such as one shown in Fig.~\ref{fig::harveri}. For each forest density, we run 10 planning trials and average the metrics. Within one trial, we to compute plans for several target trees; the number of trials is shown in Table~\ref{tab::planner_eval}.

For the forest alley scenario, we ensure the alley at least \SI{2.8}{\meter} wide. For comparison, the harvester path planning footprint has a width of \SI{2.4}{\meter} and the approach pose planning footprint is \SI{4.8}{\meter} wide at its widest point. Note that as the forest gets denser, there might not be enough space to turn the cabin, and therefore no feasible approach poses. We choose several trees (max 50) at random within the \SI{6}{\meter} distance from the forest alley middle (HEAP has a reach of about \SI{8}{\meter}) to be the targets. Black dots represent trees while target trees are colored red in Fig.~\ref{fig::forest_alley_sim}. Each target is deemed feasible if we can find a path to it within \SI{30}{\second} of planning. An example path is shown in green; the starting pose and the planned approach pose are denoted with a blue and red arrow, respectively. In this particular example, the plan requires the harvester to drive forward and turn the cabin to grab the tree encircled orange. Quantitative evaluations of the planner are shown in Fig.~\ref{fig::forest_alley_scenario}, we compute the metrics only for feasible targets.

The percentage of feasible targets (see Fig.~\ref{fig::forest_alley_success}) drops as the forest's density grows. However, one can see that the planner maintains a high success rate. Besides the success rate, we evaluate times required to generate candidate approach poses $t_{approach}$, time until the first solution inside the \ac{RRT}* planner $t_{init}$ and we show the total planning time $t_{total}$ (Fig.~\ref{fig::forest_alley_planning_times}).  One can observe that the \ac{RRT}* planner finds the first solution rather quickly (worst case in \SI{700}{\milli \second}). The most expensive part of the pipeline is the approach pose generation which in the worst case takes about \SI{7}{\second}. Planning times tend to shorten as the forest density (number of trees per $m^2$) increases since many approach poses can be discarded in the early stage of collision checking (early termination). In contrast, for low forest densities, almost all approach pose footprints have to be checked for collisions fully (no early termination). 

For the experiments presented, the planner considers 14060 approach pose candidates in total (no pruning heuristics were applied to showcase the generality of the approach). In this work, we use single-threaded implementation; however, the approach pose generation can be easily paralleled to decrease computation time. Lastly, we measured the distances between the starting pose and the planned approach poses (see Fig.~\ref{fig::forest_alley_plan_lengths}). We show the length of the first found path $d_{init}$ and the length of the optimized path $d_{final}$ (after running \ac{RRT}* for \SI{5}{\second}). Lastly, the graph shows Euclidean distance between start and finish  $d_{lb}$ which is a lower bound on the path length. We can see that the planned path is close to the lower bound in all cases. The paths tend to shorten with the increasing density since more feasible trees lie on the inside of the forest alley. Hence, \ac{HEAP} does not turn to the side, which shortens the overall path length.

\begin{table}[tbh]
\centering
\caption{Number of feasible (attainable) targets together with the total number of planning attempts for each forest density. We show the data for forest alley and unstructured forest scenario.}
\label{tab::planner_eval}
\begin{tabular}{c|c|c|c||c|c|c}
\cline{2-7}
                           & \multicolumn{3}{c||}{forest alley} & \multicolumn{3}{c|}{unstructured  forest} \\ \hline
\multicolumn{1}{|c||}{density} &
  n targets &
  \begin{tabular}[c]{@{}c@{}}n feasible\\ targets\end{tabular} &
  \begin{tabular}[c]{@{}c@{}}n (successful)\\ planning attempts\end{tabular} &
  n targets &
  \begin{tabular}[c]{@{}c@{}}n feasible\\ targets\end{tabular} &
  \multicolumn{1}{c|}{\begin{tabular}[c]{@{}c@{}}n (successful)\\ planning attempts\end{tabular}} \\ \hline \hline
\multicolumn{1}{|c||}{0.01} & 6         & 6         & (59) 60        & 23    & 23   & \multicolumn{1}{c|}{(230) 230}   \\
\multicolumn{1}{|c||}{0.03} & 18        & 18        & (180) 180       & 50    & 50   & \multicolumn{1}{c|}{(499) 500}   \\
\multicolumn{1}{|c||}{0.05} & 23        & 23        & (230) 230       & 50    & 50   & \multicolumn{1}{c|}{(498) 500}   \\
\multicolumn{1}{|c||}{0.07} & 38        & 38        & (380) 380       & 50    & 49   & \multicolumn{1}{c|}{(486) 490}   \\
\multicolumn{1}{|c||}{0.09} & 50        & 49        & (489) 490       & 50    & 44   & \multicolumn{1}{c|}{(398) 440}   \\
\multicolumn{1}{|c||}{0.1}  & 50        & 49        & (489) 490       & 50    & 36   & \multicolumn{1}{c|}{(332) 360}   \\
\multicolumn{1}{|c||}{0.11} & 50        & 48        & (477) 480       & 50    & 32   & \multicolumn{1}{c|}{(283) 320}   \\
\multicolumn{1}{|c||}{0.12} & 50        & 46        & (448) 460       & 50    & 27   & \multicolumn{1}{c|}{(269) 270}   \\
\multicolumn{1}{|c||}{0.14} & 50        & 44        & (428) 440       & 50    & 26   & \multicolumn{1}{c|}{(251) 260}   \\
\multicolumn{1}{|c||}{0.15} & 50        & 44        & (421) 440       & 50    & 23   & \multicolumn{1}{c|}{(224) 230}   \\
\multicolumn{1}{|c||}{0.2}  & 50        & 37        & (365) 370       & 50    & 9    & \multicolumn{1}{c|}{(90) 90}    \\
\multicolumn{1}{|c||}{0.25} & 50        & 30        & (283) 300        & 50    & 7    & \multicolumn{1}{c|}{(70) 70}    \\
\multicolumn{1}{|c||}{0.3}  & 50        & 26        & (255) 260       & 50    & 7    & \multicolumn{1}{c|}{(70) 70}    \\ \hline
\end{tabular}
\end{table}

As a second scenario, we evaluated planning performance while navigating an unstructured forest. The forest was generated using the same probability distributions as in the scenario above. We leave a clear area (\SI{10}{\meter} by \SI{10}{\meter}) in the middle of the map to ensure a feasible starting pose. The target trees were selected at random with a maximum of 50 trees. Again, we check the feasibility for each target by running the planner for \SI{30}{\second} and perform 10 planning trials.

Performance measures shown in Fig.~\ref{fig::forest_unst_success} - Fig.~\ref{fig::forest_unst_plan_lengths} are calculated only for feasible targets. One can observe a much faster drop in the percent of feasible targets (compared to the forest alley scenario). In most cases, our planner finds the solution rather quickly (\SI{800}{\milli \second} in the worst case). We can also notice that the success rate drops compared to the previous scenario since there is no forest alley. The planner has to find a path through narrow passages (hard for sampling-based planners). The forest density of 0.1 seems to be especially hard, i.e. the harvester can still fit through the trees, but just barely. The success rate for densities between 0.05 and 0.15 can be increased with longer planning times. For forest densities of 0.15 and higher the \ac{HEAP} harvester simply dos not fit between the trees. Hence the set of feasible targets comprises the trees around the clear patch in the middle of the simulated forest. In this case, the success rate goes again to \SI{100}{\percent}. 

To evaluate the planning performance under realistic conditions, we planned on a map of our testing field and a map of forest patch where we conducted the tests. The forest patch map with target trees is shown in Fig.~\ref{fig::planning_forest_map}. Black dots represent trees and other obstacles, while target trees are denoted with red dots. We manually selected the trees along the forest alley, discarding any tree without path computed within \SI{30}{\second} of planning. The forest density at the testing site was estimated to be about 0.14 (trees/$m^2$). The planner achieved a success rate of about \SI{98}{\percent} which is almost the same as the simulated scenario (\SI{95}{\percent}). Furthermore, the time $t_{approach}$ (\SI{1625.15}{\milli \second}) is close to the simulated one (\SI{1728.75}{\milli \second}). $t_{init}$ (\SI{120.62}{\milli \second}) is lower than the simulated one (\SI{308.36}{\milli \second}); which would indicate that it has fewer narrow passages than the simulated forest.. On the testing field, the planner achieved the success rate of \SI{100}{\percent} (omitted for the sake of brevity).

\begin{minipage}[tbh]{0.6\textwidth}
    \centering
    \includegraphics[width=\textwidth]{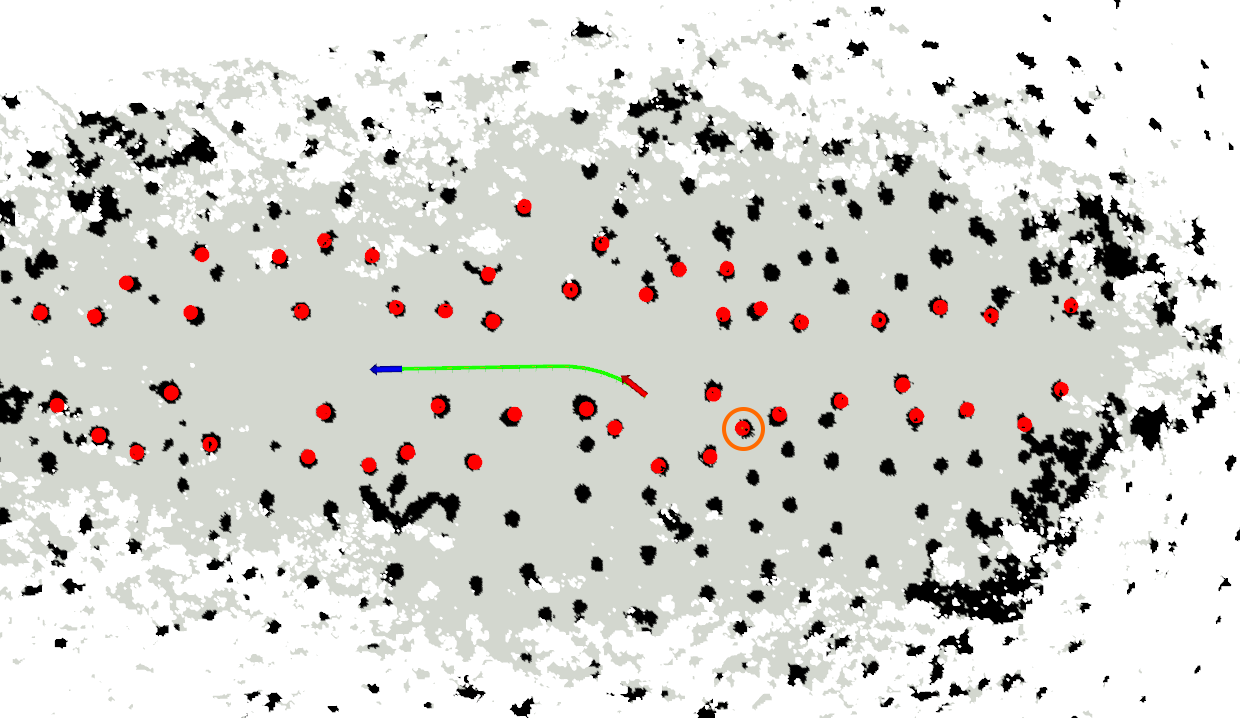}
    \captionof{figure}{Obstacle map for the forest patch scenario. Obstacles are shown in black color whereas free space is shown in gray color. The targets are denoted with red cylinders}
    \label{fig::planning_forest_map}
\end{minipage}
\hspace{0.1cm}
\begin{minipage}[tbh]{0.3\textwidth}
\centering
\captionof{table}{Planning metrics for the forest patch scenario (averaged over 10 trials).}
\begin{tabular}{|l|c|}
\hline
num targets                                                             & 51 \\ \hline
success rate                                                            & 0.98  \\ \hline
\begin{tabular}[c]{@{}l@{}}time until \\ first solution [ms]\end{tabular}    & $(120 \pm 555)$   \\ \hline
\begin{tabular}[c]{@{}l@{}}approach pose\\ generation time [ms]\end{tabular} & $(1625 \pm 621)$   \\ \hline
\begin{tabular}[c]{@{}l@{}}total planning \\ time [s]\end{tabular}          & $(6.6 \pm 0.6)$   \\ \hline
\begin{tabular}[c]{@{}l@{}}initial path \\ length [m]\end{tabular}      & $(18.38 \pm 10.86)$  \\ \hline
\begin{tabular}[c]{@{}l@{}}optimized path\\ length [m]\end{tabular}     & $(14.3 \pm 9.84)$   \\ \hline
\begin{tabular}[c]{@{}l@{}}length\\ lower bound [m]\end{tabular}            & $(13.53 \pm 9.48)$   \\ \hline
\end{tabular}
\label{tab::planning_forest_map}
\end{minipage}
\subsection{Tree Detection}
\label{sec::res_tree_detection}
\begin{figure}[tbh]
\centering
    \includegraphics[width=0.85\textwidth]{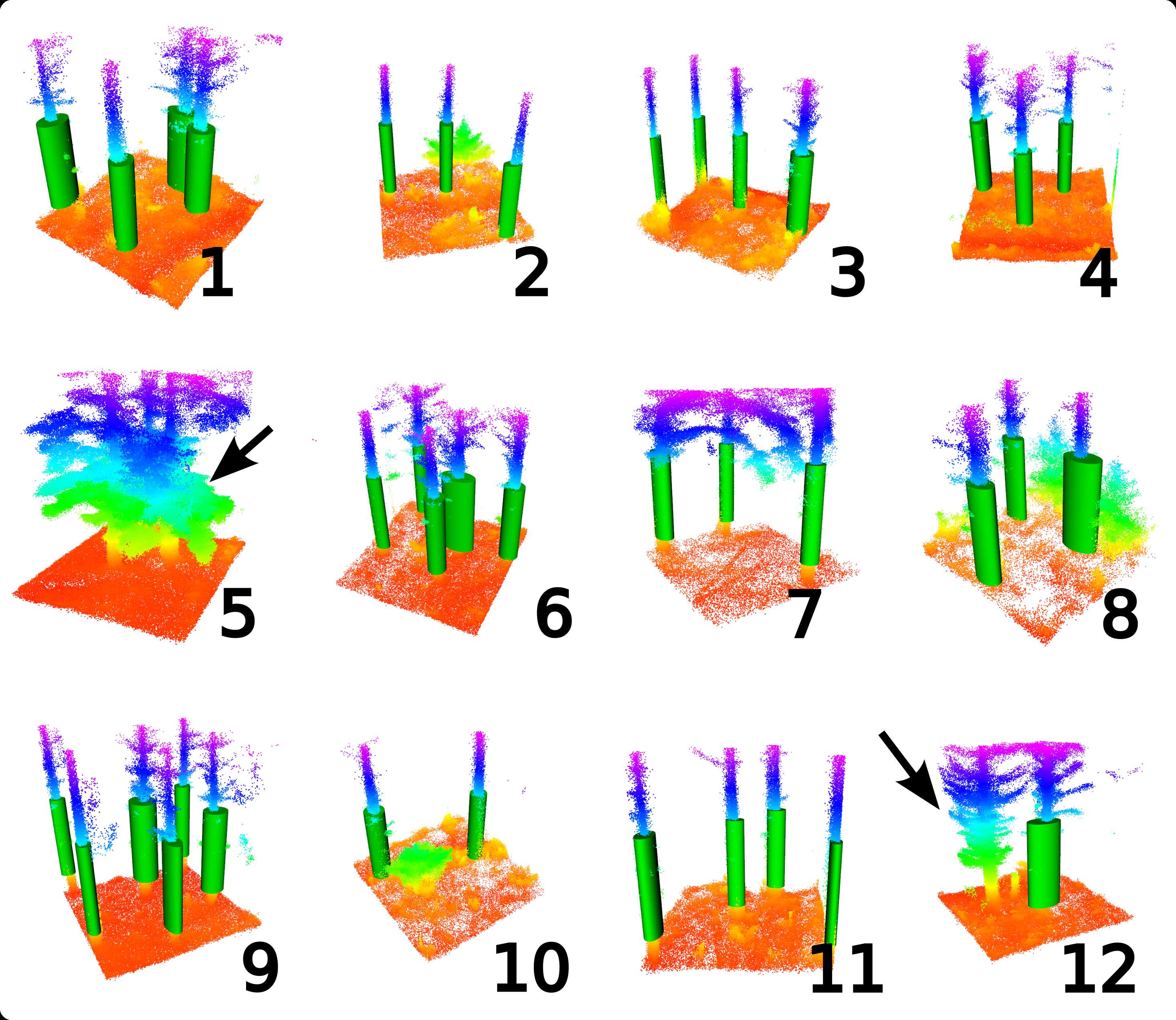}
    \caption{\SI{6}{\meter} by \SI{6}{\meter} point cloud patches from the testing site. Red corresponds to the lowest and purple to the highest elevation. Tree detections are shown with green cylinders. Tree detection has failed for snapshots 5 and 12 (marked with black arrows).}
    \label{fig::tree_detection_patch}
\end{figure}

We evaluated tree detection offline by mimicking local maps assembled from harvester sensors during the scanning maneuver. We select \SI{6}{\meter} by \SI{6}{\meter} patches inside the map shown in Fig.~\ref{fig::map_pbstream} and ask the tree detector to detect all trees inside the cropped map. The patch size roughly corresponds to the area covered by vertical Velodyne sensor frustum after the scanning maneuver. The detection procedure discarded clusters with diameter bigger than \SI{2.5}{\meter}, fewer than 1000 points or with gravity alignment score less than 0.8 (as defined in Sec.~\ref{sec::tree_detection}); the result is shown in Fig.~\ref{fig::tree_detection_patch}.  
One can observe successful tree detections even in the presence of vegetation. Heavy clutter in the scene, such as in patch five and patch twelve, causes the detector to reject segmented tree clusters, incorrectly producing false negatives. In case no trees are detected close to the target, the harvester resorts to blind grabbing at the target location (known \emph{a priori} from the map). 
We evaluated the detection accuracy on three different forest patches (the ones shown in Fig.~\ref{fig::forest_localization_pcd}, Fig.~\ref{fig::bremgarten_localization_pcd} and Fig.~\ref{fig::aligned_old_forest}). For evaluation, we selected \SI{6}{\meter} by \SI{6}{\meter} patches (about 30 of them) at random from each map and run the tree detection on them. The true number of trees was determined manually by inspecting the point cloud patch. The quantitative evaluation is shown in Table~\ref{table::detection_eval}. The fail cases and extended analysis for the tree detector are presented in the Appendix~\ref{sec::appendix_tree_detection}.  

\begin{table}[tbh]
\centering
\caption{Tree detection evaluation.}
\label{table::detection_eval}
\begin{tabular}{|l|l|l|l|l|}
\hline
map           & num trials & num trees     & recall        & precision     \\ \hline
Fig.~\ref{fig::forest_localization_pcd} & 32         & $4.13 \pm 1.91$ & $0.88 \pm 0.2$  & $0.92 \pm 0.16$ \\
Fig.~\ref{fig::aligned_old_forest}    & 33         & $4.78 \pm 2.39$ & $0.82 \pm 0.25$ & $0.94 \pm 0.11$ \\
Fig.~\ref{fig::bremgarten_localization_pcd}    & 32         & $3.28 \pm 1.71$ & $0.95 \pm 0.14$ & 1             \\ \hline \hline
total         & 97         & $4.06 \pm 1.17$ & $0.88 \pm 0.12$ & $0.95 \pm 0.07$             \\ \hline
\end{tabular}
\end{table}
\subsubsection{Tree Detector for Mission Planning}
\label{sec::res_tree_detection_mission_planner}
In this section, we present the use of a tree detector as a mission planner. Although the tree detector was designed to be used on small local maps, it can also be used to extract tree coordinates for clear-cutting missions or as a mission planning aid. We ran the tree detection offline on three forest patches of different styles, all shown in the figures below. The ground plane was filtered out, and the procedure presented in Sec.~\ref{sec::tree_detection} was used. We manually identify the trees in each point cloud to obtain the actual number of trees. 

\begin{figure*}[tbh]
\centering

     \subfloat[Mixed forest with detected trees \label{fig::hoengg_forest_mission_plan}]{
       \includegraphics[width=0.5\textwidth]{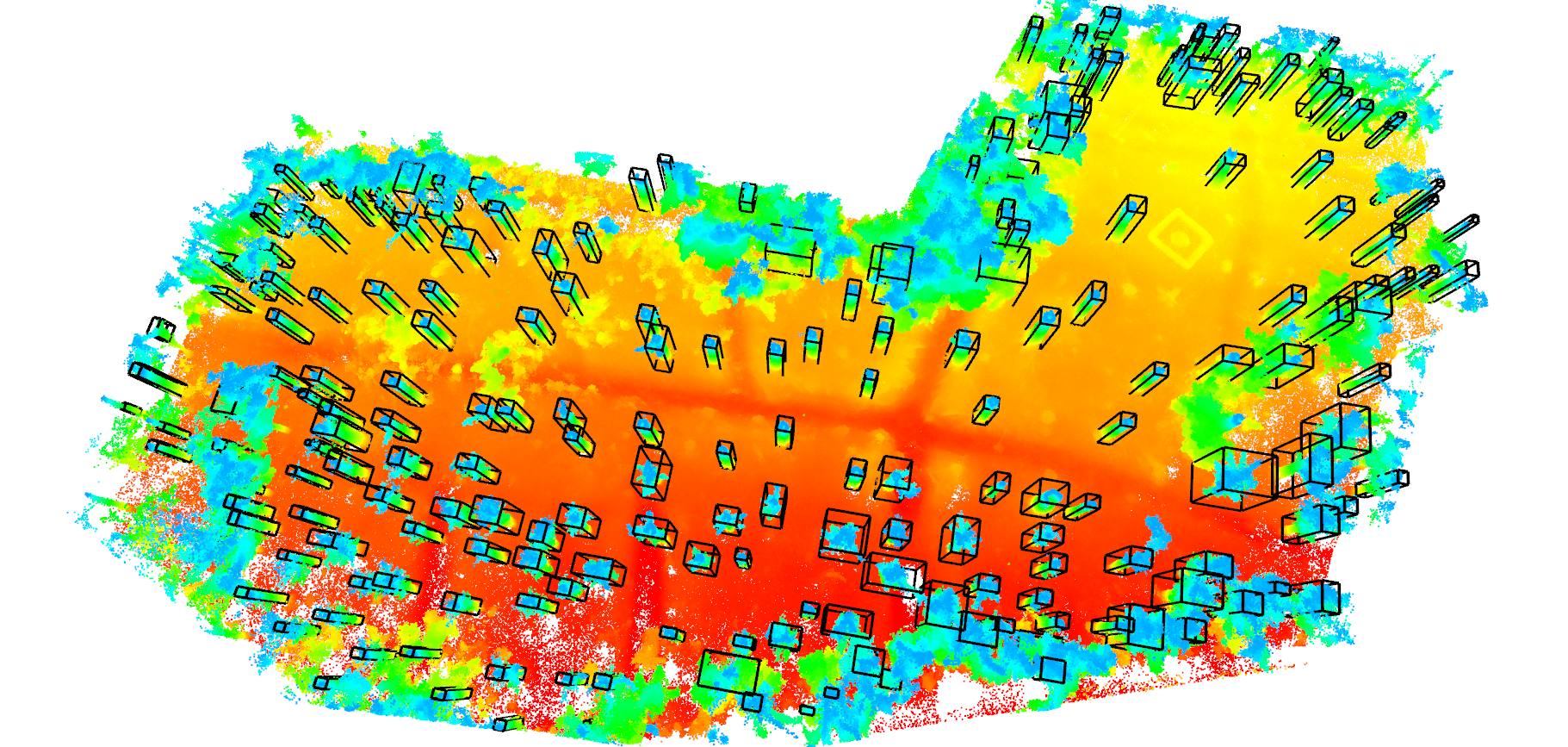}
     }
    \subfloat[Spruce forest with detected trees \label{fig::kajava_mission_plan}]{
       \includegraphics[width=0.5\textwidth]{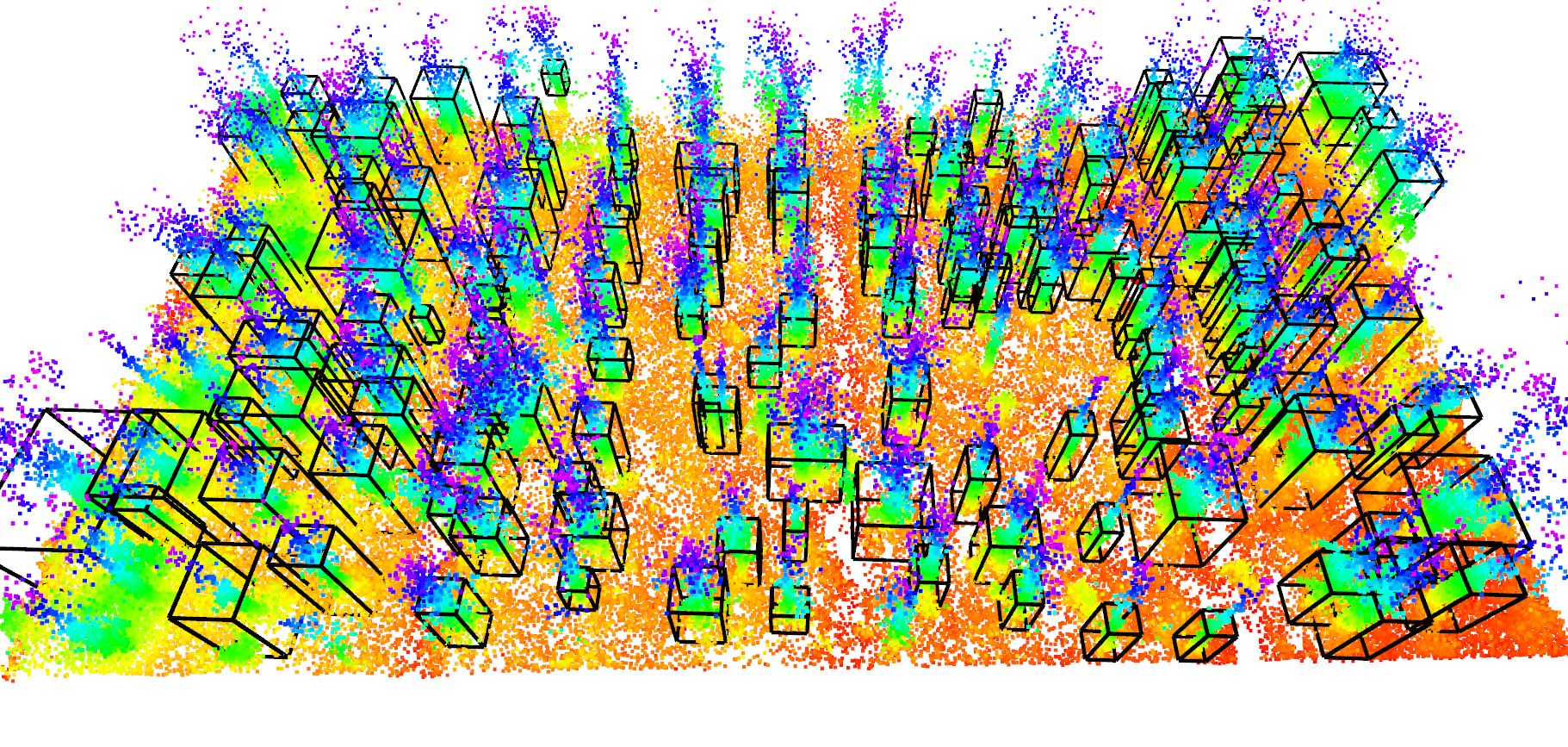}}
    \caption{Maps and tree detections for two forest patches. Detected trees are denoted with black bounding boxes. Red/yellow color corresponds to the lowest elevation and blue/purple to the highest. \textbf{\textit{Left:}} Forest with both evergreen and deciduous trees. Note the heavy clutter towards the edges of the map.  \textbf{\textit{Right:}} Forest with mainly spruce trees.}
    \label{fig::mission_plan_kajava_hoengg}
\end{figure*}

\begin{figure}[tbh]
\centering
    \includegraphics[width=0.8\textwidth]{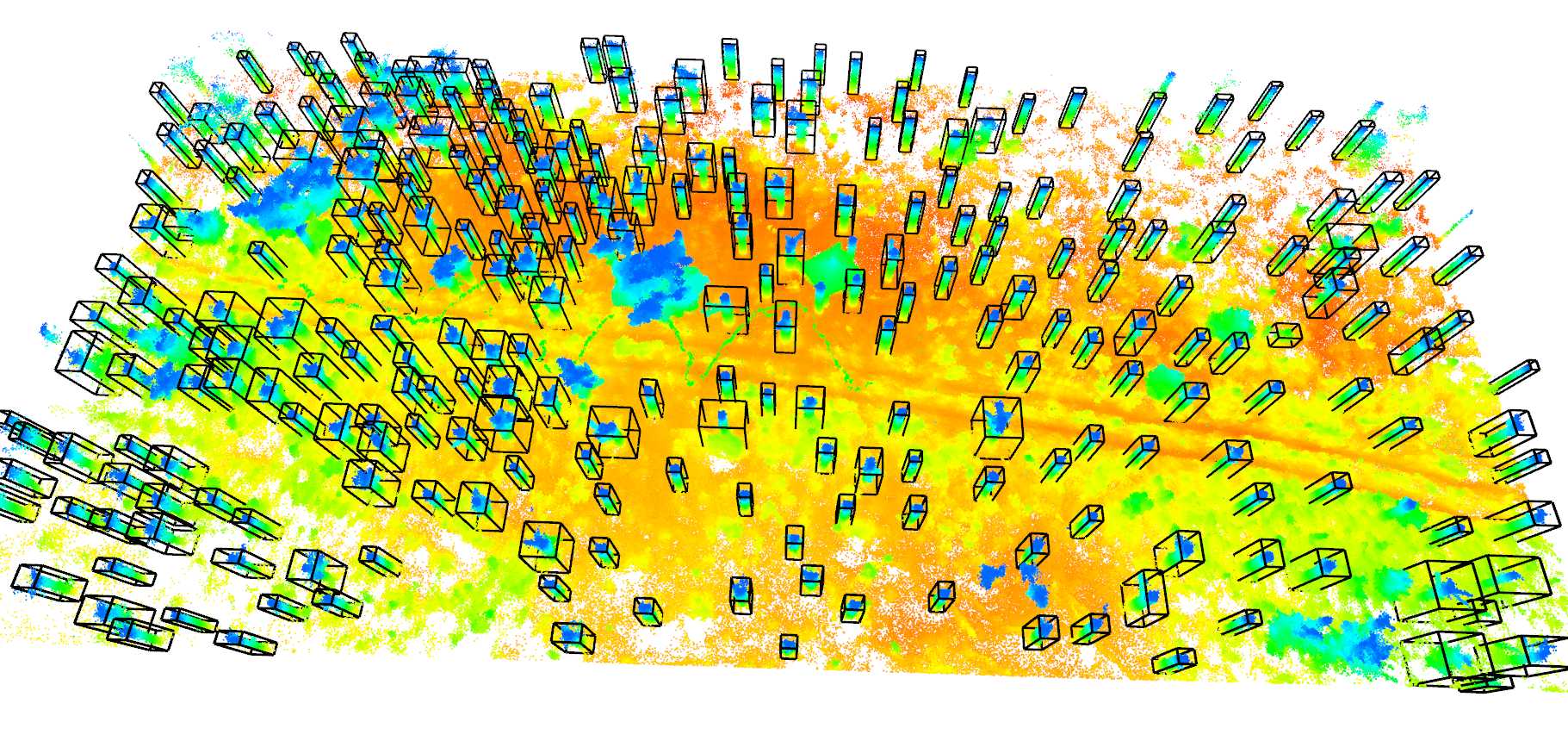}
    \caption{Forest patch and detected trees denoted with black bounding boxes. The forest patch consists of deciduous trees.}
    \label{fig::bremgarten_mission_plan}
\end{figure}

Fig.~\ref{fig::hoengg_forest_mission_plan} shows the first forest patch together with tree detections. The patch has about 246 (both evergreen and deciduous) trees, and the tree detector correctly detected 210 of them. There were 14 false positives, and 36 trees were not detected (false negatives), which amounts to a precision of \SI{94}{\percent} and recall rate of about \SI{85}{\percent}. The point cloud of the second forest patch and detected trees are shown in Fig.~\ref{fig::kajava_mission_plan}. In total, there are 163 (spruce) trees, and the tree detector correctly detected 143 of them. There were six false positives, and 20 trees were not detected, which amounts to a precision of \SI{96}{\percent} and recall rate of about \SI{88}{\percent}. The last forest patch is shown Fig.~\ref{fig::bremgarten_mission_plan}. The Forest patch has about 285 trees (all deciduous trees). The tree detector detected in total 266 trees. Four detections were false positives, whereas the number of false negatives was 19. These numbers give the tree detector precision of \SI{98}{\percent} and recall rate of \SI{93}{\percent}. For a pointcloud with 26 million points, it takes our method about \SI{300}{\second} to complete.

We want to conclude that using our tree detector as a mission planner may yield inferior results than other methods. However, the strong points of our method are that it is lightweight, available as open-source, and can be used online on the robot. For tree detection on large forest areas, one is better off using some of the more advanced approaches, such as \cite{burt2019extracting} which achieve recall rates up to \SI{96}{\percent} on tropical forest datasets. One should note, however, that the method is computationally expensive; it takes two days to process a map with 17 million points on a computer with 24 cores. Some of the best results that we have seen have been obtained using the forestry module inside the LiDAR360 commercial software for pointcloud processing. Unfortunately, LiDAR360 is not available for free.

\subsection{Localization}
\label{sec::res_localization}
\begin{figure*}[tbh]
\centering
    \subfloat[Testfield  \label{fig::hoengg_localization_pcd}]{
       \includegraphics[width=0.33\textwidth]{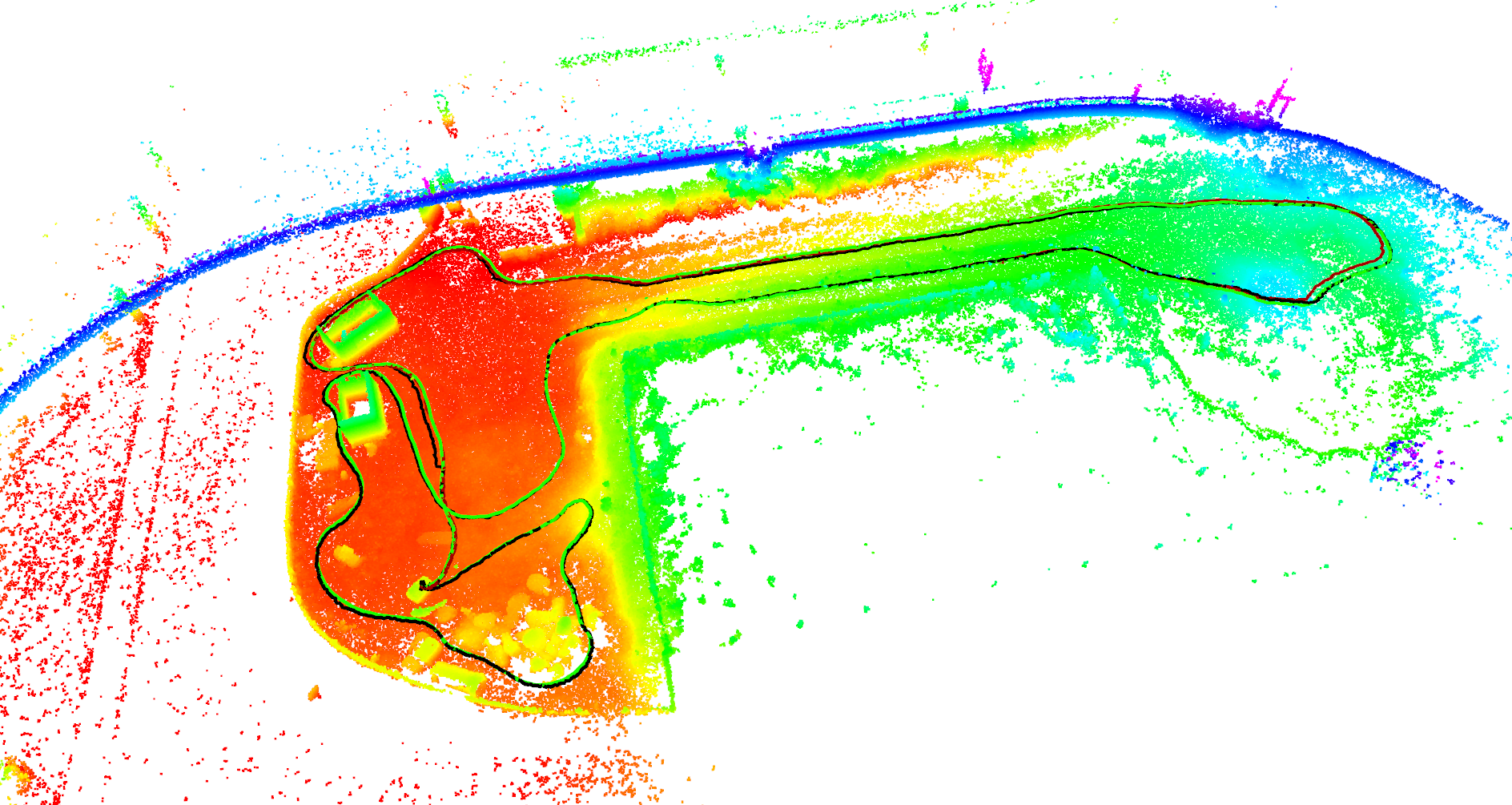}}
     \subfloat[Forest 1 \label{fig::forest_localization_pcd}]{
       \includegraphics[width=0.33\textwidth]{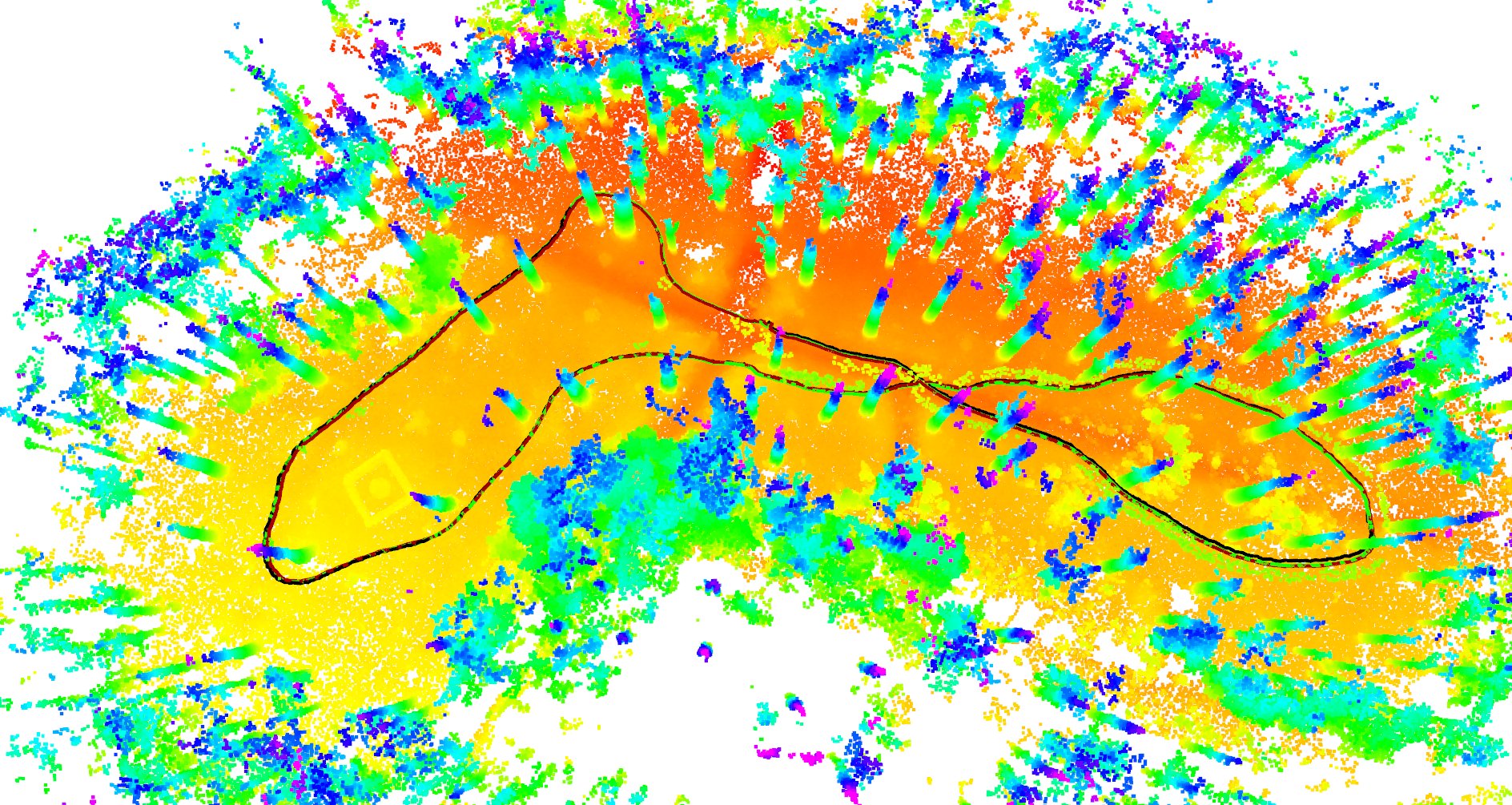}
     }
     \subfloat[Forest 2 \label{fig::bremgarten_localization_pcd}]{
       \includegraphics[width=0.33\textwidth]{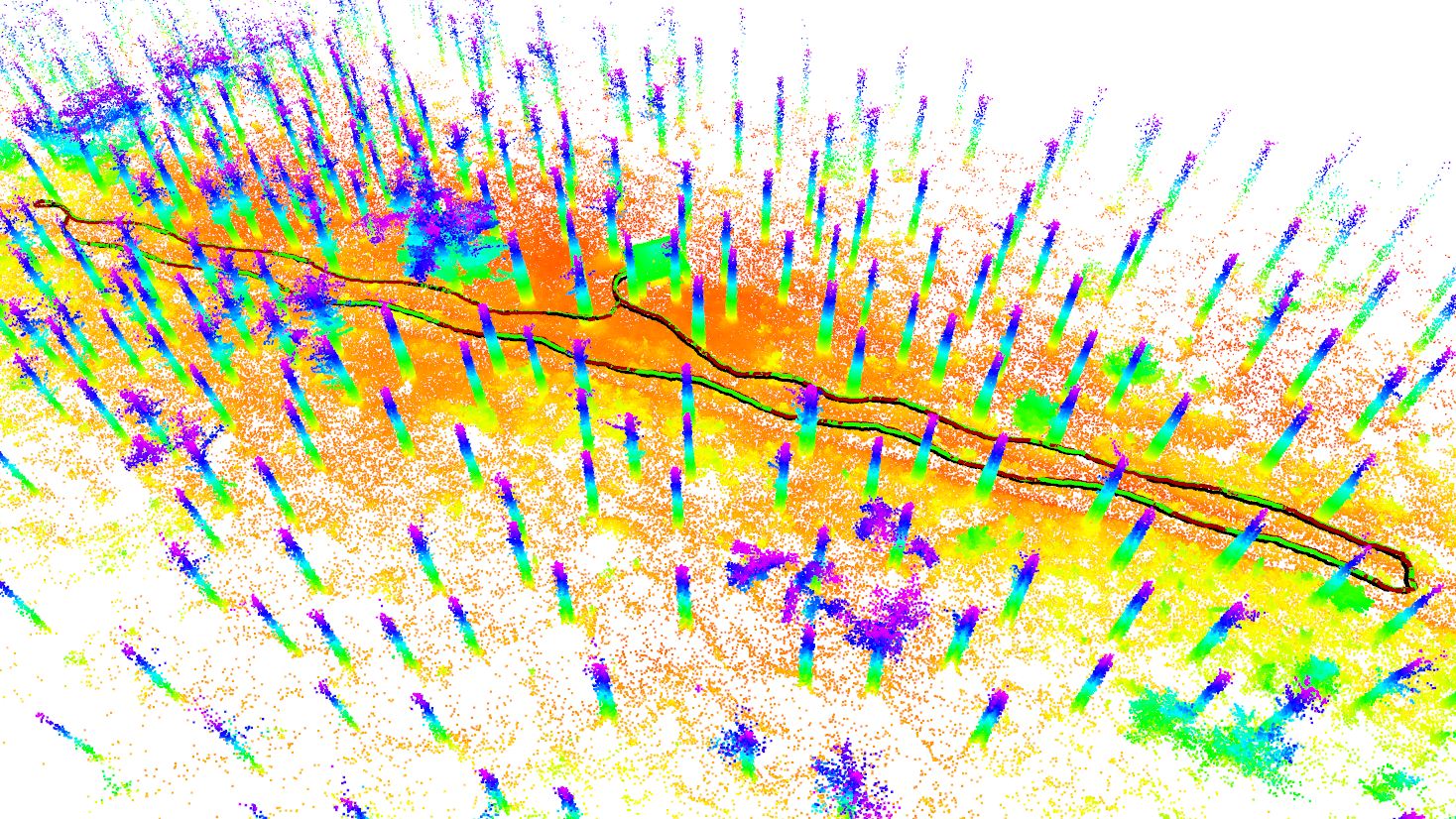}
     }
    \caption{The trajectories from the localization field experiments overlaid with the map.}
    \label{fig::localizaton_pcd}
\end{figure*}

\begin{figure*}[tbh]
\centering
    \subfloat[Testfield  \label{fig::hoengg_localization}]{
       \includegraphics[width=0.33\textwidth]{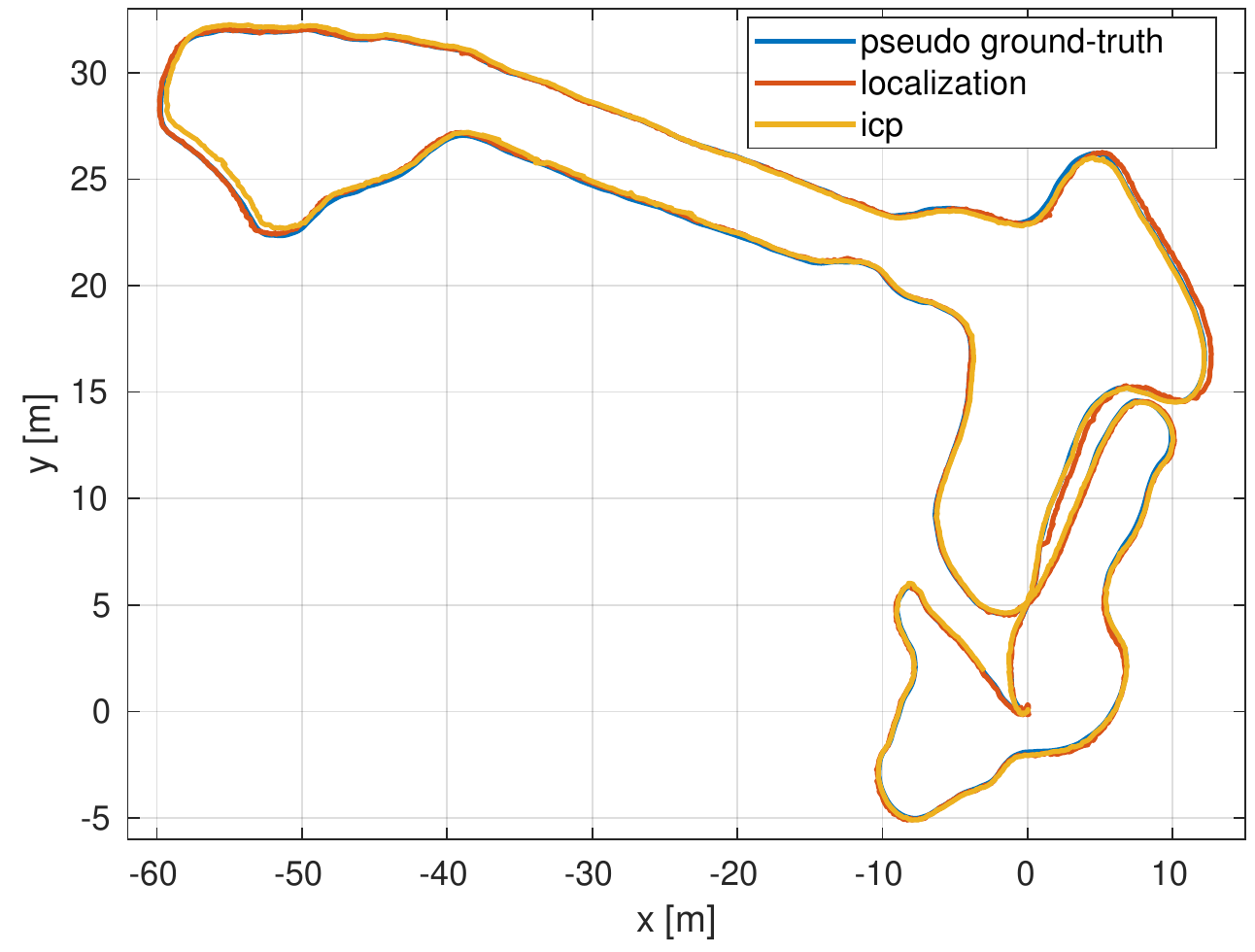}}
     \subfloat[Forest 1 \label{fig::forest_localization}]{
       \includegraphics[width=0.33\textwidth]{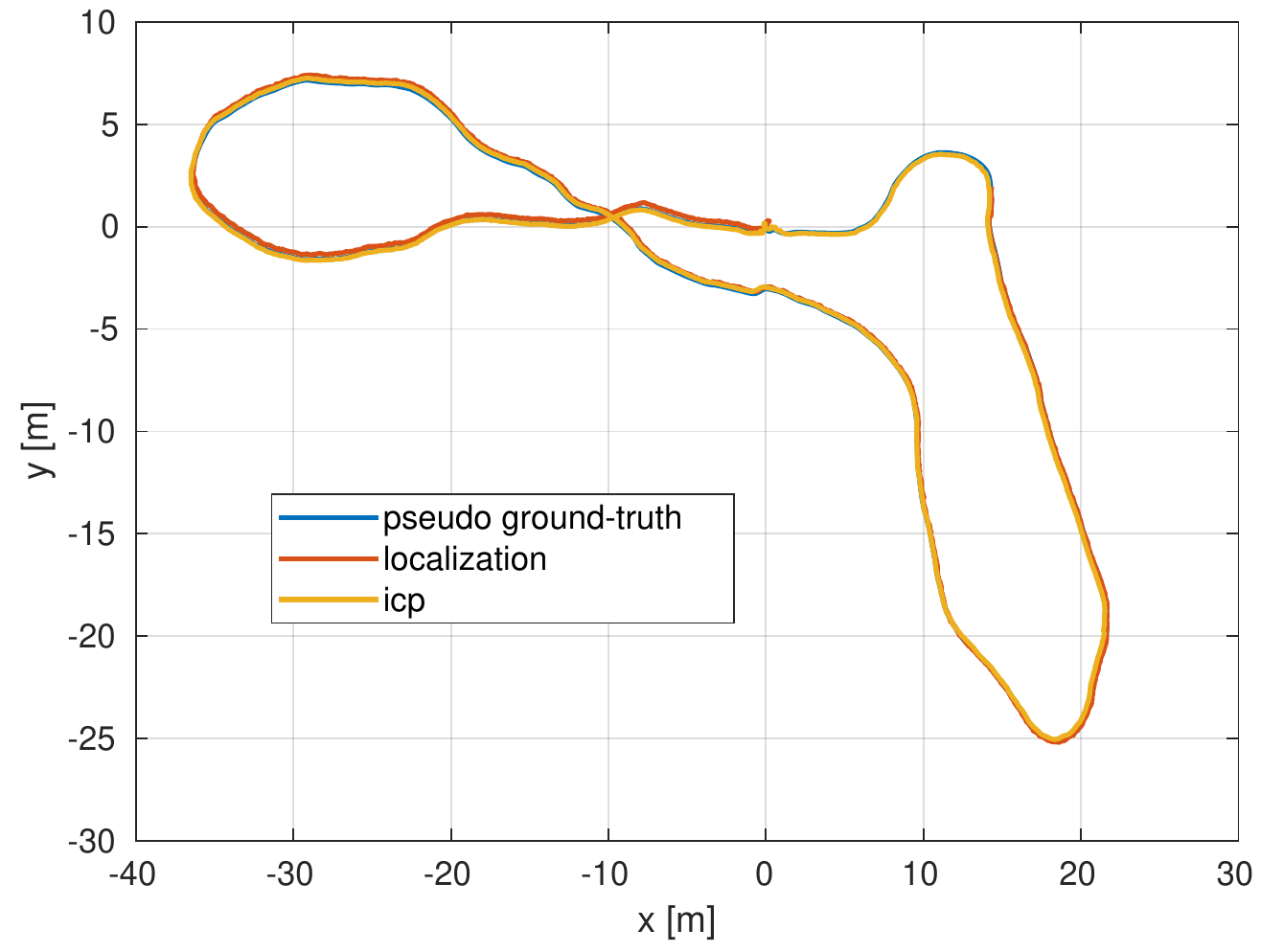}
     }
     \subfloat[Forest 2 \label{fig::bremgarten_localization}]{
       \includegraphics[width=0.33\textwidth]{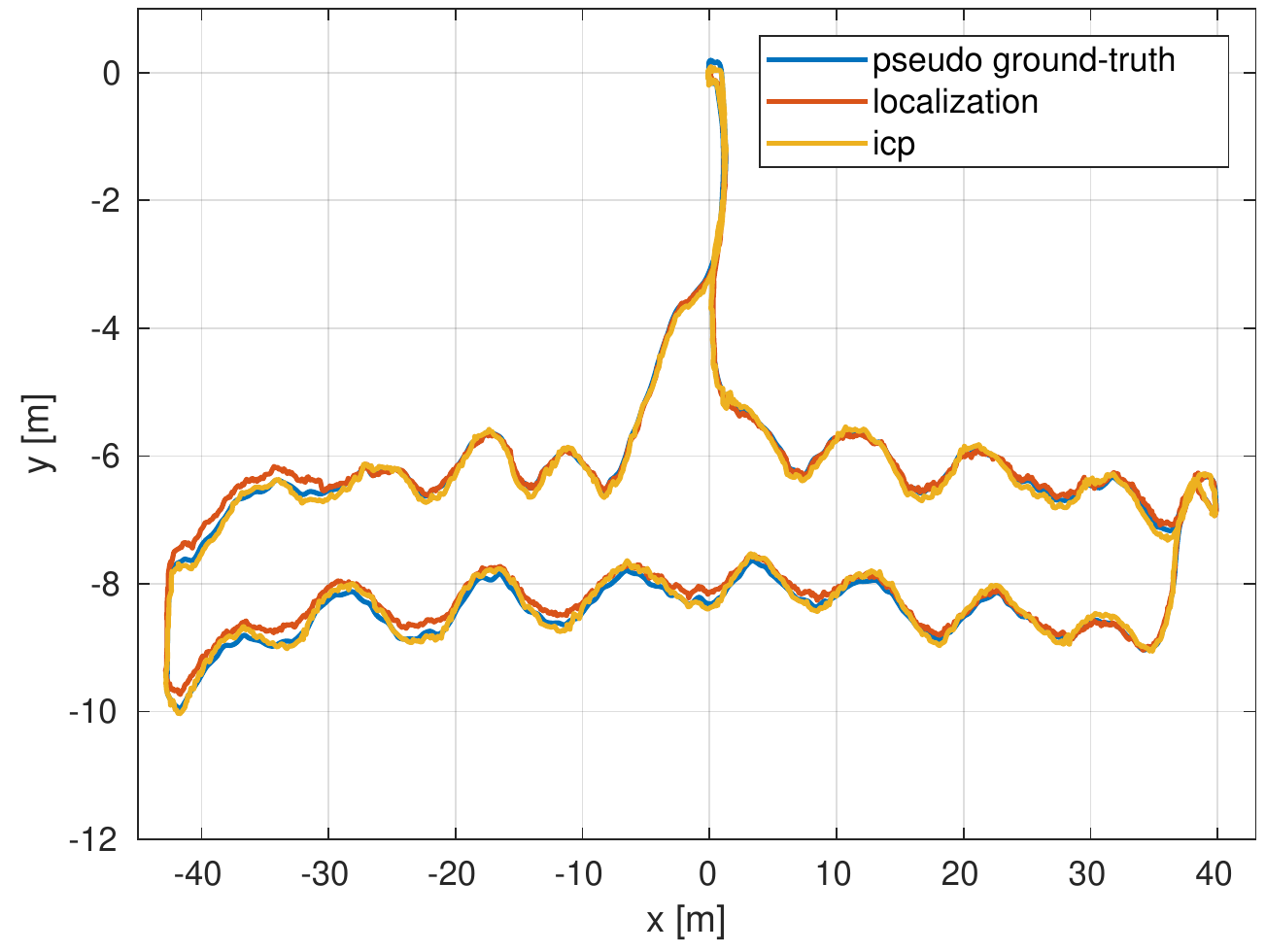}
     }
    \caption{The trajectories from the localization field experiments viewed from above. Forest 1 and Forest 2 are datasets from two different forests.}
    \label{fig::localizaton_xy}
\end{figure*}

We evaluated the localization of the proposed sensor module running Google Cartographer \ac{SLAM} in three different terrains: two forest patches and our testing field. We collect the dataset with the sensor module presented in \ref{sec::shpherds_crook}, and we run the \ac{SLAM} offline, which produces a consistent, optimized map. Subsequently, we set the Cartographer in the localization mode and tune it such that both the \ac{LIDAR} odometry and loop closures run in real-time. Note that Cartographer running in the localization mode does not try to build a global map (see \cite{catographer2017online}). We are interested in quantifying performance degradation when running in real-time localization mode without joint map and trajectory bundle adjustment. We localize in a previously built map and compare the localized pose against the bundle adjusted trajectory (pseudo-ground-truth). Paths overlaid inside the map can be seen in Fig.~\ref{fig::localizaton_pcd} while the top view plots of the trajectories can be seen in Fig.~\ref{fig::localizaton_xy}.


We evaluate the localization error as the euclidean distance between the corresponding points on the localized trajectory and pseudo ground-truth trajectory. For the test field, we obtain a translational error of $0.11\pm 0.09$ meters; the trajectory length was \SI{255.8}{\meter}. Trajectory length for the first forest dataset was \SI{158.37}{\meter}, and Cartographer localizes with the error of $0.17 \pm 0.28$ meters. Inside, the second forest trajectory was \SI{183.27} meters long, and the localization error was $0.25\pm 0.28$ meters. The presented evaluation is biased since the cartographer (the same method) is used for both map building and localization. Hence, we evaluate localization performance using a different method, the \ac{ICP} (using point to plane error metric). The \ac{ICP} is used to register \ac{LIDAR} scans in a map built using the cartographer (we do not have the ground truth). Subsequently, the translational error between the \ac{ICP} localized trajectory and the pseudo-ground truth from the cartographer is computed. For the first forest, the error is $0.085 \pm 0.038$ meters, and in the second forest, the error is $0.078 \pm 0.035$ meters. For the test field, we obtain an average error of $0.13 \pm 0.11$ between the pseud-ground truth and the \ac{ICP}.

\begin{figure*}[tbh]
\centering
    \subfloat[Wangen and der Aare (CH)  \label{fig::wangen_aerial}]{
       \includegraphics[width=0.49\textwidth]{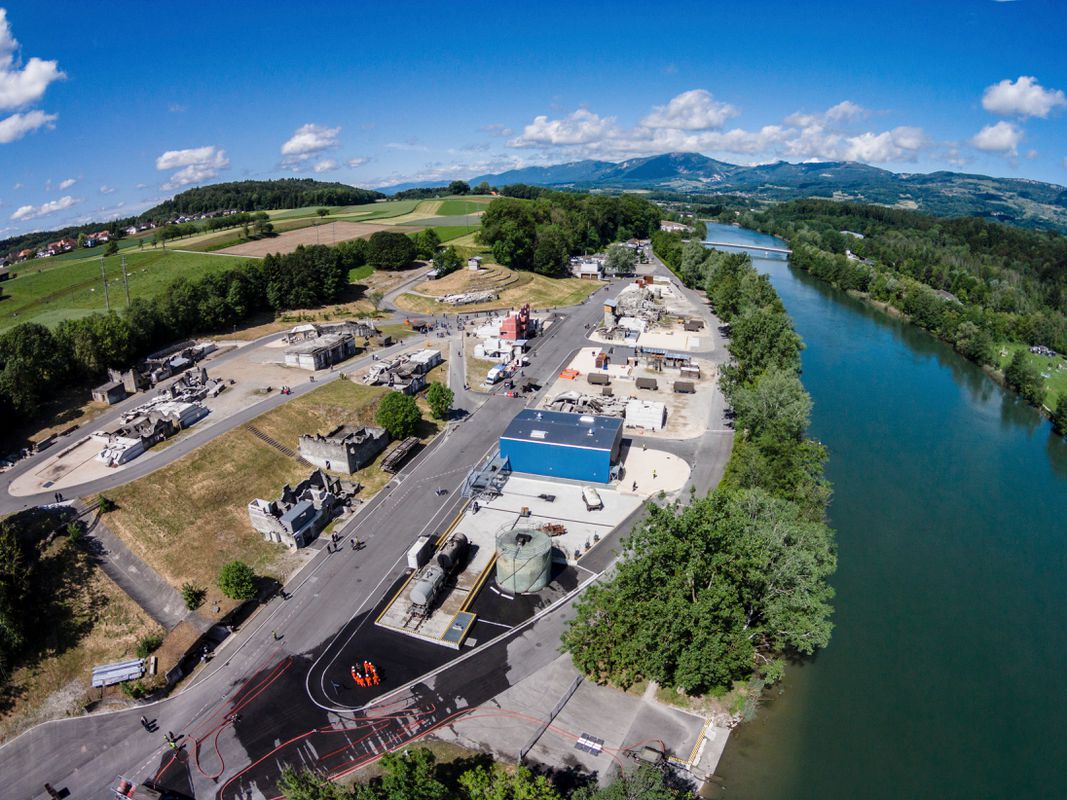}}
     \subfloat[Forest 3 \label{fig::old_forest}]{
       \includegraphics[width=0.49\textwidth]{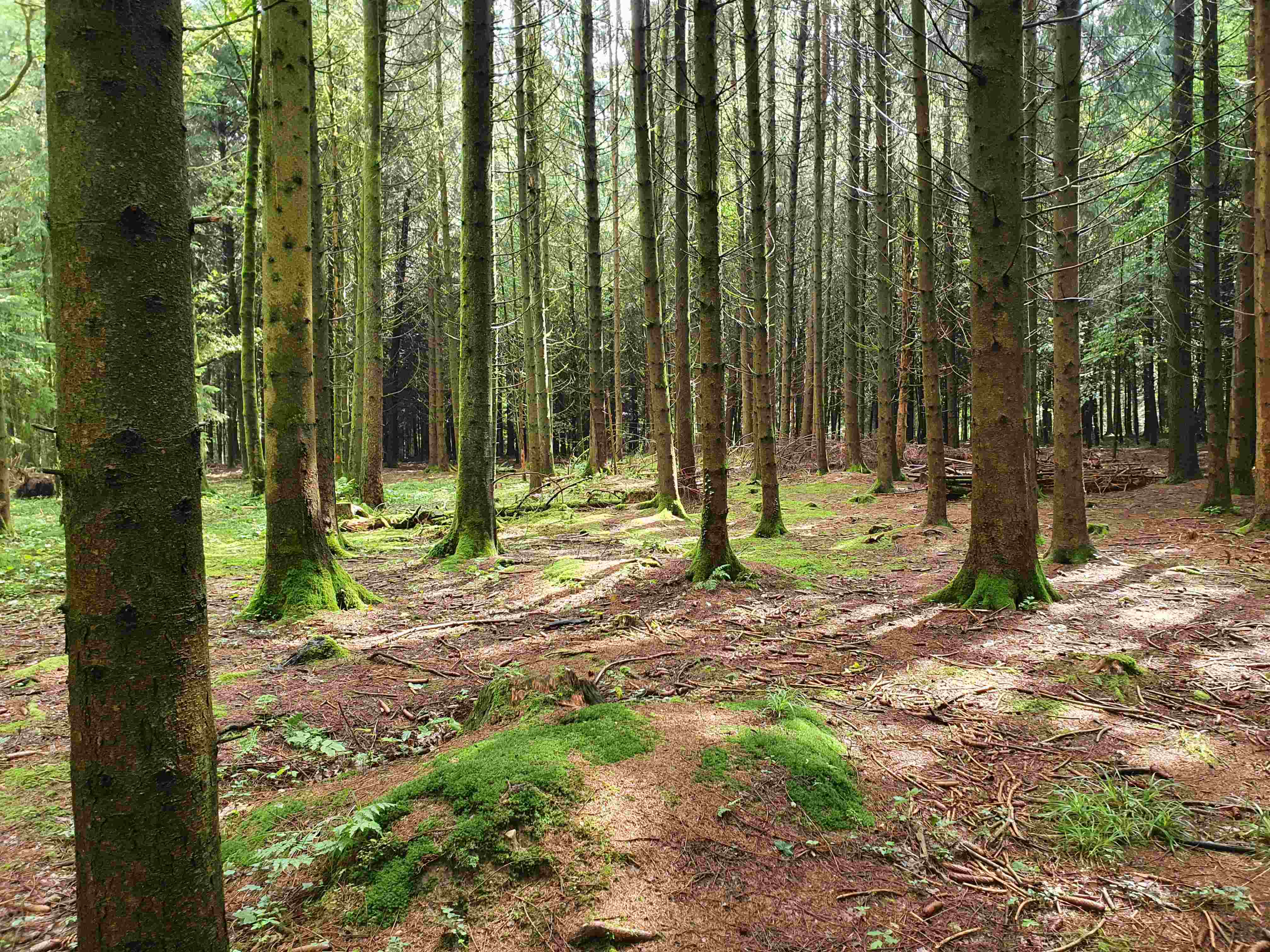}
     }
    \caption{Locations for mapping ground truth evaluation. \textbf{\textit{Left:}} Aerial photo of the emergency responders training ground, Wangen and der Aare. Image taken from \url{https://www.mediathek.admin.ch/media/image/da1c1259-8d9a-430a-893d-d325de139149} \textbf{\textit{Right:}} Forest patch where the mapping accuracy was evaluated. The forest in the photo is an old forest.}
    \label{fig::wangen_old_forest_photos}
\end{figure*}

To better understand the overall accuracy, we compare the mapping and localization performance against ground-truth data in two different environments: a forest and an urban setting (emergency services training ground in Wangen an der Aare, Switzerland). Both areas are shown in Fig.~\ref{fig::wangen_old_forest_photos}. The ground-truth data was produced using Leica's RTC 360 3D laser scanner. It is a rotating laser scanner that aligns point clouds and uses \ac{VIO} to record moves from station to station for scan pre-registration automatically. For both the forest and the training ground, the RTC360 reported sub-centimeter accuracy.

\begin{figure}[tbh]
\centering
    \includegraphics[width=0.95\textwidth]{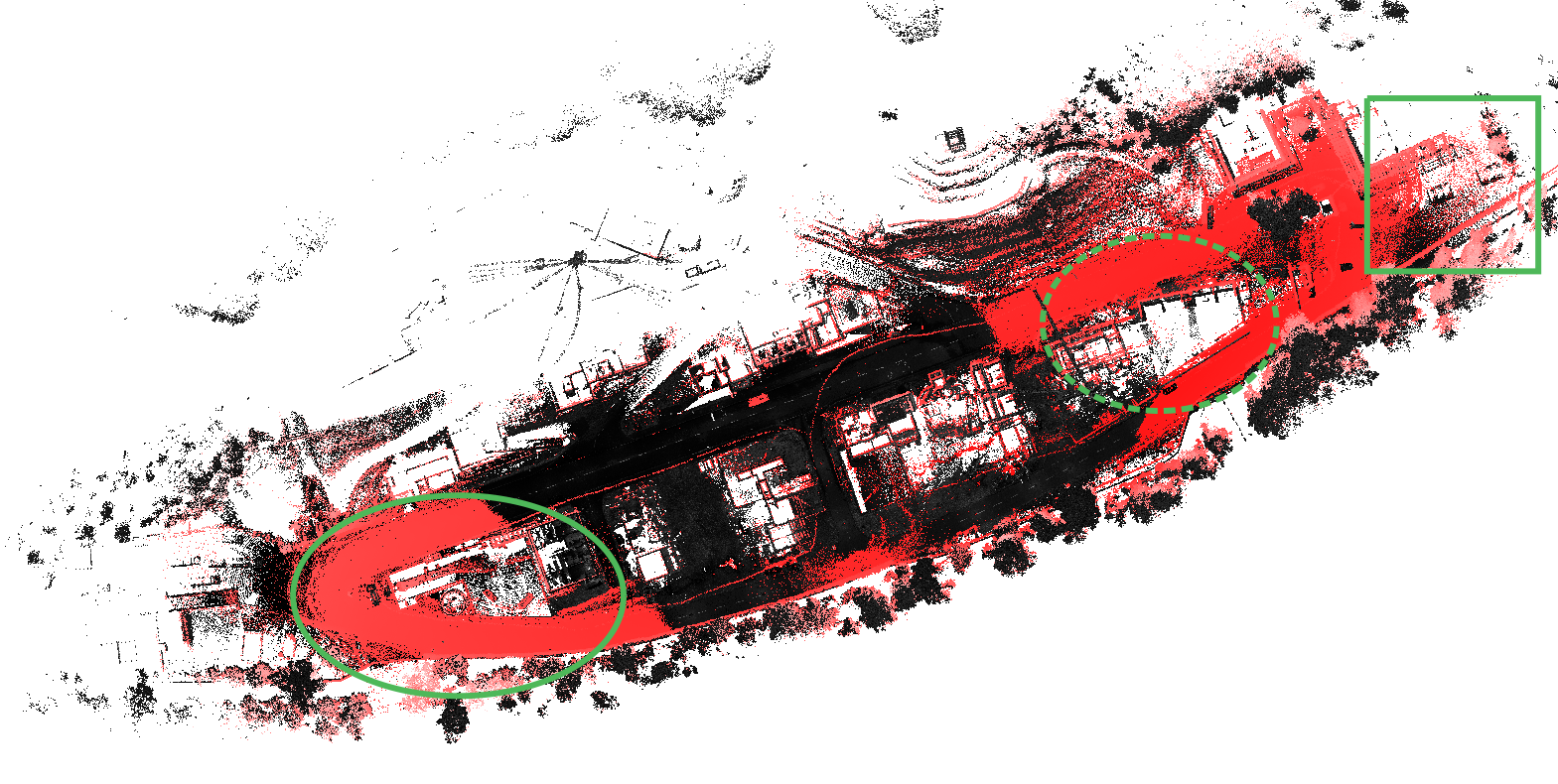}
    \caption{Map of the emergency responders training ground, Wangen an der Aare, Switzerland. The ground truth map is shown in black(from Leica RTC360), and in red, the cartographer map is shown. The green circle shows an example of good alignment, the dotted circle shows an example of bad alignment, and the green square shows an example of changes between two mapping sessions. The data was collected a couple of weeks apart.}
    \label{fig::wangen_aligned_clouds}
\end{figure}

We align the ground truth map and the cartographer map using the Cloudcompare software alignment tool. The aligned maps for the Wangen area are shown in Fig.~\ref{fig::wangen_aligned_clouds}. The area has dimensions of 294 x 577 x 66 meters (W x L x H). The RMS error from the registration was \SI{0.578}{\meter} with the average distance between the points of $0.76 \pm 0.52$ meters (see below for a discussion). To map the area using our proposed setup, about 10 min of walking around the site was required. To obtain the ground truth map, we made 55 scans which required a whole afternoon of work.

\begin{figure*}[tbh]
\centering
     \subfloat[Wangen training site \label{fig::wangen_localization}]{
       \includegraphics[width=0.49\textwidth]{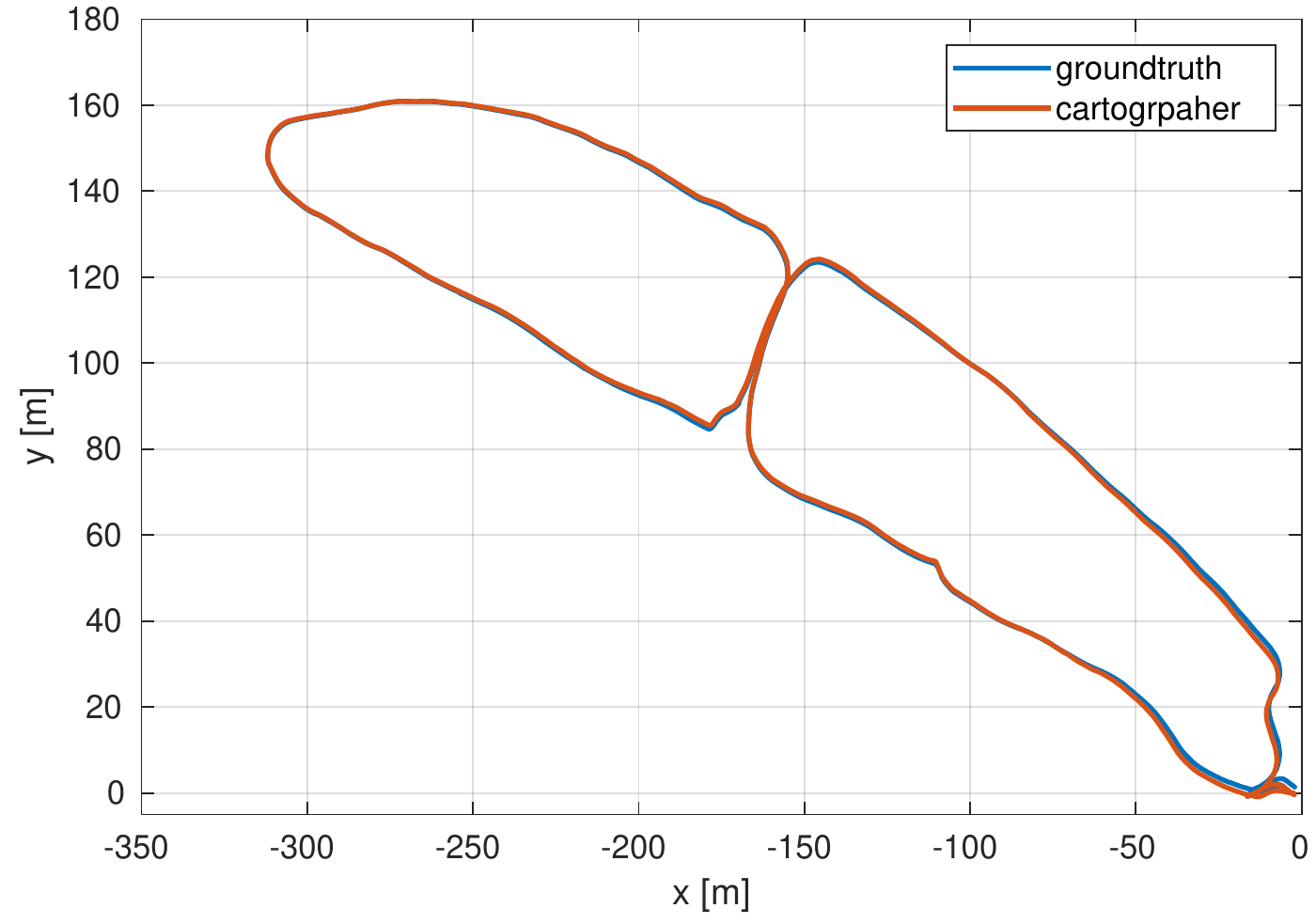}
     }
     \subfloat[Forest 3 (near Wangen) \label{fig::old_forest_localization}]{
       \includegraphics[width=0.48\textwidth]{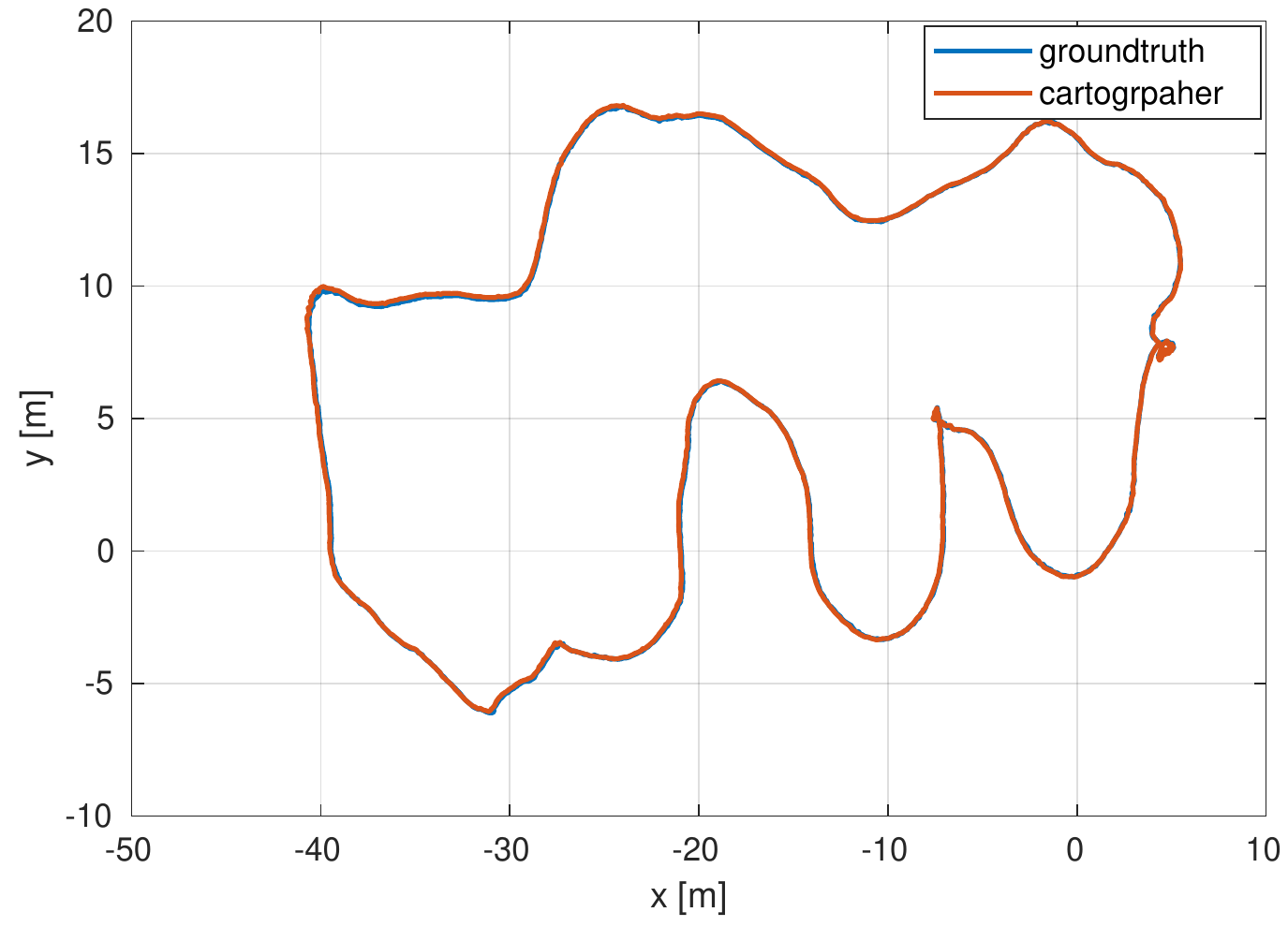}
     }
    \caption{Top view trajectory estimate from cartographer (brown) aligned with the ground truth trajectory (blue). \textbf{\textit{Left:}} Trajectory estimates for data collection at Wangen training site. Note how the error is bigger in the area where maps in Fig.~\ref{fig::wangen_aligned_clouds} are not well aligned. \textbf{\textit{Right:}} Trajectory estimates for data set collection in the forest 3.}
    \label{fig::localization_xy_groundtruth}
\end{figure*}

Besides the map quality, we also evaluate cartographer's trajectory quality (pseudo-ground truth in Fig.~\ref{fig::localizaton_pcd} and Fig.~\ref{fig::localizaton_xy}). To obtain the ground truth trajectory, we register laser scans in the ground truth map from the Leica scanner using the \ac{ICP}. Then we align the trajectories estimate from the cartographer with the ground truth trajectory (using the transform obtained during the map alignment step). The trajectory length was \SI{823}{\meter} and the overall error was $0.41 \pm 0.28$ meters, overlay visualized in Fig.~\ref{fig::wangen_localization}. This level of accuracy is acceptable for navigation; however, blind grabbing might not be accurate enough, which is why our method does an extra detection step in the local map. We note that the overall errors (both map and trajectory estimation) computed are pessimistic (worst case) because of the changes in the environment between the data collection. We observed new (or differently parked) vehicles, cranes, trailers, tents, and even a large water pool left on the main road of the training ground. All of these changes in the environment negatively influence the accuracy of scan matching and make the correct data association harder.

\begin{figure*}[tbh]
\centering
    \subfloat[Good alignment  \label{fig::wangen_well_align}]{
       \includegraphics[width=0.33\textwidth]{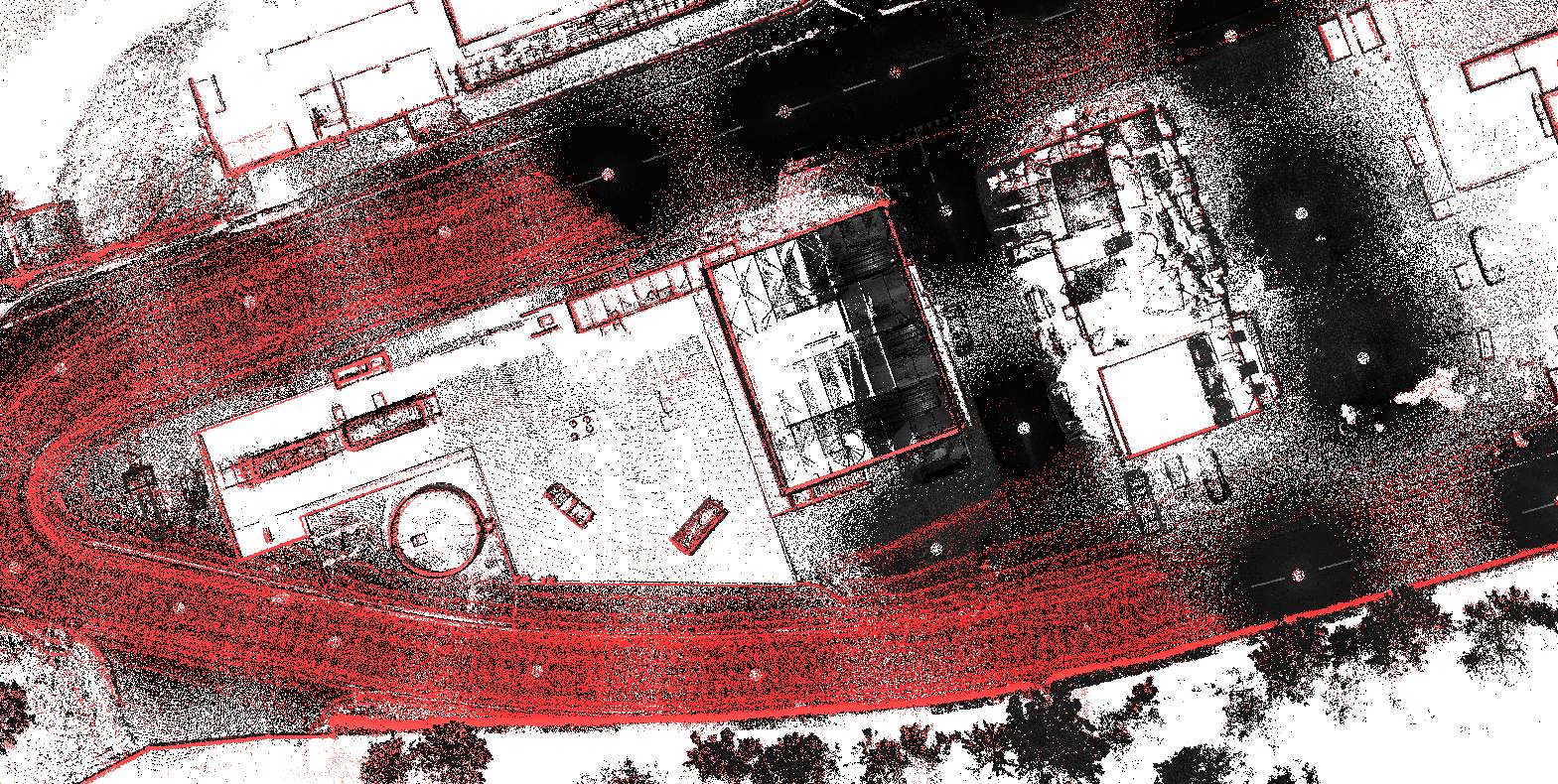}}
     \subfloat[Poor alignment \label{fig::wangen_poor_align}]{
       \includegraphics[width=0.33\textwidth]{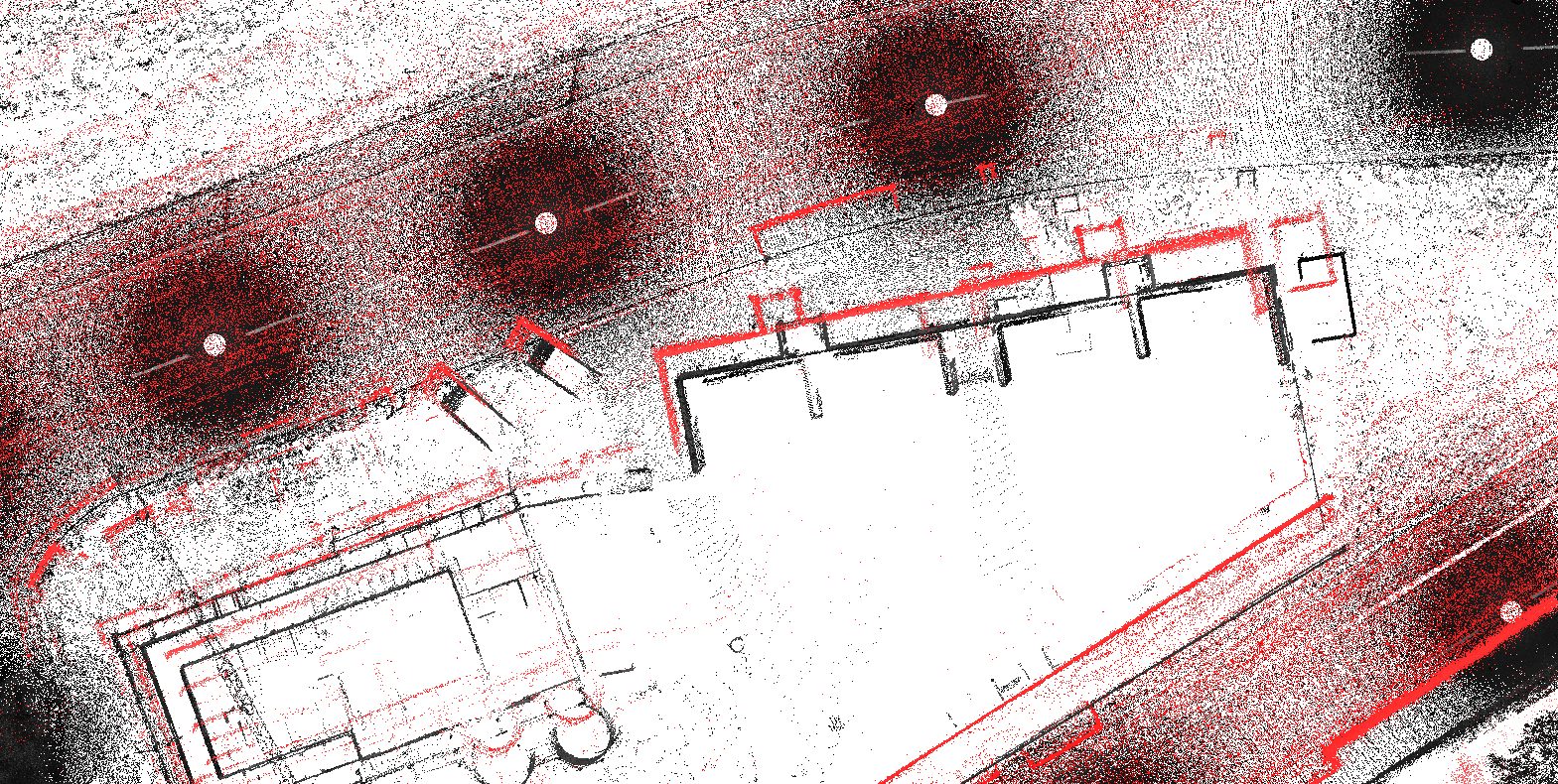}
     }
     \subfloat[Changes between data collection \label{fig::wangen_seasonal}]{
       \includegraphics[width=0.33\textwidth]{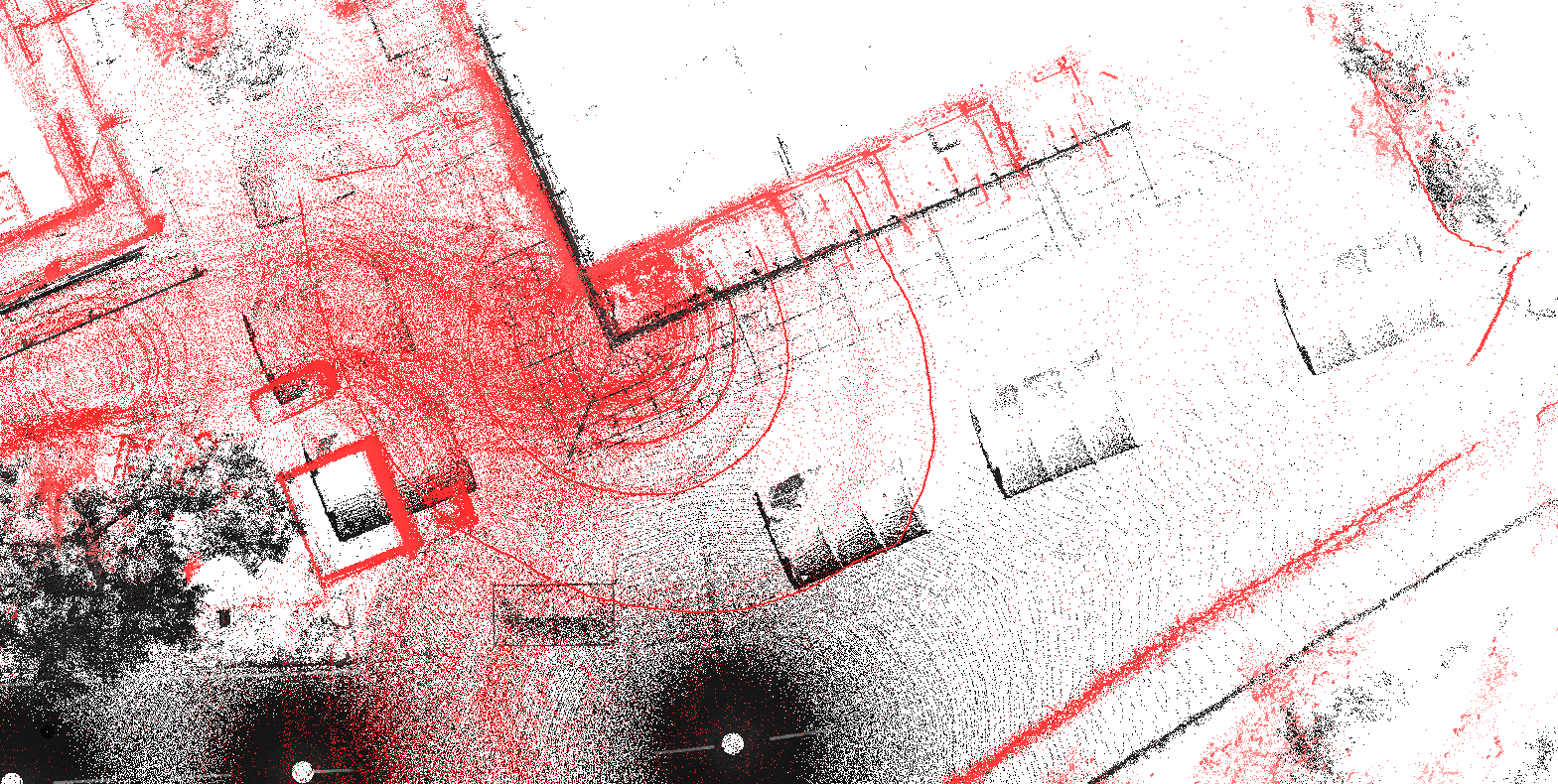}
     }
    \caption{Close up view of encircled areas from Fig.~\ref{fig::wangen_aligned_clouds}. Ground truth is shown in black and cartographer map shown in red. \textbf{\textit{Left:}} Example of good alignment. Note how building edges from different maps snap well onto each other. \textbf{\textit{Middle:}} Example of poor alignment. Note how the building walls are far apart from each other. \textbf{\textit{Right:}} Changes between the dataset collections. Note the black tents in the ground truth map and the red firefighting trailer in the cartographer map.}
    \label{fig::wangen_aligned_maps_zoomed_in}
\end{figure*}

\begin{figure*}[tbh]
\centering
    \subfloat[Registered maps  \label{fig::aligned_old_forest}]{
       \includegraphics[width=0.49\textwidth]{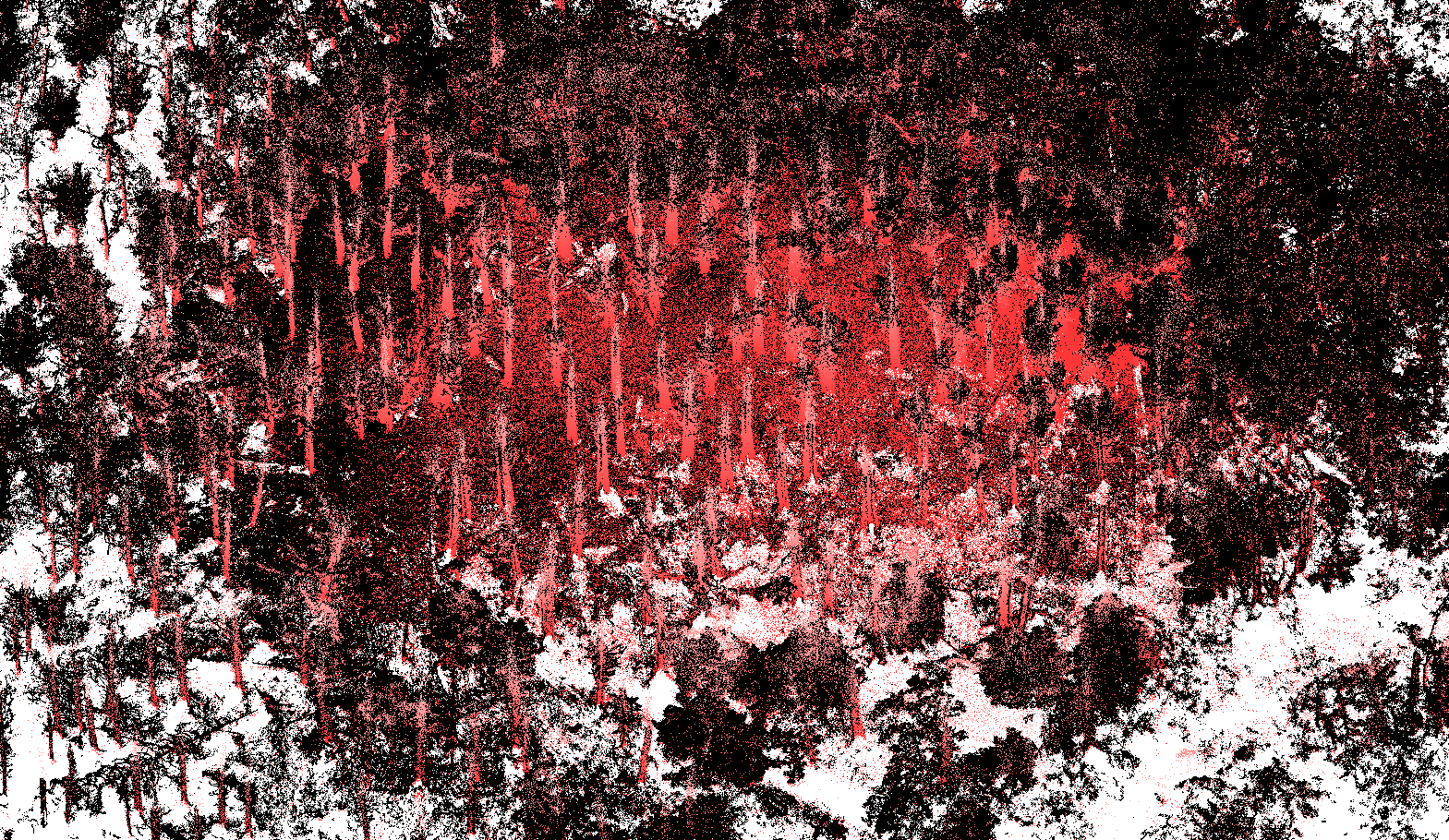}}
     \subfloat[Close up of the alignment \label{fig::well_aligned_old_forest}]{
       \includegraphics[width=0.49\textwidth]{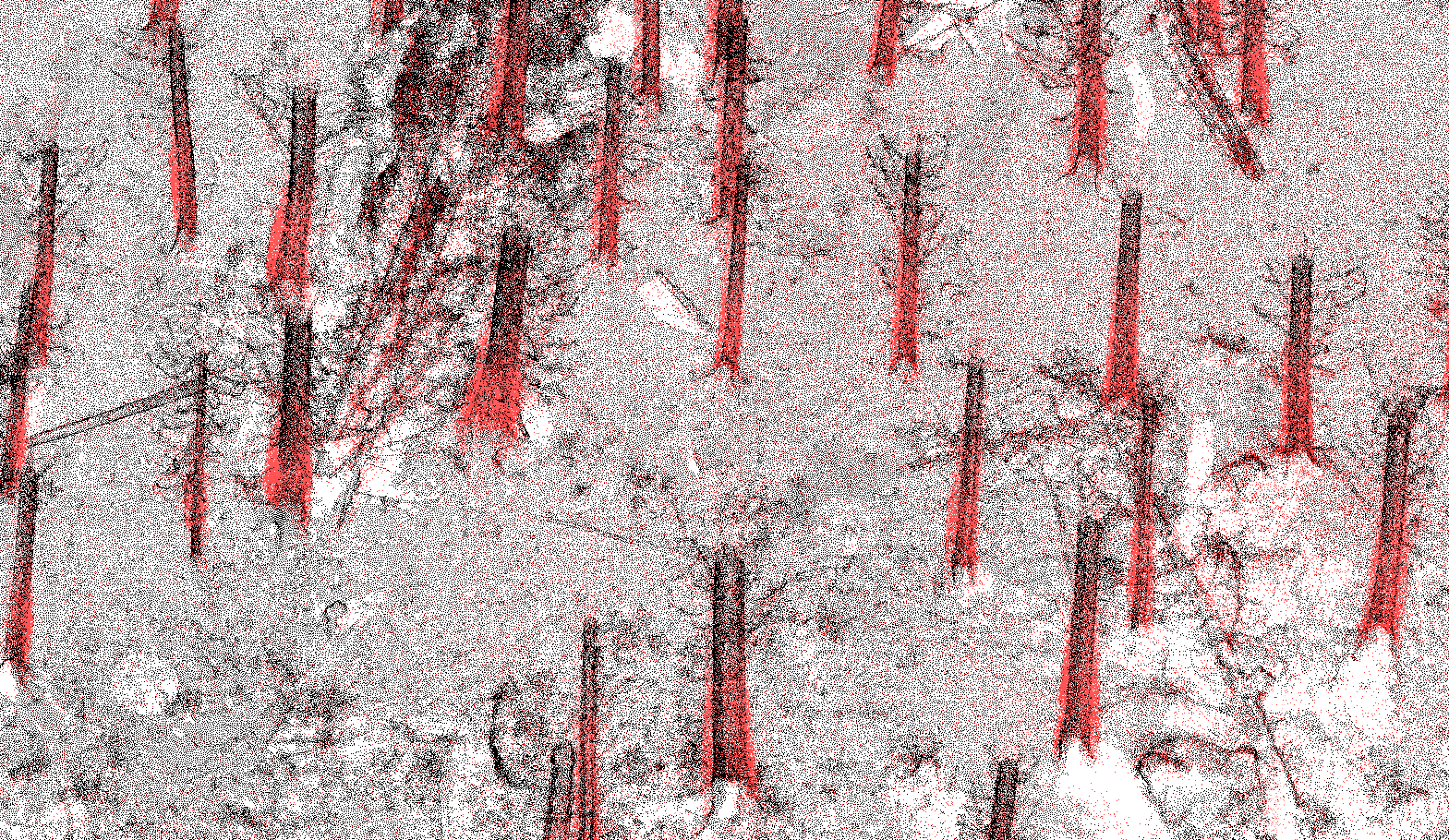}
     }
    \caption{Aligned forest maps. Ground truth map shown in black color where the cartographer map has been shown in red color. \textbf{\textit{Left:}} Top view of the aligned maps. \textbf{\textit{Right:}} Close up image of the aligned maps. Note how tree trunks nicely fit together. }
    \label{fig::registration_old_forest}
\end{figure*}

In addition to the urban environment, we collected the ground truth in two forest patches near Wangen an der Aare in Switzerland. Again, the Leica RTC360 3D scanner was used for generating the map. The cartographer map was registered against the ground truth map using Cloudcompare software. For the first forest patch, the RMS registration error was \SI{0.042}{\meter}, and the distance between registered point clouds was $0.06 \pm 0.05$ meters. Dimensions of the forest area are 243 x 257 x 43 meters (W x L x H). The aligned maps are shown in Fig.~\ref{fig::registration_old_forest}. The close-up shot shown in Fig.~\ref{fig::well_aligned_old_forest} shows a snug fit between the aligned maps. One can also notice how the cartographer map is significantly noisier than the ground truth map from the RTC scanner. This comes to no surprise since RTC360 comes with more than an order of magnitude more accurate distance measurements (less than \SI{5}{\milli \meter} at 40 meters) compared to the Ouster range sensor (\SI{5}{\centi \meter} at 40 meters).

\ac{ICP} was used to register \ac{LIDAR} scans against the ground truth map and create a ground truth trajectory. We compare it to the pseudo ground truth from the cartographer. Overall translational error was $0.05 \pm 0.03$ meters and the overlaid trajectories are shown in Fig.~\ref{fig::old_forest_localization}. This is an order of magnitude lower error compared to the urban environment, which can be explained by the fact that we collected the data in succession, and hence both maps look the same. This allows for better scan registration. Additionally, the forest areas are smaller, which also leads to better accuracy. We verified the obtained error in another forest patch with younger trees (map size 100 x 96 x 17 meters). Comparing the map with the ground truth map, we obtained similar numbers: the RMS registration error was \SI{0.078}{\meter}, the distance between the registered maps was $0.12 \pm 0.1$ meters. The error between the ground truth trajectory and pseudo-ground truth from the cartographer was $0.04 \pm 0.03$ meters over the \SI{100.34}{\meter} distance traveled (images and a map of the forest patch omitted for brevity).

Both cartographer's localization and the \ac{ICP} can run in real-time on the robot, however \ac{ICP} consumes less \ac{CPU}. The comparisons between cartographer's localization mode and pure \ac{ICP} based localization suggest the superior performance of \ac{ICP}. Hence, we would like to conclude that one could also use \ac{ICP} for localization during harvesting missions. The authors will switch to the \ac{ICP} based localization in the future. Our \ac{ICP} implementation is available as open-source for the community\footnote{\url{https://github.com/leggedrobotics/icp_localization}}.

\FloatBarrier

\subsection{Fully integrated system}
\label{sec::fully_integrated_system}
\begin{figure}[tbh]
\centering
    \includegraphics[width=\textwidth]{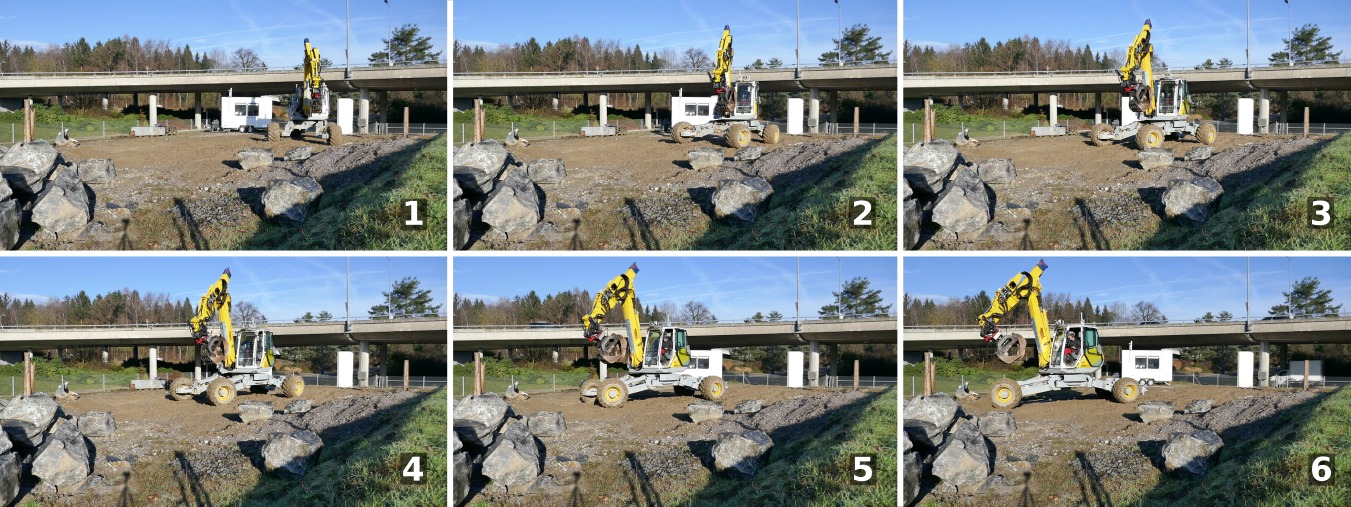}
    \caption{\ac{HEAP} approaching wooden logs at our testing field. We do not show the rest of the maneuver for the sake of brevity. The reader is encouraged to watch the video.}
    \label{fig::approach_hoengg}
\end{figure}

\begin{figure}[tbh]
\centering
    \includegraphics[width=\textwidth]{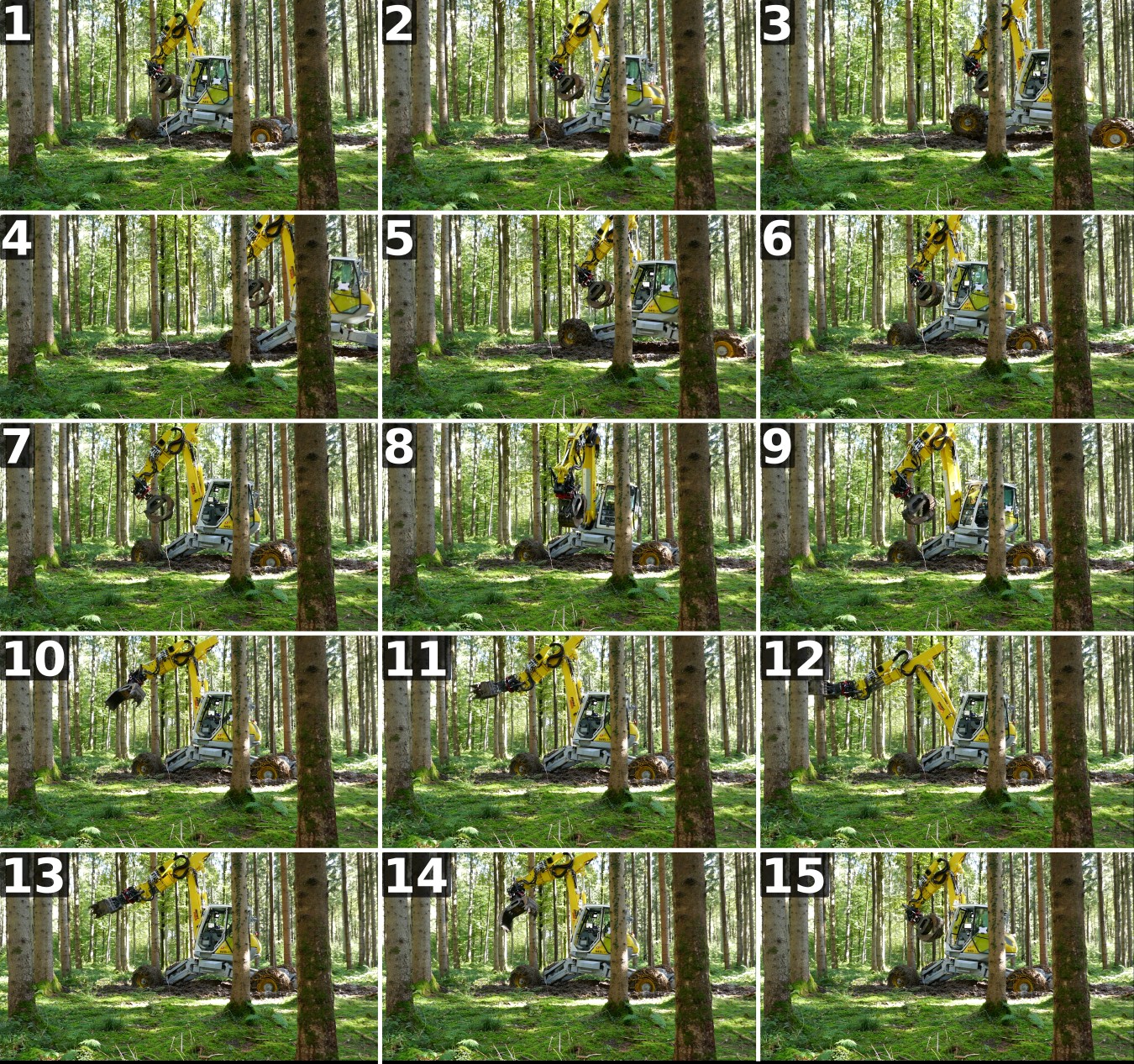}
    \caption{Snapshots of an operational cycle between grabbing two trees. \ac{HEAP} re-positioning the base close to the tree (snapshots 1-6) followed by scanning the environment and building a local map (snapshots 7-9). Finally, \ac{HEAP} grabs the tree by extending the arm (snapshots 10-12) and retracting it (snapshots 13-15). After retracting the arm, the harvester is ready to grab the next tree and whole cycle repeats. }
    \label{fig::whole_sequence}
\end{figure}

This section presents snapshots of a fully integrated system performing a harvesting mission. We tested our system on a test field where \ac{HEAP} was commanded to grab ``tree trunks''. One tree trunk is composed of three wooden logs strapped together such that they can stand straight up (see the video or Fig.~\ref{fig::approach_hoengg}). We show the approaching sequence on our test field in Fig.~\ref{fig::approach_hoengg} and omit the rest of the maneuver for the sake of brevity since the full maneuver is already shown in Fig.~\ref{fig::whole_sequence}. In addition to our test field, we have performed the second set of experiments in a small forest alley in a real forest with fully grown adult trees. We show snapshots of 1 operational cycle between grasping two trees in Fig.~\ref{fig::whole_sequence}. The reader is encouraged to watch a video accompanying this submission since it offers a better insight\footnote{\url{https://youtu.be/1FLD0djPFgU}}.

It takes about 1-2.5 minutes for the machine to complete the entire cycle, depending on how far the cabin has to turn and how far it has to drive. The cabin turns and driving are tuned conservatively (slow), and the time could be improved drastically by speeding them up. The proposed system was able to run without any changes in both the test field and the forest. Furthermore, we were able to successfully detect both tree trunks and wooden logs and perform a grab. A human operator was often inside the cabin for safety.

\FloatBarrier
\section{Conclusions \& Discussion}
\label{sec::conclusion_outlook}
We present a first attempt at performing automated forestry operations using a full-sized harvester to the best of our knowledge. The approach presented is targeted for precision harvesting. Components off our system have been evaluated individually and integrated into a complete harvesting system. We show evaluation of the control system, approach pose planner, mapping accuracy, localization accuracy, point cloud to elevation map conversion algorithm, and the tree detector.

A lightweight and versatile sensor module is presented; it can be used for mapping and later mounted on the harvester for localization. The sensor module and \ac{SLAM} system are used to map the mission area a priori. Subsequently, the \ac{HEAP} harvester can localize itself at mission time, which enables it to navigate under the forest canopy without relying on \ac{GPS} signal.

Our system can plan approach poses in confined spaces, and it can negotiate challenging terrain by combining the chassis balancing controller with the path following controller. The developed planner relies on traversability maps which we estimate from the elevation maps. Elevation maps are calculated directly from point clouds using our conversion algorithm. In this work, we rely on a human expert to specify target trees for harvesting. To combat the localization error and enable precision harvesting, we plan grasping poses in the local frame using a geometric detection algorithm operating on the point cloud assembled from the laser sensor. Lastly, we make parts of our planning\footnote{\url{https://github.com/leggedrobotics/se2_navigation}}, mapping\footnote{\url{https://github.com/ANYbotics/grid_map/tree/master/grid_map_pcl}}, localization\footnote{\url{https://github.com/leggedrobotics/icp_localization}} and tree detection\footnote{\url{https://github.com/leggedrobotics/tree_detection}} software stack open source for the community.

Each of the modules (e.g., planning, mapping) has been first tested and benchmarked individually before integrating them into a complete precision harvesting mission.  For example, we first tested planning and control with \ac{RTK} \ac{GPS} before adding the \ac{SLAM} system. Stepwise integration was essential to isolate problems. We also note that having a simulation environment was beneficial, since we could first test the functionality in simulation and then focus on tuning on the machine. E.g. the behavior of the state machine can be almost completely verified in simulation. Another lesson learned from this work was that state machines get complicated when one wants to include recovery behavior (e.g. re-plan if no solution found). In the future, it would be beneficial too look into using behavior trees \cite{winter2010integrating} which are typically more compact and have implicitly integrated recovery behaviors.

Experimental verification of our approach during the wet season was unfortunate, and the biggest problem occurring was that the machine would occasionally get stuck in the mud. This problem was somewhat mitigated by using chains and putting wooden logs and branches in the mud to improve traction. We do not see this as a limitation of our system since forestry operations typically find a place during dry seasons.

We point out that pure geometry-based localization (or detection) can exhibit failures in the forest. For example, a crowd of humans observing the machine can look very similar to tree trunks in a point cloud, which on one occasion caused the \ac{SLAM} to produce spurious pose estimates. Furthermore, we had a few cases of sun rays directly shining on our camera, which caused the visual odometry to diverge. Subsequently, the whole state estimation pipeline diverged despite the pure \ac{LIDAR} odometry working well. This seeks an approach that can adequately detect sensor degeneration to ensure robust mapping and localization in all cases. We also believe that it is important to perform a quantitative evaluation of different \ac{SLAM} systems under the same conditions in forest areas. This would involve testing different approaches (e.g. LOAM variants, Cartographer, \ac{ICP}) in different forest styles with a high accuracy ground-truth map (e.g. from the 3D scanner). Hence, the community could focus the development efforts on the most promising approach.

An immediate improvement to the localization system is to use \ac{ICP} as opposed to Cartographer's localization mode (as discussed in Section~\ref{sec::res_localization}). Other improvements will focus on utilizing complementary sensors where applicable and implementing health checking systems to mitigate state estimation problems. For example, one could run visual and \ac{LIDAR} based \ac{SLAM} to increase robustness since these two kinds of sensors are complementary to each other (similar as in \cite{khattak2020complementary}). Furthermore, traversability estimation should use visual information since pure geometry can be misleading in natural environments. Tree detection in the local map could also benefit from using visual information. In case of thick vegetation or heavy branch clutter, one could also look into using radar sensors for detecting a grabbing pose on a tree stem. Robust treatment of heavy clutter would be especially beneficial, since this is the main reason for the tree detection system failure.

We plan to improve the approach pose planner by incorporating the footprint adaptation and adjusting the number of generated approach poses based on the environment, as noted in Section \ref{sec::planning}. Furthermore, it would be interesting to benchmark different planners (e.g., \ac{PRM} or hybrid A* algorithm \cite{dolgov2010path}). Additionally, the implementation of approach pose generation can be parallelized. For tree felling, the arm planner can be enhanced to include the tree geometry once the gripper is holding a tree trunk. Lastly, the tree trunk weight could be included in the arm controller to improve the arm plan tracking.

Besides improving the mission execution and motion planners, an exciting area to pursue is to develop a more intelligent mission planner. Instead of just relying on the order of execution provided by the user, one could optimize some objective (e.g., minimal time). Such a planner becomes especially relevant in a scenario involving multiple harvesters.

Apart from algorithmic improvements, one crucial aspect for future forestry missions is to have rugged sensors. We used an umbrella to protect our sensor module from water during the experiments (see Fig.~\ref{fig::menzi_forest}, in the back of the machine); however, this is not a viable solution for a fully autonomous harvester operating in a harsh environment. Furthermore, the sensors must be protected from branches, vegetation clutter, and branch debris falling onto the machine. Since harvester machines need to deal with clutter and vegetation, it is advisable to have redundant sensors such that localization can be robust to occlusions.

Incorporating the improvements outlined above will bring us one step closer to fully automated precision forestry that will increase yields and relieve humans from tedious labor.

\FloatBarrier
\section*{Acknowledgement}
This research was partially supported by the European Research Council (ERC) under the European Union’s Horizon 2020 research and innovation programme grant agreement No 852044 and through the SNSF National Centre of Competence in Digital Fabrication (NCCR dfab). We would also like to thank Alexander Reske for his help with Leica RTC 360 3D laser scanner during the data collection. We also thank to Silvere and Simo Gr\"{o}hn for their help with Cartographer.

\appendix
\section{Appendix}
\subsection{Point Cloud to Elevation Map Conversion}
\label{sec::apendix_pointcloud_to_map}
We analyze how the point cloud conversion algorithm presented in Section~\ref{sec::mapping} reacts to clutter and vegetation. The data was collected with a handheld sensor, and we build small maps of forest patches. The forest patches vary in dimension between 17x18 meters up to 29x32 meters (width x length). We convert the point cloud to the elevation map and show the results below. Unfortunately, we cannot provide rigorous quantitative analysis since the ground truth (actual ground elevation) is extremely hard to collect in the presence of vegetation. Furthermore, it is hard to define the ground truth since there are cases where we want more than just ground information in the elevation map (e.g., trees, stumps). However, looking at the images and maps presented below, one can draw some conclusions about the conversion algorithm's behavior.

\begin{figure*}[tbh]
\centering
    \subfloat[Forest patch \label{fig::patch1}]{
       \includegraphics[width=0.27\textwidth]{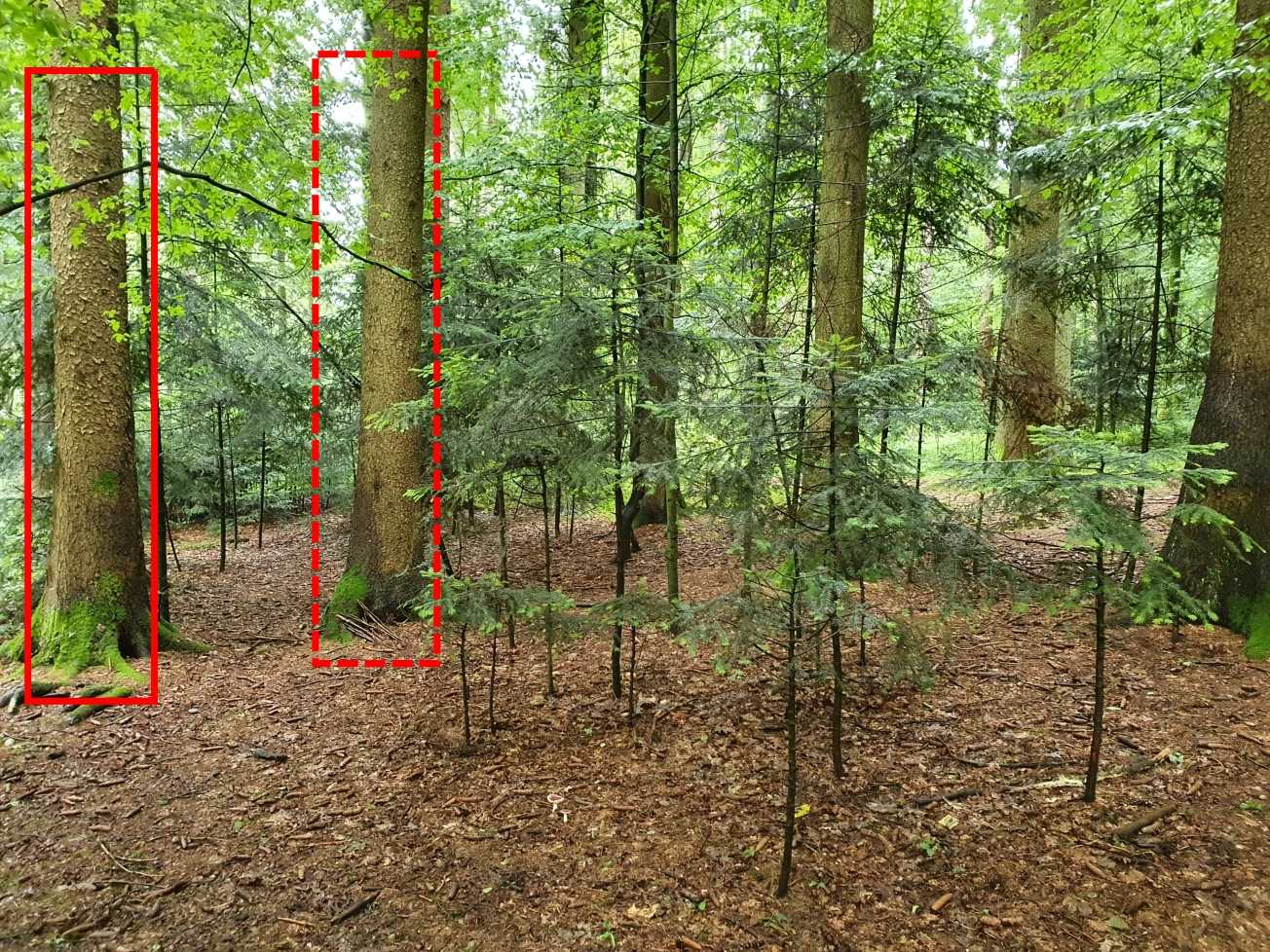}}
     \subfloat[Elevation map \label{fig::map1}]{
       \includegraphics[width=0.36\textwidth]{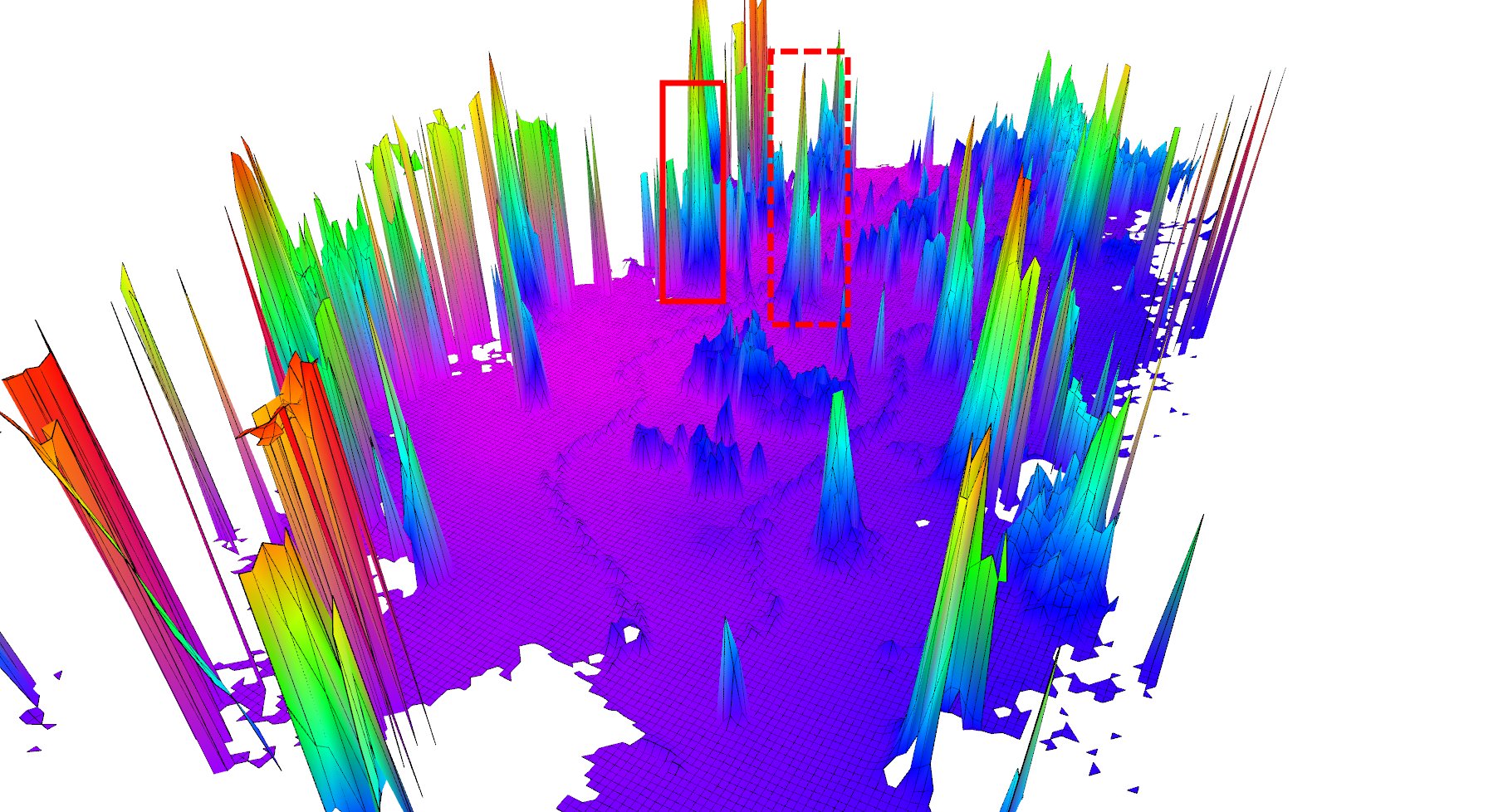}}
     \subfloat[Elevation map + point cloud \label{fig::overlay1}]{
       \includegraphics[width=0.36\textwidth]{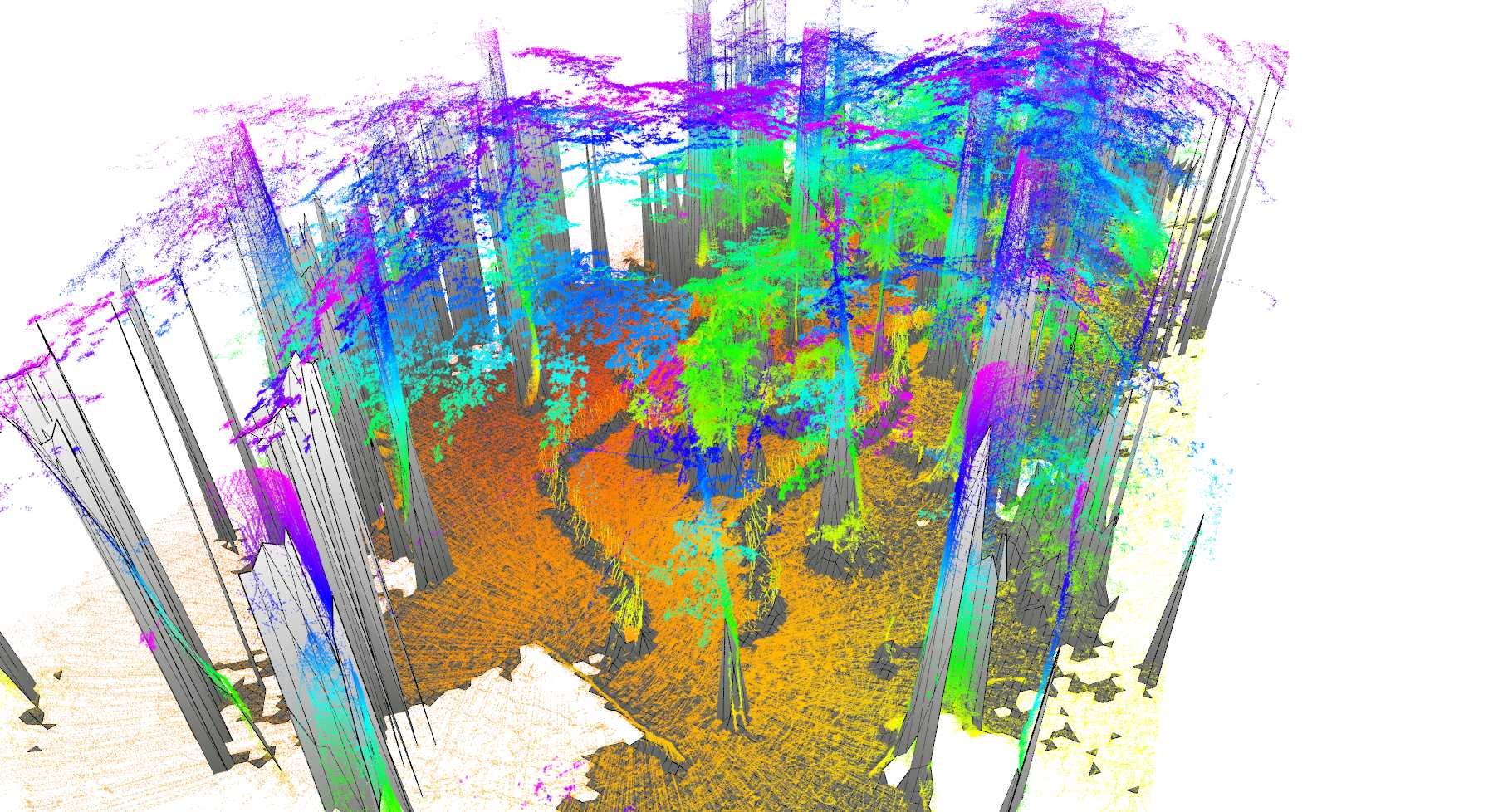}}
    \caption{Scene with no clutter, but higher branch density compared to the previous scene. Besides the big tree trunks many smaller trees with trunk diameter less than \SI{10}{\centi \meter} are present. One can clearly see a limitation of our algorithm, since many of the smaller trees are not well captured in the map at \SI{10}{\centi \meter} resolution.}
    \label{fig::conversion1}
\end{figure*}

\begin{figure*}[tbh]
\centering
    \subfloat[Forest patch \label{fig::patch3}]{
       \includegraphics[width=0.27\textwidth]{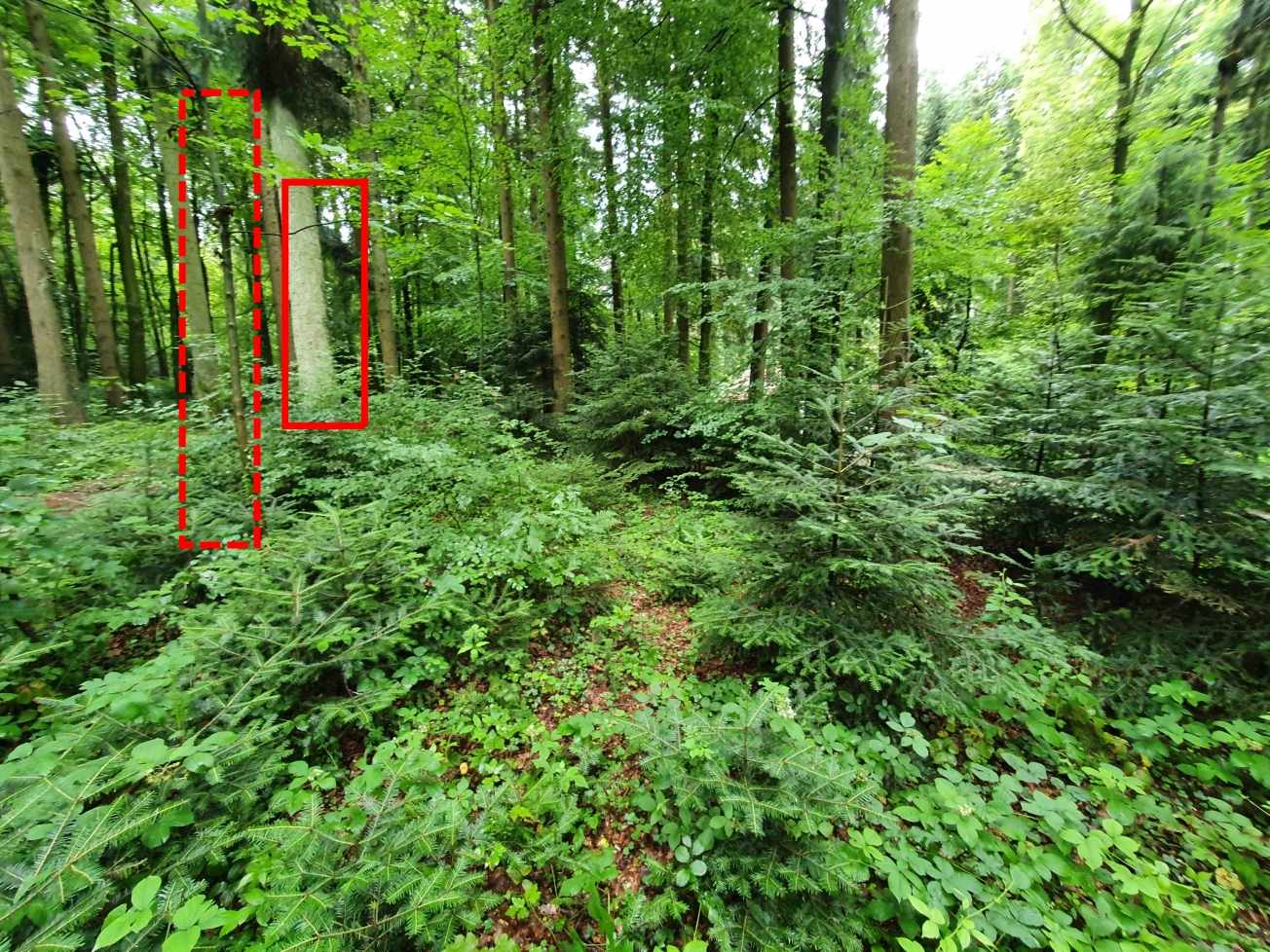}}
     \subfloat[Elevation map \label{fig::map3}]{
       \includegraphics[width=0.36\textwidth]{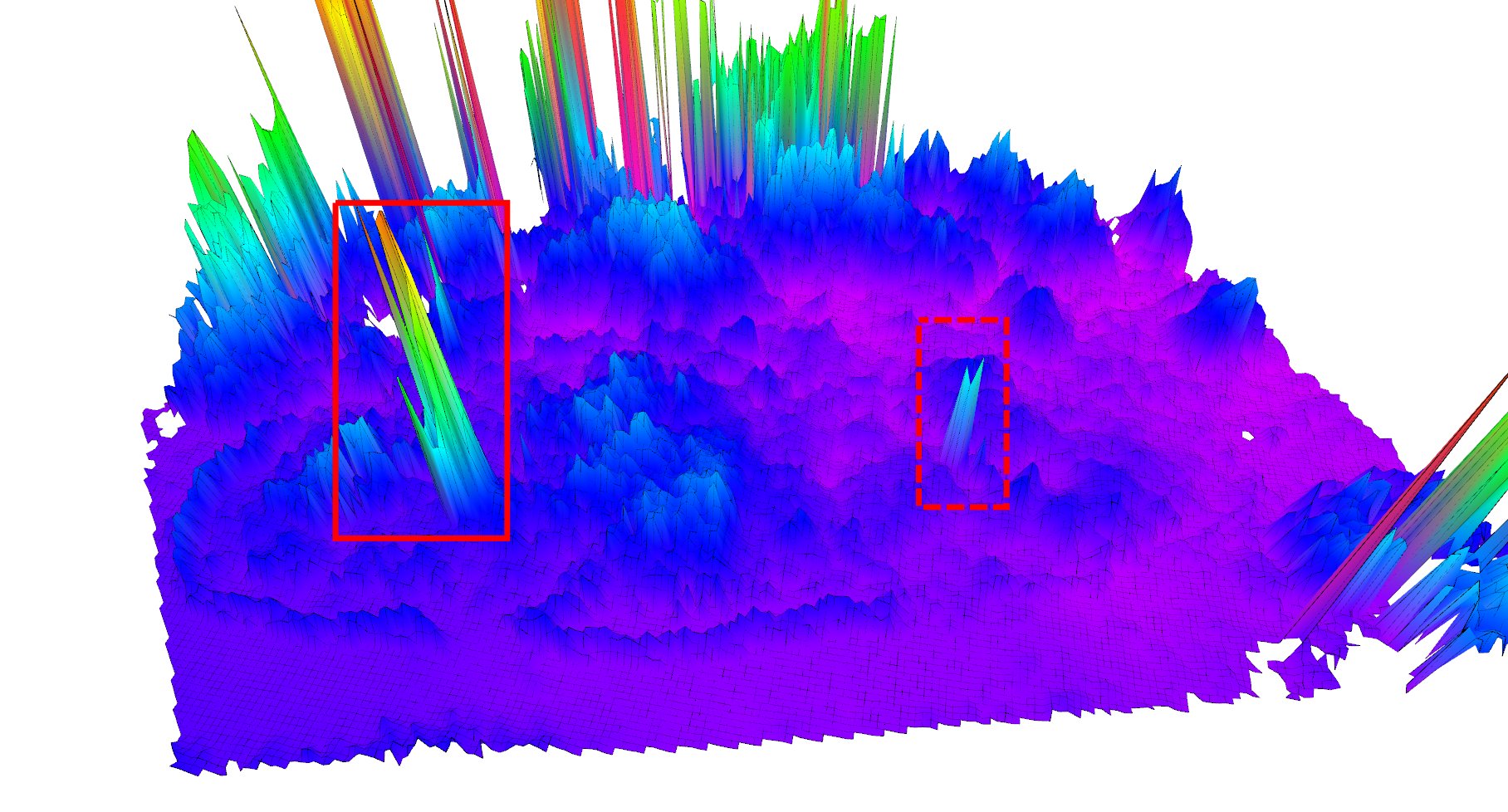}}
     \subfloat[Elevation map + point cloud \label{fig::overlay3}]{
       \includegraphics[width=0.36\textwidth]{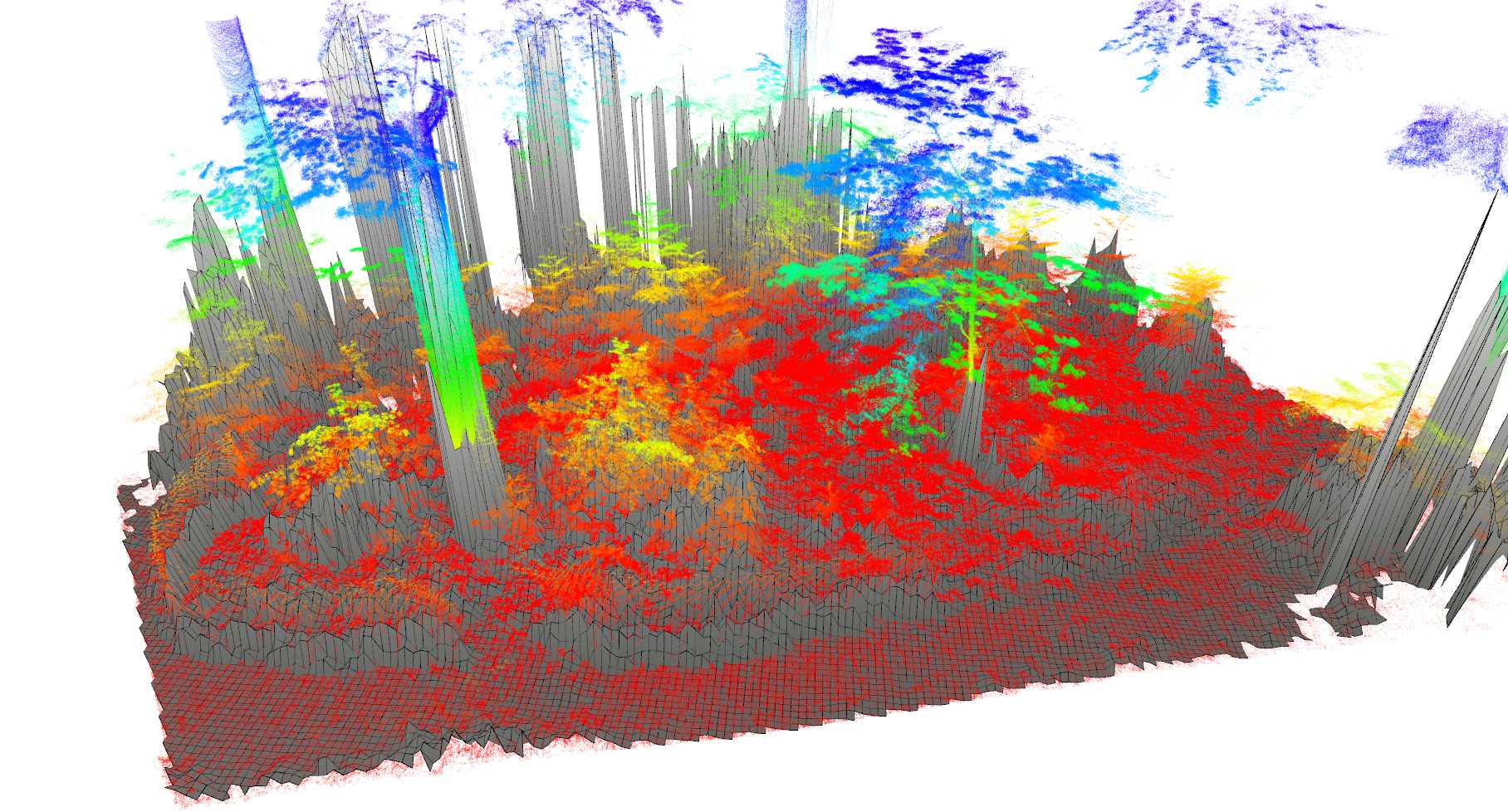}}
    \caption{High clutter scene with lots of vegetation. We can see that the thick vegetation corrupts the elevation map and causes bumps and roughness in it. Thick tree trunks are still well visible in the map.}
    \label{fig::conversion3}
\end{figure*}

All the elevation maps shown below have a resolution of \SI{10}{\centi \meter}. One can observe (especially from Fig.~\ref{fig::conversion1}) that thin tree trunks present a problem for the elevation map and are often not well captured in it. One could increase the map's resolution to capture them better; however, there is a trade-off between detail and computation time. Furthermore, to avoid holes and artifacts, one needs denser point clouds with the increasing elevation map resolution (smaller cell size). Dense branches and low and medium clutter do not seem to pose a big problem, and the resulting elevation maps have no or low artifacts. Thick tree trunks (\SI{30}{\centi \meter} diameter and more) are visible well in the map, and we did not experience any problems with the tree canopy or branches. This holds as long as forest ground is captured in the point cloud. One limitation of the approach is the ability to deal with heavy clutter (e.g., Fig.~\ref{fig::conversion3}). Indeed thick vegetation and branches touching the ground will often appear as blobs in the elevation map.

\begin{figure*}[tbh]
\centering
    \subfloat[Map computed with higher cluster tolerance \label{fig::patch3_high_cluster_tolerance}]{
       \includegraphics[width=0.45\textwidth]{figures/data3_map_rainbow.jpg}}
     \subfloat[Map computed with lower cluster tolerance \label{fig::patch3_low_cluster_tolerance}]{
       \includegraphics[width=0.45\textwidth]{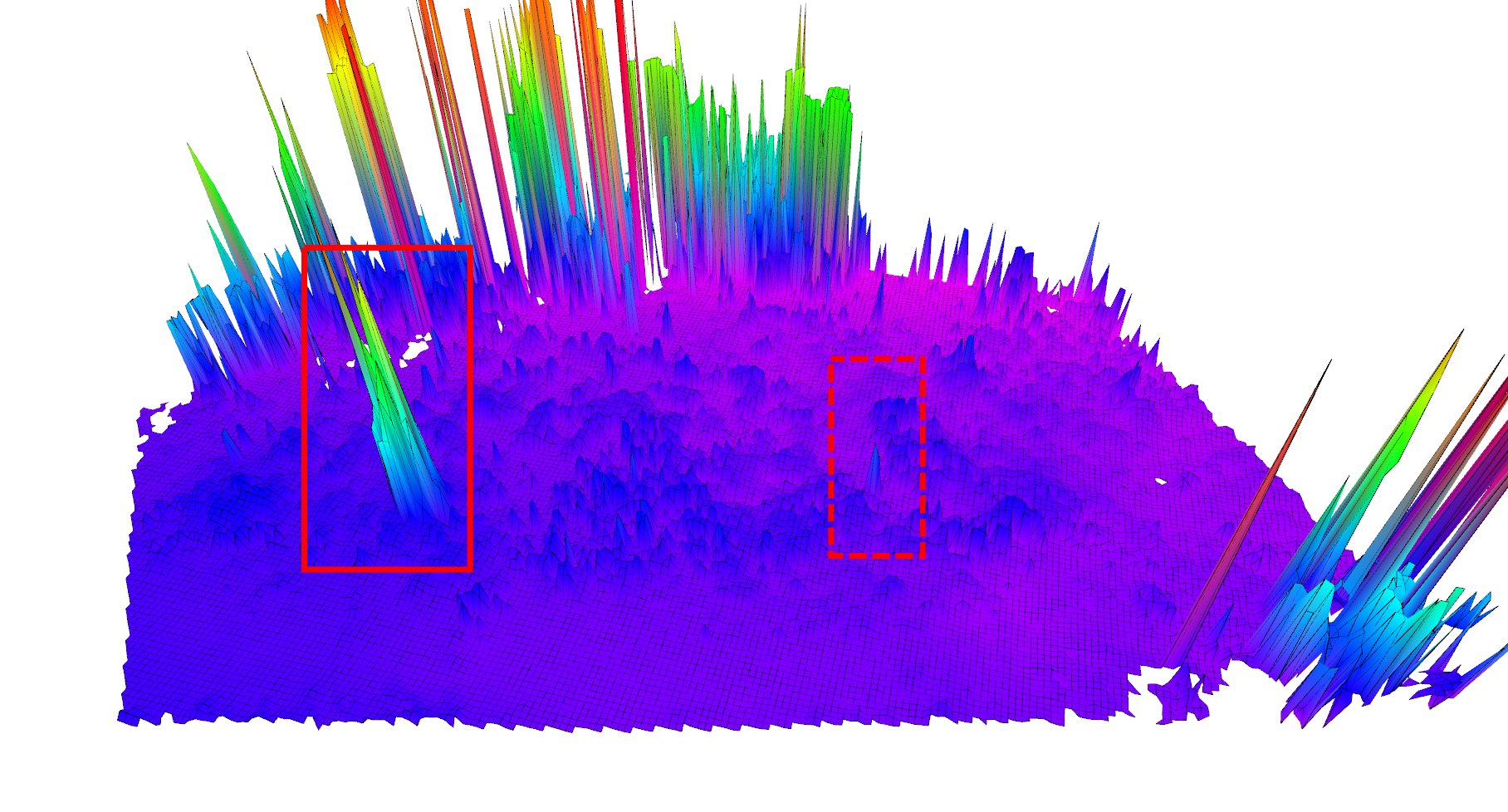}}
    \caption{Elevation maps computed using our point cloud to elevation map conversion algorithm with different cluster tolerance parameter. One can observe that the lower cluster tolerance better filters out the vegetation and clutter.}
    \label{fig::cluster_tolerance_effect}
\end{figure*}

In Fig.~\ref{fig::cluster_tolerance_effect} we illustrate the influence of a cluster tolerance parameter on the elevation map. Both maps correspond to the cluttered forest patch shown in Fig.~\ref{fig::patch3}. One can note that the lower value of cluster tolerance better filters out the clutter and recovers the ground elevation. However, there is a trade-off since it also filters smaller trees out of the map (denoted with a red rectangle). Large tree trunks remain unaffected; in case of a cluttered environment and larger tree trunks, smaller values are recommended. In a case the point cloud is very dense, smaller values usually yield better results.

\begin{figure*}[tbh]
\centering
    \subfloat[Forest patch \label{fig::patch7}]{
       \includegraphics[width=0.27\textwidth]{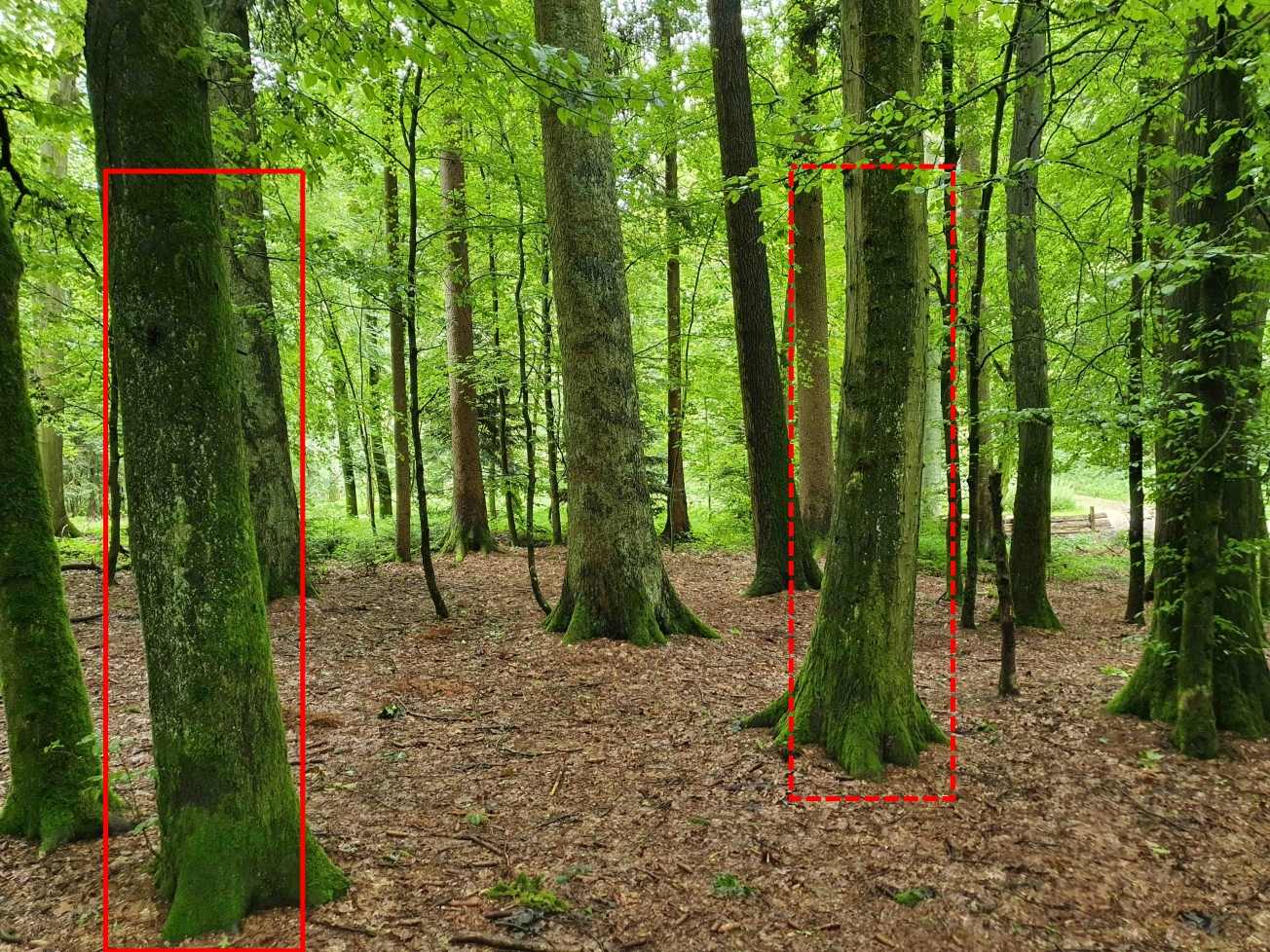}}
     \subfloat[Elevation map \label{fig::map7}]{
       \includegraphics[width=0.36\textwidth]{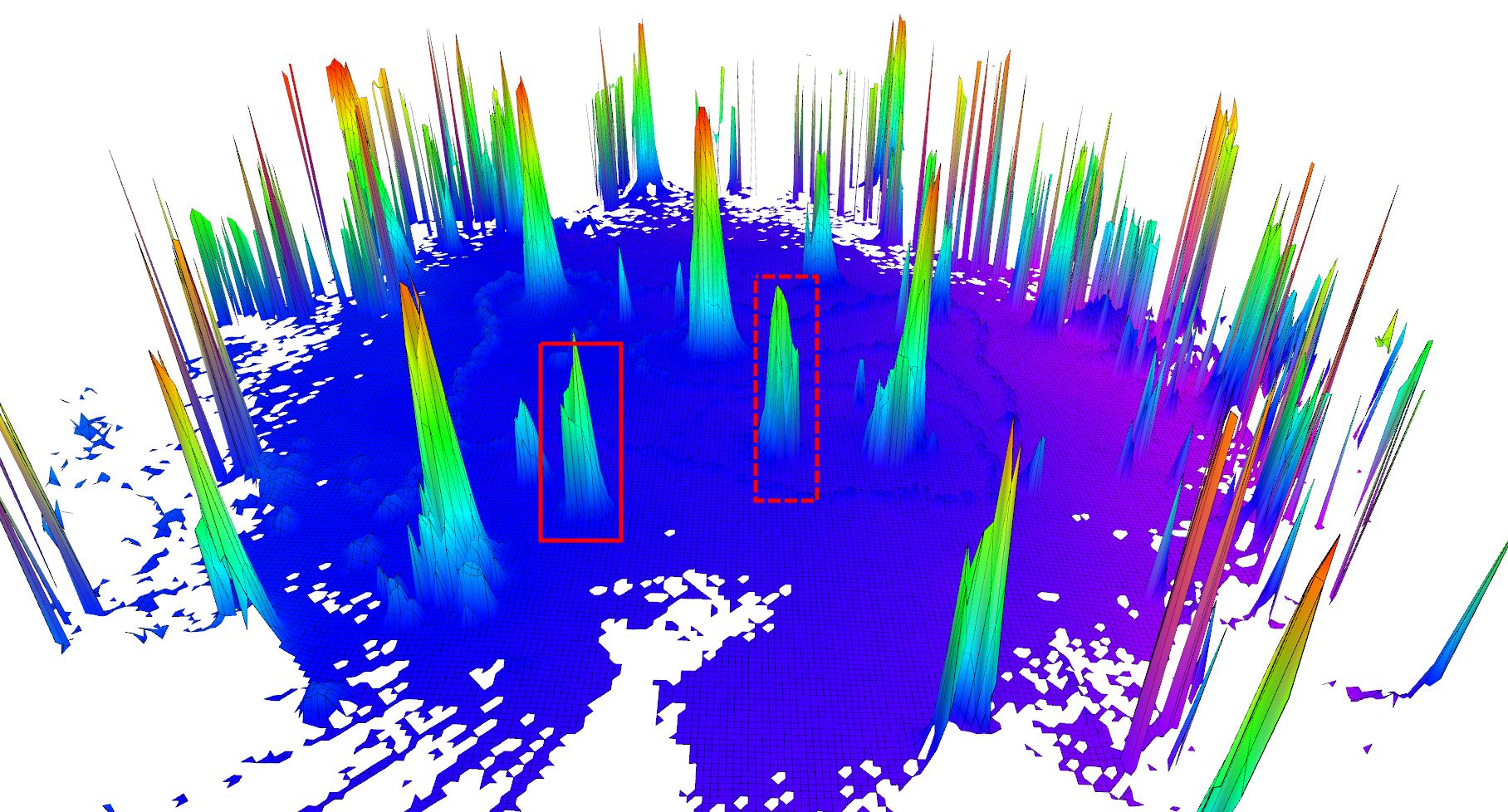}}
     \subfloat[Elevation map + point cloud \label{fig::overlay7}]{
       \includegraphics[width=0.36\textwidth]{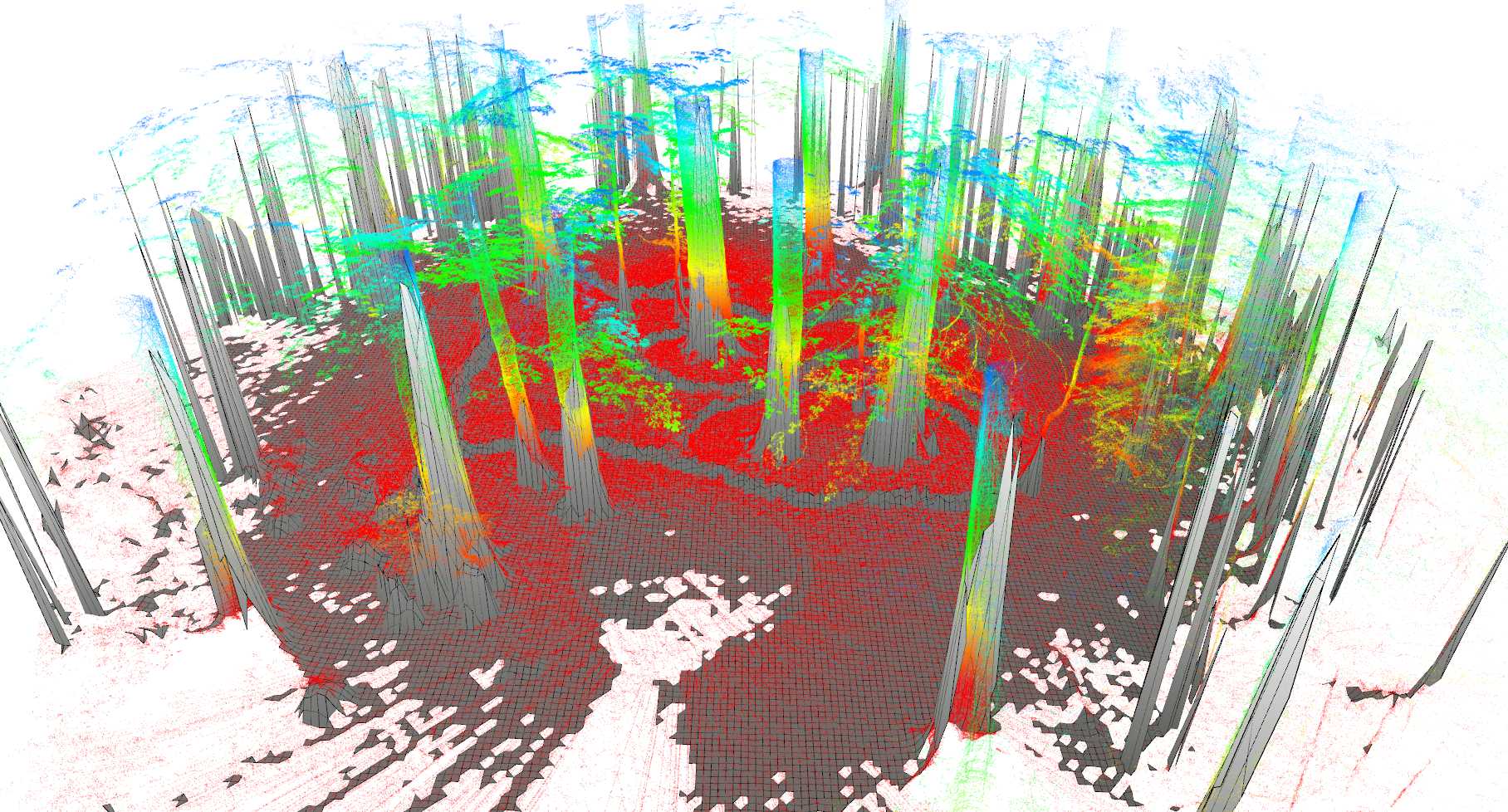}}
    \caption{Forest scene with large tree trunks, no clutter and low branch density. The elevation map nicely captures the tree trunks and the forest ground.}
    \label{fig::conversion7}
\end{figure*}
\begin{figure*}[tbh]
\centering
    \subfloat[Forest patch \label{fig::patch6}]{
       \includegraphics[width=0.27\textwidth]{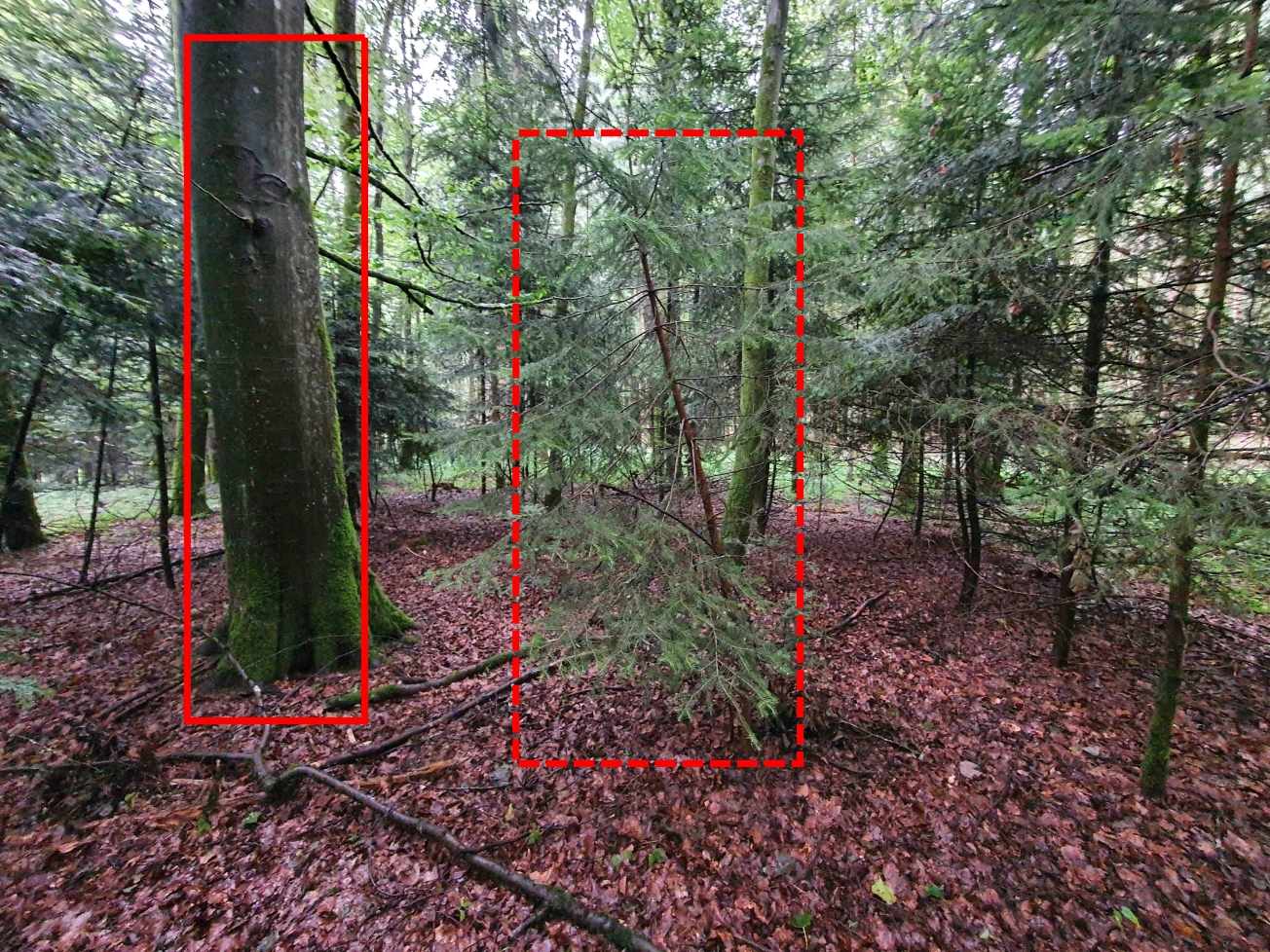}}
     \subfloat[Elevation map \label{fig::map6}]{
       \includegraphics[width=0.36\textwidth]{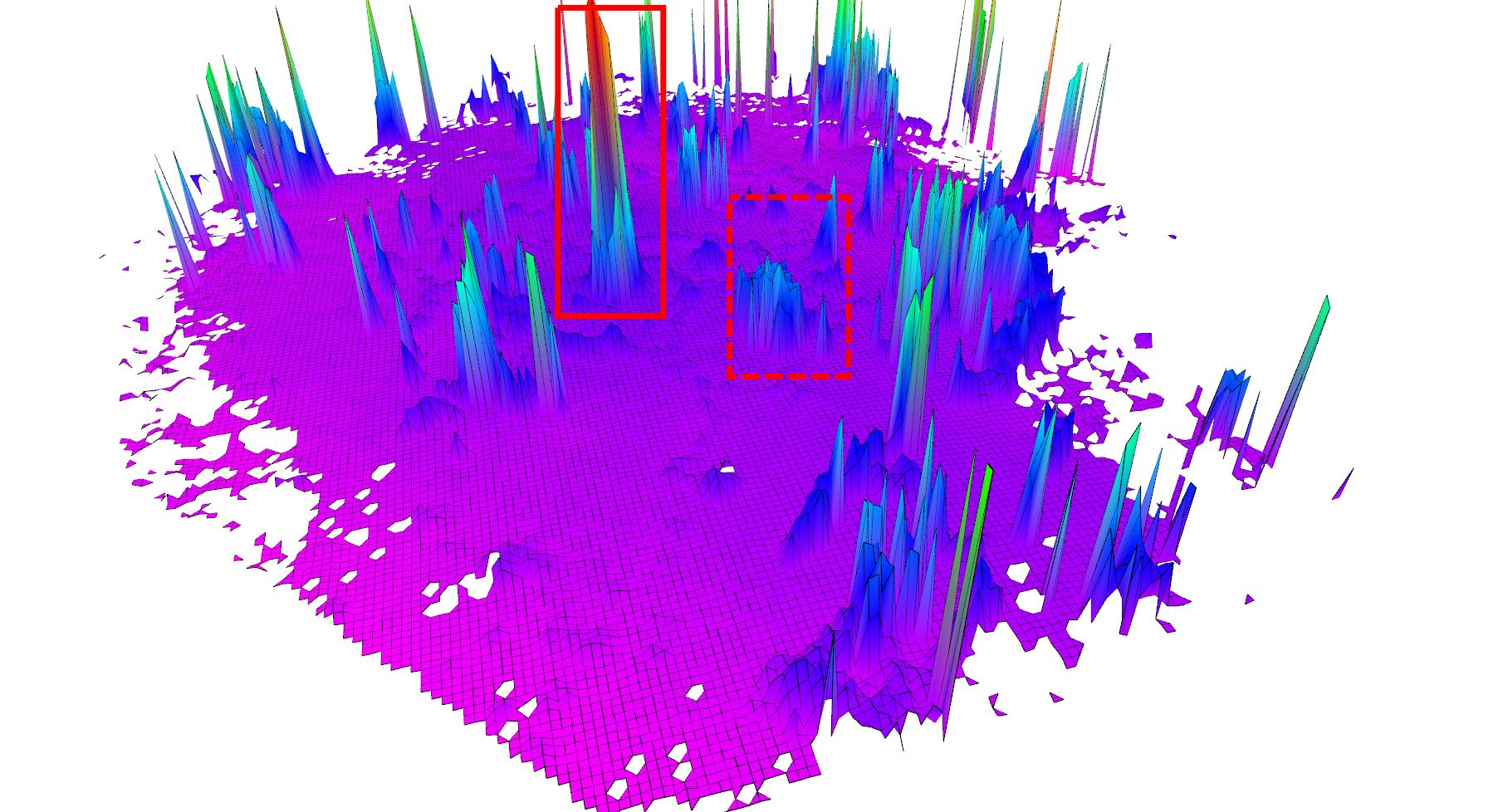}}
     \subfloat[Elevation map + point cloud \label{fig::overlay6}]{
       \includegraphics[width=0.36\textwidth]{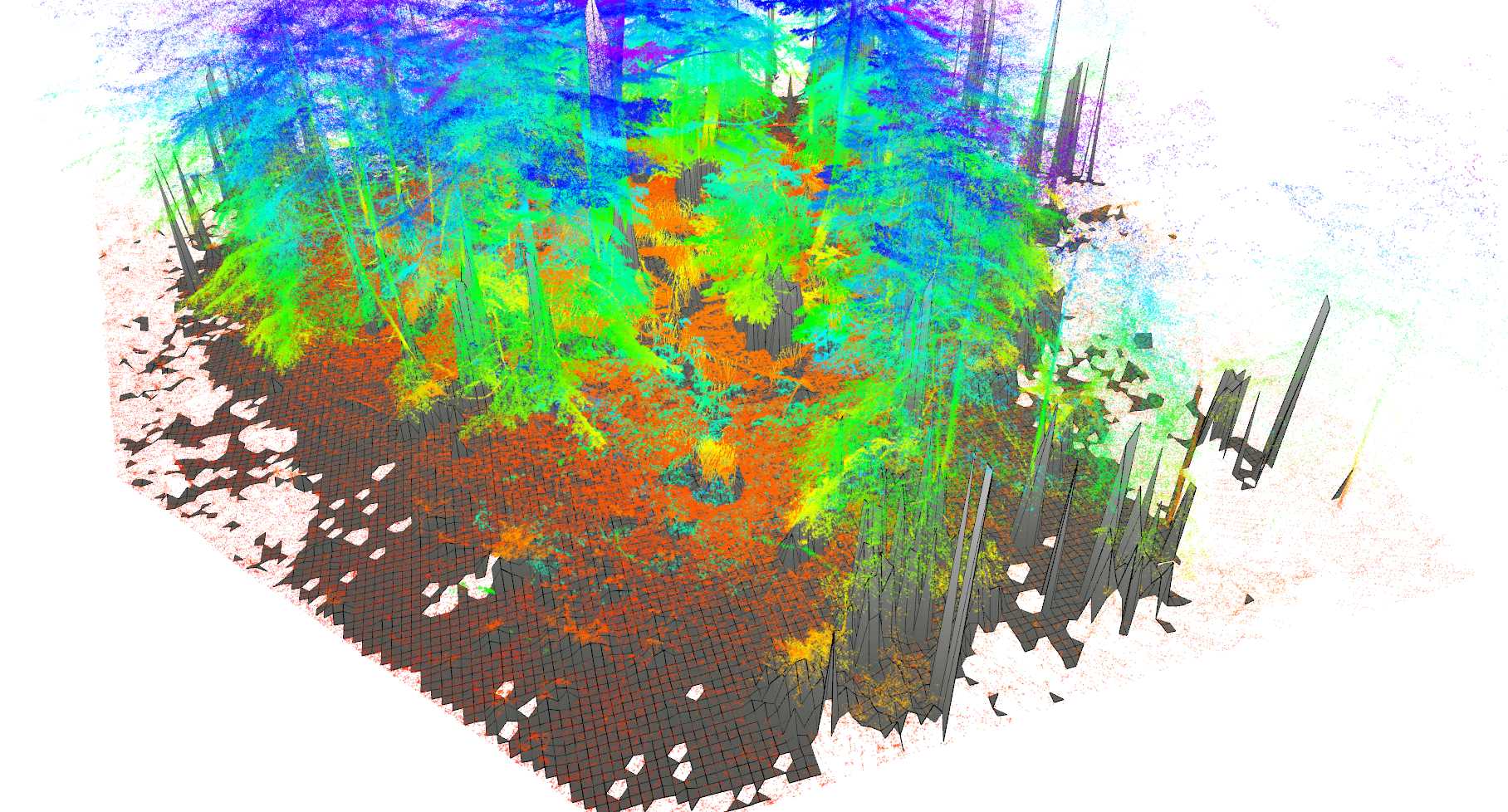}}
    \caption{Cluttered scene without vegetation. Compared to the Fig.~\ref{fig::conversion1},  tree trunks in this scene are thicker and better captured in the elevation map. High branch density does not seem to pose the problem for the conversion algorithm as long as there is enough ground data in the point cloud. }
    \label{fig::conversion6}
\end{figure*}
\begin{figure*}[tbh]
\centering
    \subfloat[Forest patch \label{fig::patch2}]{
       \includegraphics[width=0.27\textwidth]{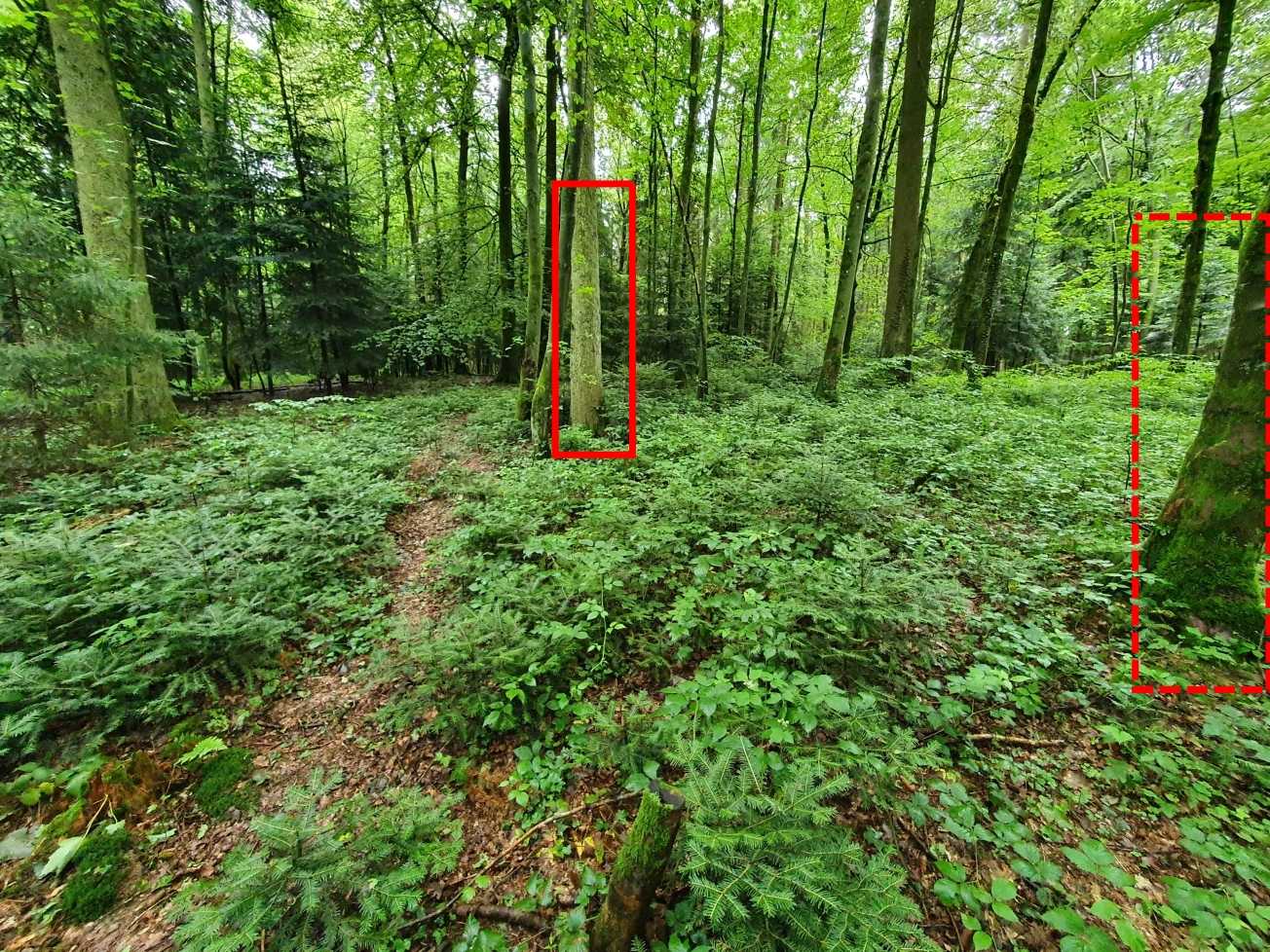}}
     \subfloat[Elevation map \label{fig::map2}]{
       \includegraphics[width=0.36\textwidth]{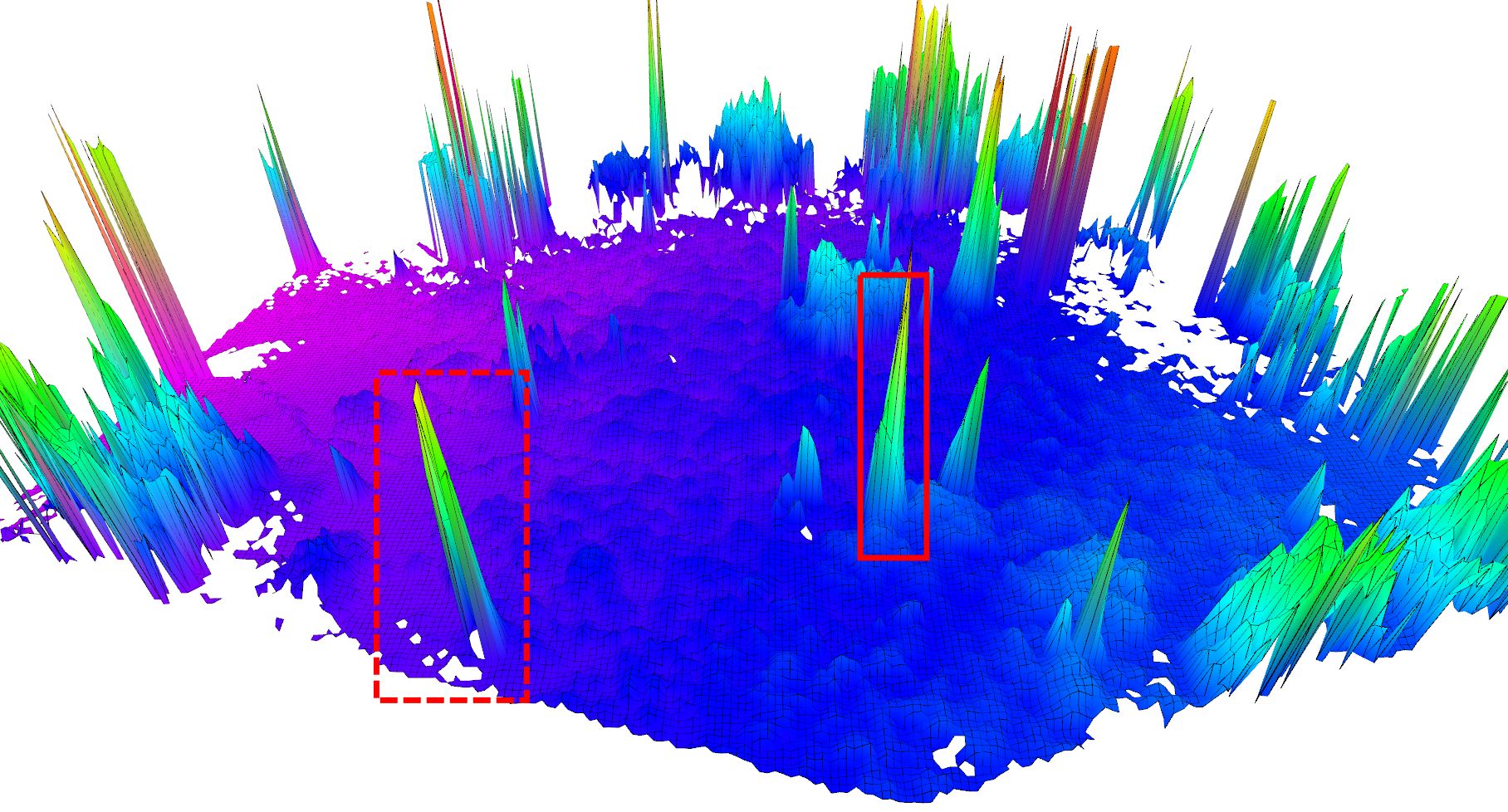}}
     \subfloat[Elevation map + point cloud \label{fig::overlay2}]{
       \includegraphics[width=0.36\textwidth]{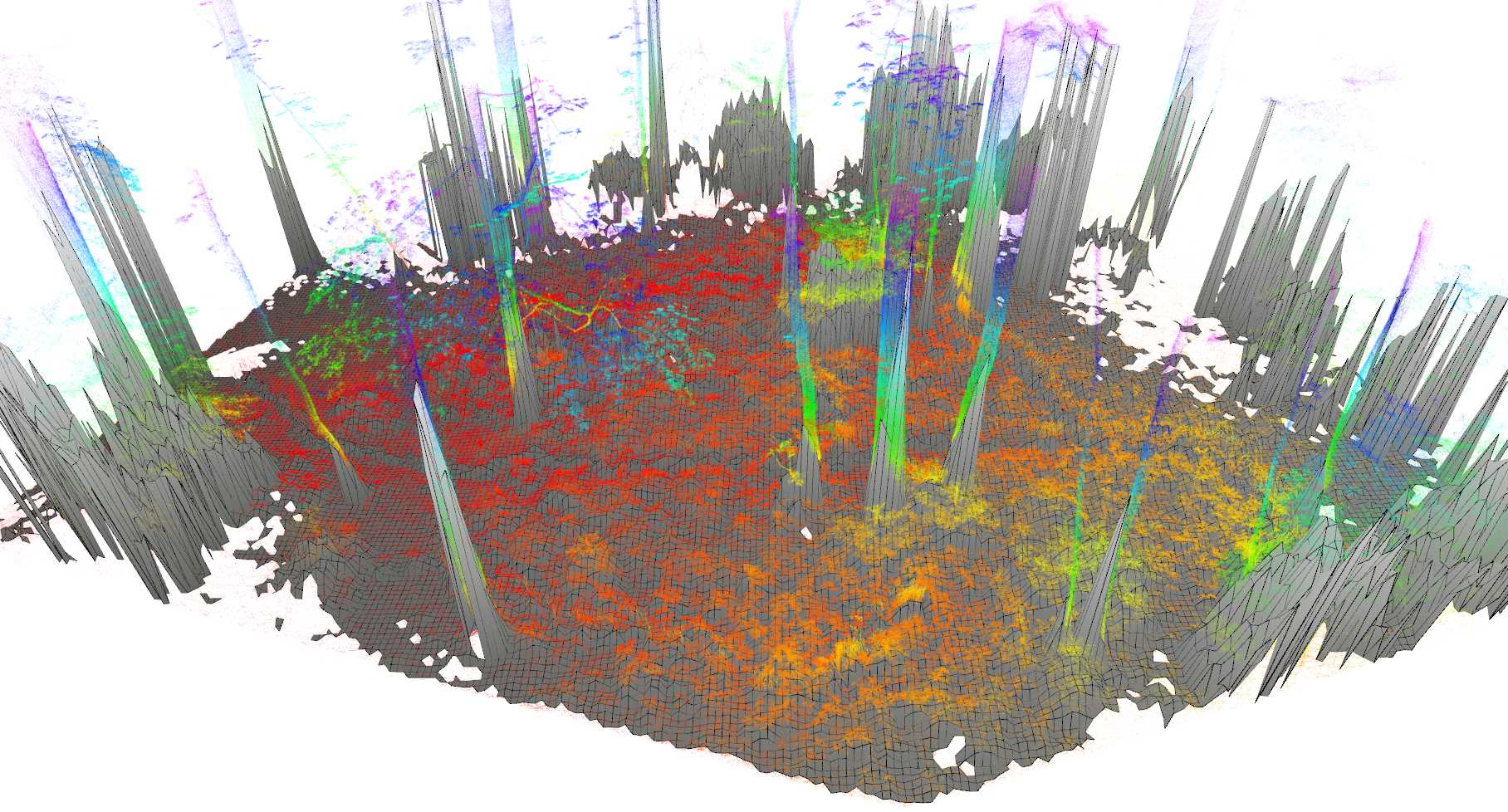}}
    \caption{Scene with low vegetation and low clutter. One can see that the elevation map does not contain any artifacts. The vegetation causes the map to be somewhat rougher compared to the forest ground without the vegetation. }
    \label{fig::conversion2}
\end{figure*}
\begin{figure*}[tbh]
\centering
    \subfloat[Forest patch \label{fig::patch4}]{
       \includegraphics[width=0.27\textwidth]{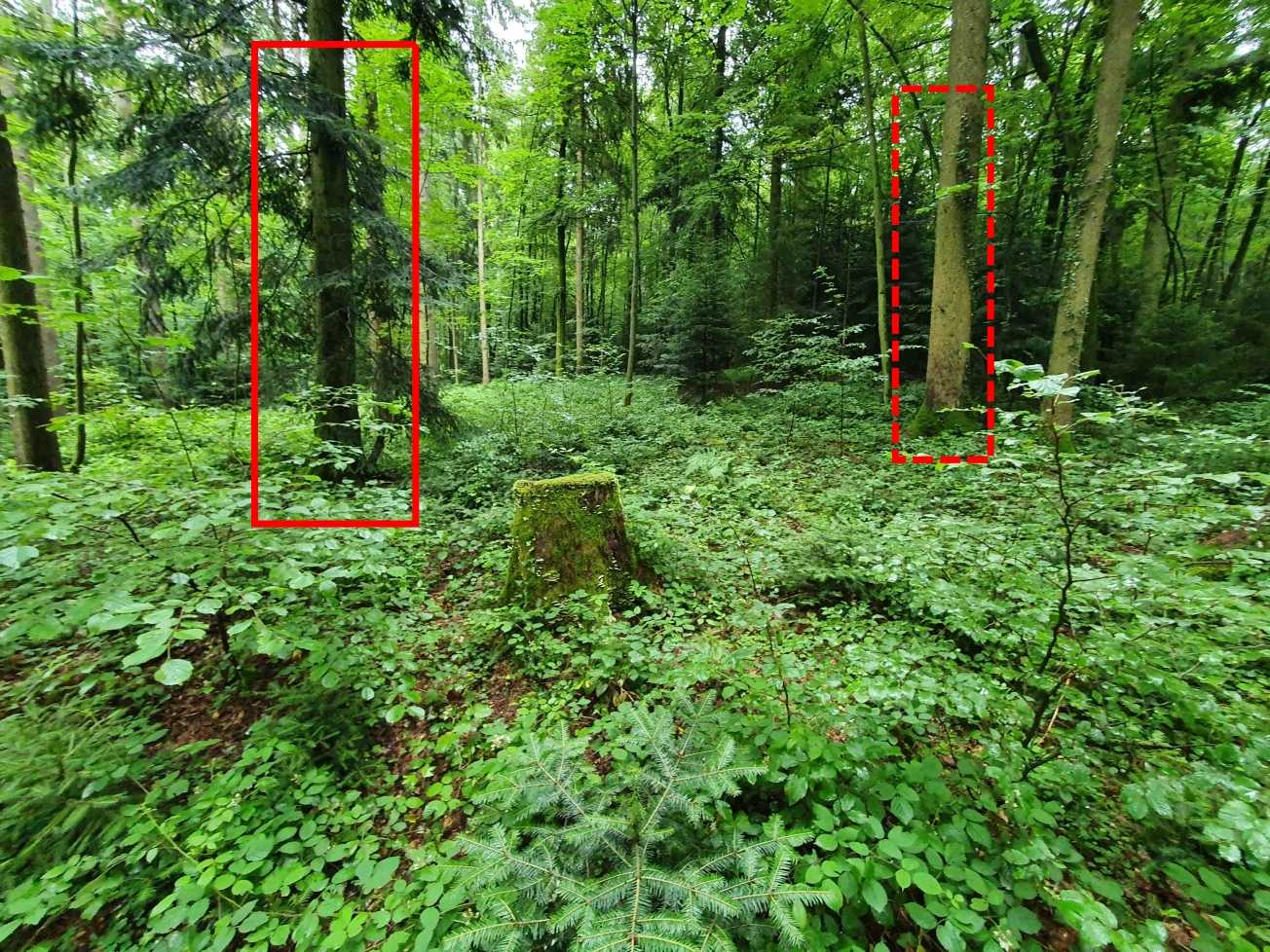}}
     \subfloat[Elevation map \label{fig::map4}]{
       \includegraphics[width=0.36\textwidth]{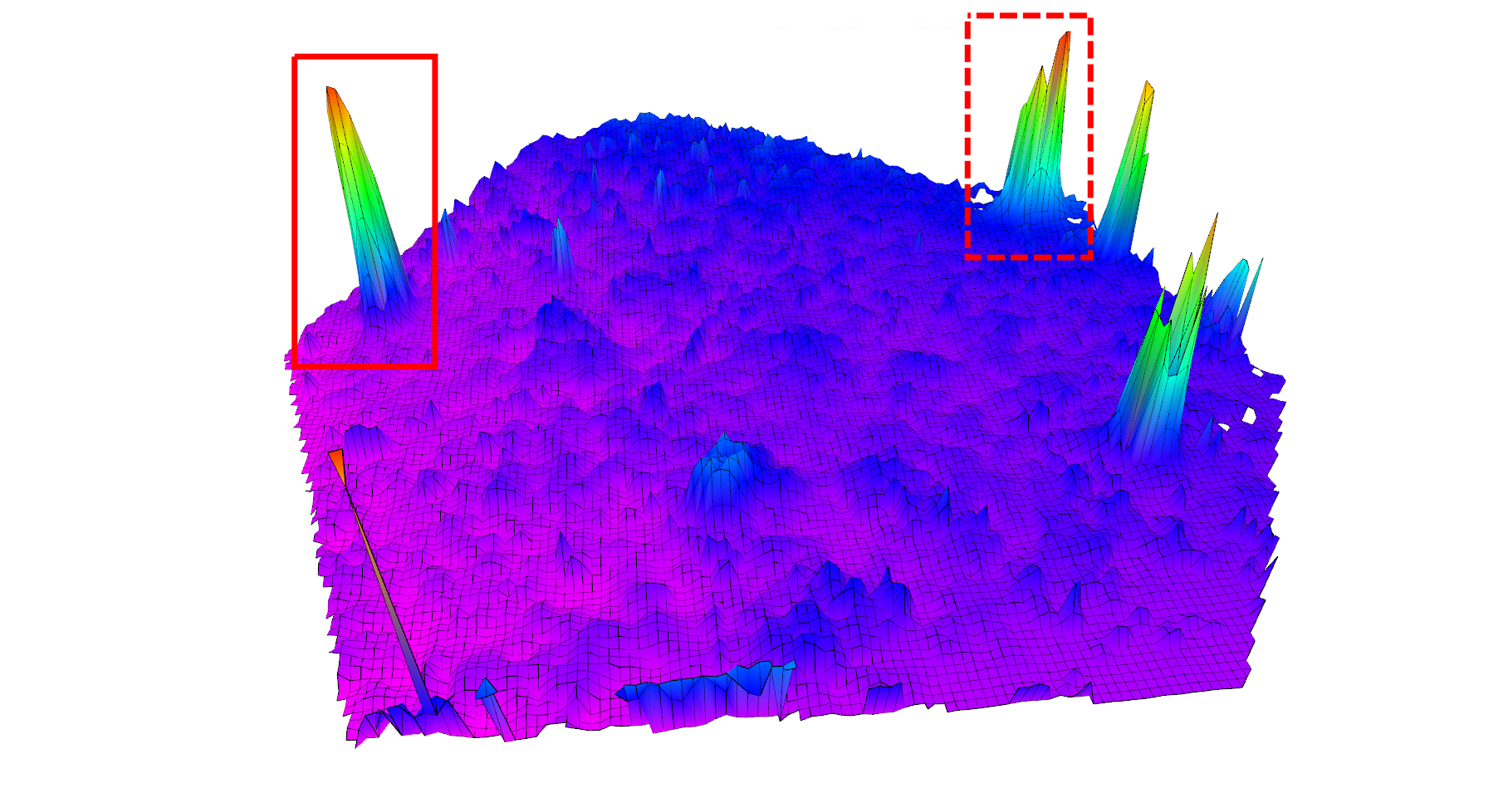}}
     \subfloat[Elevation map + point cloud \label{fig::overlay4}]{
       \includegraphics[width=0.36\textwidth]{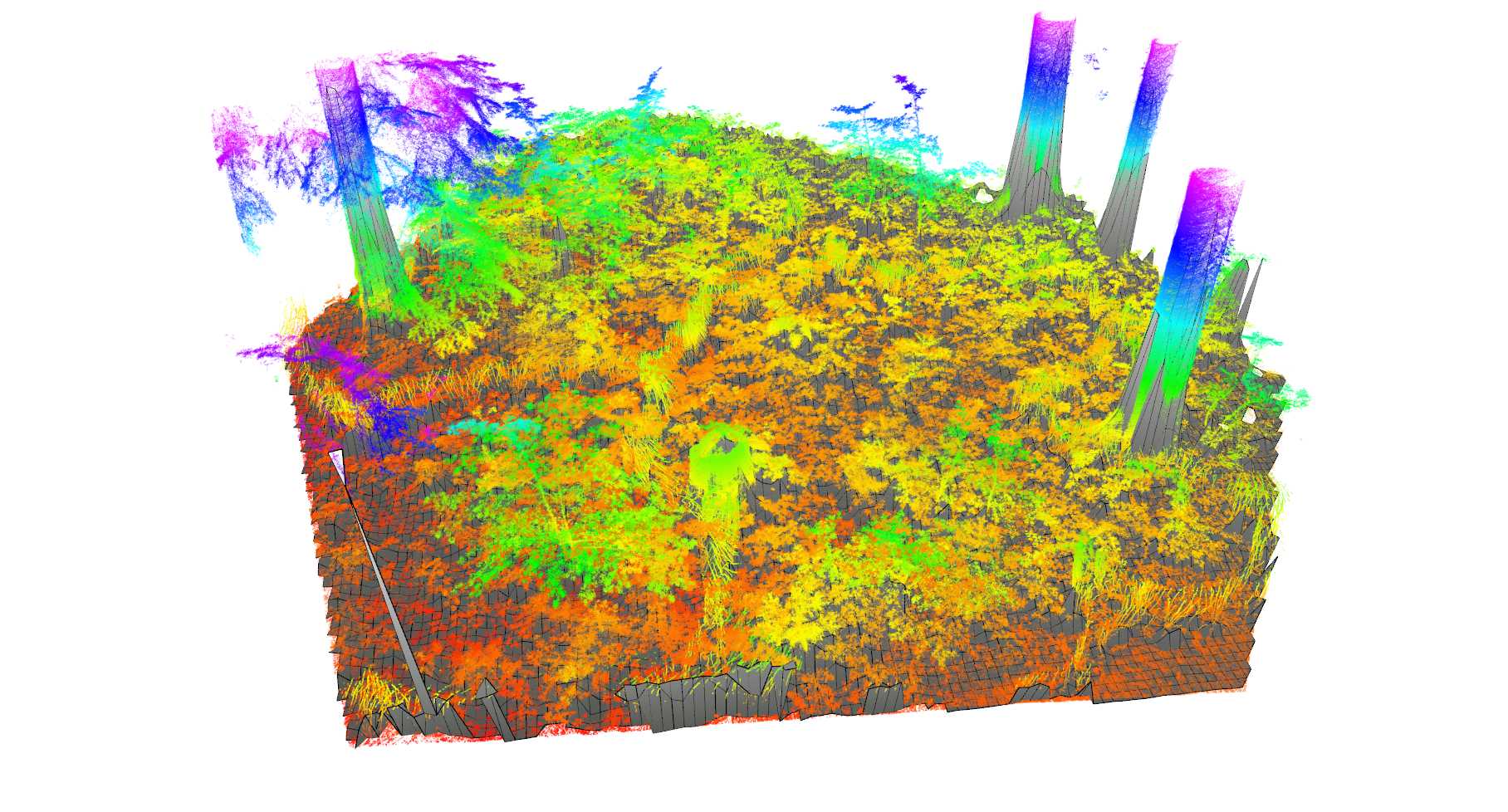}}
    \caption{Medium clutter forest patch with ground vegetation. We can see that the surface of the elevation map looks somewhat rougher compared to the Fig.~\ref{fig::conversion2}. The algorithm successfully captures the tree trunks and filters out some thin trees.}
    \label{fig::conversion4}
\end{figure*}
\begin{figure*}[tbh]
\centering
    \subfloat[Forest patch \label{fig::patch9}]{
       \includegraphics[width=0.27\textwidth]{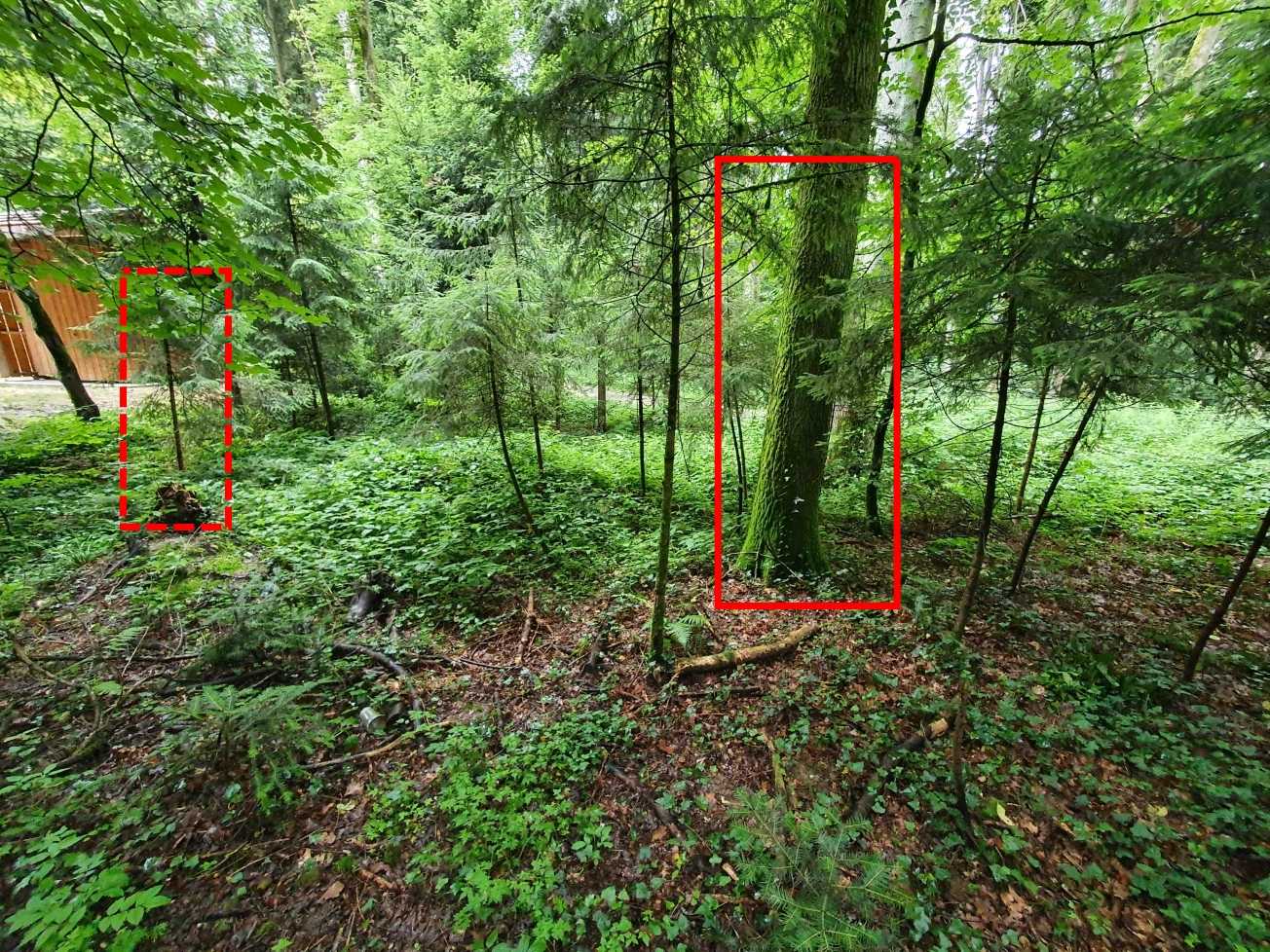}}
     \subfloat[Elevation map \label{fig::map9}]{
       \includegraphics[width=0.36\textwidth]{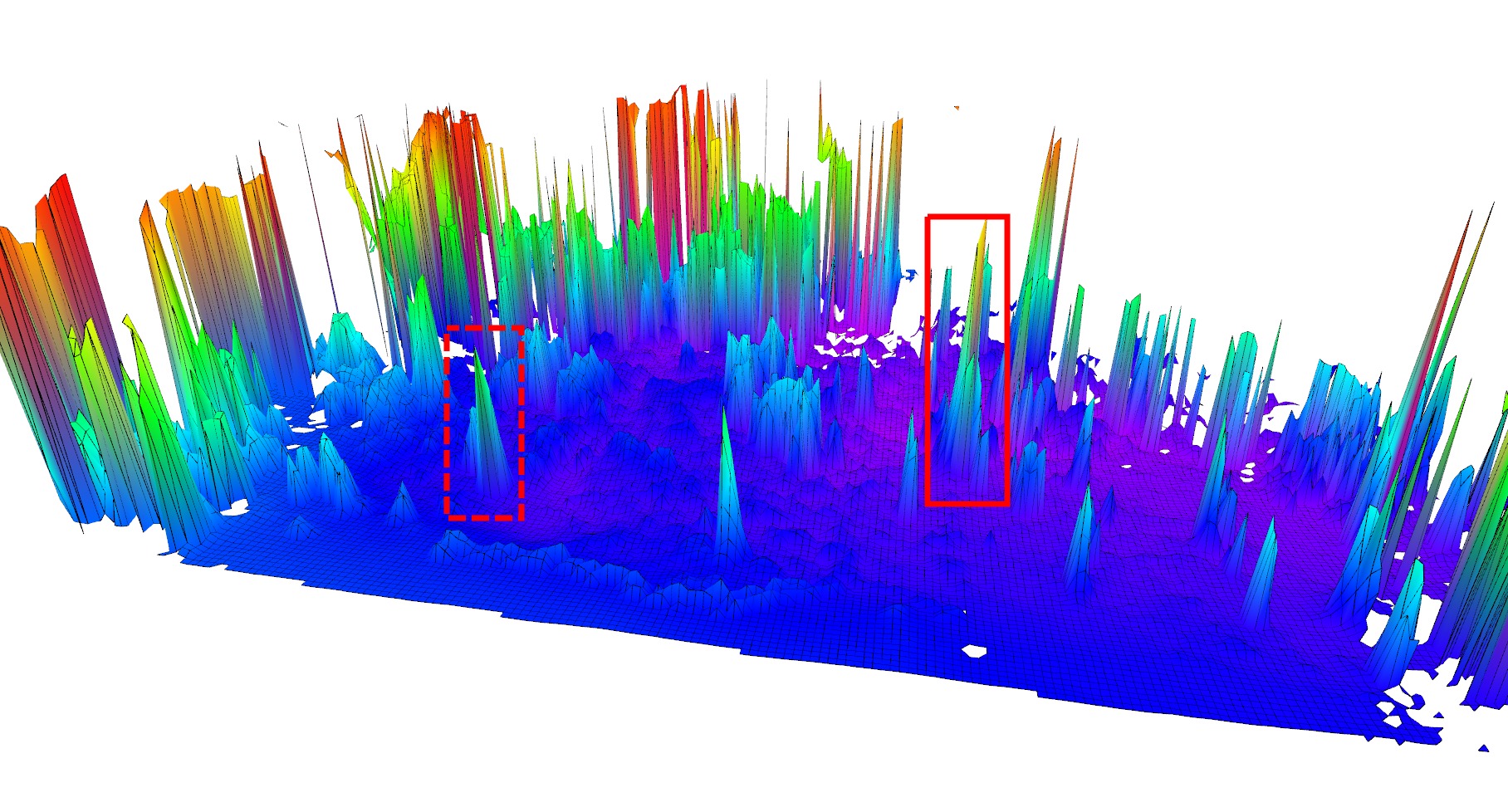}}
     \subfloat[Elevation map + point cloud \label{fig::overlay9}]{
       \includegraphics[width=0.36\textwidth]{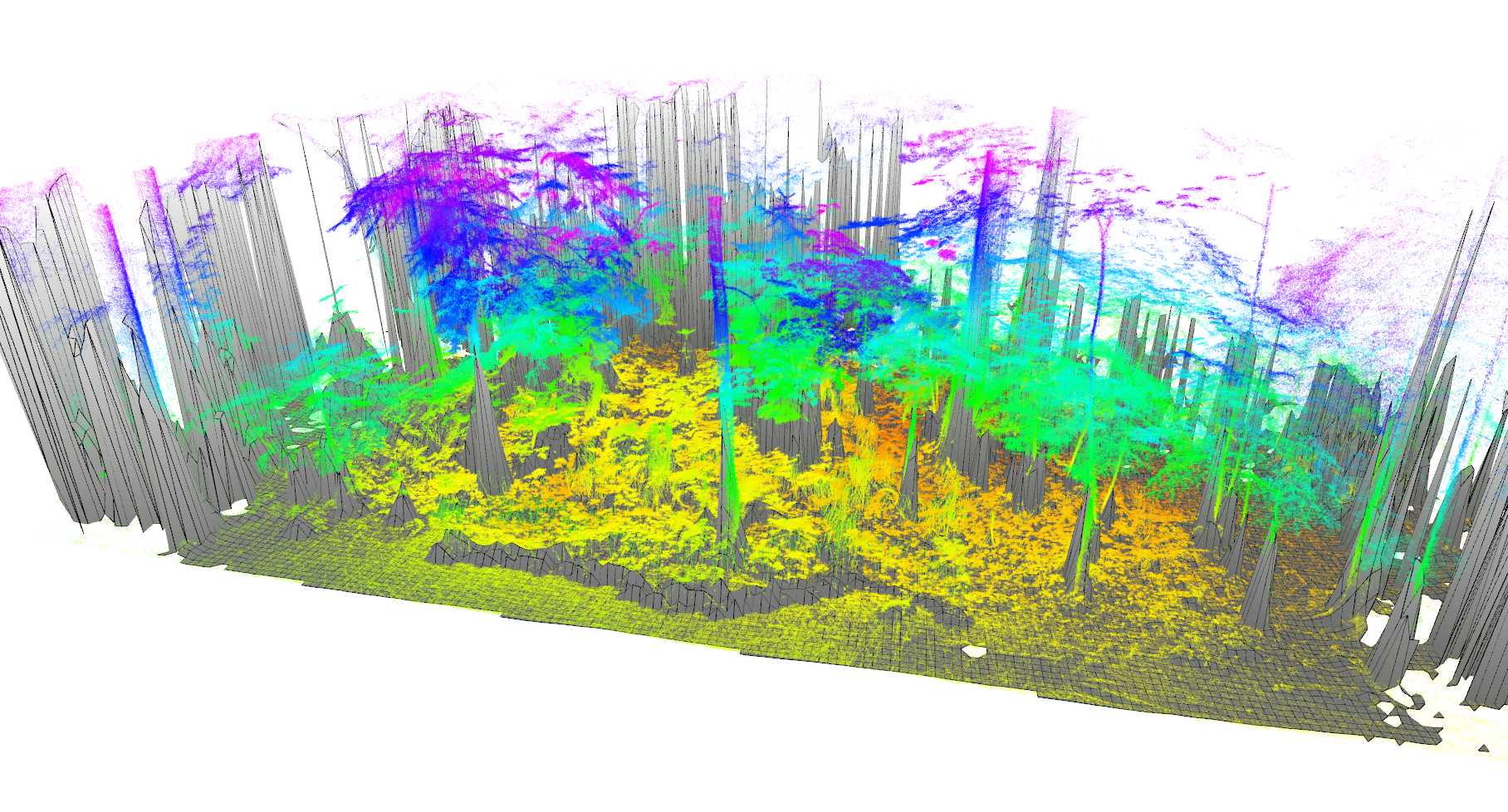}}
    \caption{Combined scene with medium vegetation clutter and trees. Branches reaching all the way to the ground effectively inflate the trunk of the tree which can be seen in the middle of the elevation map (big blobs to the left of the tree trunk).}
    \label{fig::conversion9}
\end{figure*}

\FloatBarrier
\subsection{Tree Detection from Local Maps}
\label{sec::appendix_tree_detection}

\begin{figure}[tbh]
\centering
    \includegraphics[width=0.99\textwidth]{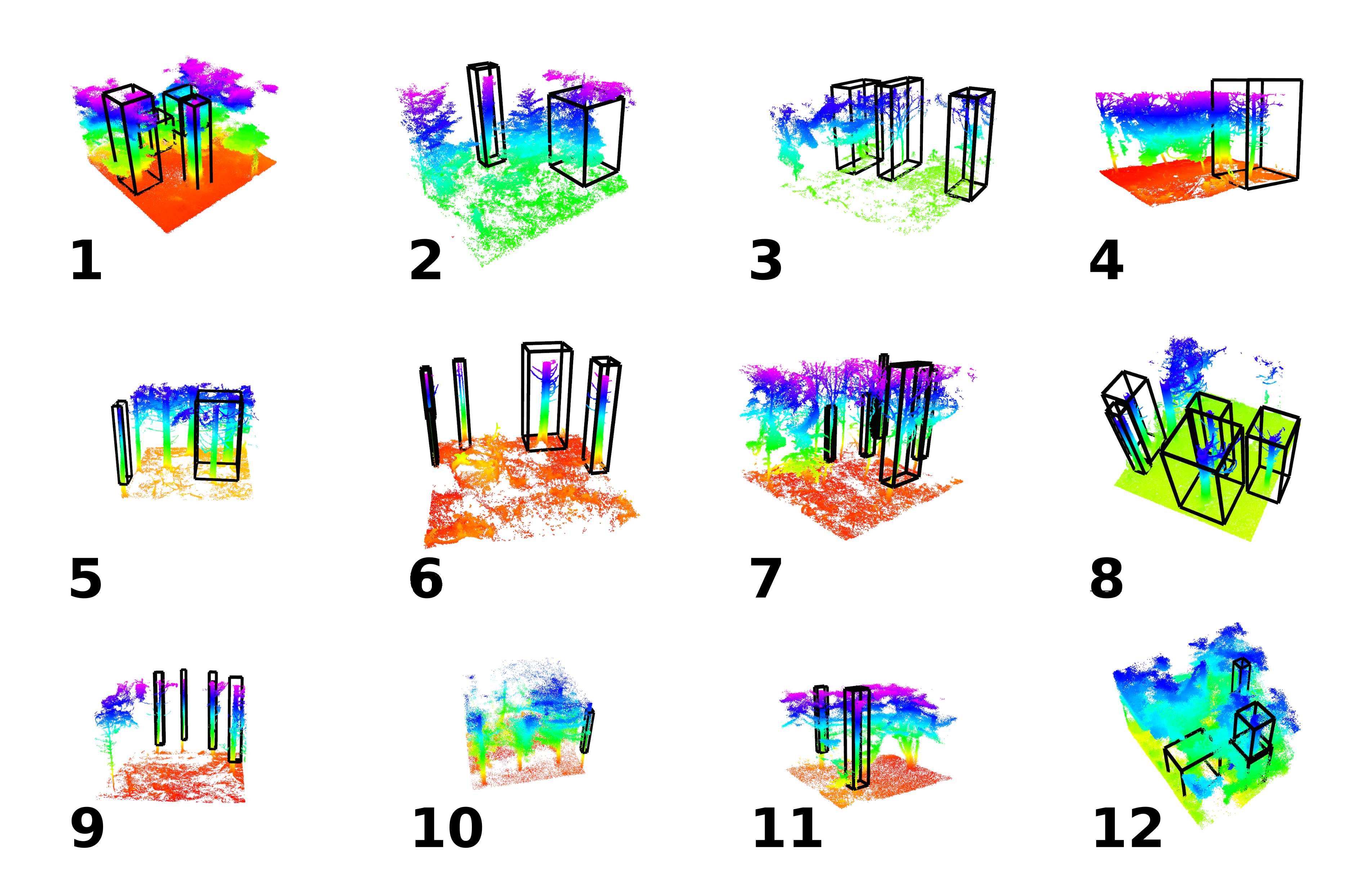}
    \caption{\SI{6}{\meter} by \SI{6}{\meter} point cloud patches where detection algorithm failed. Red corresponds to the lowest and purple to the highest elevation. Tree detections are shown with transparent boxes. The patches come from three different maps that were used for detection accuracy evaluation (see Sec.~\ref{sec::res_tree_detection}. }
    \label{fig::tree_detection_failures}
\end{figure}

In this section, we present some tree detection failure cases on the local map patches. Example failure cases are shown in Fig.~\ref{fig::tree_detection_failures}. The most common cause of failure is a false negative when no tree is detected, and there is one in the map.  False-negative examples can be seen in snapshot 1,2,3,4,5,7,8,9,10,11 and 12. What all these fail cases have in common is the presence of vegetation and clutter in the scene. Some patches are so cluttered (e.g., 1 and 12) that it is extremely challenging for a human to find the trees in the scene. Since our approach extracts Euclidean clusters, any branches or canopy that touches other trees presents a problem. Touching canopies (from two different trees) cause the algorithm to extract two or more trees (together with all the branches) as one cluster. Subsequently, the whole blob of points will be rejected because it is either too wide or the gravity alignment is not vertical enough.  
Ultimately, since our detection approach is purely geometric, it runs into limitations in the presence of high clutter and vegetation. The clutter problem could be mitigated by changing the cluster tolerance in the Euclidean cluster extraction phase. If the point cloud is very dense smaller values (less than \SI{0.1}{\meter}) usually yield better results. Values that are too small will result in a single tree being split into multiple clusters, whereas large values will result in multiple trees merged into a single cluster. This calls for a detection procedure involving a tree model, which would help filter out the branch clutter (e.g., RANSAC). A learning-based approach could also be an option. Alternatively, one could use a different sensor modality that will penetrate through the clutter and allow for direct tree trunk detection (e.g., a radar).

The second failure mode is mistaking a branch for a tree or detecting the same tree twice (false positive). Examples of such behavior can be seen in snapshots 6 and 12. This type of failure is not as common as the false negative. One could extend our approach with simple sanity checks (e.g., if two clusters have similar x and y coordinates than most likely they belong to the same tree) to increase the performance. Another option would be to filter out branches and vegetation, which can be mistaken for a tree. A visual sensor would also help in this case since it is relatively easy to distinguish between vegetation and a tree trunk for a human.
\FloatBarrier
\bibliographystyle{apalike}
\bibliography{frExampleRefs}

\end{document}